\documentclass[preprint,sort&compress,nopreprintline]{elsarticle}

\usepackage[colorlinks=true]{hyperref}
\usepackage{lineno}
\usepackage{graphicx}    
\usepackage{mathtools}
\usepackage{amsmath,amssymb}
\usepackage{multicol}
\usepackage{multirow}
\usepackage{comment}
\usepackage{xcolor}
\usepackage{array}
\usepackage{booktabs}
\usepackage{color,soul}

\usepackage[a4paper, total={6in, 8in}]{geometry}

\usepackage{algorithm}
\usepackage{algpseudocodex}

\usepackage{float}
\usepackage{subcaption} 

\renewcommand\hl[1]{#1} 

\newif\ifdraft
\drafttrue

\definecolor{redtext}{RGB}{255, 100, 100}
\definecolor{bluebox}{RGB}{100, 100, 255}

\usepackage{fvextra}
\DefineVerbatimEnvironment{PromptBlock}{Verbatim}{
    fontsize=\footnotesize,
    breaklines=true,
    breakanywhere=true,
    frame=single,
    framesep=3mm
}

\modulolinenumbers[1]

\newcommand{\targetjournal}{Automation in Construction}
\journal{\targetjournal}
\myfooter[L]{Preprint submitted to \targetjournal}

\begin{document}

\begin{frontmatter}

\title{AstraLOD3: Zero-shot multimodal agentic reconstruction of LOD3 building models}

\author[mymainaddress]{B.G. Pantoja-Rosero\corref{mycorrespondingauthor}}
\cortext[mycorrespondingauthor]{Corresponding author}
\ead{bryan.pr@ntu.edu.sg}

\address[mymainaddress]{Digital Intelligence for Structures and Construction (DISC) Laboratory, School of Civil and Environmental Engineering, Nanyang Technological University, Singapore}


\begin{abstract}

Automated LOD3 building modeling typically relies on purpose-built geometric or learning-based pipelines, limiting flexibility across heterogeneous buildings and input evidence conditions.
This study investigates whether Astra, a general-purpose multimodal foundation model, can address these limitations through zero-shot reconstruction of LOD3 building models within an agentic framework under bounded autonomy.
AstraLOD3 combines multi-view images, calibrated cameras, and a filtered sparse SfM point cloud with a natural-language reconstruction specification, while the Astra agent dynamically selects and executes computational procedures using Python and Blender.
Across 35 runs, including 24 benchmark buildings, AstraLOD3 achieved a mean FRDS of 0.9647 and geometric agreement comparable to that of previous purpose-built methods. Controlled ablations further revealed the effects of reconstruction guidance, evidence modalities, model configuration, and run-to-run variability.
The results demonstrate that structured LOD3 reconstruction can be formulated as a constrained agentic process rather than as a fixed pipeline. 
Future work will investigate adaptive refinement, user-guided correction, task-specific specialization, and damage-aware reconstruction.

\end{abstract}


\end{frontmatter}

\clearpage


\section{Introduction}
\label{sec:intro}

The digitalization of the built environment is increasingly important for supporting more efficient management of buildings and infrastructure throughout their life cycle. Digital representations of physical assets can facilitate activities ranging from planning and construction to operation, inspection, maintenance, renovation, and ultimately deconstruction and reuse~\cite{Lu2020DigitalTwinAnomaly, Gordon2023DeconstructionReuse}. By providing structured information about existing assets, such representations can support their visualization, documentation, analysis, and integration with data-driven workflows for asset management and decision-making~\cite{Zhao2022DigitalTwinOM}. As the number and diversity of existing assets requiring digital documentation continue to grow, scalable approaches for generating reliable digital representations of the built environment are therefore becoming increasingly important.

Among the different forms of digital representation, lightweight and geometrically structured 3D models provide an attractive compromise between geometric fidelity and model complexity. Rather than preserving all captured information indiscriminately, level-of-detail (LOD) representations can retain the geometric features required for a particular application while abstracting details that provide limited additional value for that purpose~\cite{Biljecki2016LODSpecification}. Such representations offer explicit and editable surfaces and architectural components, manageable geometric complexity, and efficient visualization and computation, while also reducing storage and data-transfer requirements. Moreover, organizing building geometry into identifiable components, such as walls, roofs, windows, doors, balconies, and other architectural elements, provides an object-level structure to which additional semantic or application-specific information can subsequently be associated~\cite{Ochmann2016ParametricReconstruction}. These characteristics make lightweight structured models suitable as geometric foundations for a range of downstream workflows, including building information modeling, digital twins, asset documentation and management, numerical analysis, and urban-scale applications~\cite{PantojaRosero2024FEM,Ariss2025SeismicAssessment}.

Structured digital models of existing buildings have traditionally been produced through manual or semi-manual CAD modeling, a process that can require substantial human effort to interpret the physical asset and reproduce its geometry~\cite{Thomson2015AutomaticGeometry}. Reality-capture technologies have progressively reduced the need for direct manual measurement by providing dense visual and geometric observations of existing environments. In particular, image-based photogrammetry and Structure-from-Motion (SfM) can recover three-dimensional scene geometry together with calibrated camera poses from sets of overlapping images~\cite{Schonberger2016SfM}, while laser scanning and LiDAR systems can directly acquire point-cloud representations of the observed geometry~\cite{Pu2009KnowledgeReconstruction}. Nevertheless, reality capture itself remains constrained by factors such as accessibility, occlusions, acquisition time, sensor deployment, and incomplete observational coverage~\cite{Previtali2014OccludedBuilding}. More importantly, the resulting images, camera geometry, and point clouds constitute observations of the physical asset rather than directly usable structured building models, and must therefore be transformed into compact and semantically organized representations suitable for subsequent engineering and built-environment applications~\cite{Xiong2013SemanticBuildingModels}.

Transforming reality-capture observations into structured building models nevertheless remains a challenging task. Conventional automated reconstruction methods commonly rely on combinations of data preprocessing, segmentation, geometric primitive extraction, model fitting, regularization, and component-specific reconstruction procedures~\cite{Huang2022City3D}. Although such constraints can provide effective solutions for particular reconstruction scenarios, their performance often depends on predefined thresholds, geometric priors, and assumptions regarding building configuration, which can reduce flexibility when processing buildings with different architectural characteristics or observations affected by noise, sparsity, occlusion, and incomplete coverage~\cite{Monszpart2015RAPTER}. Learning-based approaches can replace some manually designed reconstruction stages by learning geometric or semantic relationships from data~\cite{Chen2022DeepImplicitBuildings}; however, they generally require task-specific training data and their performance remains dependent on the distributions represented during training~\cite{Liu2024Point2Building}. Consequently, generating detailed structured models from heterogeneous and incomplete observations often requires specialized algorithms, prior assumptions, and multiple processing stages, while the inference and completion of architectural elements that are only partially observed remain particularly challenging.

Recent advances in multimodal foundation models provide a fundamentally different mechanism for addressing such heterogeneous reconstruction tasks. In contrast to conventional learning-based approaches designed and trained for a specific prediction task, general-purpose multimodal models can combine visual and language information within a common instruction-following framework~\cite{Liu2023VisualInstruction}. When coupled with code generation and external computational tools, such models can translate high-level instructions into executable procedures and invoke available software functions~\cite{Liang2023CodePolicies}. In an agentic setting, these capabilities can support dynamic task decomposition, execution of intermediate operations, and iterative evaluation and refinement of their outputs~\cite{Shinn2023Reflexion}. Such systems can therefore operate not only as predictors of predefined outputs but also as goal-directed computational agents capable of selecting, composing, and revising executable actions according to observations obtained during task execution~\cite{Wang2024CodeAct}. These emerging capabilities create the possibility of replacing parts of rigid, purpose-built reconstruction pipelines with flexible, evidence-driven workflows in which the procedure required to transform heterogeneous observations into a structured 3D representation is determined dynamically during execution.

Recent studies have begun to explore large language and multimodal foundation models for building-model generation and editing, BIM interaction, and Scan-to-BIM automation. For example, agentic frameworks have been used to transform natural-language descriptions into editable BIM models~\cite{Du2026Text2BIM}, dynamically query and modify IFC models through tool-augmented reasoning~\cite{Gao2026IFCAgent}, and combine point-cloud instance segmentation with large-language-model knowledge for automated Scan-to-BIM generation~\cite{Liu2025ScanToBIMLLM}. However, the extent to which a general-purpose multimodal foundation model can operate directly as a geometric reconstruction agent remains insufficiently understood. It remains unclear whether a general-purpose multimodal model can autonomously reconstruct building-scale LOD3 models from calibrated multi-view images, camera geometry, and sparse three-dimensional observations. In particular, it is unknown whether this can be achieved zero-shot, without task-specific training or fine-tuning, and without a fixed, purpose-built reconstruction algorithm. Moreover, the influence of the reconstruction-specification guidance provided to the agent, the availability of different input modalities, and the configuration of the underlying foundation model on reconstruction performance has not been systematically characterized. Establishing these capabilities and limitations is necessary to determine whether agentic foundation models can provide a viable and generalizable alternative paradigm for automated building reconstruction rather than serving only as tools for isolated modeling demonstrations.

To investigate this paradigm, this study introduces AstraLOD3, a zero-shot multimodal agentic framework for generating lightweight and semantically structured LOD3 building models. AstraLOD3 uses Astra GPT~\cite{OpenAI2026Astra}, a general-purpose multimodal foundation model, without task-specific training or fine-tuning. Its inputs comprise multi-view images and their associated Structure-from-Motion outputs, including camera calibration parameters, camera poses, and a filtered sparse point cloud, together with a reconstruction specification defining the required model characteristics, constraints, and outputs. From this information, the agent dynamically plans and executes the reconstruction process by generating computational procedures, selecting and using available Python libraries and Blender tools, interpreting the available visual and geometric evidence, and iteratively evaluating and refining the generated geometry. The resulting models consist of geometrically simplified but architecturally meaningful components organized within a semantic object hierarchy and are accompanied by reproducible procedural code and structured reconstruction metadata. Rather than evaluating the approach through isolated demonstrations, AstraLOD3 is systematically investigated across 24 building reconstruction cases, including a controlled synthetic case, together with complementary experiments examining the effects of reconstruction-specification guidance, available input modalities, Astra model configuration, and run-to-run variability. Its reconstruction performance is additionally evaluated using geometric and image-based metrics and compared with previously published purpose-built approaches for automated LOD model generation. Together, these experiments are designed to assess the accuracy, robustness, autonomy, and practical limitations of general-purpose multimodal foundation models as an alternative paradigm for structured building reconstruction.

The main contributions of this study are fourfold. First, it introduces AstraLOD3, a zero-shot multimodal agentic framework that uses the general-purpose Astra GPT foundation model to autonomously transform multi-view imagery and associated SfM information into lightweight LOD3 building models without task-specific training or fine-tuning. Beyond geometric reconstruction, the framework produces architecturally meaningful components organized within a semantic object hierarchy. It also records the evidence supporting reconstructed elements and whether their geometry is observed or inferred, while generating reproducible procedural code and structured reconstruction metadata. Second, the study systematically characterizes the behavior of the agentic reconstruction paradigm. Controlled experiments examine the effect of the guidance encoded in the reconstruction specification, the availability of different input modalities, the Astra model configuration, and independent repeated execution. Third, the framework is evaluated across 24 building reconstruction cases, including a controlled synthetic case. Complementary geometric and image-based metrics, together with comparisons against previously published purpose-built LOD reconstruction approaches, provide quantitative evidence of its capabilities and limitations. Fourth, the study contributes an open input--output benchmark comprising the reconstruction inputs, prompting protocols, generated LOD3 models, reconstruction outputs, and evaluation results. This provides a reproducible reference for future research on multimodal foundation models and agentic 3D reconstruction for the built environment.

\section{Background and Problem Formulation}
\label{sec:problem_formulation}

This section establishes the methodological context and formal reconstruction problem addressed by AstraLOD3. It first summarizes conventional geometry-driven and learning-based approaches for generating structured building models, followed by recent developments in foundation-model-based and agentic 3D model generation. Based on this background, the final subsection formulates the reconstruction problem considered in this study and positions AstraLOD3 with respect to these existing approaches.

\subsection{Automated reconstruction of structured building models}

The generation of structured digital models of existing buildings has traditionally relied on manual or semi-automated interpretation of survey data within CAD environments \cite{Tang2010AsBuiltBIM,Esfahani2021ScanToBIM}. The increasing availability of laser scanning, photogrammetry, and computer-vision techniques has progressively supported the transition toward scan-to-model and related automated reconstruction workflows. In these workflows, geometric and visual observations are processed to recover explicit building geometry, architectural components, and semantic information~\cite{Tang2022BIMGeneration,Justo2021ScanToBIM}. Although manual and semi-automated modeling remain important when detailed interpretation or intervention is required, current research increasingly seeks to reduce this effort by automating object recognition, geometric reconstruction, regularization, and semantic organization from point clouds and images \cite{Czerniawski2020AutomatedDigitalModeling, Mehranfar2024DataDriven}. These developments have led to a broad range of purpose-built reconstruction methods that combine geometric processing, optimization, domain knowledge, and, increasingly, learned visual or geometric information.

Within this broader landscape, a major family of automated reconstruction methods relies on explicitly defined geometric primitives, constraints, and optimization procedures to transform point clouds into compact building representations. Planar segments can be extracted from the observations, intersected to generate candidate surfaces, and jointly selected through optimization to obtain regularized polygonal models, as exemplified by the PolyFit formulation \cite{Nan2017PolyFit}. Related approaches fit parameterized building components directly to measured point clouds or combine geometric extraction with domain-specific constraints to recover structured models \cite{Rausch2021ShapePose}. Such methods commonly rely on assumptions such as planarity, verticality, parallelism or orthogonality, together with thresholds and objective-function weights controlling primitive extraction, data fidelity, model complexity, and point support. For example, a previous LOD3 reconstruction approach combined Structure-from-Motion, point-cloud filtering, planar primitive extraction, and PolyFit-based optimization to generate lightweight LOD2 building envelopes that were subsequently enriched to LOD3 using image-derived information~\cite{PantojaRosero2022LOD3}. These approaches provide strong geometric control and compact outputs, but their reconstruction procedures, geometric assumptions, and parameter settings are generally specified in advance for the intended reconstruction task.

Learning-based methods have increasingly been incorporated into building reconstruction to automate the recognition and interpretation of geometric and semantic information from images and point clouds. In many cases, learned perception is integrated within hybrid workflows that retain explicit geometric processing and domain knowledge. For example, Scan2LoD3 combines multimodal observations, building priors, ray-based geometric reasoning, and probabilistic inference to reconstruct semantic LOD3 building models~\cite{Wysocki2023Scan2LoD3}. Other approaches have extended learning toward larger parts of the reconstruction process. Deep-learning-based point-cloud segmentation has been combined with parametric reconstruction for automated BIM generation~\cite{Mahmoud2024AutomatedBIM}. Learned hierarchical segmentation and skeleton-graph prediction have also been used to reconstruct semantically organized LOD3 parametric models~\cite{Zuo2023HierarchicalReconstruction}. Learned visual information can also be transferred to reconstructed geometry through calibrated image-to-3D mapping; this strategy has been used to associate architectural openings with building envelopes and was subsequently extended to map damage information into damage-augmented digital twins \cite{PantojaRosero2023DADT}. These methods demonstrate the increasing capability of learned models to automate perception and parts of geometric reconstruction. However, their processing architecture, target semantic classes, and interaction between learned and geometric components are generally defined in advance. Their performance and accessible classes also remain influenced by the availability and representativeness of task-specific training data.

Overall, purpose-built reconstruction methods have progressively enabled the generation of increasingly detailed and information-rich building representations by combining geometric processing, optimization, learned perception, and domain-specific knowledge. This progression has extended beyond exterior LOD3 models toward representations that integrate additional semantic information and interior geometry. For example, LOD4 building models have been generated by separately processing exterior and interior SfM reconstructions using planar primitives and subsequently registering them through image-based correspondences~\cite{PantojaRosero2024LOD4}. Such developments demonstrate that specialized reconstruction pipelines can provide robust and geometrically controlled solutions for complex modeling tasks. At the same time, across both geometry-driven and learning-based approaches, the sequence of processing operations, geometric assumptions, target semantic information, and interaction between individual algorithms are largely established during method development. This motivates the investigation of more flexible reconstruction strategies in which part of the computational procedure can instead be formulated dynamically according to the available observations and the required modeling objective.

\subsection{Foundation models and agentic 3D generation}

General-purpose foundation models extend conventional task-specific learning by providing broad pretrained capabilities that can be adapted to new downstream problems without training a separate model for each fixed input--output mapping. More importantly for complex computational tasks, their role can extend beyond generating textual or visual responses when they are coupled with external tools and executable actions. Approaches such as ReAct and Toolformer demonstrated that language models can dynamically select actions or external tools, incorporate the resulting information, and continue the task according to its evolving state \cite{Yao2023ReAct,Schick2023Toolformer}. This capability has subsequently been extended to multimodal settings, where models can combine visual and textual information with specialized tools for perception, generation, and information retrieval \cite{Liu2024LLaVAPlus}. In this setting, heterogeneous information such as images, code, structured files, and tool-accessible geometric data can be incorporated within a common task-solving process. Agents can then decompose a task into intermediate operations, select and execute computational procedures, inspect their outputs, and iteratively refine the solution toward a specified objective.

Foundation models and generative AI have also been increasingly explored for automatic 3D content generation. Direct generative approaches can synthesize three-dimensional representations from textual or visual inputs, for example by optimizing neural representations from text prompts \cite{Poole2023DreamFusion} or predicting 3D representations directly from single images using large-scale learned reconstruction models \cite{Hong2024LRM}. A complementary research direction represents 3D generation as a programmatic task, allowing language or multimodal models to interact with established modeling environments through executable code. SceneCraft, for example, converts textual scene descriptions into Blender Python programs and uses rendered visual feedback to iteratively refine the generated scene \cite{Hu2024SceneCraft}, while MeshCoder learns to transform point clouds into structured and editable Blender Python scripts \cite{Dai2025MeshCoder}. Compared with directly generated neural or mesh representations, programmatic generation provides an explicit and editable description of the modeling operations and enables generated geometry to be regenerated, inspected, and subsequently modified through the corresponding executable procedure.

Agentic foundation-model approaches have recently begun to appear in building modeling and reconstruction. BIMgent uses multimodal large language models as computer-use agents within BIM software~\cite{Deng2025BIMgent}. The agents plan and execute modeling operations through the graphical interface, demonstrating the potential of foundation models for automating extended building-modeling tasks. More directly related to reconstruction, an LLM-enabled multi-agent framework has been proposed for zero-shot Scan-to-BIM and Scan-to-Graph generation from point clouds, integrating query interpretation, geometric inference, topology-aware IFC generation, semantic graph construction, and multi-stage validation~\cite{Pan2026MultiAgentScanToBIM}. A particularly close development is SVI2LOD3 \cite{Kanna2026SVI2LoD3}, which employs an agent-driven workflow combining large language and visual models with expert-designed programs to reconstruct façade openings from volunteered street-view imagery. The method uses zero-shot visual segmentation and geometric projection to enrich existing LOD2 CityGML buildings with windows and doors while preserving semantic and partonomic relationships in the resulting LOD3 representation. This formulation is closely related to the objectives considered here because it combines foundation models, tool-based execution, zero-shot perception, and semantic LOD3 generation; however, its reconstruction task starts from an existing LOD2 building model and primarily focuses on façade-opening enrichment. In contrast, the problem addressed by AstraLOD3 considers the generation of the complete lightweight building model directly from multi-view imagery and SfM-derived camera and sparse geometric information, including the reconstruction of the building envelope, roof configuration, architectural components, and their semantic organization.

\subsection{Problem formulation and positioning of AstraLOD3}

The reconstruction problem considered in this work is formulated as the transformation of a multimodal evidence set \(\mathcal{I}\) and a reconstruction specification \(\mathcal{R}\) into a lightweight, semantically structured LOD3 building model \(\mathcal{M}\). The available evidence is defined as
$$
\mathcal{I}=\{\mathcal{V},\mathcal{C},\mathcal{P}\},
$$
where \(\mathcal{V}\) denotes the available multi-view images, \(\mathcal{C}\) contains the calibrated camera intrinsics and poses associated with the registered images, and \(\mathcal{P}\) is the filtered sparse SfM point cloud. The reconstruction specification \(\mathcal{R}\) defines the modeling objective, evidence rules, geometric and semantic constraints, execution limits, required outputs, and the level of reconstruction guidance provided to the agent. The target representation is expressed as
$$
\mathcal{M}=\{\mathcal{G},\mathcal{S},\mathcal{H},\mathcal{E}\},
$$
where \(\mathcal{G}\) represents the lightweight building geometry, \(\mathcal{S}\) the semantic architectural components, \(\mathcal{H}\) their hierarchical relationships, and \(\mathcal{E}\) the supporting evidence and provenance associated with each component, including whether its geometry is directly observed or inferred. The reconstruction objective can therefore be expressed as
$$
F:(\mathcal{I},\mathcal{R})\rightarrow\mathcal{M},
$$
subject to consistency with the available geometric and image observations, preservation of architecturally relevant components, and compact geometric representation.

In AstraLOD3, the mapping \(F\) is not implemented as a fixed reconstruction algorithm or a task-specific trained network. Instead, the framework operates under bounded autonomy. The reconstruction specification $\mathcal{R}$ defines the objective, evidence rules, geometric and semantic requirements, execution constraints, procedural guidance, and required outputs. Within these boundaries, the agent determines how the requirements are computationally realized. It can inspect the available observations, decompose the reconstruction task into intermediate operations, and generate and execute code. It can also select computational libraries and Blender operations, construct geometric hypotheses, and evaluate intermediate results against the supplied evidence. The reconstruction procedure is therefore instantiated dynamically during execution rather than being completely encoded in advance by the method developer. Importantly, this autonomy remains constrained by the reconstruction specification. Architectural components must be supported by the supplied visual or geometric evidence, and the generated model must remain in the SfM coordinate system. The specified output requirements must also be satisfied, while validation and geometric refinement remain subject to predefined limits. Consequently, AstraLOD3 shifts part of the reconstruction logic from explicit algorithm design before execution to agent-directed selection and composition of computational procedures during execution, while retaining externally defined objectives and constraints.

Within this context, AstraLOD3 investigates whether a general-purpose multimodal foundation model can dynamically organize and execute building-scale LOD3 reconstruction from calibrated multi-view evidence without task-specific training or fine-tuning. In contrast to purpose-built reconstruction pipelines, the framework does not prescribe a complete sequence of reconstruction algorithms or rely on a dedicated learned model trained to map the available observations directly to the desired building representation. Instead, the general-purpose agent is provided with the reconstruction evidence \(\mathcal{I}\) and specification \(\mathcal{R}\), and is responsible for selecting, generating, and coordinating the computational procedures required to recover the building envelope, roof configuration, openings, and other architecturally relevant components while satisfying the defined geometric and semantic constraints. The term \emph{zero-shot} is therefore used here at the reconstruction-task level: no task-specific training or fine-tuning, building-specific pretrained reconstruction pipeline, externally supplied 3D assets, or access to ground-truth geometry is used during reconstruction. It does not imply that the underlying general-purpose foundation model is itself untrained; rather, the study evaluates whether its broad pretrained multimodal and tool-use capabilities can be transferred directly to a previously unseen building-reconstruction task. Accordingly, AstraLOD3 is not positioned as a replacement for a specific geometric or learning-based reconstruction algorithm. Instead, it provides an alternative computational formulation in which heterogeneous reconstruction operations are selected and combined dynamically under bounded autonomy. These operations ultimately produce a lightweight and semantically structured LOD3 representation.

\section{AstraLOD3 Methodology}
\label{sec:methodology}

Building on the problem formulation introduced in Section~\ref{sec:problem_formulation}, AstraLOD3 provides a computational framework for transforming multimodal reconstruction evidence $\mathcal{I}$ and a human-defined reconstruction specification $\mathcal{R}$ into a lightweight and semantically structured LOD3 representation $\mathcal{M}$. The framework uses GPT-6 Astra~\cite{OpenAI2026Astra} (hereafter Astra), a general-purpose multimodal foundation model, through the Codex IDE extension in Visual Studio Code. Codex provides the agentic interface through which the model can access the reconstruction workspace, inspect and modify files, generate and execute code, and interact with external computational tools.

\begin{figure}[H]
    \centering
    \includegraphics[width=\textwidth]{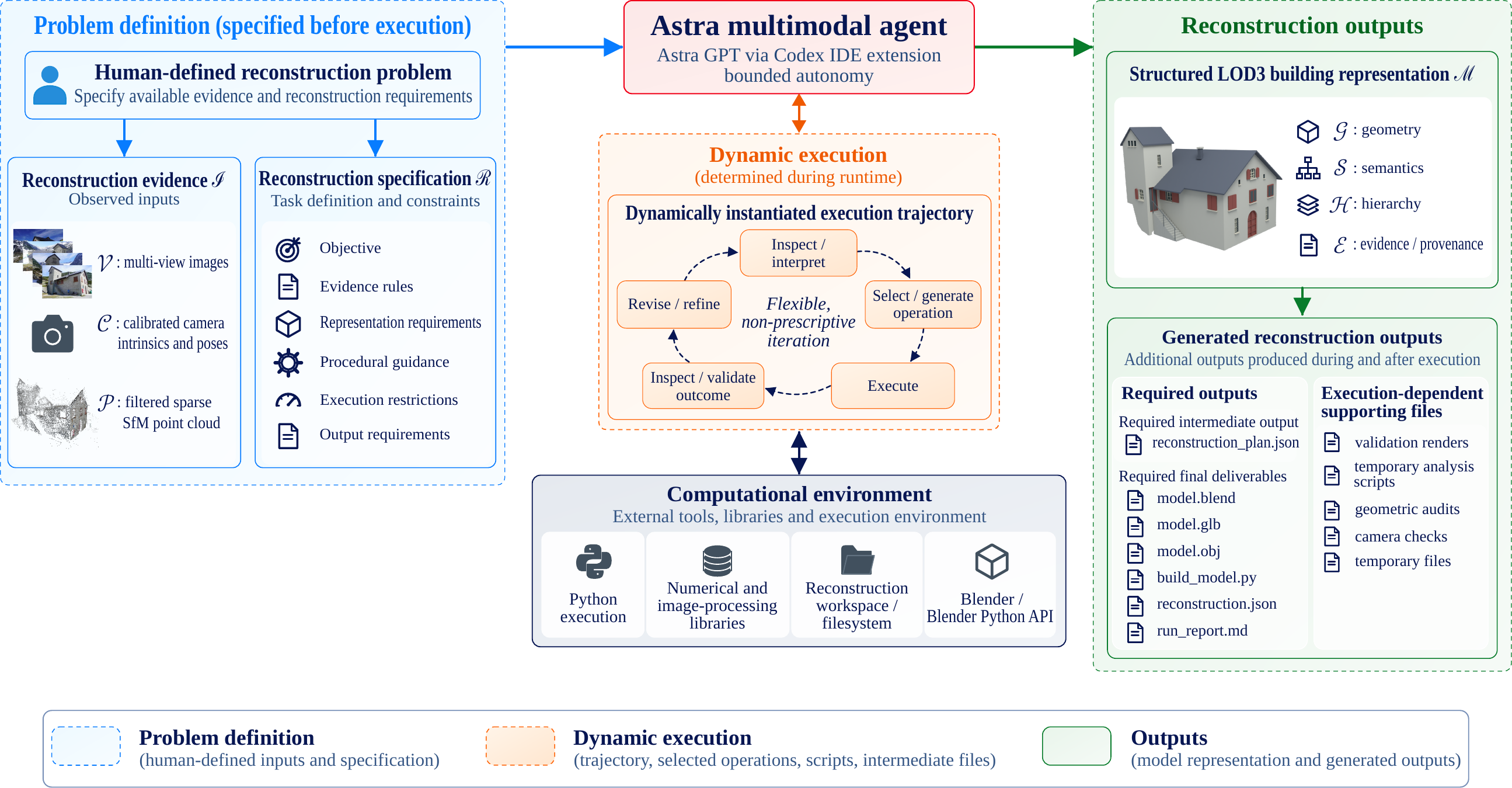}
    \caption{Overview of the AstraLOD3 framework. The reconstruction problem is defined before execution through the multimodal evidence $\mathcal{I}$ and reconstruction specification $\mathcal{R}$. The evidence is provided as files within the reconstruction workspace, whereas $\mathcal{R}$ is communicated to Astra GPT as a natural-language task prompt. Operating through Codex in Visual Studio Code and under bounded autonomy, the agent interacts with the available computational environment to inspect the evidence, execute computational operations, and generate the building model. The reconstruction trajectory is determined dynamically within the specified constraints, producing the structured LOD3 representation $\mathcal{M}$, the required outputs, and execution-dependent supporting files.}
    \label{fig:astralod3_framework}
\end{figure}

Figure~\ref{fig:astralod3_framework} summarizes the overall framework. Two complementary human-defined inputs specify the reconstruction problem. The reconstruction evidence $\mathcal{I}$ is made available to the agent within the reconstruction workspace and comprises multi-view images $\mathcal{V}$, calibrated camera information $\mathcal{C}$, and a filtered sparse SfM point cloud $\mathcal{P}$. The reconstruction specification $\mathcal{R}$ is communicated as a natural-language task prompt defining the reconstruction objective and the boundaries within which the agent operates, including the admissible evidence, geometric and semantic requirements, representation constraints, procedural guidance, execution restrictions, and required outputs. Together, $\mathcal{I}$ and $\mathcal{R}$ define the reconstruction problem without completely prescribing the computational procedure used to solve it.

Operating through Codex, the Astra agent works under bounded autonomy and interacts with an external computational environment providing access to Python execution, numerical and image-processing libraries, filesystem operations, and Blender~\cite{BlenderFoundation2026PythonAPI} for programmatic 3D modeling. During execution, the agent may inspect and transform the available data, formulate intermediate geometric interpretations, generate task-specific scripts, construct and modify geometry, inspect intermediate results, and select additional computational operations according to the evolving state of the reconstruction. Consequently, the exact sequence of operations is not represented as a fixed algorithmic pipeline; instead, the computational trajectory is determined dynamically within the constraints defined by $\mathcal{R}$. AstraLOD3 therefore separates the reconstruction elements specified before execution from the computational procedures determined dynamically for each building and set of observations.

\subsection{Multimodal reconstruction evidence}
\label{sec:reconstruction_evidence}

The reconstruction evidence is defined as
\begin{equation}
    \mathcal{I}
    =
    \{\mathcal{V},\mathcal{C},\mathcal{P}\},
    \label{eq:reconstruction_evidence}
\end{equation}
where $\mathcal{V}$ denotes the set of multi-view images, $\mathcal{C}$ the calibrated camera information associated with the images registered by SfM, and $\mathcal{P}$ a filtered sparse SfM point cloud of the target building. The camera information and sparse three-dimensional geometry are obtained before agent execution by processing the image set with COLMAP-based SfM~\cite{Schonberger2016SfM}. The SfM reconstruction estimates camera intrinsics and poses for the registered images together with a sparse set of triangulated 3D points. The images, calibrated camera information, and subsequently filtered sparse point cloud are stored in the reconstruction workspace and made accessible to the Astra agent through Codex.

More specifically, the visual evidence is represented as
\begin{equation}
    \mathcal{V}
    =
    \{I_i\}_{i=1}^{N},
\end{equation}
where $I_i$ denotes the $i$th input image. Let $\mathcal{J}\subseteq\{1,\ldots,N\}$ denote the subset of images registered by the SfM reconstruction. The associated calibrated camera information is expressed as
\begin{equation}
    \mathcal{C}
    =
    \{K_i,R_i,t_i\}_{i\in\mathcal{J}},
\end{equation}
where $K_i$ contains the camera intrinsics and $(R_i,t_i)$ denotes the corresponding world-to-camera extrinsic transformation recovered by SfM. The sparse point-cloud evidence is represented as
\begin{equation}
    \mathcal{P}
    =
    \{(\mathbf{p}_j,\mathbf{c}_j)\}_{j=1}^{M},
\end{equation}
where $\mathbf{p}_j\in\mathbb{R}^{3}$ is the position of the $j$th retained sparse SfM point and $\mathbf{c}_j$ denotes its associated RGB color.

Before being placed in the reconstruction workspace, the original sparse SfM point cloud is filtered to reduce points associated with dispersed outliers and regions outside the dominant building extent. The filtering consists of statistical outlier removal based on the mean distance from each point to its $k$ nearest neighbours, using a threshold of the form $\mu_d+\alpha\sigma_d$, followed by robust cropping within a percentile-based bounding box aligned with the principal directions of the point cloud using principal component analysis. The resulting point set constitutes $\mathcal{P}$.

The three modalities provide complementary information for reconstruction. The image set $\mathcal{V}$ supplies visual evidence regarding architectural configuration and component appearance, including fa\c{c}ade organization, openings, roof form, projections, and other visible elements. The calibrated camera information $\mathcal{C}$ establishes the projective relationship between registered image observations and the three-dimensional SfM reference frame. It allows observations from different views to be related geometrically and enables the generated model to be projected into the corresponding images. The sparse point cloud $\mathcal{P}$ provides explicit three-dimensional geometric support for sufficiently reconstructed parts of the building, particularly major surfaces and the overall envelope, together with the color information associated with the retained SfM points.

The camera information and sparse point cloud share the coordinate system established by the SfM reconstruction. The reconstruction specification $\mathcal{R}$ requires AstraLOD3 to treat this SfM frame as the common geometric reference and to keep the generated geometry expressed in the same coordinate system. No independent scaling, rotation, or translation of the reconstructed building is allowed. This preserves the geometric relationship among the images, calibrated cameras, sparse point cloud, and generated model and provides a common reference for subsequent validation and quantitative evaluation.

\subsection{Reconstruction specification}
\label{sec:reconstruction_specification}

The reconstruction specification $\mathcal{R}$ is encoded as a natural-language task prompt provided to the Astra agent at the beginning of each reconstruction run. While the reconstruction evidence $\mathcal{I}$ is made available as files within the reconstruction workspace, $\mathcal{R}$ defines how this evidence should be used and establishes the requirements and constraints under which the agent operates. It specifies the reconstruction objective, evidence-use rules, geometric and semantic representation requirements, degree of procedural guidance, execution restrictions, and required outputs. These elements are summarized in Table~\ref{tab:reconstruction_specification}. Rather than prescribing a complete reconstruction algorithm, $\mathcal{R}$ defines the boundaries of the reconstruction problem while leaving the specific computational procedures used to satisfy them to the agent.

\begin{table}[H]
\centering
\caption{Main elements of the AstraLOD3 reconstruction specification $\mathcal{R}$.}
\label{tab:reconstruction_specification}
\small
\begin{tabular}{p{0.28\columnwidth}p{0.64\columnwidth}}
\hline
\textbf{Element} & \textbf{Role in the reconstruction} \\
\hline
Objective &
Defines the target lightweight LOD3 representation and the main reconstruction priorities. \\

Evidence rules &
Defines which observations may support architectural components and how incomplete or unobservable geometry should be treated. \\

Representation requirements &
Defines the architectural scope, geometric simplicity, semantic organization, and preservation of the SfM reference frame. \\

Procedural guidance &
Provides high-level reconstruction stages and refinement limits without prescribing the algorithms used within them. \\

Execution restrictions &
Limits access to external resources, ground truth, external assets, and manual correction during reconstruction. \\

Output requirements &
Defines the required model files, executable reconstruction code, structured metadata, and run documentation. \\
\hline
\end{tabular}
\end{table}

The reconstruction objective is to generate a detail-enriched but lightweight LOD3 representation of the observed building. The specification prioritizes consistency with the supplied SfM geometry and image observations under the calibrated cameras, preservation of architecturally relevant components, and compact geometric representation. Representative components include the main building envelope, fa\c{c}ade planes, roof geometry, openings, simplified frames, balconies, parapets, overhangs, chimneys, and other major architectural projections when supported by the available evidence. Elements that do not contribute meaningfully to the architectural representation, such as vegetation, vehicles, people, cables, furniture, individual bricks, or very small decorative details, are excluded.

A central element of $\mathcal{R}$ is the evidence rule. Architectural components are not introduced solely because they would normally be expected for a particular building type or architectural style. Each modeled component must instead be supported by at least one supplied image or by the available three-dimensional geometry. 
This requirement does not prevent geometric completion where observations are incomplete, provided that inferred geometry is kept geometrically simple and distinguished from directly observed geometry. When part of the geometry is genuinely unobservable but a representation is required to obtain a coherent building model, the agent is instructed to adopt the simplest geometrically reasonable interpretation rather than introducing unsupported architectural detail. The resulting reconstruction metadata records whether major component geometry is directly observed or inferred.

The specification also constrains the form of the generated geometry. Low-complexity procedural representations are preferred, including planar surfaces, extrusions, Boolean openings, repeated or instanced elements, and simple profile geometry, while unnecessarily dense meshes are discouraged. The generated model is also required to remain in the SfM reference frame defined by the reconstruction evidence. Together, these requirements promote compact, structured, and editable architectural geometry rather than a dense surface reproduction of the observed scene.

Procedural guidance forms another component of $\mathcal{R}$. The reconstruction specification organizes the task into six high-level stages comprising evidence inspection, reconstruction planning, programmatic geometry generation, validation, iterative refinement, and finalization, with a maximum of three geometry-refinement iterations. These stages prescribe the intermediate objectives of the reconstruction rather than the algorithms or computational operations used to achieve them; their realization is determined by the agent during execution, as described in Section~\ref{sec:agentic_process}.

The reconstruction specification also defines the execution restrictions and required outputs for each run. The agent is restricted to the supplied reconstruction data and the software available through the computational environment and is not permitted to access external web resources, building-specific pretrained reconstruction pipelines, external 3D assets, ground-truth or evaluation geometry, or manual user corrections during reconstruction. In addition, $\mathcal{R}$ requires a common set of model representations, executable reconstruction code, structured semantic and provenance information, and run documentation. The complete reconstruction specification used for the reference experiments is provided in~\ref{app:reconstruction_specification}. The resulting model representation and generated outputs are described in Section~\ref{sec:generated_outputs}.

\subsection{Agentic reconstruction process}
\label{sec:agentic_process}

The reconstruction specification organizes the AstraLOD3 process into six high-level stages: evidence inspection, reconstruction planning, programmatic geometry generation, validation, iterative refinement, and finalization. These stages define the main objectives to be achieved during reconstruction, but they do not prescribe the algorithms or computational operations used within them. Operating through Codex, the Astra agent determines how each stage is carried out. It inspects the available evidence, generates and executes task-specific code, and uses the computational tools available in the reconstruction workspace. Subsequent operations are adapted according to the intermediate results obtained.

\paragraph{Evidence inspection}
The agent first inspects the available images, calibrated camera information, and sparse point cloud to establish an initial interpretation of the building. This includes identifying the principal building surfaces, approximate envelope and roof configuration, and visible architectural components relevant to the target LOD3 representation. The computational operations used for this inspection are not predefined. Depending on the available evidence, the agent may generate scripts for point-cloud analysis, inspect selected images, visualize geometric information, estimate local reference directions, or apply geometric fitting procedures considered useful for characterizing the building.

\paragraph{Reconstruction planning}
Based on the initial inspection, the agent generates a structured reconstruction plan before creating the building model. This plan is saved as \texttt{reconstruction\_plan.json} within the reconstruction workspace and records the interpreted building planes and dimensions, roof configuration, architectural component inventory, and the image evidence supporting major components. The reconstruction plan therefore provides an explicit intermediate representation between interpretation of the observations and implementation of the building geometry in Blender.

\paragraph{Programmatic geometry generation}
The agent then generates Python code to construct the building programmatically in Blender. Geometry is created through executable procedures rather than through manual interaction with the Blender interface. The agent determines how individual architectural components are represented and may employ planar meshes, solids, extrusions, Boolean operations, repeated elements, profiles, or other low-complexity geometric constructions according to the observed building configuration. The resulting \texttt{build\_model.py} program must be capable of regenerating the complete model from an empty Blender scene.

\paragraph{Validation}
After the initial geometry is generated, representative calibrated cameras are imported into Blender and the model is rendered from the corresponding viewpoints. The resulting renders are inspected against the associated source photographs to identify inconsistencies in overall proportions, component placement, roof geometry, openings, and other reconstructed elements. The specific cameras selected for validation may depend on the building configuration, but the selected validation views are retained for the subsequent refinement cycles to provide a consistent basis for comparison.

\paragraph{Iterative refinement}
Following the initial validation, the agent may perform up to three geometry-refinement iterations. For each refinement, the agent modifies \texttt{build\_model.py}, regenerates the model from an empty Blender scene, renders the same validation cameras, and inspects the remaining geometric inconsistencies. The refinement trajectory is therefore not fixed in advance but emerges from the intermediate reconstruction and validation results. Depending on the case, the agent may modify geometric parameters, component placement, opening dimensions, roof geometry, Boolean constructions, or other aspects of the procedural model considered inconsistent with the available evidence.

\paragraph{Finalization}
After reconstruction and refinement are completed, the agent generates the required final model representations, executable reconstruction code, structured semantic and provenance information, and run documentation. Additional supporting files may also be produced according to the computational procedures followed during the reconstruction; unlike the required outputs, their type and organization can vary between runs. The resulting model representation and generated outputs are described in Section~\ref{sec:generated_outputs}.

\subsection{Model representation and generated outputs}
\label{sec:generated_outputs}

As defined in Section~\ref{sec:problem_formulation}, the target reconstruction $\mathcal{M}$ combines lightweight building geometry, semantic architectural information, component organization, and the evidence and provenance associated with the reconstructed elements. 
In the implemented framework, these elements are distributed across the generated geometric models, executable reconstruction code, and structured metadata rather than being represented by a single mesh alone. The geometric representation contains the reconstructed architectural form, while the accompanying metadata records, for each major component, its identifier, semantic class, dimensions, position, supporting evidence references when available, and whether its geometry is directly observed or inferred.

The reconstruction specification requires a standardized set of final deliverables for each reconstruction run:
\begin{itemize}
    \item \texttt{model.blend}, containing the Blender reconstruction;
    \item \texttt{model.glb}, providing a portable binary glTF representation;
    \item \texttt{model.obj}, providing a conventional mesh representation;
    \item \texttt{build\_model.py}, containing the executable procedural reconstruction;
    \item \texttt{reconstruction.json}, containing component-level semantic, geometric, and provenance information; and
    \item \texttt{run\_report.md}, documenting the data used, reconstruction decisions, refinement process, geometry statistics, and unresolved ambiguities.
\end{itemize}
This common set of outputs provides a consistent basis for storing, inspecting, and evaluating reconstructions across different building cases.

The generated \texttt{build\_model.py} provides an executable and inspectable representation of the geometric solution produced during a particular reconstruction run. When executed from an empty Blender scene, the script regenerates the corresponding final model without manual modeling operations. This property concerns the re-execution of an individual generated geometric solution and should be distinguished from deterministic repeatability of the complete agentic reconstruction process. Independent executions of AstraLOD3 with the same reconstruction evidence and reconstruction specification may follow different intermediate computational trajectories and may produce slightly different geometric solutions.

Beyond the required outputs, the agent may generate additional supporting files according to the needs of each reconstruction. These may include point-cloud analysis scripts, image rectifications, local coordinate representations, geometric audits, validation renders and overlays, camera checks, temporary Blender files, or other intermediate analysis products. Their type, number, and organization are not standardized and may vary between runs. AstraLOD3 therefore fixes the required reconstruction outputs while allowing the intermediate computational procedures and supporting files to adapt to the available evidence and the reconstructed building.

\section{Experimental design}
\label{sec:experimental_design}

The experimental evaluation combines a broad building benchmark with a set of controlled experiments designed to examine different aspects of AstraLOD3. First, the reference configuration is evaluated across 24 buildings with different architectural characteristics and observation conditions. Controlled experiments then investigate the effect of the guidance provided through the reconstruction specification $\mathcal{R}$, the modalities available in the reconstruction evidence $\mathcal{I}$, and the Astra model configuration by comparing the \emph{Light}, \emph{Medium}, \emph{High}, and \emph{Extra-High} settings. Finally, an independent repeated reconstruction is used to examine run-to-run variability under the same experimental conditions. Together, these experiments comprise 35 reconstruction runs. The benchmark and reference configuration are introduced first, followed by the controlled experiments, evaluation metrics, and comparison protocol with previous reconstruction approaches.

\subsection{Benchmark cases and reference configuration}
\label{sec:benchmark_cases}

The main benchmark comprises 24 building cases, listed in Table~\ref{tab:benchmark_cases}. Twenty-three correspond to real buildings represented by multi-view image and SfM datasets, covering different envelope configurations, roof geometries, opening arrangements, architectural styles, levels of occlusion, and degrees of geometric irregularity or damage. Building~24 is a controlled synthetic case originally modeled in Blender and subsequently processed through the same image-based SfM workflow used for the real cases.

\begin{table}[H]
\centering
\caption{Building cases used in the main AstraLOD3 benchmark. The number of images corresponds to the photographs made available to the agent; the number registered by SfM and associated with calibrated camera information may be smaller. The final column identifies cases with corresponding models from the previous LOD3 reconstruction study~\cite{PantojaRosero2022LOD3} or the Damage-Augmented Digital Twin (DADT) study~\cite{PantojaRosero2023DADT}.}
\label{tab:benchmark_cases}
\small
\begin{tabular}{clcc}
\hline
\textbf{ID} & \textbf{Building} & \textbf{Images} & \textbf{Previous work} \\
\hline
01 & Petrinja building              & 131 & DADT \\
02 & Parish house                   & 110 & DADT \\
03 & Petrinja school                & 72  & DADT \\
04 & Country house                  & 91  & DADT \\
05 & School                         & 75  & LOD3 \\
06 & UNIL                           & 148 & LOD3 \\
07 & Ozcan                          & 54  & LOD3 \\
08 & Bianco                         & 47  & LOD3 \\
09 & Paiano                         & 65  & LOD3 \\
10 & Wyss                           & 52  & LOD3 \\
11 & Gerlmerbahn                    & 37  & LOD3 \\
12 & Church Sela                    & 51  & -- \\
13 & Church Popusko                 & 95  & -- \\
14 & Church Gora                    & 35  & DADT \\
15 & House Aegerten                 & 108 & -- \\
16 & House Huttwil                  & 98  & -- \\
17 & House Eriswil                  & 129 & -- \\
18 & UNIL 2                         & 188 & -- \\
19 & Country house 2                & 57  & DADT \\
20 & \'Ecole Roseaux                & 83  & -- \\
21 & Gymnase de Chamblandes         & 28  & -- \\
22 & UNIL 3                         & 136 & -- \\
23 & Country house 3                & 67  & DADT \\
24 & Synthetic house                & 41  & -- \\
\hline
\end{tabular}
\end{table}

For the 24 reference reconstructions, the reconstruction evidence corresponds to the complete multimodal configuration introduced in Section~\ref{sec:reconstruction_evidence},
\begin{equation}
    \mathcal{I}_{\mathrm{VCP}}
    =
    \{\mathcal{V},\mathcal{C},\mathcal{P}\},
    \label{eq:reference_evidence}
\end{equation}
where $\mathcal{V}$ denotes the images, $\mathcal{C}$ the calibrated camera information, and $\mathcal{P}$ the filtered sparse SfM point cloud. The reconstruction specification described in Section~\ref{sec:reconstruction_specification} is used as the reference specification and is denoted as \emph{Normal} in the controlled experiments. The reference Astra model configuration is \emph{High}; the \emph{Light}, \emph{Medium}, \emph{High}, and \emph{Extra-High} configurations are compared separately in the model-configuration sensitivity experiment.

Fourteen of the 24 buildings have corresponding models from previous work. Seven were previously reconstructed using the purpose-built LOD3 reconstruction approach~\cite{PantojaRosero2022LOD3}, while seven correspond to models reported in the Damage-Augmented Digital Twin (DADT) study~\cite{PantojaRosero2023DADT}. These overlapping cases enable direct comparison between AstraLOD3 and the previous reconstruction methods using common evaluation metrics.

\subsection{Controlled ablation and sensitivity experiments}
\label{sec:controlled_experiments}

Building~11, Gerlmerbahn, is used as the common case for the controlled ablation and sensitivity experiments. Its reference reconstruction is already included among the 24 benchmark cases and corresponds to the complete evidence configuration $\mathcal{I}_{\mathrm{VCP}}$, the \emph{Normal} reconstruction specification, and the \emph{High} Astra model configuration. Additional runs vary one experimental factor while keeping the remaining applicable conditions fixed. The experiments examine the amount of guidance encoded in the reconstruction specification $\mathcal{R}$, the modalities available through the reconstruction evidence $\mathcal{I}$, and the Astra model configuration. A separate repeated reconstruction of Building~03 is used to examine run-to-run variability. The complete experimental design is summarized in Table~\ref{tab:ablation_design}.

\begin{table}[H]
\centering
\caption{Controlled ablation and sensitivity experiments. The reference configuration corresponds to Building~11 reconstructed using the \emph{Normal} specification, the complete evidence configuration $\mathcal{I}_{\mathrm{VCP}}$, and the \emph{High} Astra model configuration.}
\label{tab:ablation_design}
\small
\begin{tabular}{p{0.23\columnwidth}p{0.31\columnwidth}p{0.37\columnwidth}}
\hline
\textbf{Varied factor} & \textbf{Configurations} & \textbf{Fixed conditions} \\
\hline

Guidance in $\mathcal{R}$ &
\emph{Normal}, \emph{Short}, \emph{Nano} &
Building~11; $\mathcal{I}_{\mathrm{VCP}}$; \emph{High} Astra model configuration \\

Evidence in $\mathcal{I}$ &
$\mathcal{I}_{\mathrm{VCP}}$, $\mathcal{I}_{\mathrm{VC}}$,
$\mathcal{I}_{\mathrm{VP}}$, $\mathcal{I}_{\mathrm{CP}}$,
$\mathcal{I}_{\mathrm{V}}$, $\mathcal{I}_{\mathrm{P}}$ &
Building~11; \emph{Normal} specification adapted only to the available modalities; \emph{High} Astra model configuration \\

Astra model config &
\emph{Light}, \emph{Medium}, \emph{High}, \emph{Extra-High} &
Building~11; $\mathcal{I}_{\mathrm{VCP}}$; \emph{Normal} specification \\

Run-to-run variability &
Two independent executions &
Building~03; $\mathcal{I}_{\mathrm{VCP}}$; \emph{Normal} specification; \emph{High} Astra model configuration \\
\hline
\end{tabular}
\end{table}

\paragraph{Reconstruction-specification guidance}
Three variants of $\mathcal{R}$ are considered to investigate how the level of human-provided guidance affects the reconstruction. The \emph{Normal} specification corresponds to the complete formulation described in Section~\ref{sec:reconstruction_specification} and reproduced in~\ref{app:reconstruction_specification}. In addition to the reconstruction objective, evidence and representation requirements, execution restrictions, and required outputs, it explicitly organizes the task into the six stages of evidence inspection, reconstruction planning, model generation, camera-based validation, iterative refinement, and finalization. The \emph{Short} specification retains the reconstruction objective, architectural scope, evidence rules, geometric representation requirements, camera constraints, execution restrictions, and required outputs, but removes the prescribed staged workflow. The \emph{Nano} specification further reduces the information provided in $\mathcal{R}$, retaining the reconstruction task, available inputs, principal zero-shot and data-access restrictions, and required outputs while leaving most geometric, semantic, and procedural decisions unspecified. The three variants therefore represent progressively less prescriptive reconstruction specifications rather than simply prompts of different textual length.

\paragraph{Evidence-modality ablation}
The contribution of the different reconstruction modalities is examined by varying the contents of $\mathcal{I}$. Based on the notation introduced in Section~\ref{sec:reconstruction_evidence}, the six evaluated configurations are defined as
\begin{equation}
\begin{aligned}
    \mathcal{I}_{\mathrm{VCP}} &= \{\mathcal{V},\mathcal{C},\mathcal{P}\}, &
    \mathcal{I}_{\mathrm{VC}}  &= \{\mathcal{V},\mathcal{C}\}, &
    \mathcal{I}_{\mathrm{VP}}  &= \{\mathcal{V},\mathcal{P}\}, \\
    \mathcal{I}_{\mathrm{CP}}  &= \{\mathcal{C},\mathcal{P}\}, &
    \mathcal{I}_{\mathrm{P}}   &= \{\mathcal{P}\}, &
    \mathcal{I}_{\mathrm{V}}   &= \{\mathcal{V}\}.
\end{aligned}
\label{eq:evidence_configurations}
\end{equation}
For each configuration, the \emph{Normal} reconstruction specification is adapted only where necessary to remain consistent with the available evidence. Instructions that explicitly require a withheld modality are removed; for example, when images or camera information are unavailable, the corresponding image-inspection or camera-validation instructions are omitted. The remaining reconstruction objective, representation requirements, execution restrictions, and output requirements are retained whenever applicable. This design allows the agent to operate consistently with the evidence actually provided rather than receiving instructions that depend on unavailable inputs. The six configurations cover the relevant combinations containing visual evidence, geometric evidence, or both. A camera-only configuration $\{\mathcal{C}\}$ is not considered because calibrated intrinsics and poses describe the imaging geometry but provide no direct observations of the building appearance or surface geometry from which a meaningful reconstruction could be inferred.

\paragraph{Astra model-configuration sensitivity}
The sensitivity of the reconstruction to the Astra model configuration is examined using the \emph{Light}, \emph{Medium}, \emph{High}, and \emph{Extra-High} configurations. For these runs, Building~11, the complete reconstruction evidence $\mathcal{I}_{\mathrm{VCP}}$, and the \emph{Normal} reconstruction specification are kept unchanged. \emph{High} corresponds to the reference Astra configuration used for the 24-building benchmark, while \emph{Light}, \emph{Medium}, and \emph{Extra-High} provide alternative model configurations for evaluating how the selected configuration affects the resulting reconstruction and computational effort.

\paragraph{Run-to-run variability}
Run-to-run variability is examined using Building~03, Petrinja school. Among the 24 reference reconstructions, this case required the longest processing time and represents one of the more architecturally extensive cases, with long façades, numerous repeated openings, projecting entrances, and a complex roof configuration. It was therefore selected as a demanding case with an extended reconstruction trajectory and substantial opportunities for agent-selected analysis, modeling, validation, and refinement decisions. Because repeating the complete benchmark multiple times would substantially increase the computational cost of the study, the variability analysis is limited to this case. A second reconstruction is initiated independently using the same $\mathcal{I}_{\mathrm{VCP}}$, \emph{Normal} specification, and \emph{High} Astra model configuration, allowing differences in processing time, computational trajectory, and final reconstruction to be examined under unchanged experimental conditions.\\

\noindent To summarize, starting from the 24 reference reconstructions, the controlled study adds two runs using the \emph{Short} and \emph{Nano} specifications, which progressively reduce the guidance encoded in $\mathcal{R}$, five runs that vary the modalities included in $\mathcal{I}$ relative to the complete $\mathcal{I}_{\mathrm{VCP}}$ configuration, and three runs that replace the reference \emph{High} Astra model configuration with \emph{Light}, \emph{Medium}, or \emph{Extra-High}. One additional independent reconstruction of Building~03 is included for the run-to-run variability analysis. Together, these experiments result in 35 reconstruction runs.

\subsection{Evaluation metrics}
\label{sec:evaluation_metrics}

The reconstructions are evaluated using complementary image-space and three-dimensional geometric measures. Image-space agreement is quantified using the Facade Re-projection Dice Score (FRDS)~\cite{PantojaRosero2022LOD3}, 
while geometric consistency with the SfM point cloud is evaluated using the symmetric squared Chamfer distance~\cite{Fan2017PointSet} and the Inliers of Model Fidelity (IMF) metric~\cite{PantojaRosero2022LOD3}. Processing time is additionally recorded for each run. 

\paragraph{Facade Re-projection Dice Score (FRDS)}
For an evaluation view $k$, let $A_k$ denote the binary reference building mask and $B_k$ the mask obtained by projecting the reconstructed model into the same image. FRDS is computed as

\begin{equation}
    \mathrm{FRDS}_k
    =
    \frac{2|A_k\cap B_k|}
    {|A_k|+|B_k|},
    \qquad
    \overline{\mathrm{FRDS}}
    =
    \frac{1}{K}
    \sum_{k=1}^{K}\mathrm{FRDS}_k ,
    \label{eq:frds}
\end{equation}

where $K$ is the number of valid evaluation views. The reported building-level FRDS corresponds to the mean across these views. FRDS ranges from 0 to 1, with larger values indicating greater agreement between the projected reconstruction and the reference masks.

\paragraph{Symmetric squared Chamfer distance}

Let $P=\{\mathbf{p}_i\}_{i=1}^{N}$ denote the geometric coordinates of the filtered SfM point cloud $\mathcal{P}$ and $Q=\{\mathbf{q}_j\}_{j=1}^{M}$ a point set sampled from the reconstructed model surface. Their symmetric squared Chamfer distance is defined as

\begin{equation}
    D_{\mathrm{CD}}(P,Q)
    =
    \frac{1}{2}
    \left[
    \frac{1}{N}
    \sum_{i=1}^{N}
    \min_{\mathbf{q}_j\in Q}
    \left\|
    \mathbf{p}_i-\mathbf{q}_j
    \right\|_2^2
    +
    \frac{1}{M}
    \sum_{j=1}^{M}
    \min_{\mathbf{p}_i\in P}
    \left\|
    \mathbf{q}_j-\mathbf{p}_i
    \right\|_2^2
    \right].
    \label{eq:chamfer}
\end{equation}

The two terms respectively measure point-cloud-to-model and model-to-point-cloud agreement. Lower values indicate greater geometric consistency.

\paragraph{Inliers of Model Fidelity (IMF)}
IMF provides a robust one-sided measure of the agreement between the SfM observations and the reconstructed model. In the implementation used here, the distance from each SfM point to the reconstructed model is evaluated against the sampled finite model surface,

\begin{equation}
    d_i
    =
    \min_{\mathbf{q}_j\in Q}
    \left\|
    \mathbf{p}_i-\mathbf{q}_j
    \right\|_2 .
    \label{eq:imf_distance}
\end{equation}

Let $\mu_d$ and $\sigma_d$ denote the mean and standard deviation of the resulting distances. The inlier set is defined as
\begin{equation}
    \Omega
    =
    \left\{
    i : d_i < \mu_d+\sigma_d
    \right\},
\end{equation}
and IMF is computed as

\begin{equation}
    \mathrm{IMF}
    =
    \frac{1}{|\Omega|}
    \sum_{i\in\Omega}d_i .
    \label{eq:imf}
\end{equation}

Lower IMF values indicate closer agreement between the inlier SfM observations and the reconstructed surface. Unlike the symmetric Chamfer distance, IMF is one-sided and reduces the influence of the larger-distance tail through the inlier criterion.

Both geometric metrics are expressed in the coordinate scale of the corresponding SfM reconstruction. They are therefore compared directly only for the same building or between models that have been explicitly brought to the same scale.

\paragraph{Processing time}
Processing time is recorded from the beginning of each reconstruction run until generation of the final outputs. It represents the elapsed time of the complete agentic process, including agent execution, generated computational operations, Blender execution, validation, and refinement. Because these operations can differ between runs, processing time is not expected to depend only on the number of input images.\\

\noindent Together, FRDS, $D_{\mathrm{CD}}$, IMF, and processing time provide the quantitative basis for the experimental analysis. For buildings with corresponding models from previous work, the same geometric evaluation is applied after the required scale adjustment, as described in the following subsection.

\subsection{Comparison with previous reconstruction approaches}
\label{sec:previous_work_comparison}

For 14 benchmark buildings, corresponding reconstructions are available from the earlier LOD3~\cite{PantojaRosero2022LOD3} and DADT~\cite{PantojaRosero2023DADT} studies. These cases are used to compare AstraLOD3 with the two purpose-built reconstruction approaches. The archived models are evaluated using the same surface-sampling and nearest-neighbour procedures applied to the AstraLOD3 models, while the previously reported FRDS values are retained for image-space comparison.

Note that the point-cloud preprocessing differs between the previous and present studies. The previous evaluations used manually cleaned SfM point clouds in which observations not associated with the building were removed, whereas the present study uses the automatic statistical and spatial filtering described in Section~\ref{sec:reconstruction_evidence}. Consequently, the geometric metrics follow the same formulation but are not evaluated against identical point sets. The automatically filtered clouds can retain residual non-building observations, and this difference should be considered when interpreting the comparisons.

Additionally, because some archived models are represented at a different scale from the SfM reconstruction used in the present experiments, their geometric metrics are converted to the current reconstruction scale. Let

\begin{equation}
    s
    =
    \frac{d_{\mathrm{old}}}{d_{\mathrm{current}}}
\end{equation}
denote the scale factor obtained from corresponding reference dimensions. The corrected metrics are then

\begin{equation}
    \mathrm{IMF}_{\mathrm{scaled}}
    =
    \frac{\mathrm{IMF}_{\mathrm{old}}}{s},
    \qquad
    D_{\mathrm{CD,scaled}}
    =
    \frac{D_{\mathrm{CD,old}}}{s^2}.
    \label{eq:previous_scale}
\end{equation}

\noindent The scale correction enables case-by-case comparison within a common coordinate scale, while differences in point-cloud preprocessing are retained as part of the respective reconstruction workflows and are considered in the interpretation of the results in Section~\ref{sec:comparison_results}.

\section{Results and discussion}
\label{sec:results_discussion}

All 35 reconstruction runs produced the required final outputs. Following the experimental design introduced in Section~\ref{sec:experimental_design}, the analysis first considers the 24-building reference benchmark and its comparison with the previous purpose-built reconstruction approaches. The controlled experiments are then used to examine the effect of the guidance encoded in $\mathcal{R}$, the modalities available in $\mathcal{I}$, the Astra model configuration, and run-to-run variability. Finally, the execution behavior observed across the experiments is discussed together with the main limitations of the framework.

\subsection{Overall reconstruction performance}
\label{sec:overall_results}

\begin{figure}[H]
    \centering

    \begin{subfigure}[t]{0.16\textwidth}
        \centering
        \includegraphics[width=\linewidth]{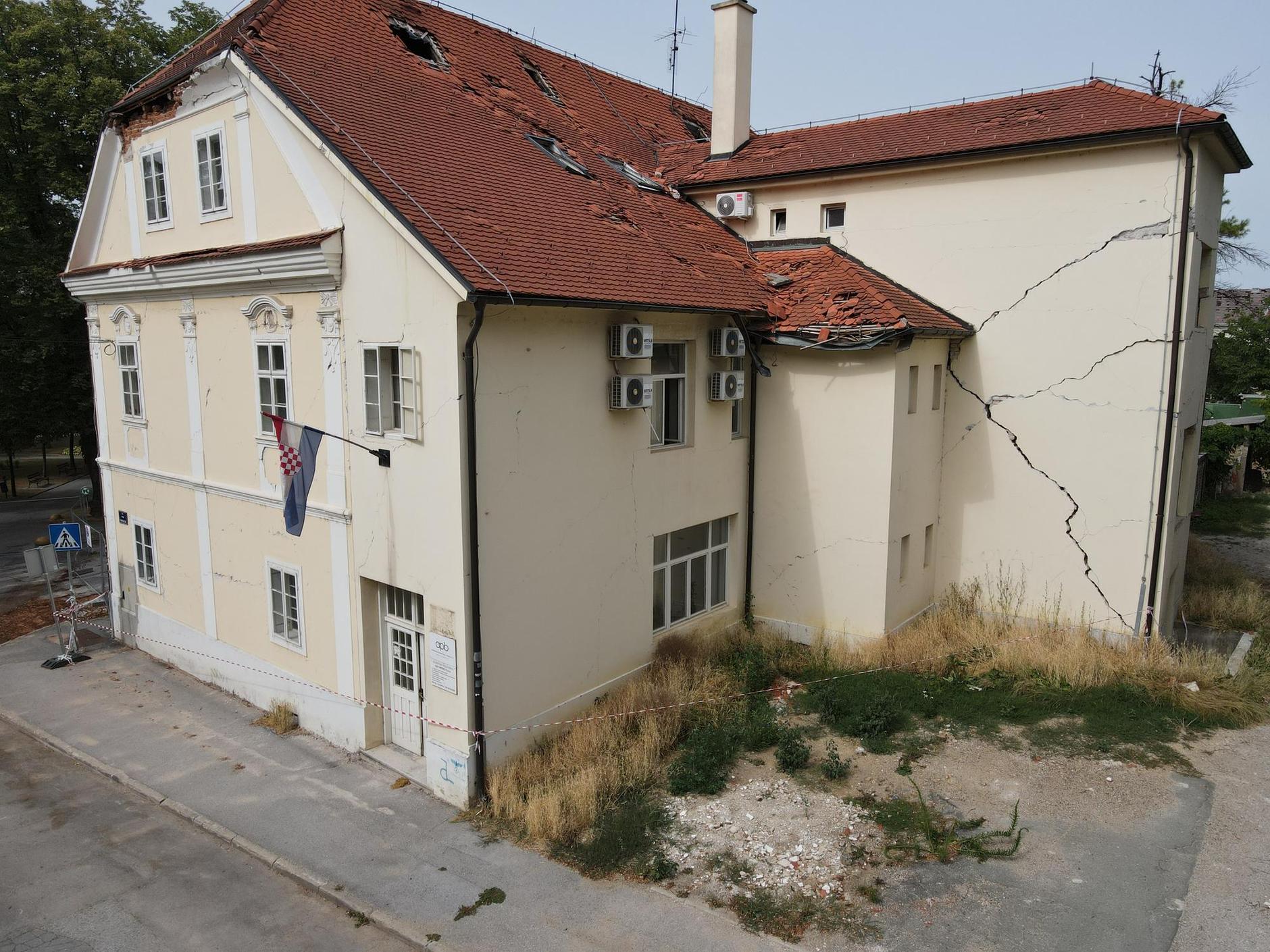}
    \end{subfigure}\hfill%
    \begin{subfigure}[t]{0.16\textwidth}
        \centering
        \includegraphics[width=\linewidth]{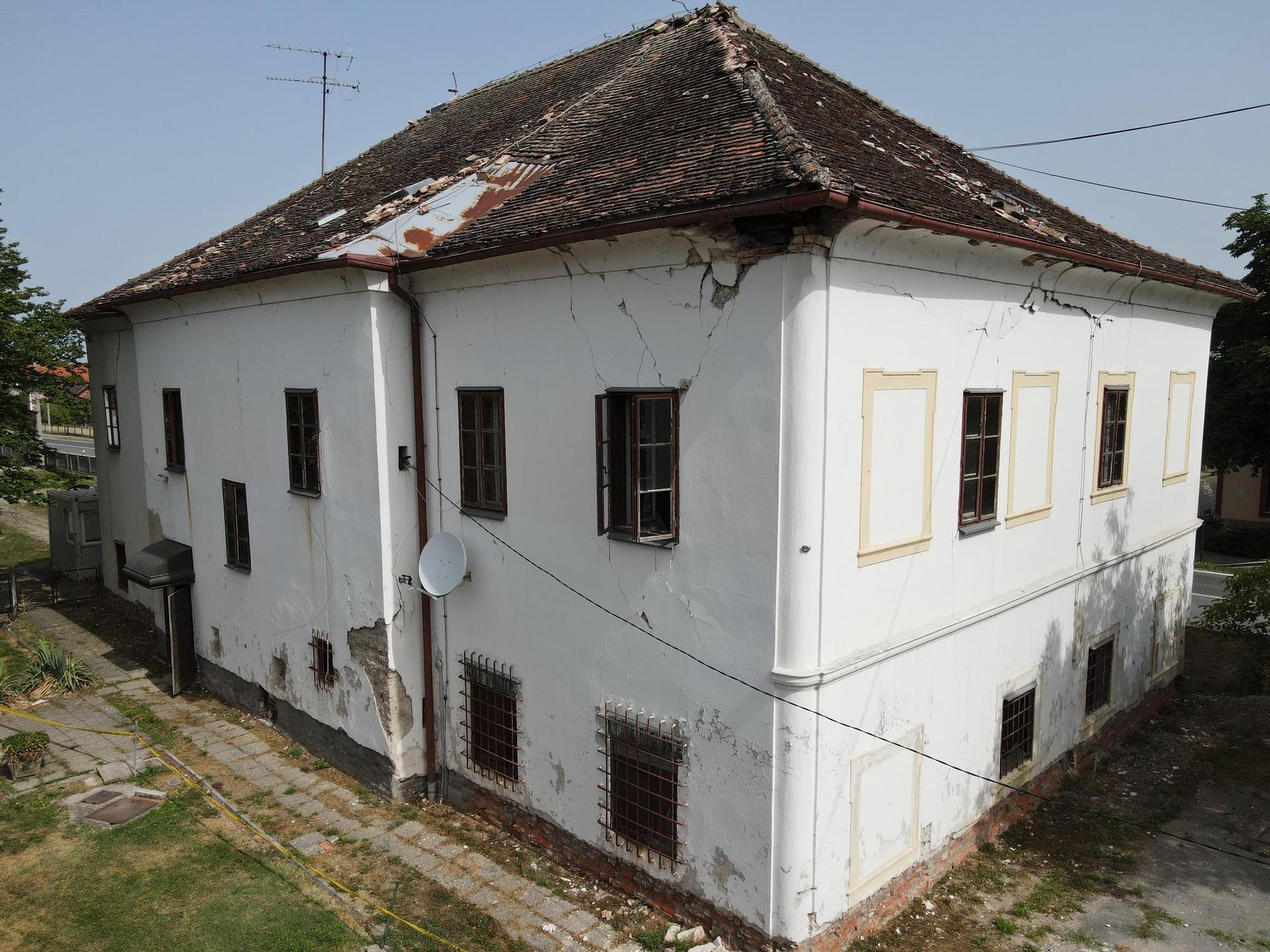}
    \end{subfigure}\hfill%
    \begin{subfigure}[t]{0.16\textwidth}
        \centering
        \includegraphics[width=\linewidth]{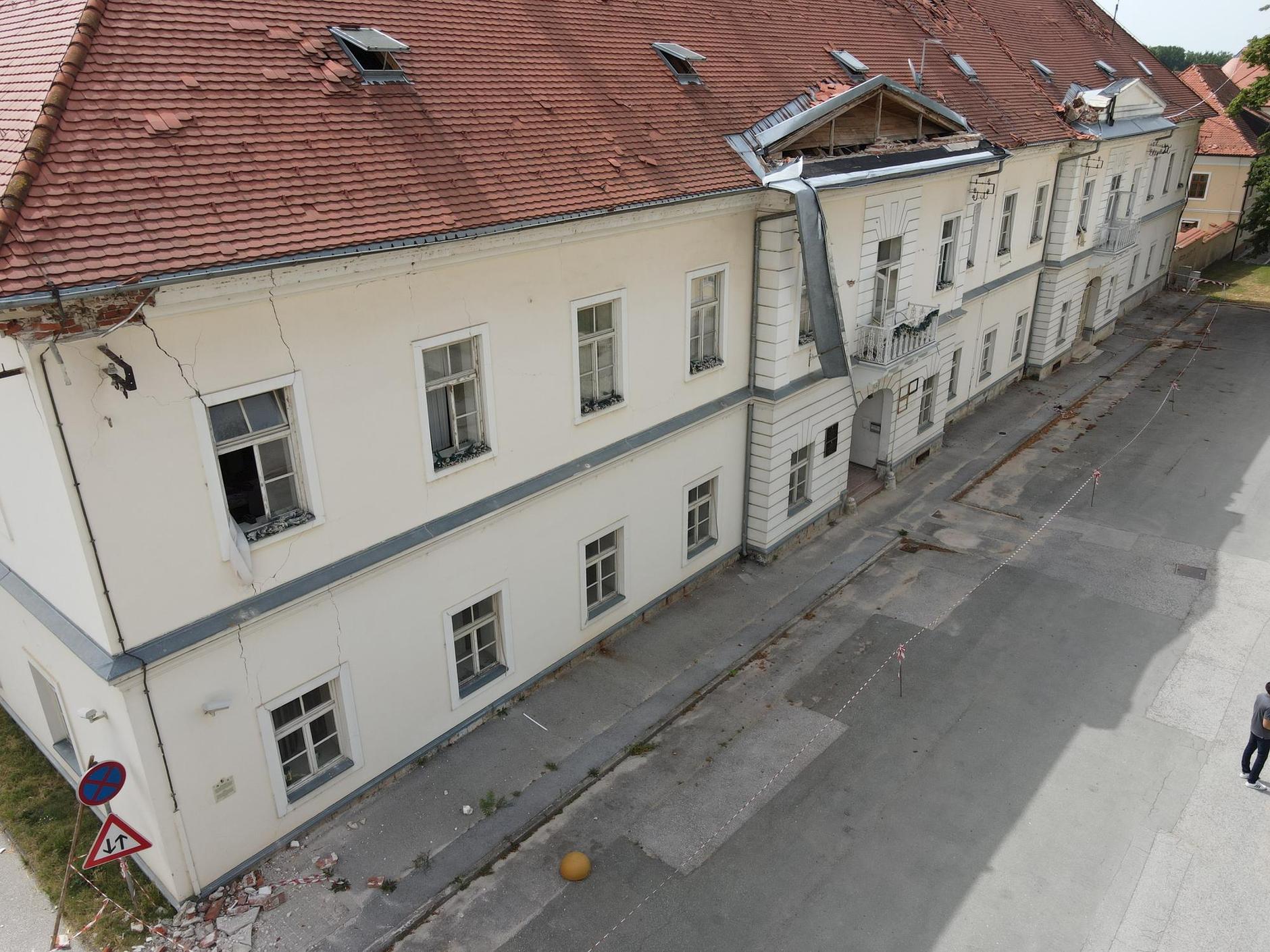}
    \end{subfigure}\hfill%
    \begin{subfigure}[t]{0.16\textwidth}
        \centering
        \includegraphics[width=\linewidth]{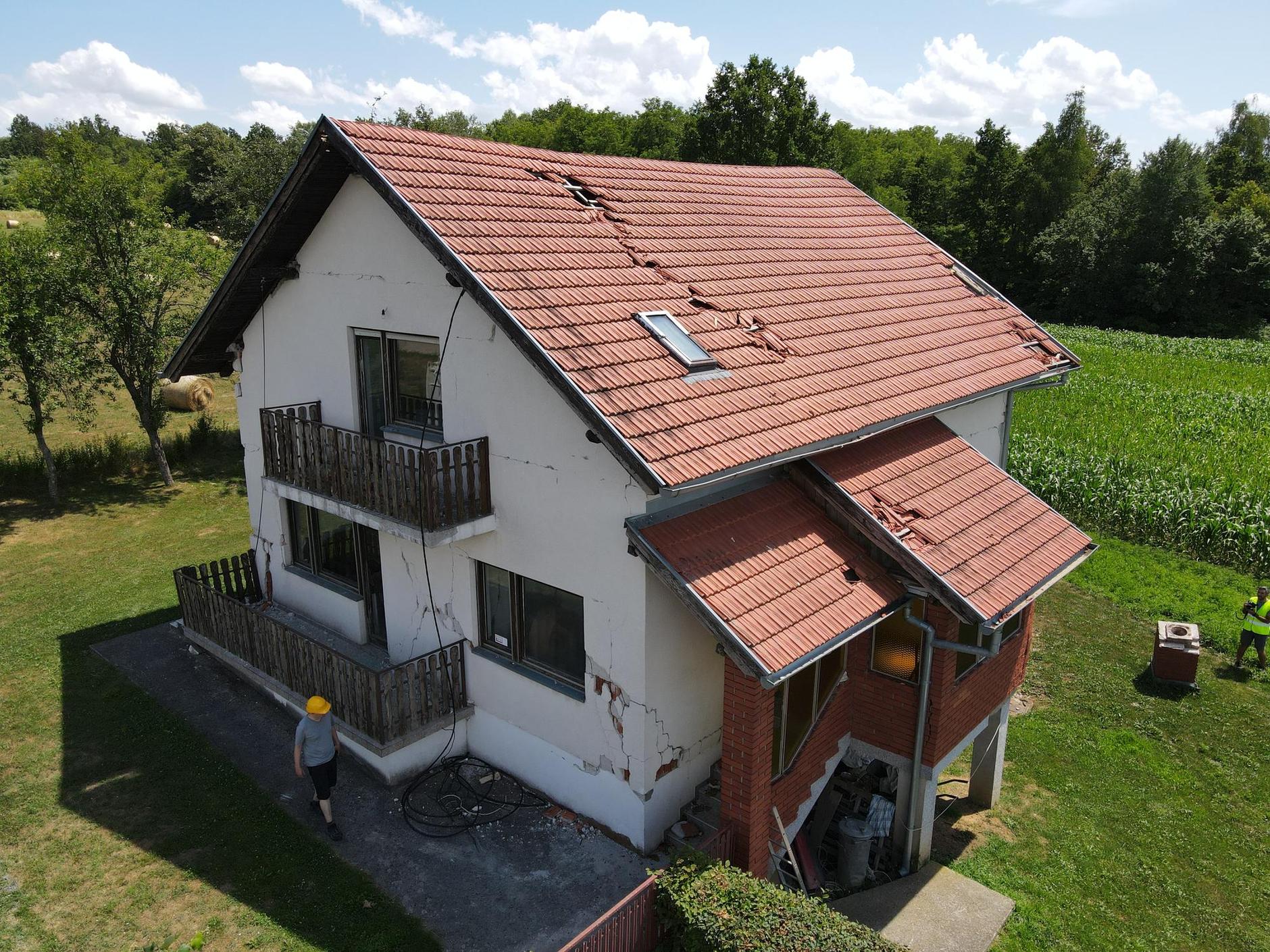}
    \end{subfigure}\hfill%
    \begin{subfigure}[t]{0.16\textwidth}
        \centering
        \includegraphics[width=\linewidth]{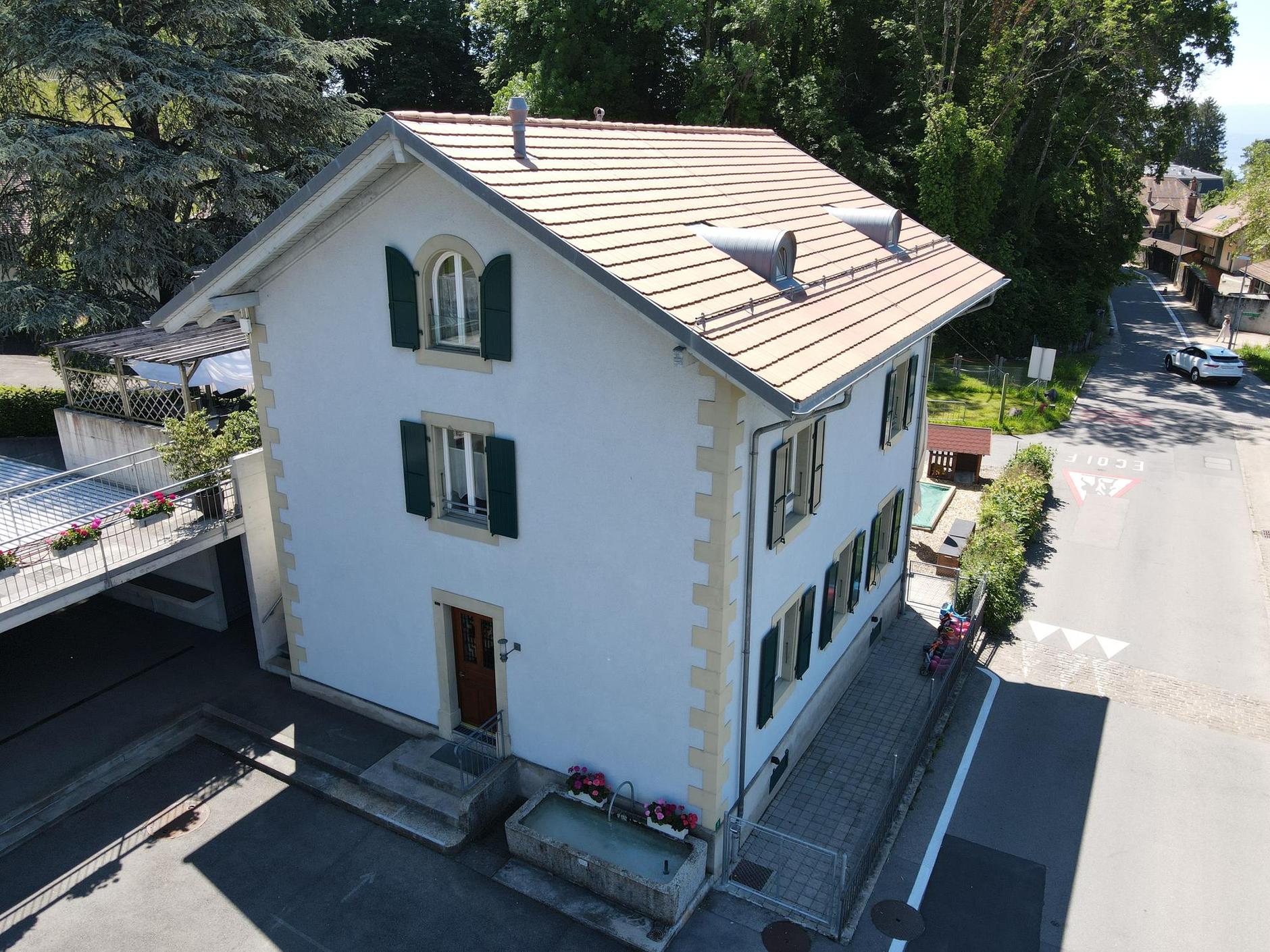}
    \end{subfigure}\hfill%
    \begin{subfigure}[t]{0.16\textwidth}
        \centering
        \includegraphics[width=\linewidth]{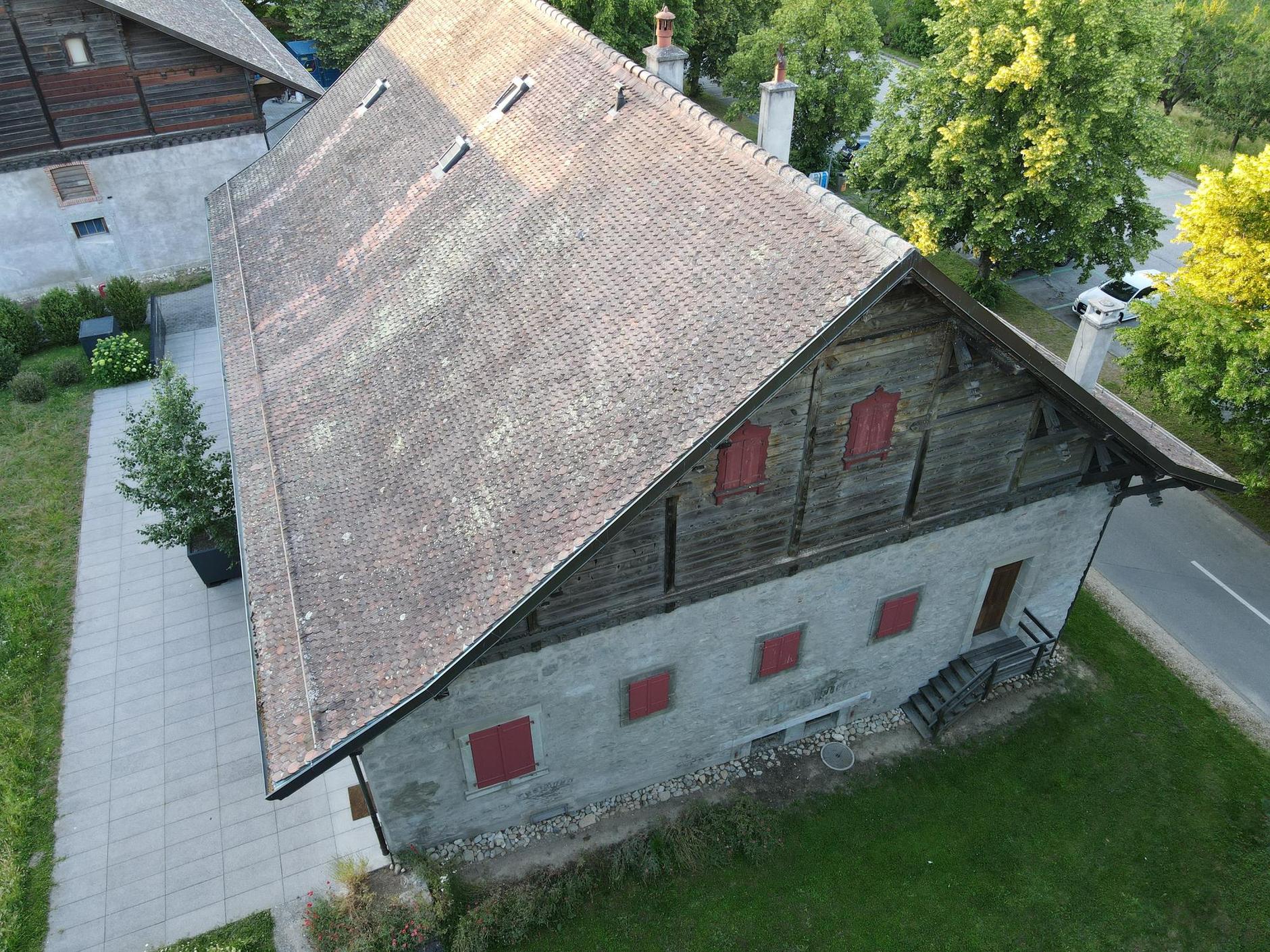}
    \end{subfigure}

    \par\smallskip

    \begin{subfigure}[t]{0.16\textwidth}
        \centering
        \includegraphics[width=\linewidth]{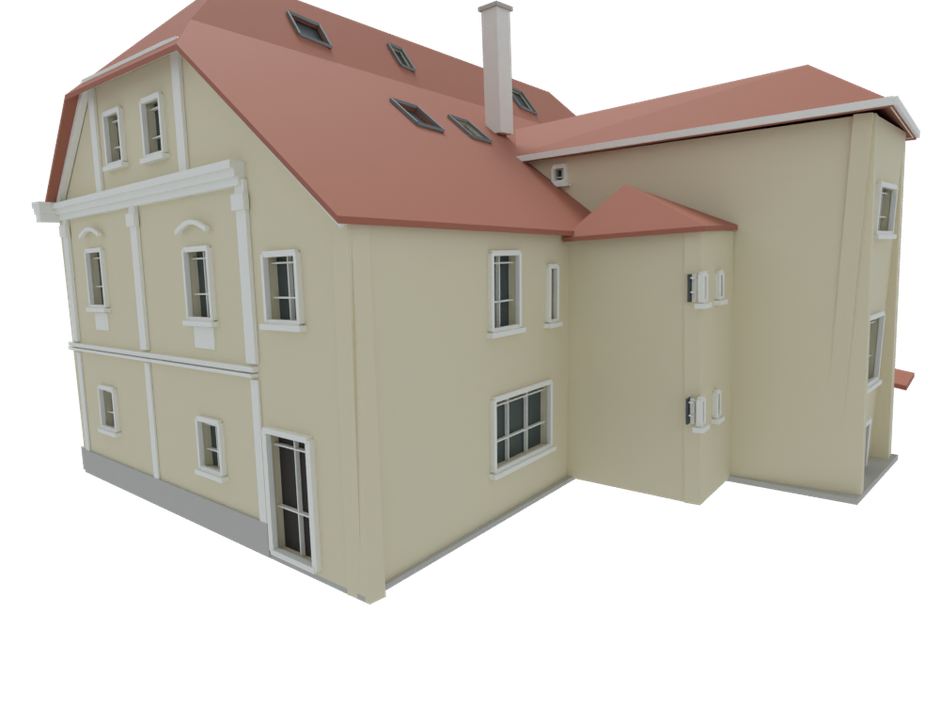}
        \caption*{01}
    \end{subfigure}\hfill%
    \begin{subfigure}[t]{0.16\textwidth}
        \centering
        \includegraphics[width=\linewidth]{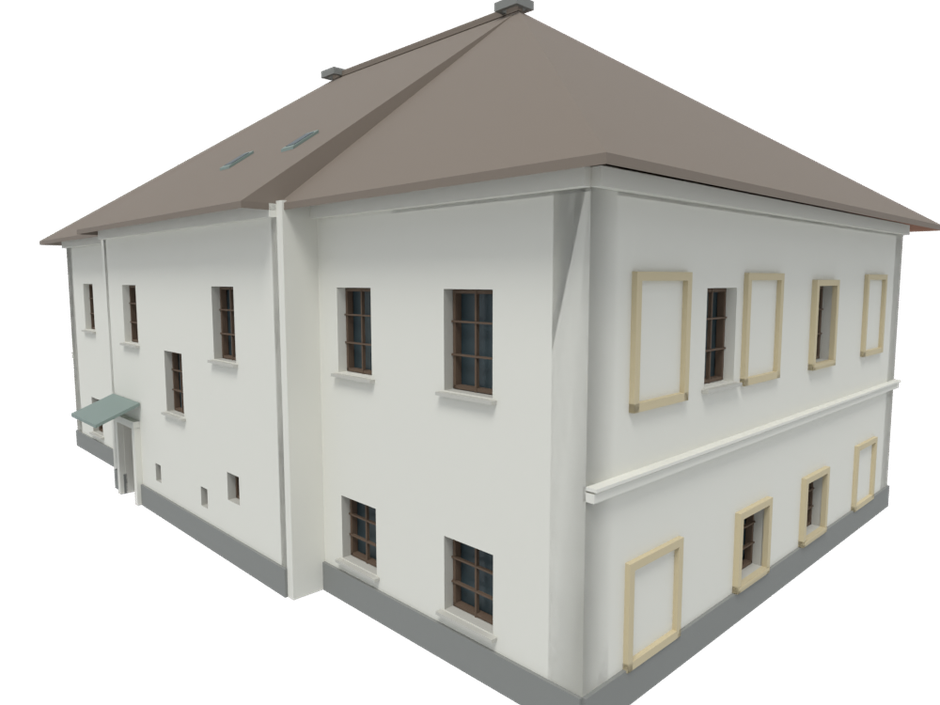}
        \caption*{02}
    \end{subfigure}\hfill%
    \begin{subfigure}[t]{0.16\textwidth}
        \centering
        \includegraphics[width=\linewidth]{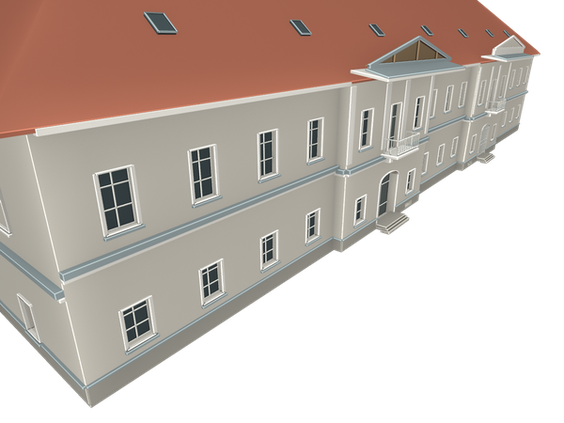}
        \caption*{03}
    \end{subfigure}\hfill%
    \begin{subfigure}[t]{0.16\textwidth}
        \centering
        \includegraphics[width=\linewidth]{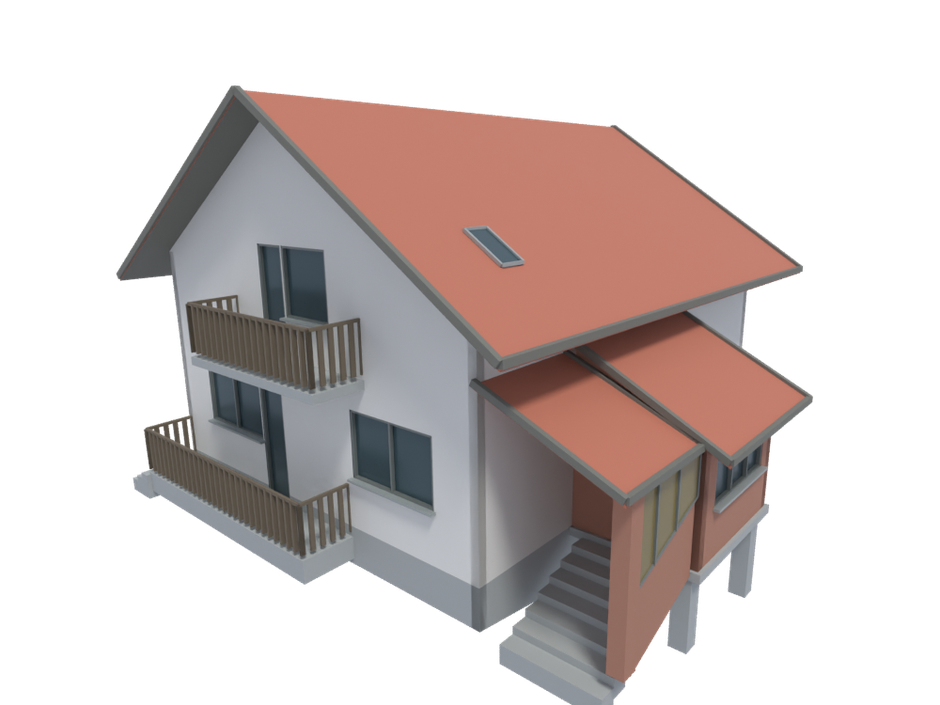}
        \caption*{04}
    \end{subfigure}\hfill%
    \begin{subfigure}[t]{0.16\textwidth}
        \centering
        \includegraphics[width=\linewidth]{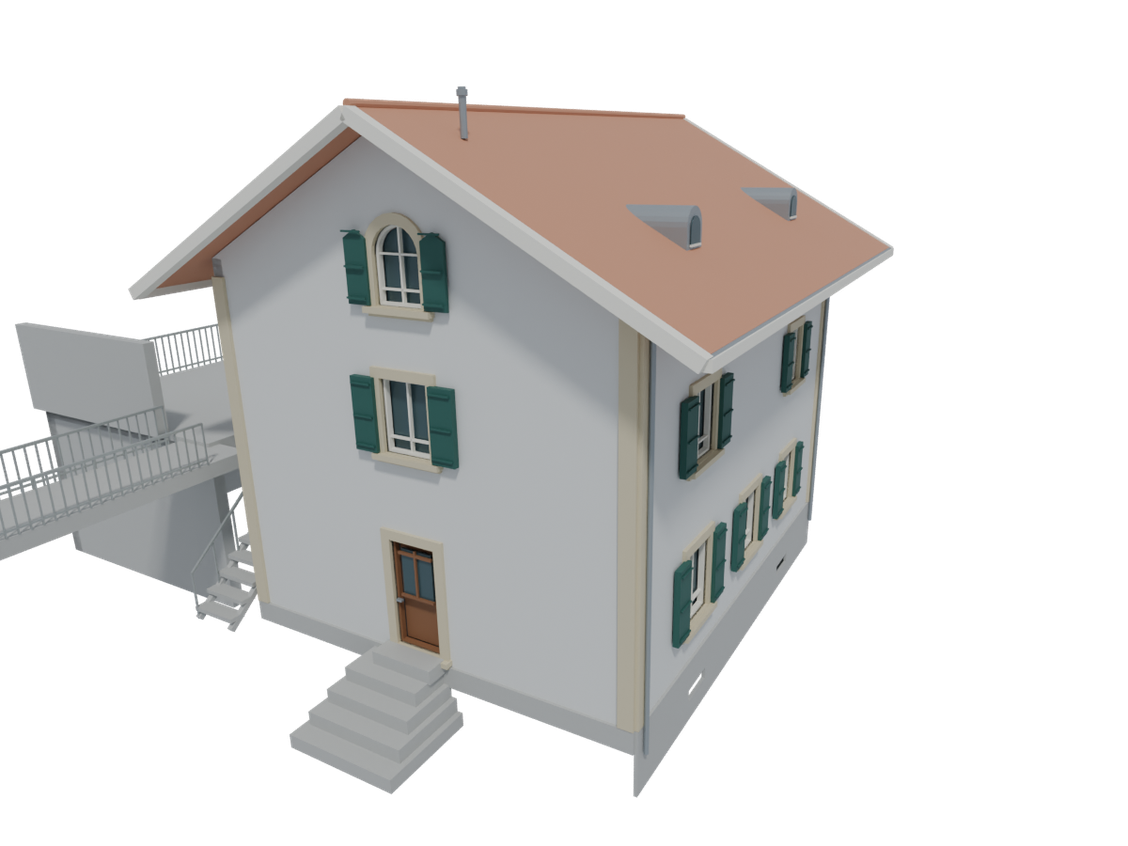}
        \caption*{05}
    \end{subfigure}\hfill%
    \begin{subfigure}[t]{0.16\textwidth}
        \centering
        \includegraphics[width=\linewidth]{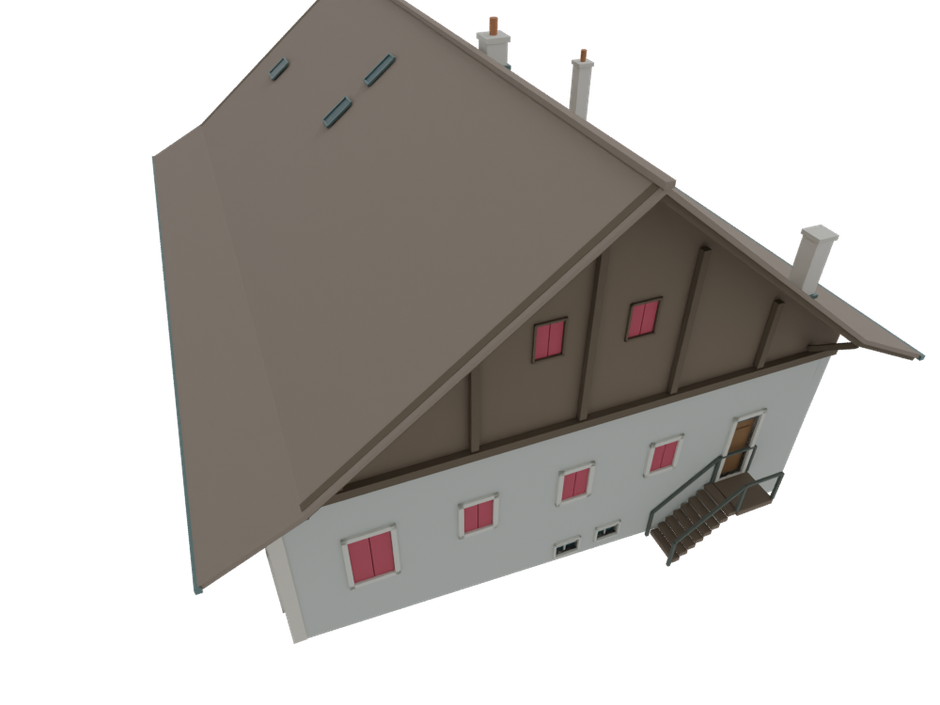}
        \caption*{06}
    \end{subfigure}

    \par\medskip

    \begin{subfigure}[t]{0.16\textwidth}
        \centering
        \includegraphics[width=\linewidth]{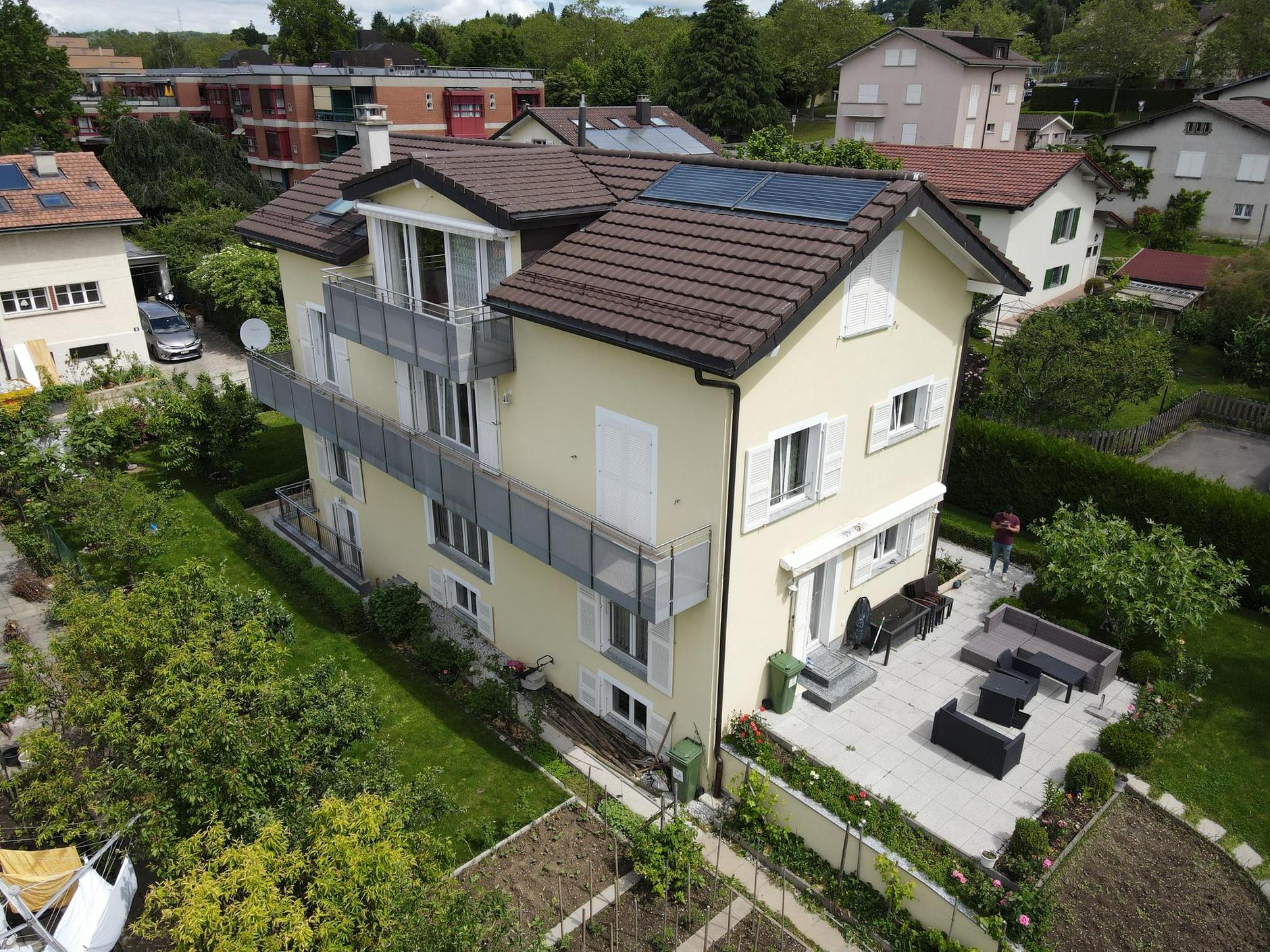}
    \end{subfigure}\hfill%
    \begin{subfigure}[t]{0.16\textwidth}
        \centering
        \includegraphics[width=\linewidth]{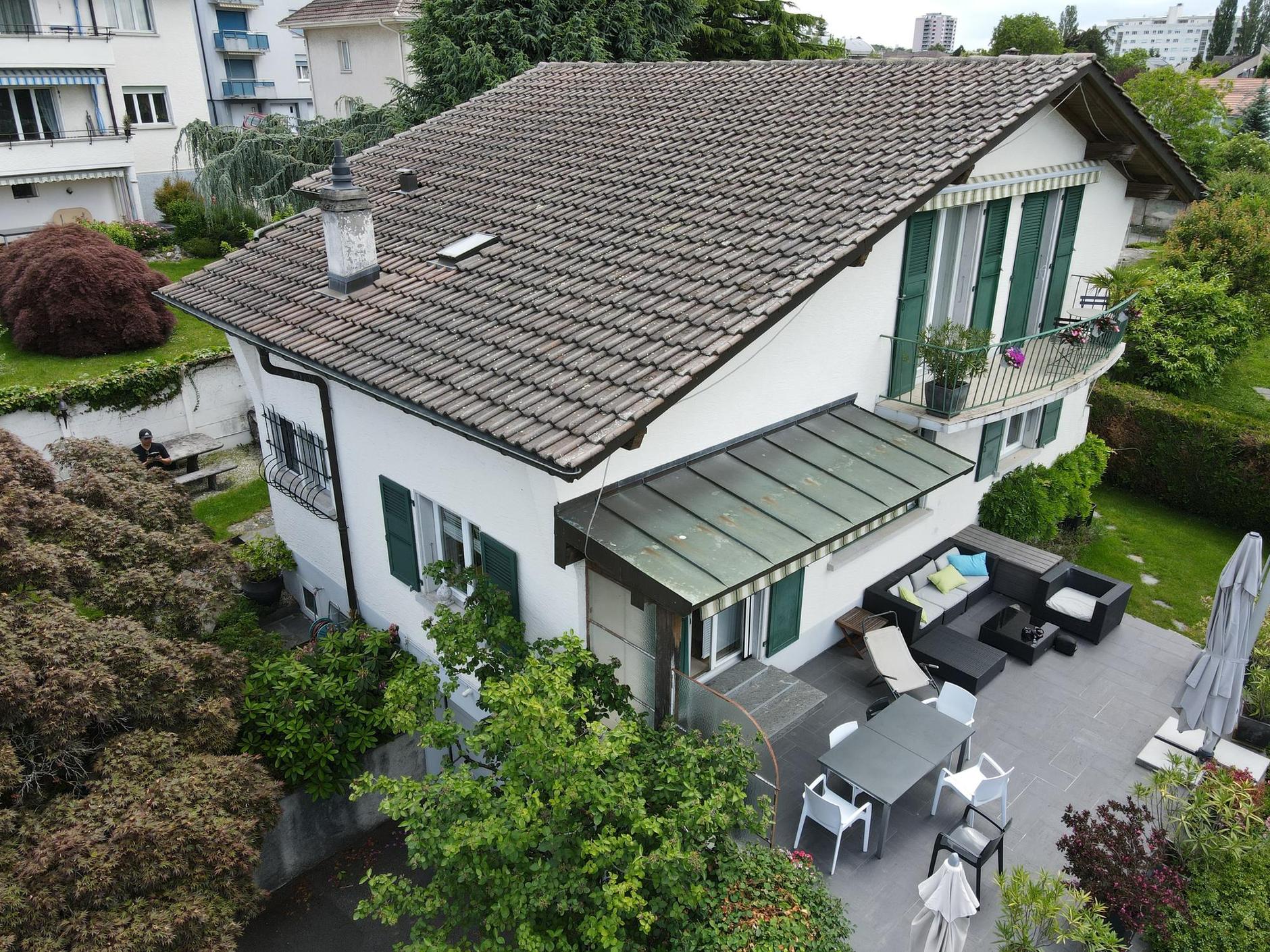}
    \end{subfigure}\hfill%
    \begin{subfigure}[t]{0.16\textwidth}
        \centering
        \includegraphics[width=\linewidth]{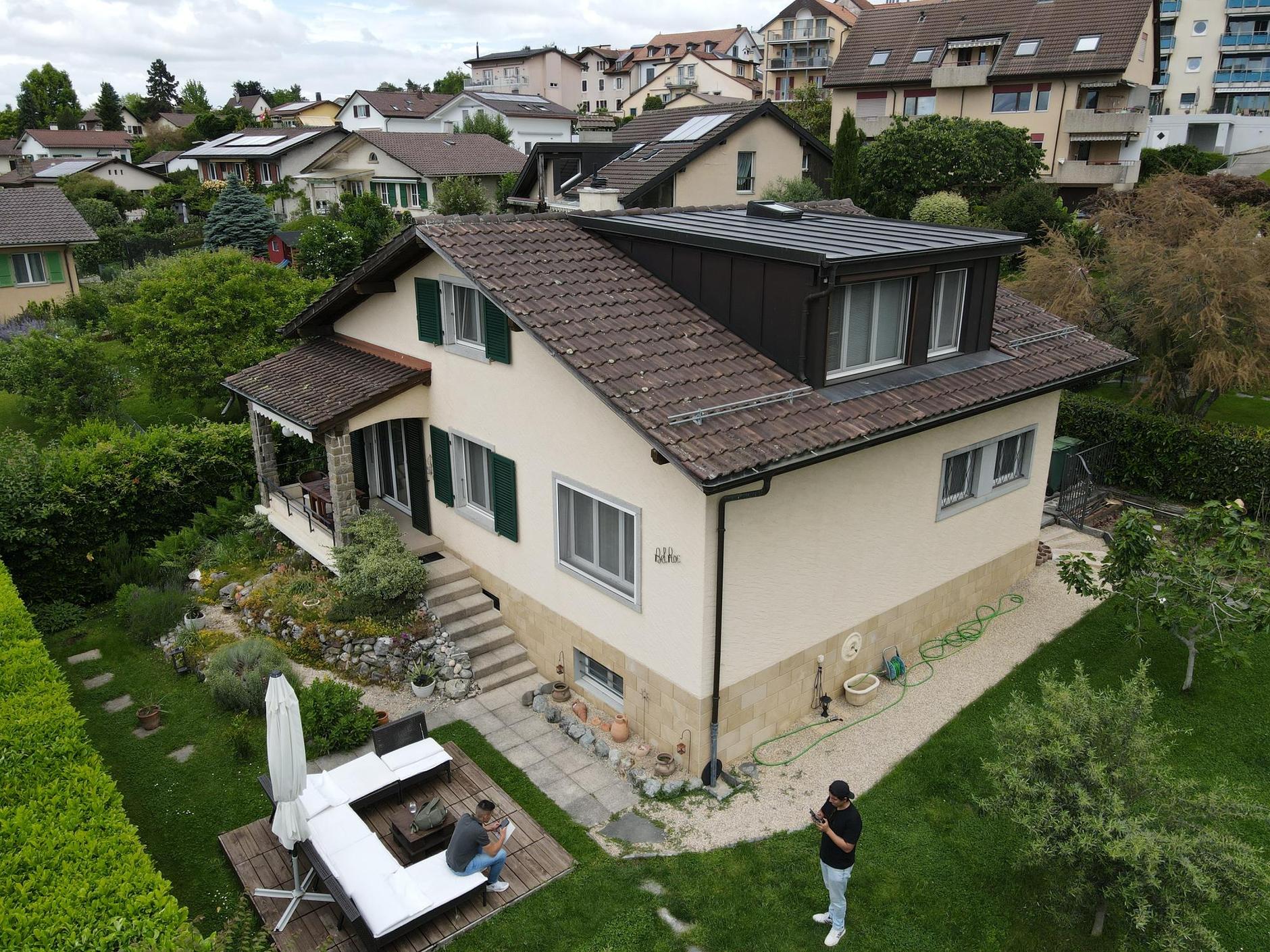}
    \end{subfigure}\hfill%
    \begin{subfigure}[t]{0.16\textwidth}
        \centering
        \includegraphics[width=\linewidth]{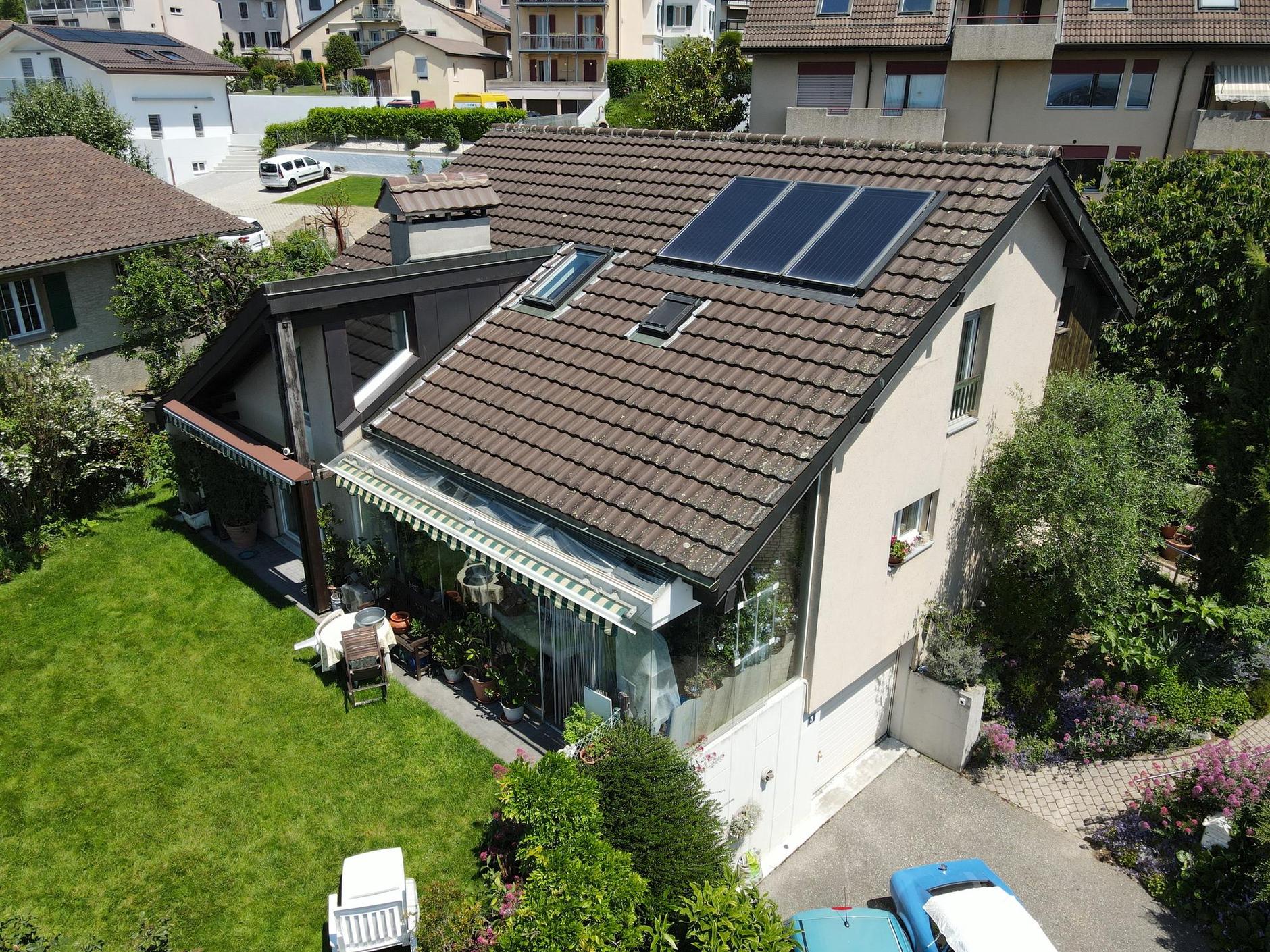}
    \end{subfigure}\hfill%
    \begin{subfigure}[t]{0.16\textwidth}
        \centering
        \includegraphics[width=\linewidth]{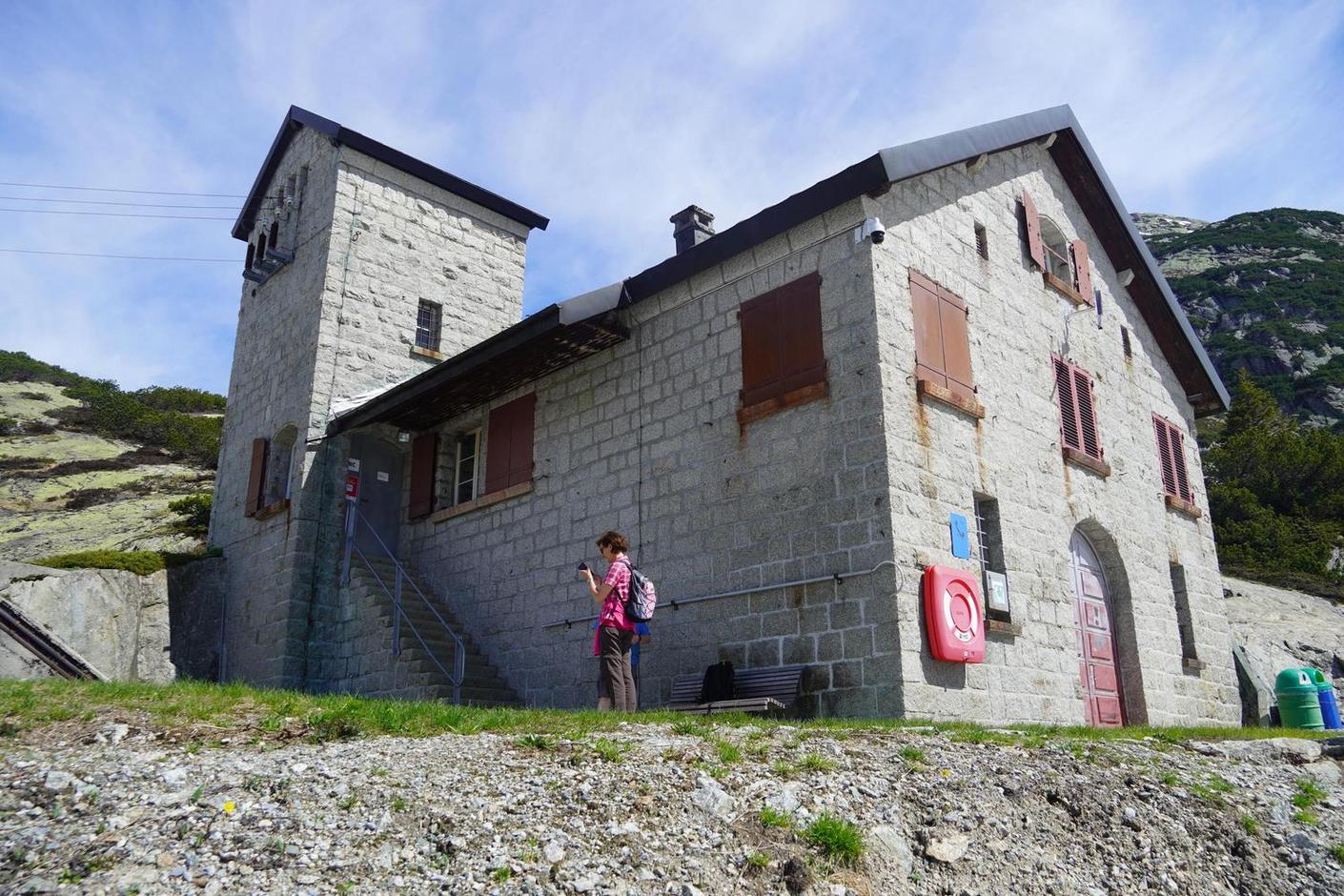}
    \end{subfigure}\hfill%
    \begin{subfigure}[t]{0.16\textwidth}
        \centering
        \includegraphics[width=\linewidth]{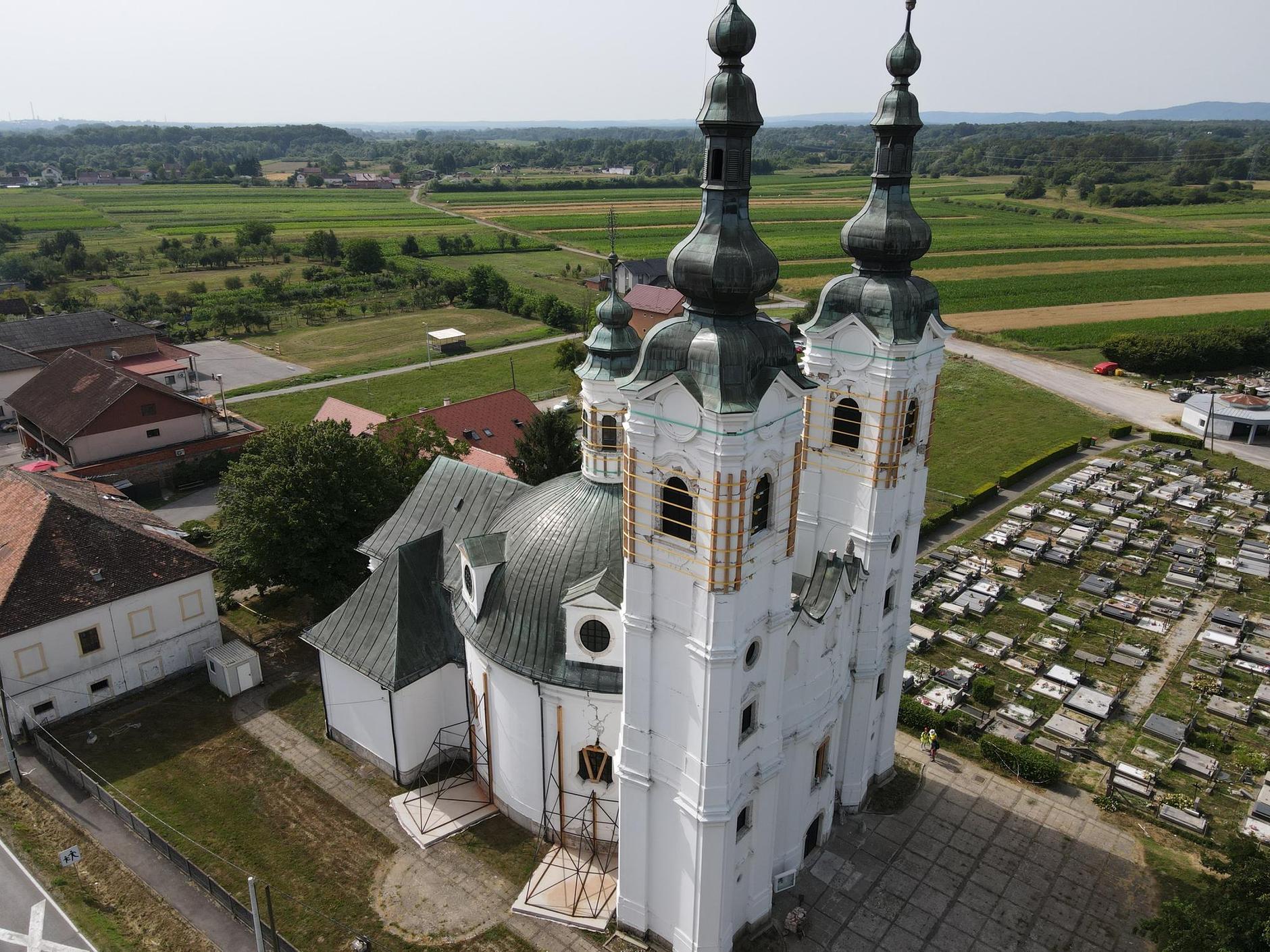}
    \end{subfigure}

    \par\smallskip

    \begin{subfigure}[t]{0.16\textwidth}
        \centering
        \includegraphics[width=\linewidth]{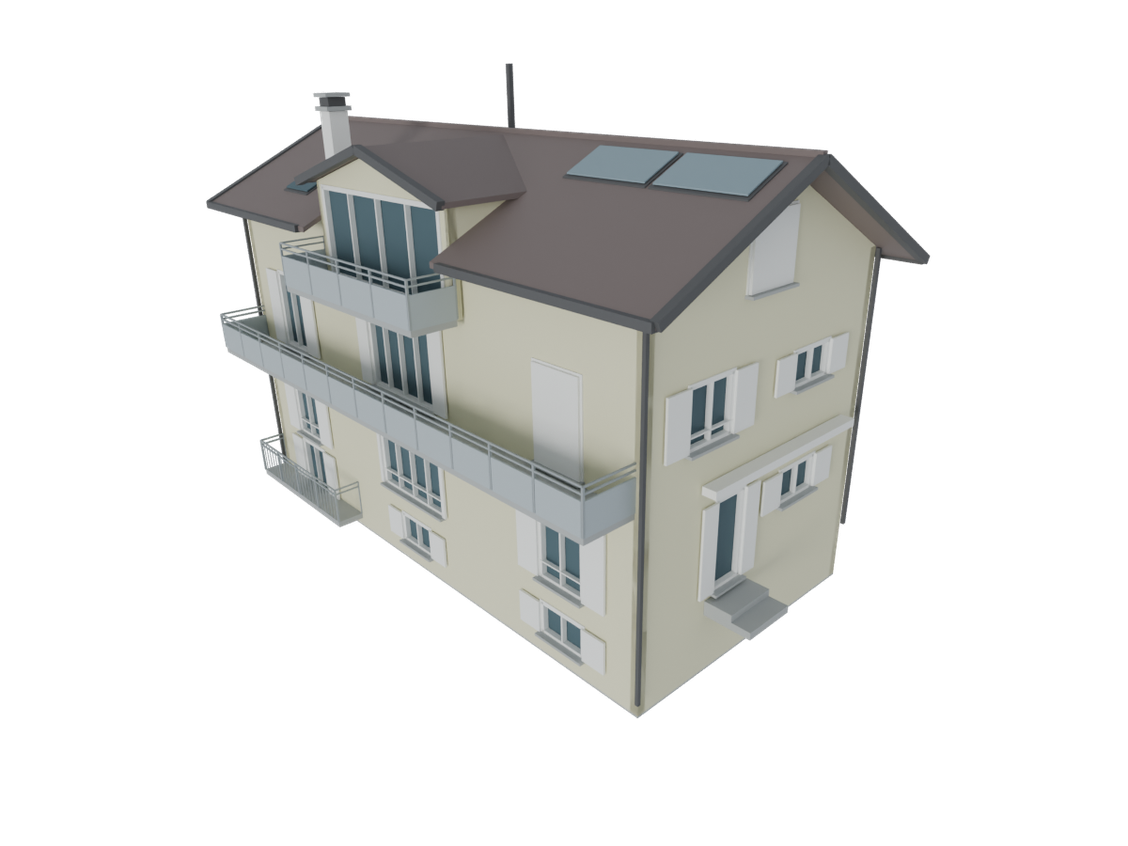}
        \caption*{07}
    \end{subfigure}\hfill%
    \begin{subfigure}[t]{0.16\textwidth}
        \centering
        \includegraphics[width=\linewidth]{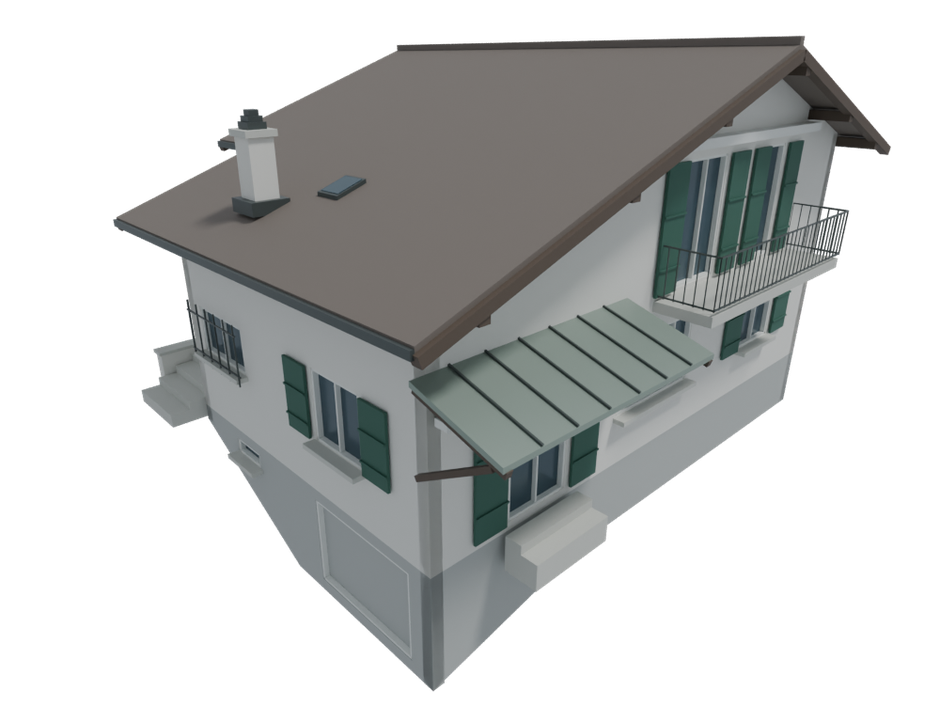}
        \caption*{08}
    \end{subfigure}\hfill%
    \begin{subfigure}[t]{0.16\textwidth}
        \centering
        \includegraphics[width=\linewidth]{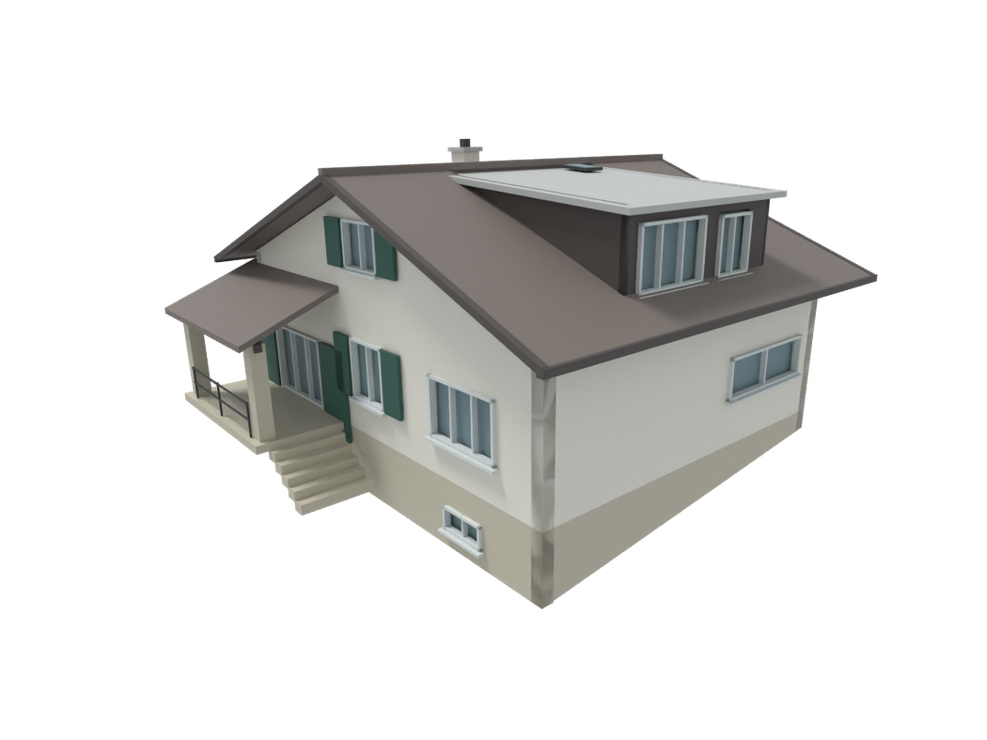}
        \caption*{09}
    \end{subfigure}\hfill%
    \begin{subfigure}[t]{0.16\textwidth}
        \centering
        \includegraphics[width=\linewidth]{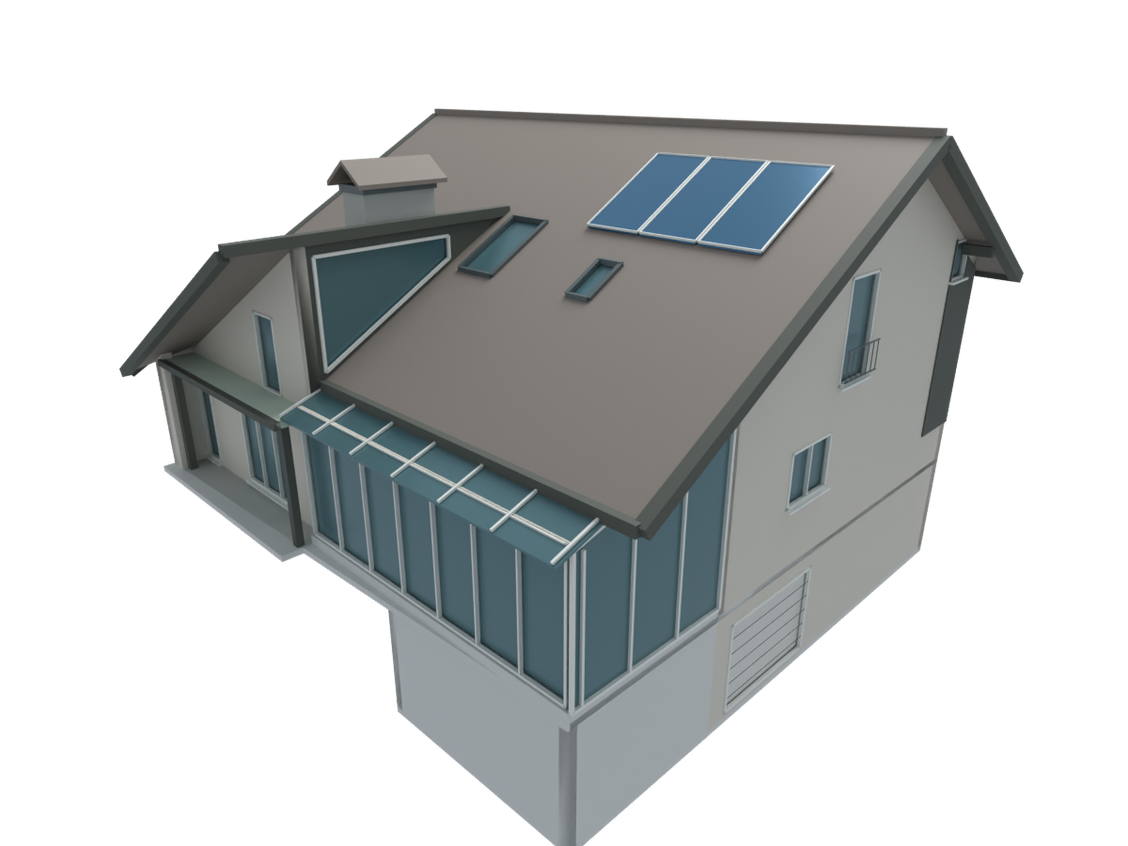}
        \caption*{10}
    \end{subfigure}\hfill%
    \begin{subfigure}[t]{0.16\textwidth}
        \centering
        \includegraphics[width=\linewidth]{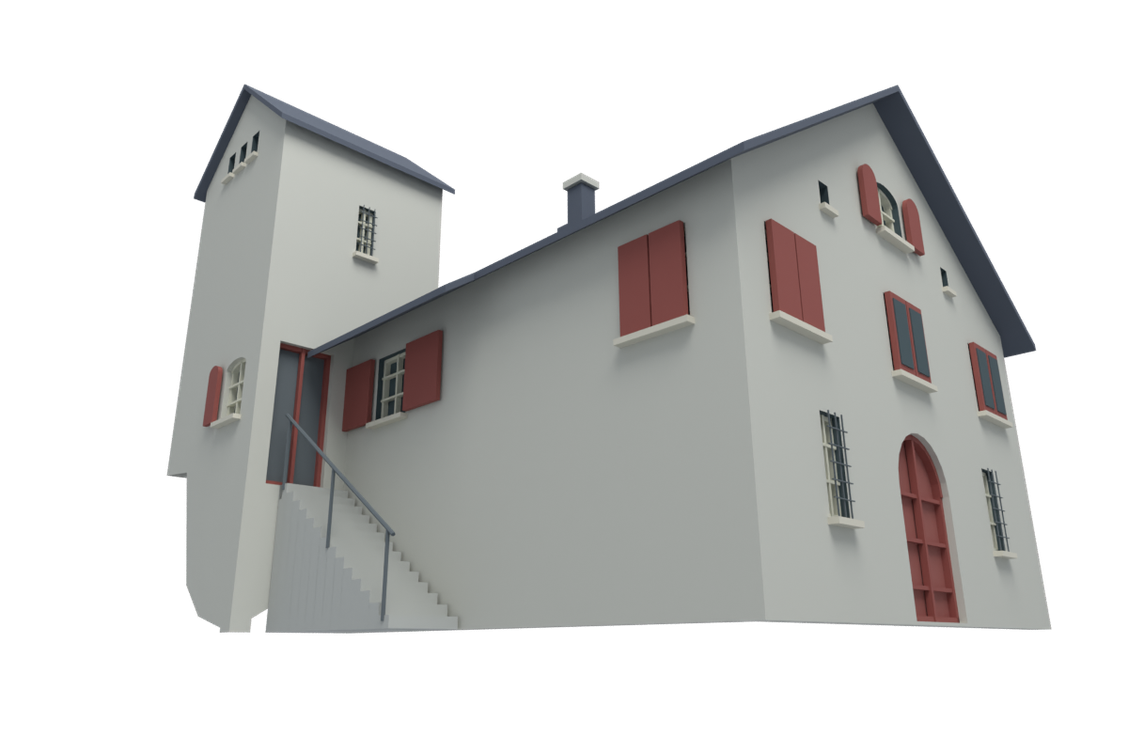}
        \caption*{11}
    \end{subfigure}\hfill%
    \begin{subfigure}[t]{0.16\textwidth}
        \centering
        \includegraphics[width=\linewidth]{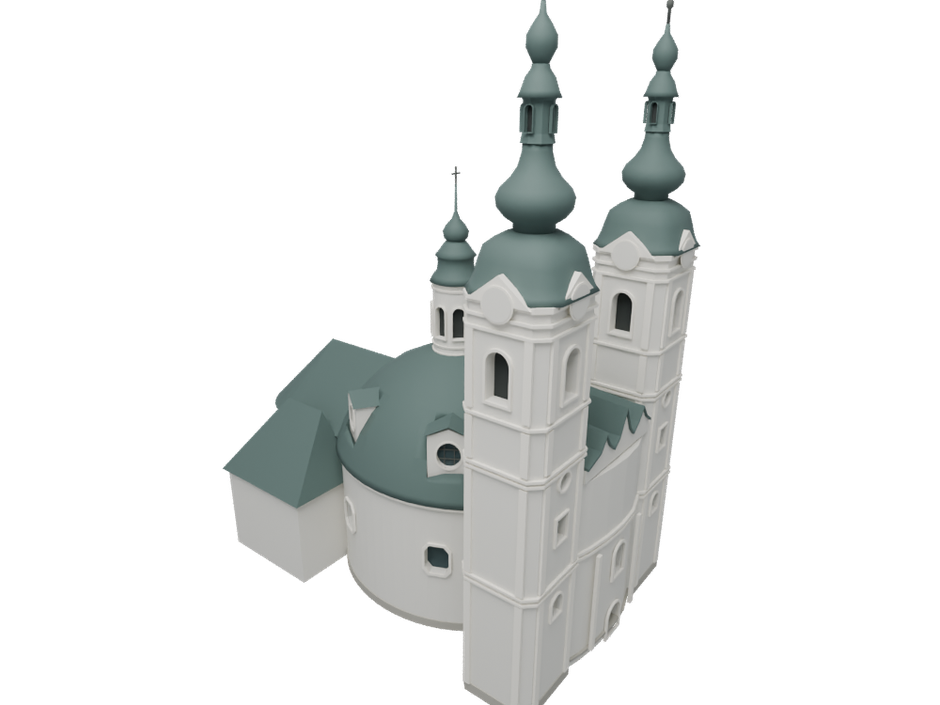}
        \caption*{12}
    \end{subfigure}

    \par\medskip

    \begin{subfigure}[t]{0.16\textwidth}
        \centering
        \includegraphics[width=\linewidth]{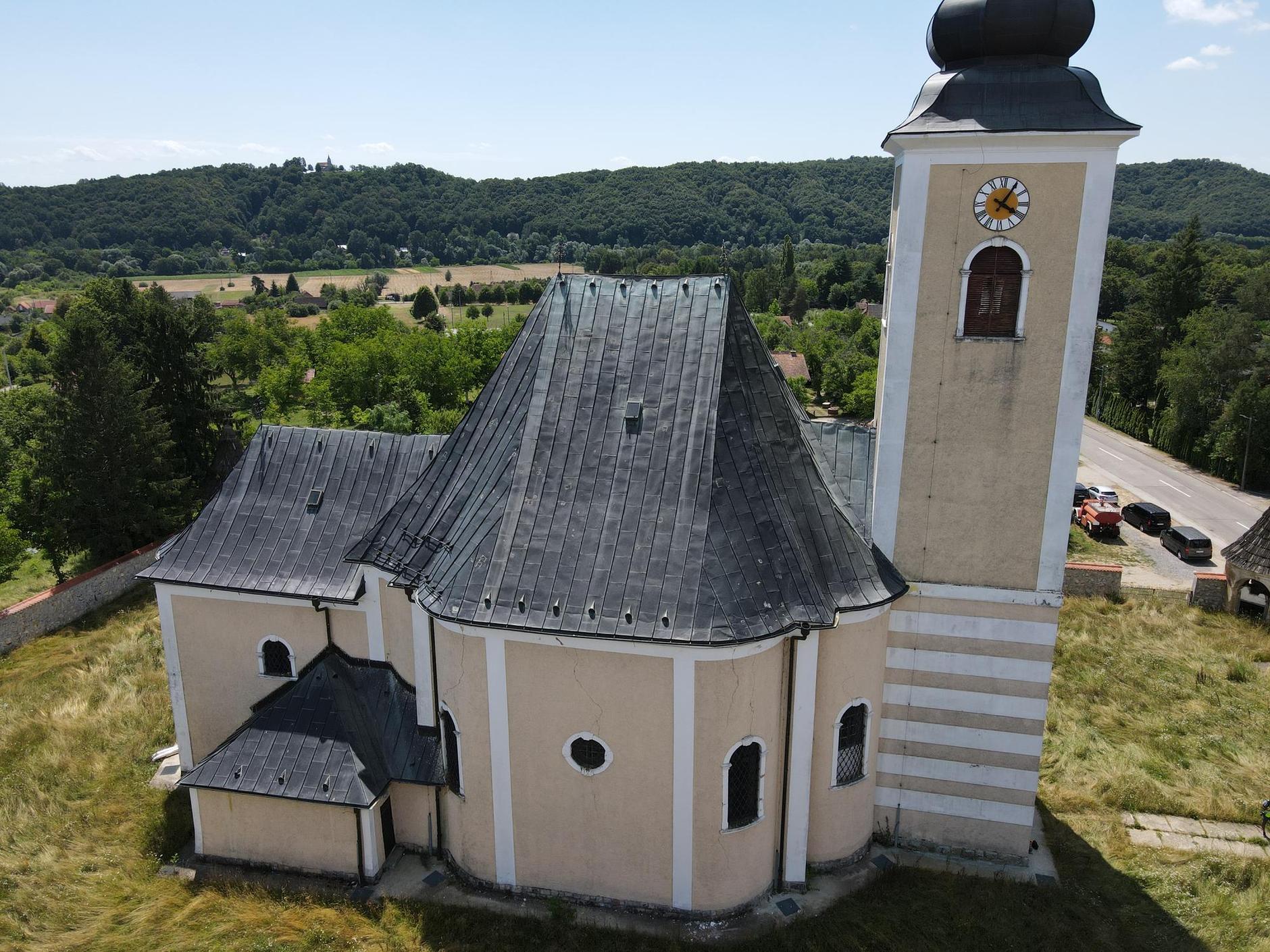}
    \end{subfigure}\hfill%
    \begin{subfigure}[t]{0.16\textwidth}
        \centering
        \includegraphics[width=\linewidth]{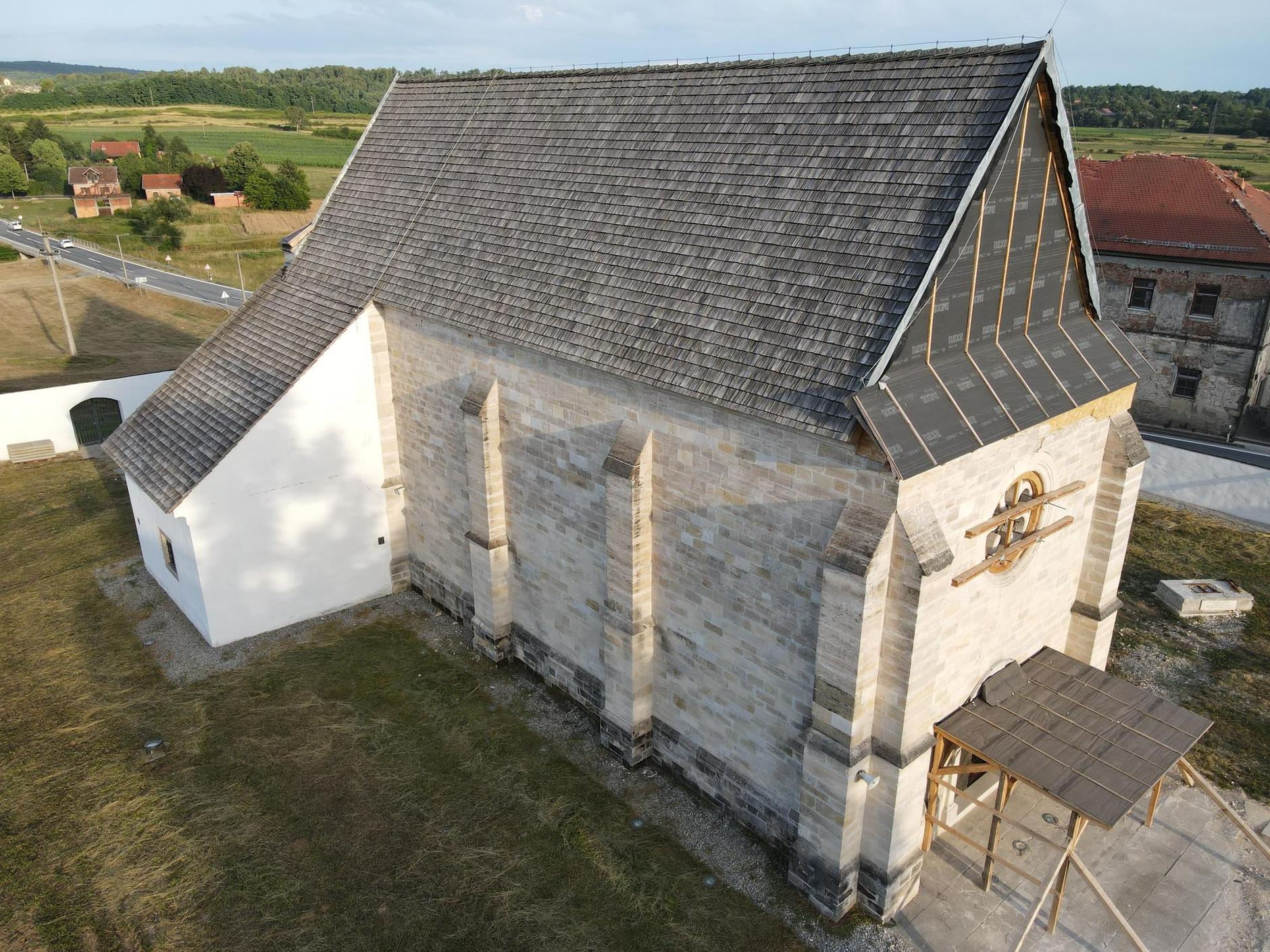}
    \end{subfigure}\hfill%
    \begin{subfigure}[t]{0.16\textwidth}
        \centering
        \includegraphics[width=\linewidth]{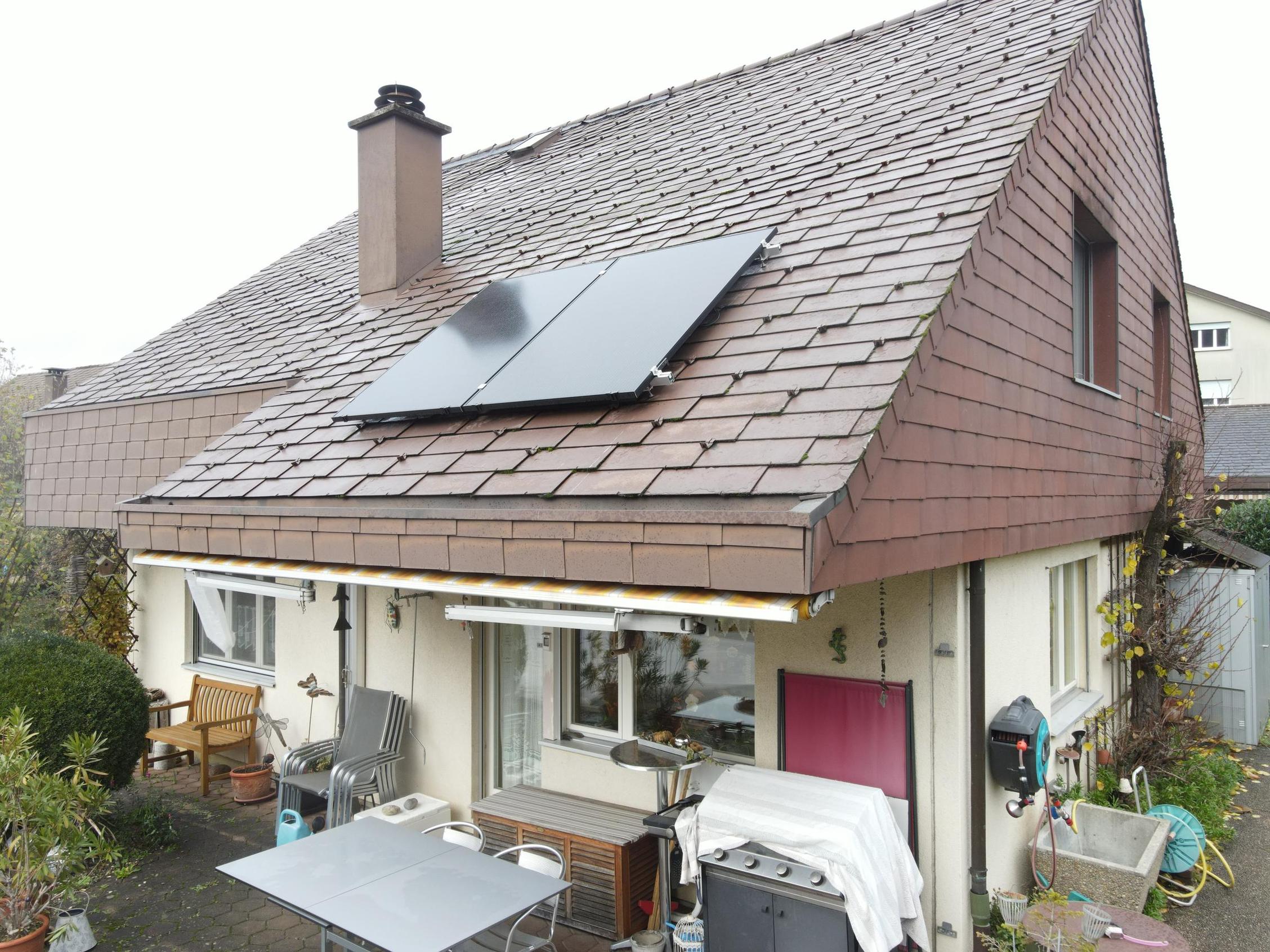}
    \end{subfigure}\hfill%
    \begin{subfigure}[t]{0.16\textwidth}
        \centering
        \includegraphics[width=\linewidth]{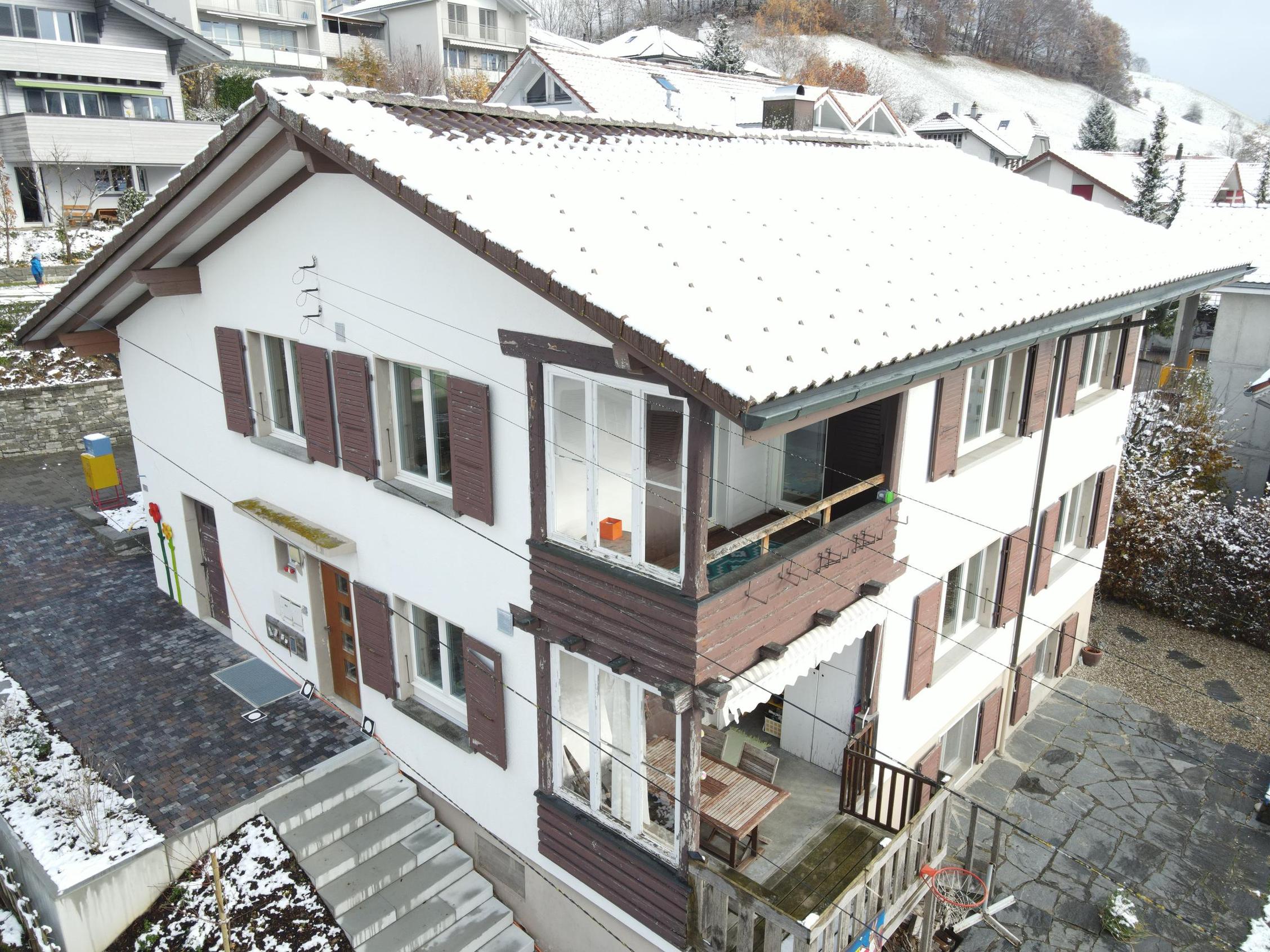}
    \end{subfigure}\hfill%
    \begin{subfigure}[t]{0.16\textwidth}
        \centering
        \includegraphics[width=\linewidth]{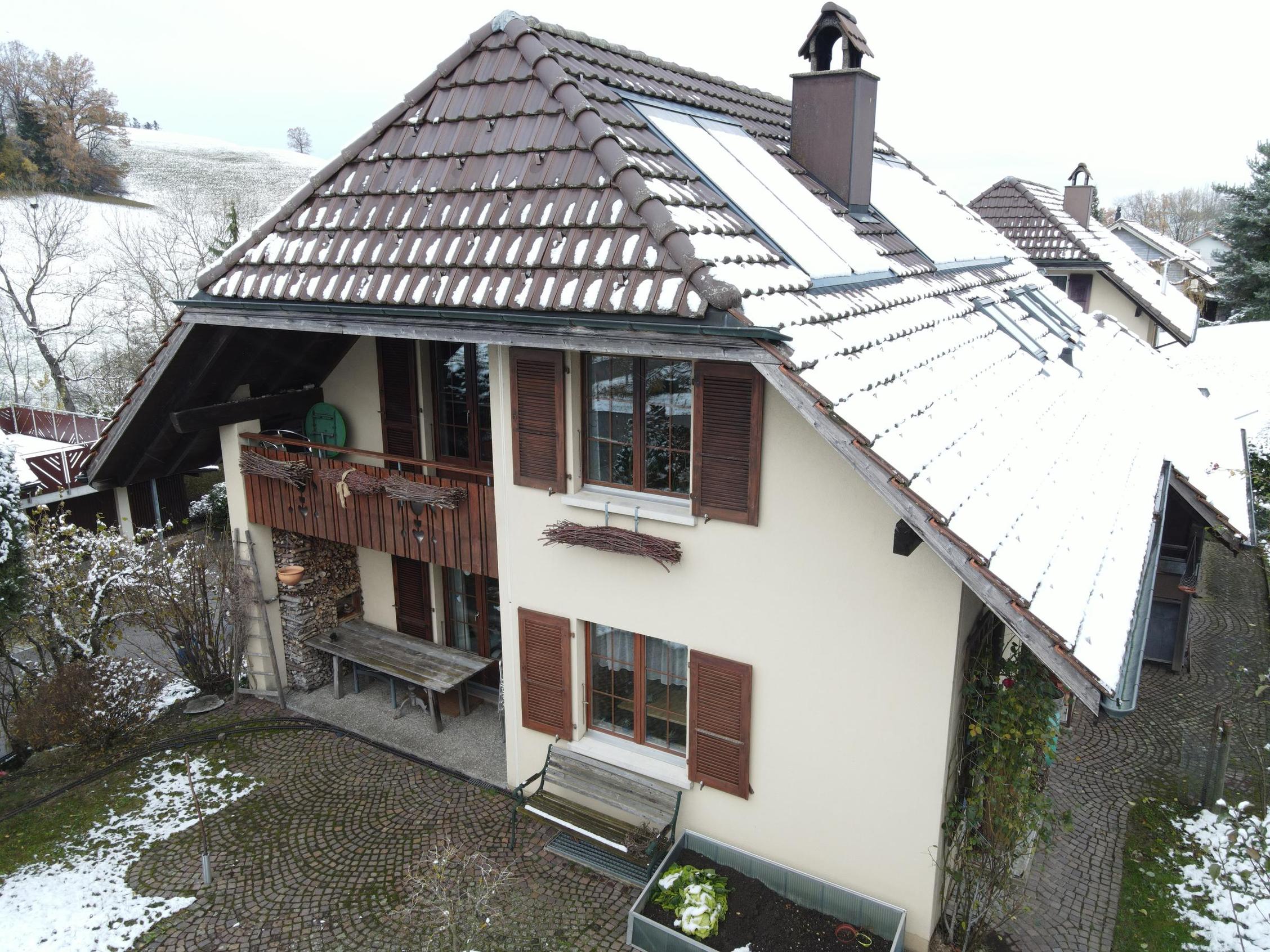}
    \end{subfigure}\hfill%
    \begin{subfigure}[t]{0.16\textwidth}
        \centering
        \includegraphics[width=\linewidth]{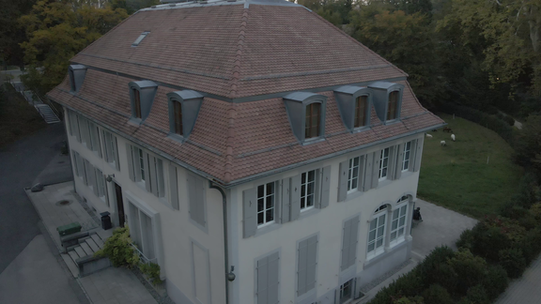}
    \end{subfigure}

    \par\smallskip

    \begin{subfigure}[t]{0.16\textwidth}
        \centering
        \includegraphics[width=\linewidth]{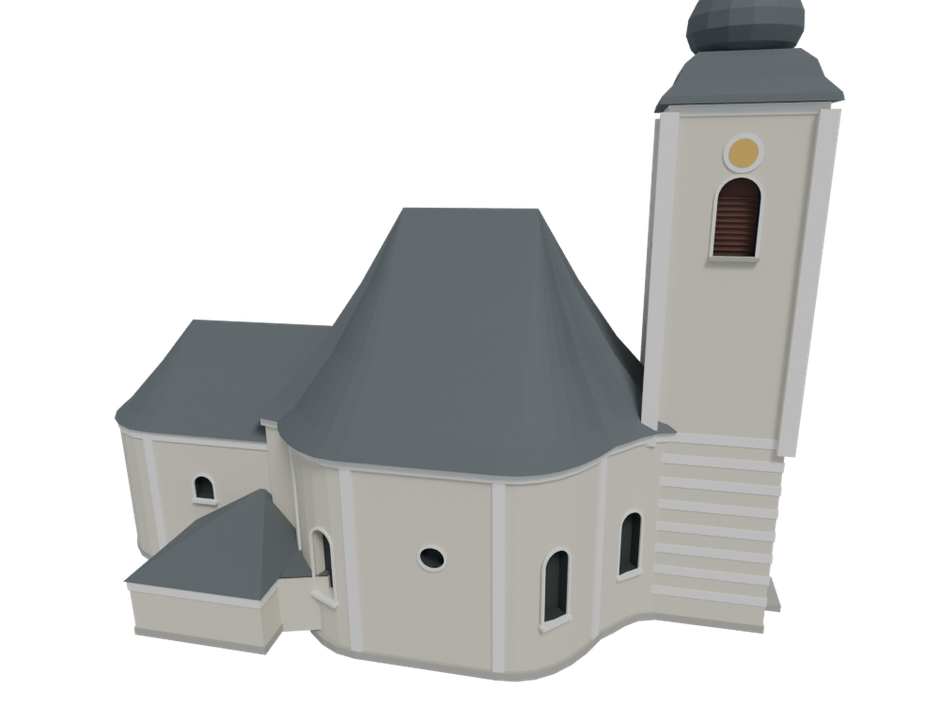}
        \caption*{13}
    \end{subfigure}\hfill%
    \begin{subfigure}[t]{0.16\textwidth}
        \centering
        \includegraphics[width=\linewidth]{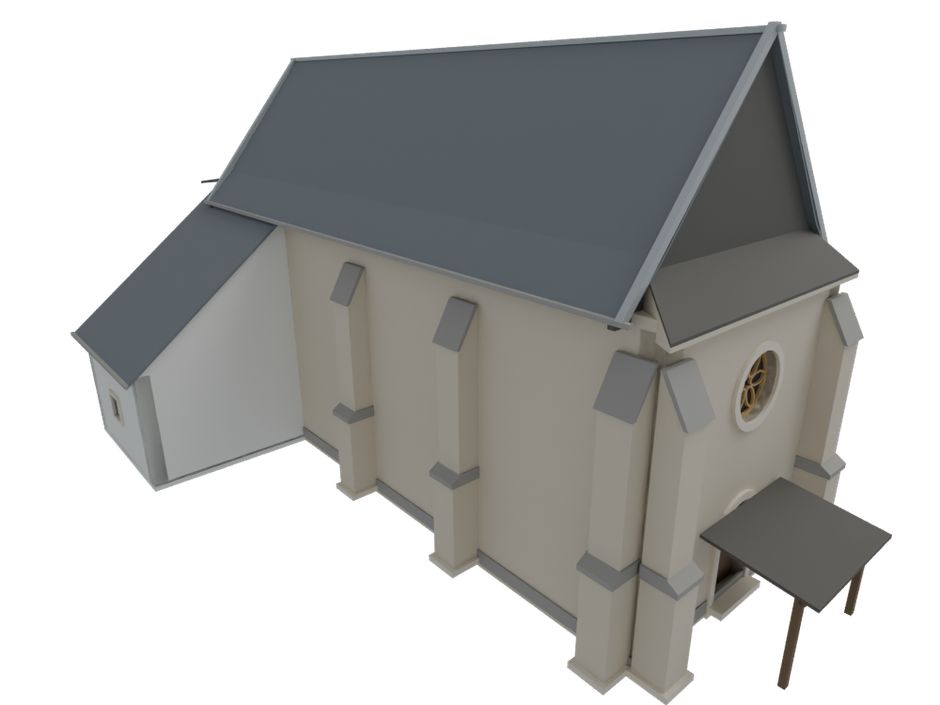}
        \caption*{14}
    \end{subfigure}\hfill%
    \begin{subfigure}[t]{0.16\textwidth}
        \centering
        \includegraphics[width=\linewidth]{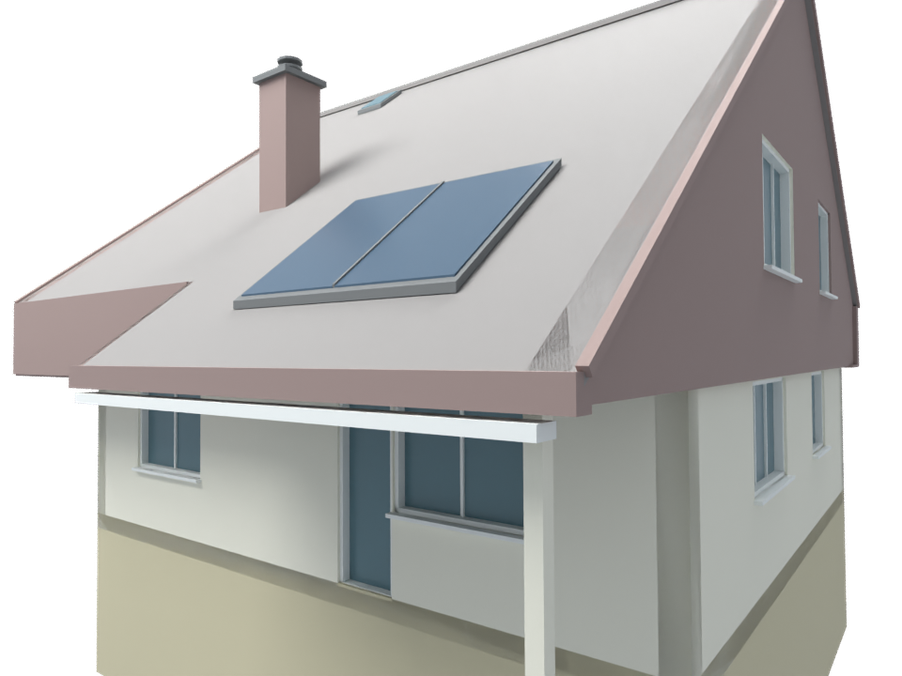}
        \caption*{15}
    \end{subfigure}\hfill%
    \begin{subfigure}[t]{0.16\textwidth}
        \centering
        \includegraphics[width=\linewidth]{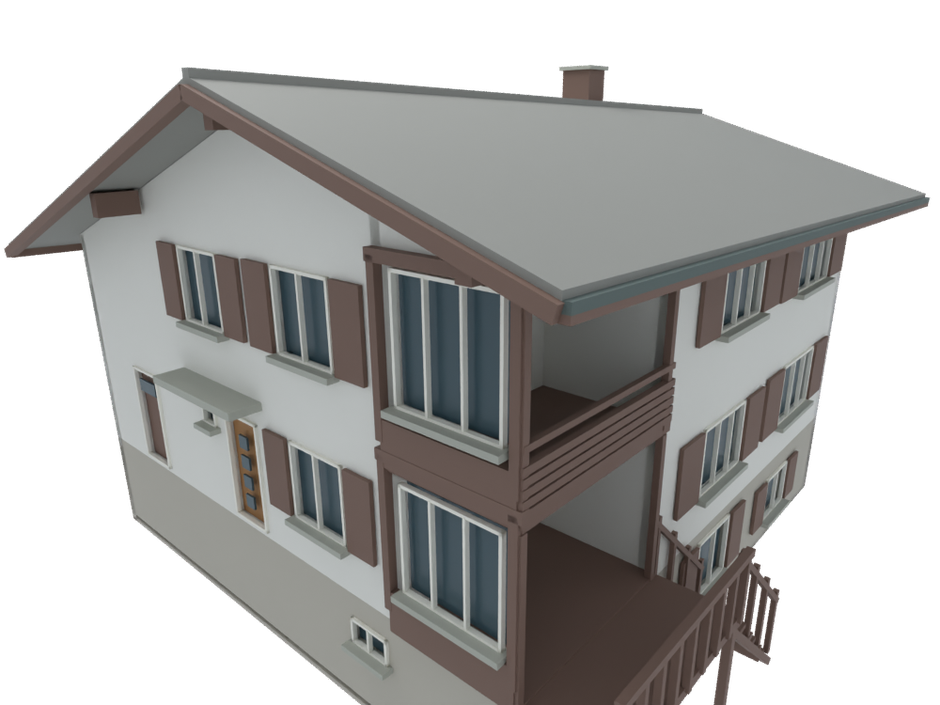}
        \caption*{16}
    \end{subfigure}\hfill%
    \begin{subfigure}[t]{0.16\textwidth}
        \centering
        \includegraphics[width=\linewidth]{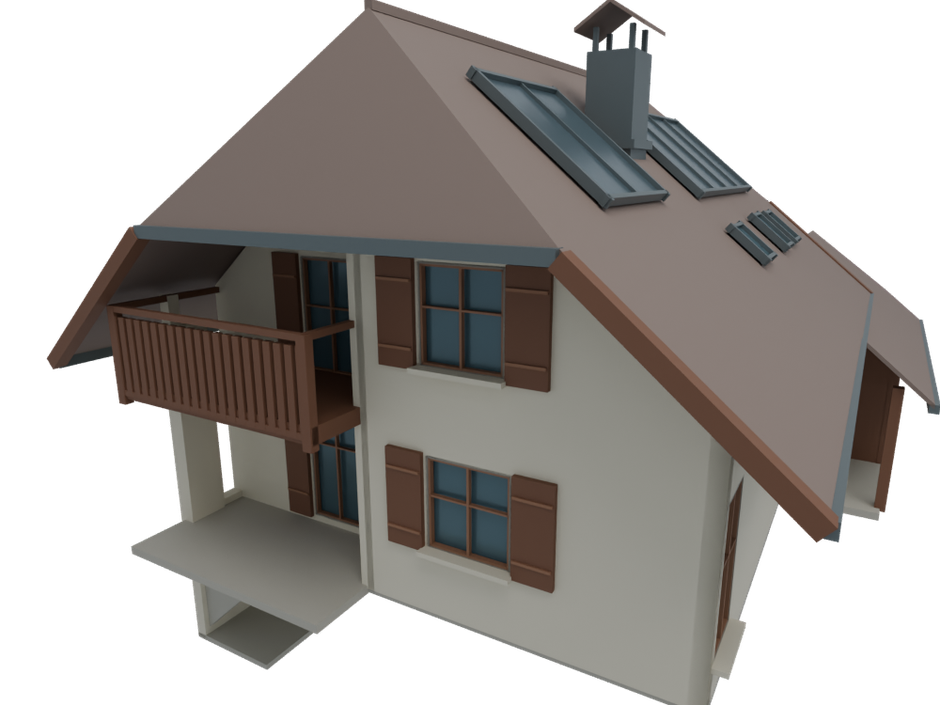}
        \caption*{17}
    \end{subfigure}\hfill%
    \begin{subfigure}[t]{0.16\textwidth}
        \centering
        \includegraphics[width=\linewidth]{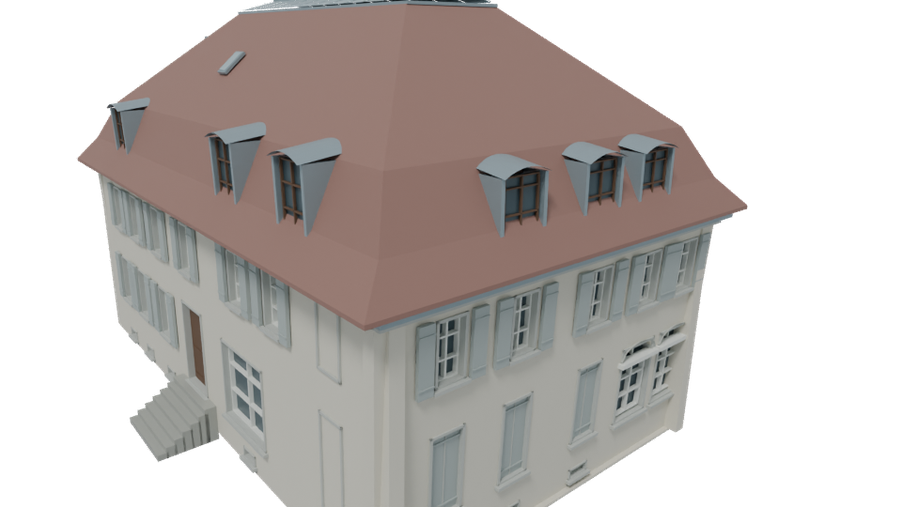}
        \caption*{18}
    \end{subfigure}

    \par\medskip

    \begin{subfigure}[t]{0.16\textwidth}
        \centering
        \includegraphics[width=\linewidth]{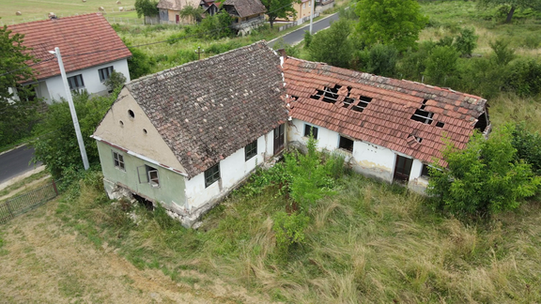}
    \end{subfigure}\hfill%
    \begin{subfigure}[t]{0.16\textwidth}
        \centering
        \includegraphics[width=\linewidth]{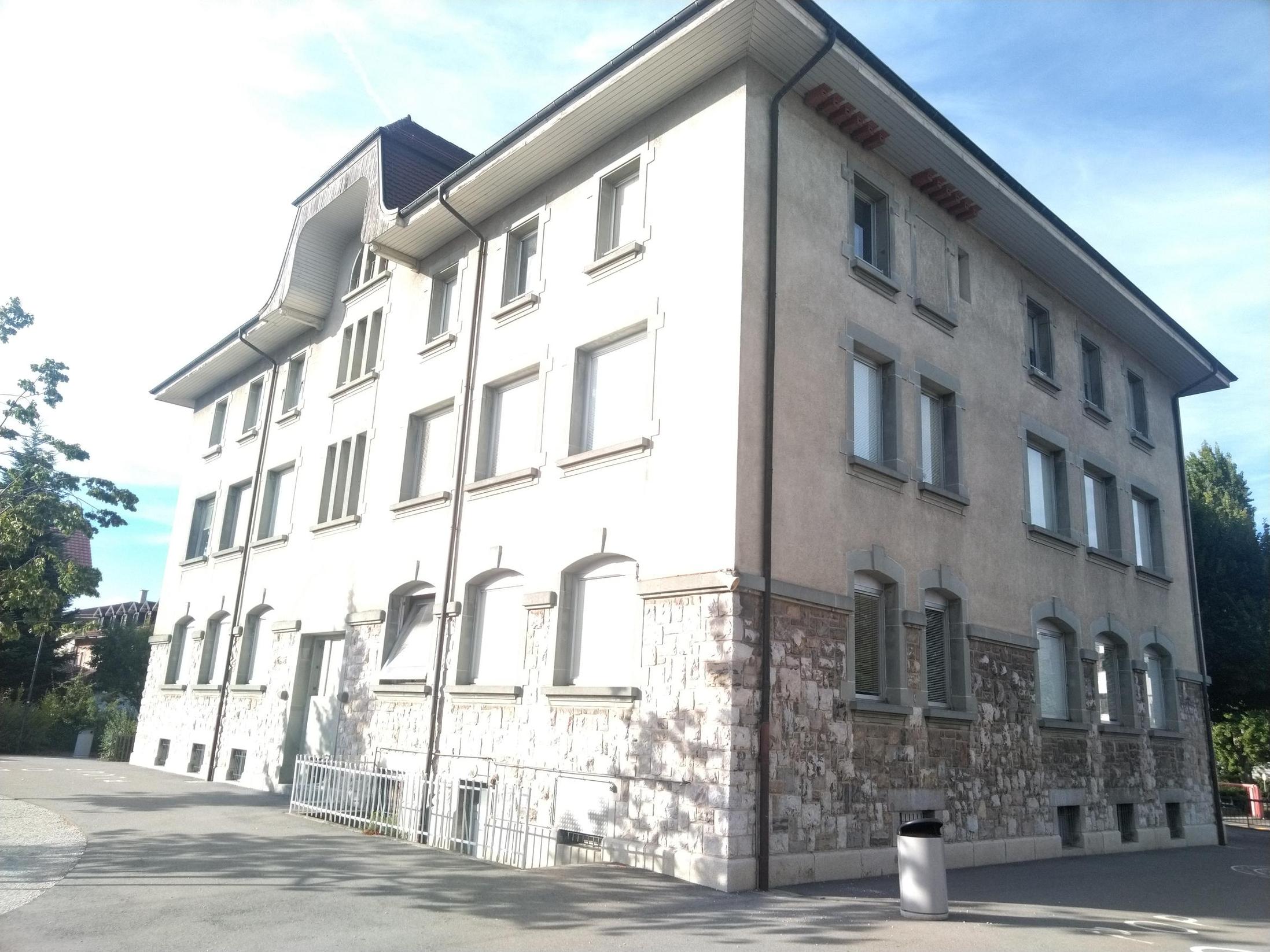}
    \end{subfigure}\hfill%
    \begin{subfigure}[t]{0.16\textwidth}
        \centering
        \includegraphics[width=\linewidth]{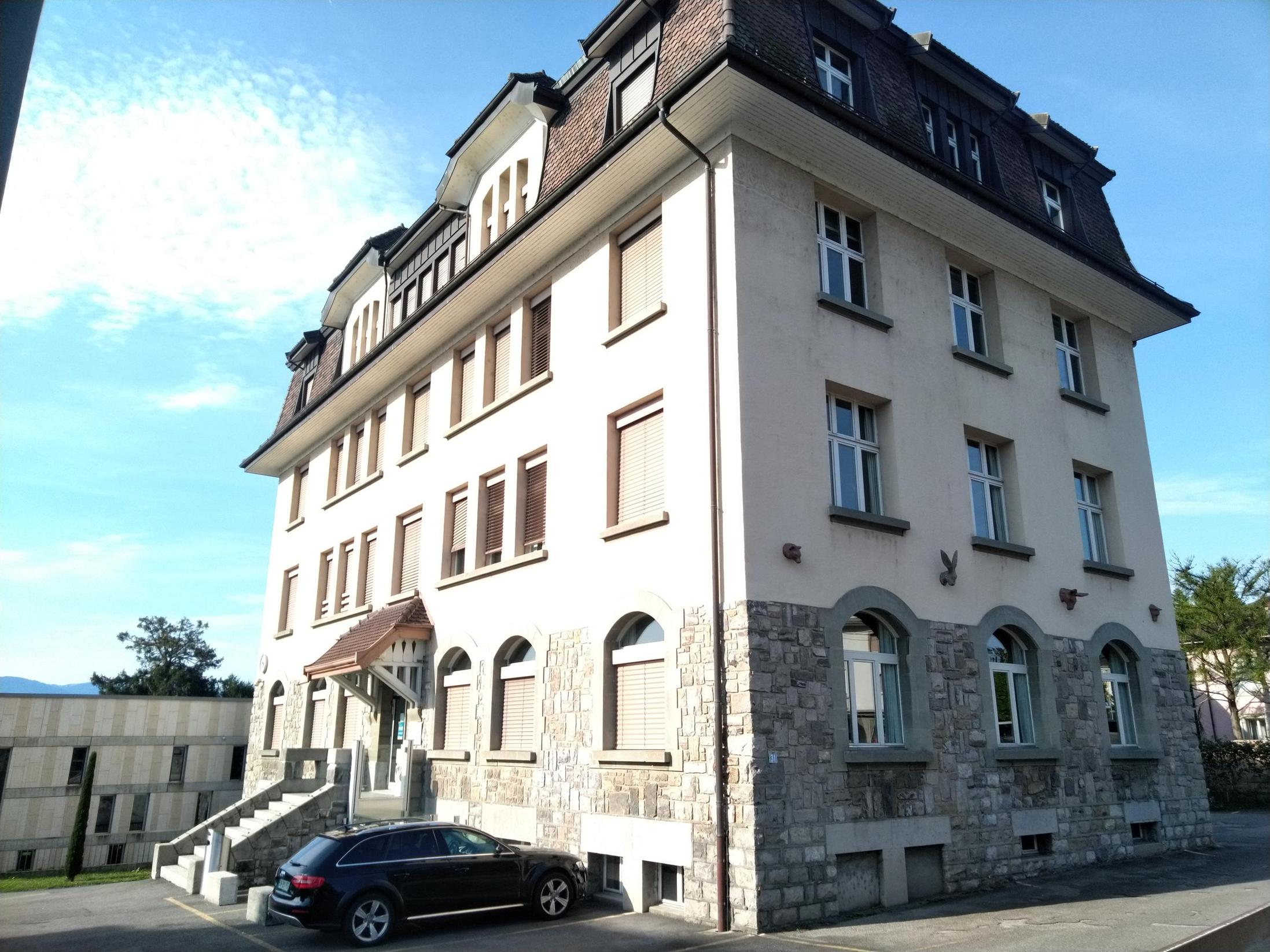}
    \end{subfigure}\hfill%
    \begin{subfigure}[t]{0.16\textwidth}
        \centering
        \includegraphics[width=\linewidth]{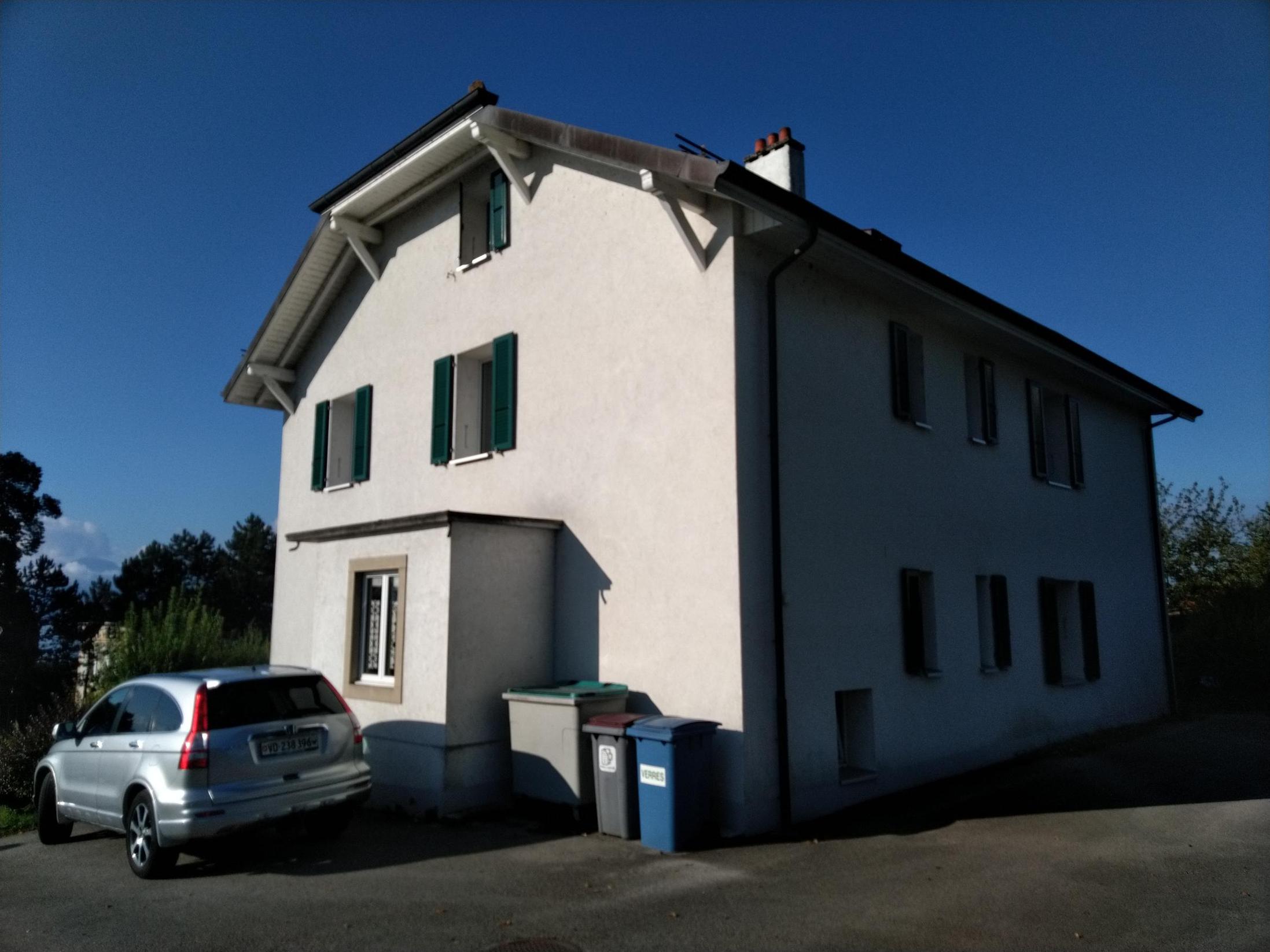}
    \end{subfigure}\hfill%
    \begin{subfigure}[t]{0.16\textwidth}
        \centering
        \includegraphics[width=\linewidth]{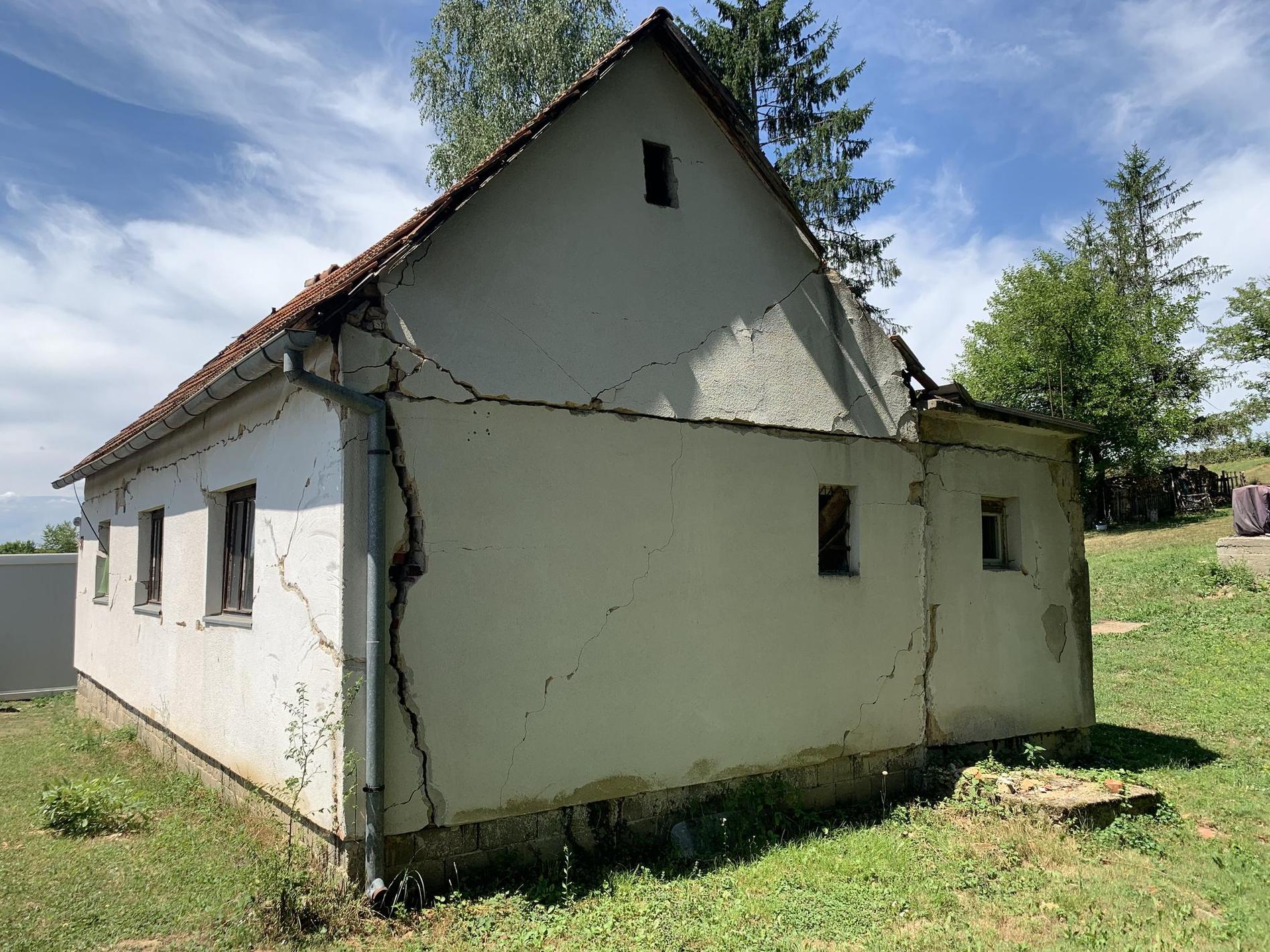}
    \end{subfigure}\hfill%
    \begin{subfigure}[t]{0.16\textwidth}
        \centering
        \includegraphics[width=\linewidth]{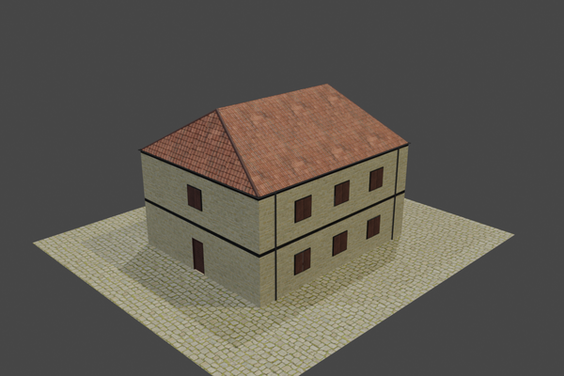}
    \end{subfigure}

    \par\smallskip

    \begin{subfigure}[t]{0.16\textwidth}
        \centering
        \includegraphics[width=\linewidth]{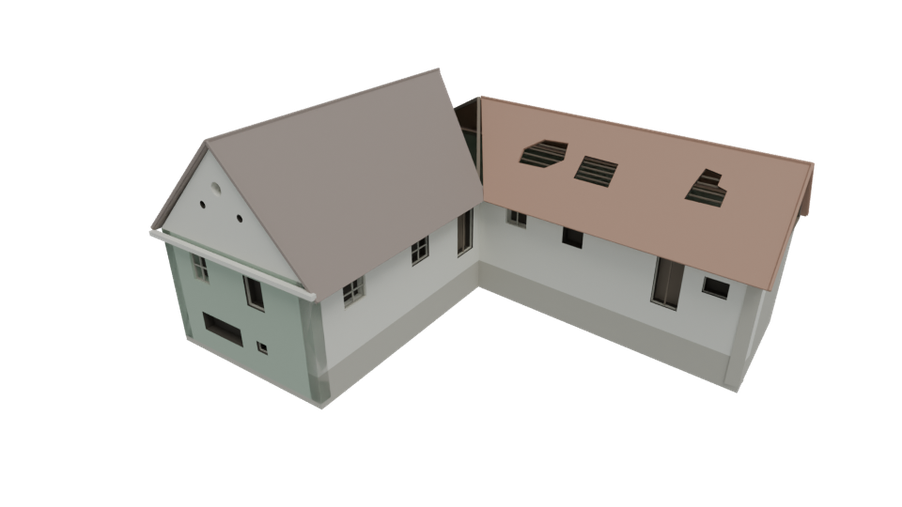}
        \caption*{19}
    \end{subfigure}\hfill%
    \begin{subfigure}[t]{0.16\textwidth}
        \centering
        \includegraphics[width=\linewidth]{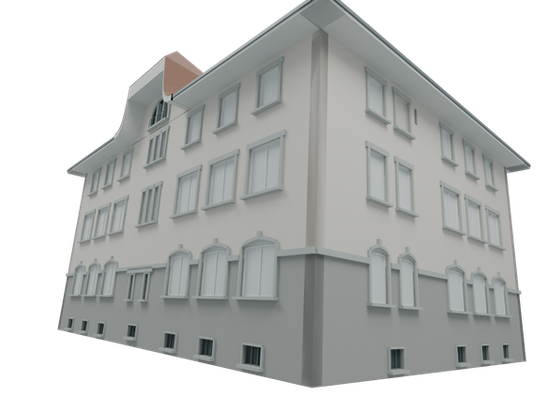}
        \caption*{20}
    \end{subfigure}\hfill%
    \begin{subfigure}[t]{0.16\textwidth}
        \centering
        \includegraphics[width=\linewidth]{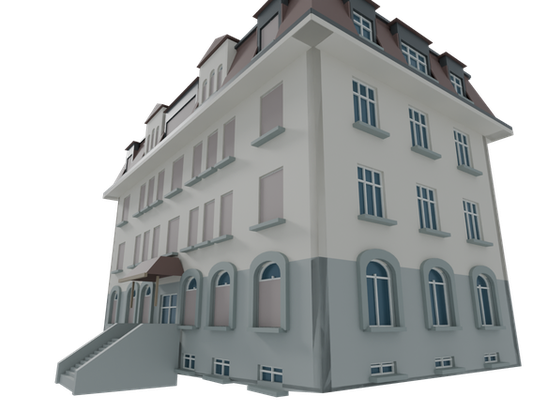}
        \caption*{21}
    \end{subfigure}\hfill%
    \begin{subfigure}[t]{0.16\textwidth}
        \centering
        \includegraphics[width=\linewidth]{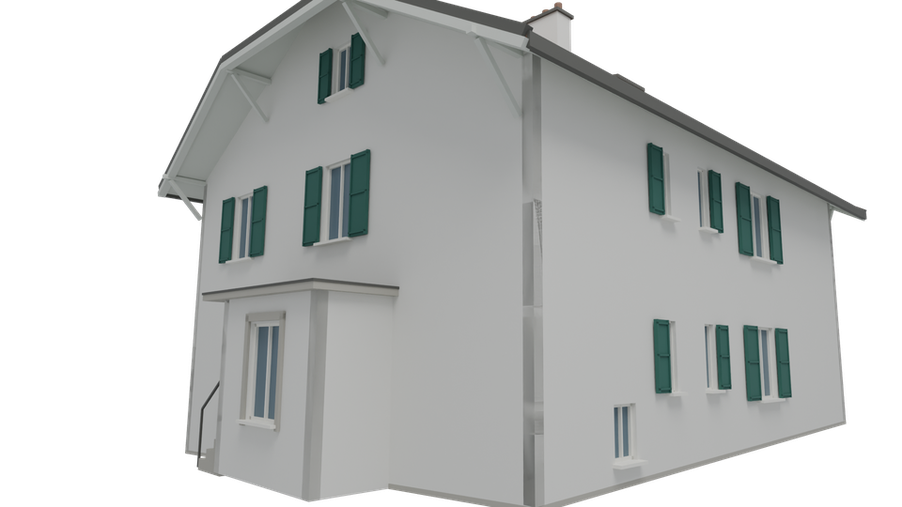}
        \caption*{22}
    \end{subfigure}\hfill%
    \begin{subfigure}[t]{0.16\textwidth}
        \centering
        \includegraphics[width=\linewidth]{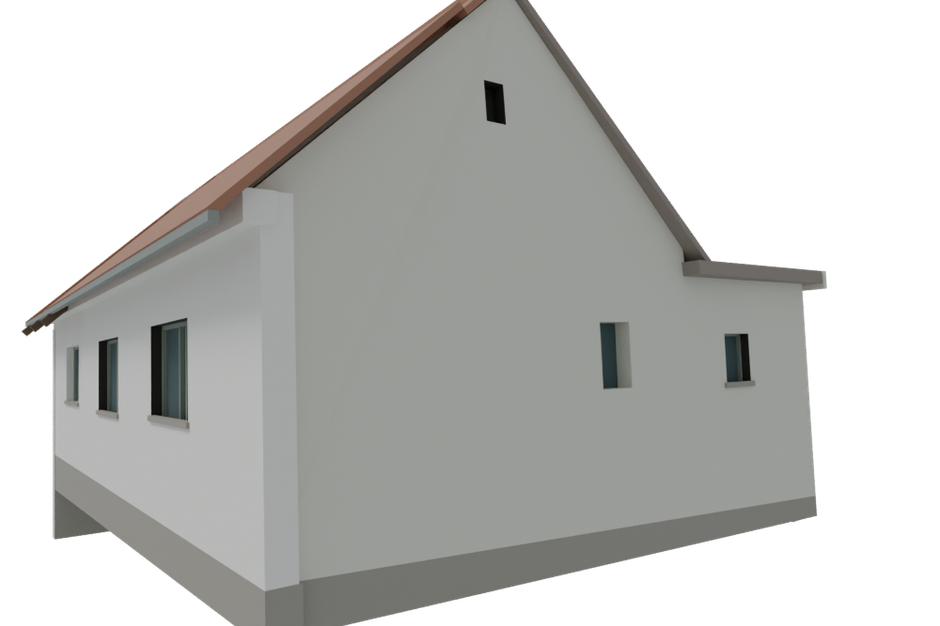}
        \caption*{23}
    \end{subfigure}\hfill%
    \begin{subfigure}[t]{0.16\textwidth}
        \centering
        \includegraphics[width=\linewidth]{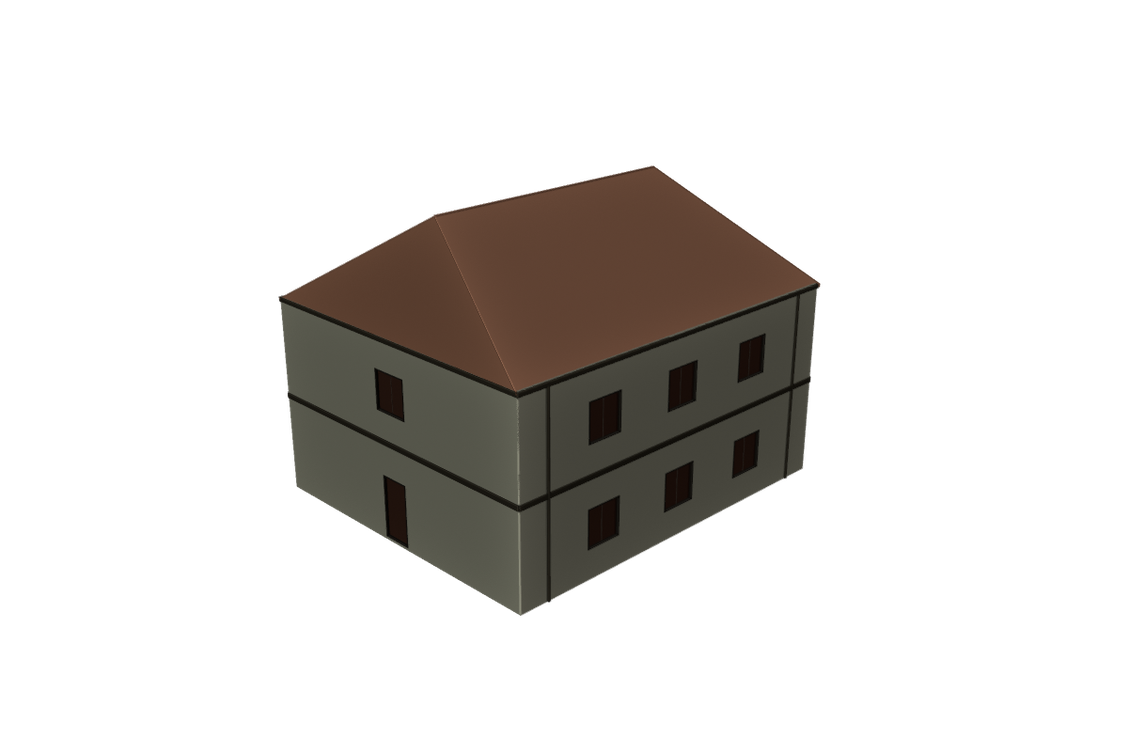}
        \caption*{24}
    \end{subfigure}

    \caption{Reference AstraLOD3 reconstructions for the 24 benchmark buildings. Each case shows a representative source image and the reconstruction rendered from the corresponding calibrated camera. All models were generated using the complete reconstruction evidence $\mathcal{I}_{\mathrm{VCP}}$, the \emph{Normal} reconstruction specification, and the Astra \emph{High} model configuration.}
    \label{fig:benchmark_models}
    
\end{figure}

Table~\ref{tab:benchmark_results} summarizes the quantitative results for the 24 reference reconstructions, while Figure~\ref{fig:benchmark_models} presents the corresponding source images and generated models. Across the benchmark, FRDS ranges from 0.9126 to 0.9937, with a mean of 0.9647 and a median of 0.9671. All cases obtain an FRDS above 0.90, and 19 of the 24 exceed 0.95. These results indicate that the reconstructed models generally preserve the projected building extent across buildings with substantially different architectural configurations and observation conditions.

\begin{table}[H]
\centering
\caption{Quantitative results for the 24 reference AstraLOD3 reconstructions. Higher FRDS and lower $D_{\mathrm{CD}}$ and IMF indicate greater agreement. The geometric metrics are expressed in the SfM scale of each individual dataset and are therefore not directly comparable as physical errors across different buildings.}
\label{tab:benchmark_results}
\small
\begin{tabular}{clcccc}
\hline
\textbf{ID} & \textbf{Building} & \textbf{Time} &
\textbf{FRDS $\uparrow$} &
\textbf{$D_{\mathrm{CD}}\downarrow$} &
\textbf{IMF $\downarrow$} \\
\hline
01 & Petrinja building              & 32:41 & 0.9766 & 0.0424 & 0.0476 \\
02 & Parish house                   & 21:24 & 0.9841 & 0.0199 & 0.0228 \\
03 & Petrinja school                & 38:47 & 0.9873 & 0.0188 & 0.0234 \\
04 & Country house                  & 23:33 & 0.9718 & 0.0123 & 0.0400 \\
05 & School                         & 23:38 & 0.9646 & 0.0426 & 0.0515 \\
06 & UNIL                           & 18:08 & 0.9814 & 0.0026 & 0.0222 \\
07 & Ozcan                          & 21:22 & 0.9126 & 0.0272 & 0.0515 \\
08 & Bianco                         & 23:39 & 0.9581 & 0.0723 & 0.0584 \\
09 & Paiano                         & 17:47 & 0.9644 & 0.0163 & 0.0426 \\
10 & Wyss                           & 24:47 & 0.9473 & 0.0380 & 0.0458 \\
11 & Gerlmerbahn                    & 20:00 & 0.9556 & 0.0161 & 0.0177 \\
12 & Church Sela                    & 23:33 & 0.9646 & 0.0146 & 0.0273 \\
13 & Church Popusko                 & 21:15 & 0.9696 & 0.0249 & 0.0501 \\
14 & Church Gora                    & 22:32 & 0.9763 & 0.0274 & 0.0377 \\
15 & House Aegerten                 & 18:28 & 0.9540 & 0.0529 & 0.0699 \\
16 & House Huttwil                  & 21:19 & 0.9536 & 0.1024 & 0.1682 \\
17 & House Eriswil                  & 19:55 & 0.9325 & 0.0421 & 0.0691 \\
18 & UNIL 2                         & 22:43 & 0.9830 & 0.0075 & 0.0277 \\
19 & Country house 2                & 20:55 & 0.9346 & 0.0372 & 0.0364 \\
20 & \'Ecole Roseaux                & 26:49 & 0.9906 & 0.0439 & 0.0239 \\
21 & Gymnase de Chamblandes         & 20:35 & 0.9749 & 0.0337 & 0.0314 \\
22 & UNIL 3                         & 26:53 & 0.9777 & 0.0697 & 0.0872 \\
23 & Country house 3                & 16:15 & 0.9448 & 0.0071 & 0.0265 \\
24 & Synthetic house                & 12:32 & 0.9937 & 0.0168 & 0.0482 \\
\hline
\end{tabular}
\end{table}

Processing time ranges from 12:32 to 38:47, with a mean of approximately 22:29 and a median of 21:23. It does not vary directly with the number of input images. For example, UNIL~2 contains 188 images and is reconstructed in 22:43, whereas Petrinja school contains 72 images and requires 38:47. The processing time therefore depends not only on the amount of input evidence but also on the computational operations selected during execution, the complexity of the geometric interpretation, and the validation and refinement performed for each building.

Building~24 provides a useful controlled reference. Under the comparatively ideal conditions of the synthetic dataset, the reconstruction obtains the highest FRDS of the benchmark, 0.9937, indicating an almost complete agreement between the projected model and the reference building masks. Its $D_{\mathrm{CD}}$ and IMF values should not, however, be compared directly with those of the other buildings because each SfM reconstruction has an independent scale. Moreover, even within an individual case, the geometric metrics are influenced by the composition of the sparse point cloud. For Building~24, the SfM cloud contains points belonging to the surrounding ground, which is excluded from the architectural reconstruction and therefore contributes to the measured point-cloud-to-model discrepancy. The opposite situation can also occur when architectural surfaces visible in the images, such as weakly textured roof regions, are reconstructed in the model but contain few corresponding SfM points. The geometric metrics should therefore be interpreted as measures of consistency with the available sparse reconstruction rather than as complete measures of architectural correctness.

\begin{figure}[H]
    \centering

    \begin{tabular}{
        >{\centering\arraybackslash}m{0.04\textwidth}
        >{\centering\arraybackslash}m{0.2\textwidth}
        >{\centering\arraybackslash}m{0.2\textwidth}
        >{\centering\arraybackslash}m{0.2\textwidth}
        >{\centering\arraybackslash}m{0.2\textwidth}
    }
        \textbf{ID} &
        \textbf{Image} &
        \textbf{Model} &
        \textbf{Contour overlay} &
        \textbf{Point-cloud overlay} \\[2mm]

        04 &
        \includegraphics[width=\linewidth]{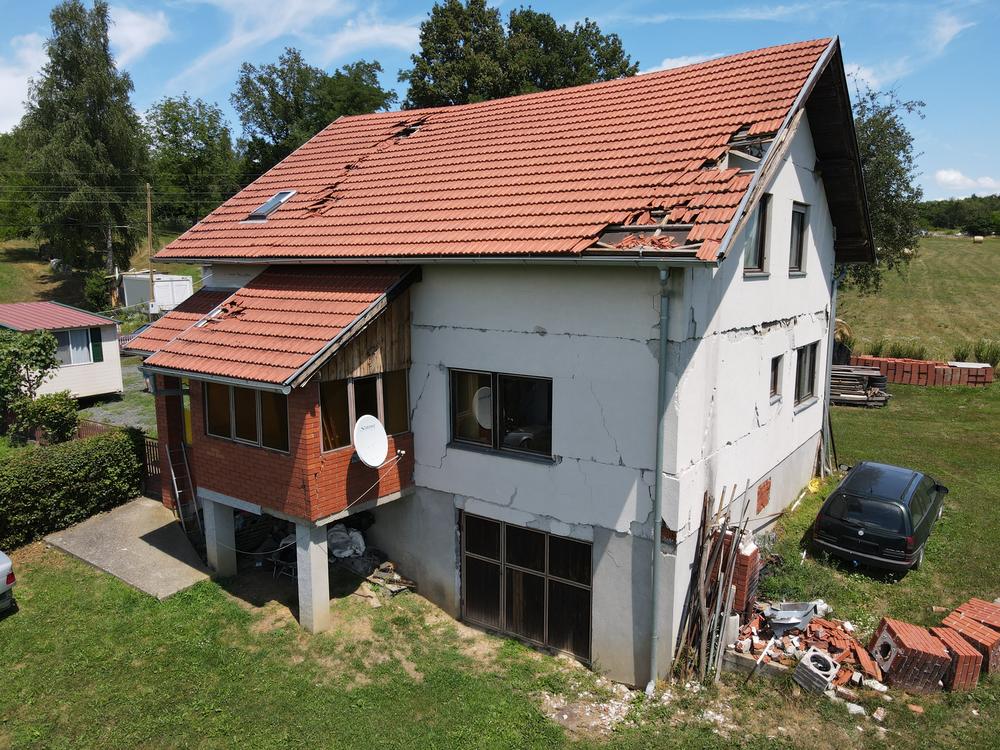} &
        \includegraphics[width=\linewidth]{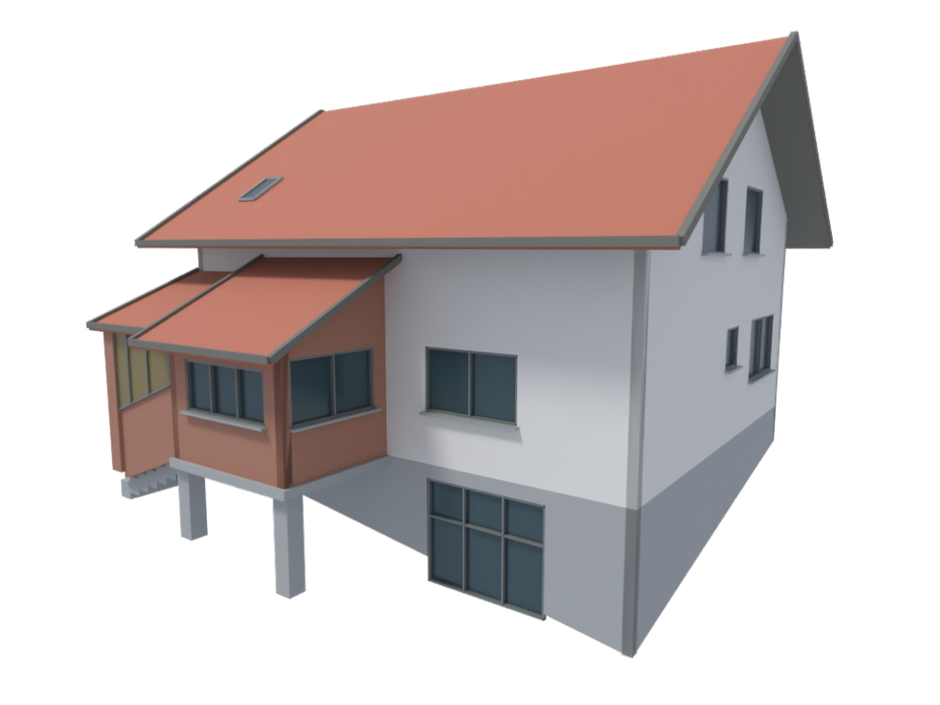} &
        \includegraphics[width=\linewidth]{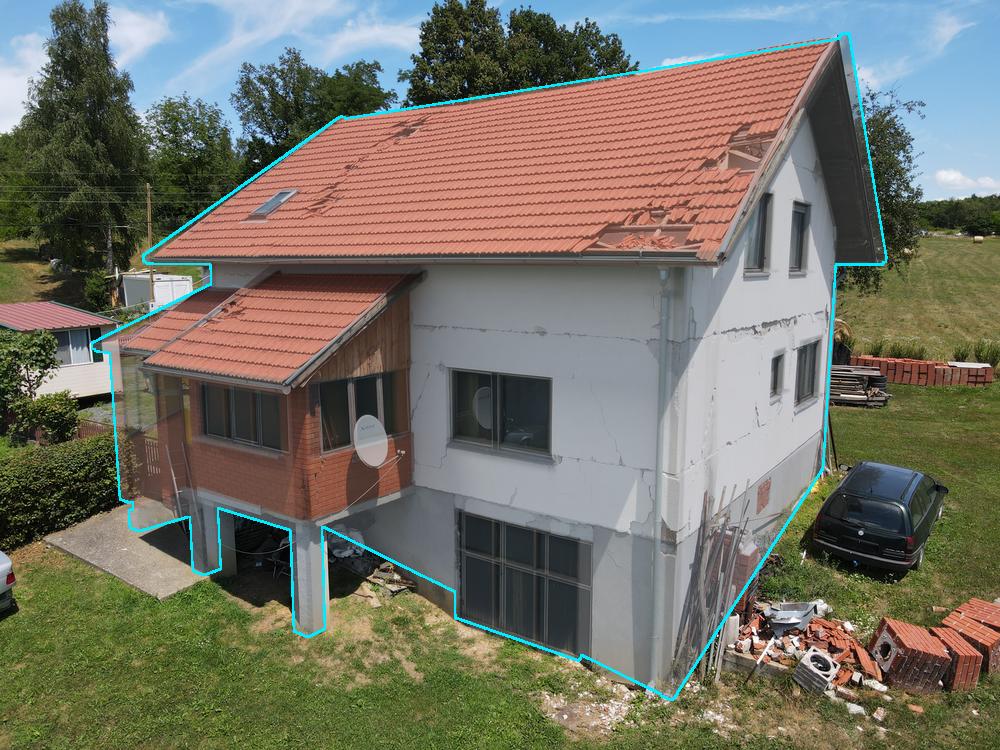} & 
        \includegraphics[width=\linewidth]{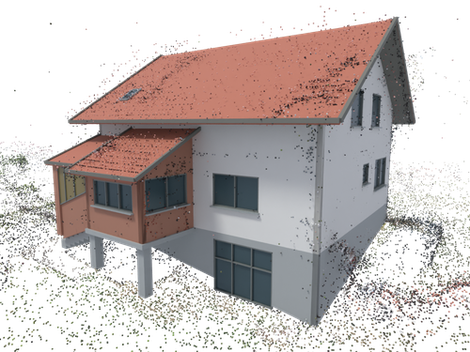}\\

        05 &
        \includegraphics[width=\linewidth]{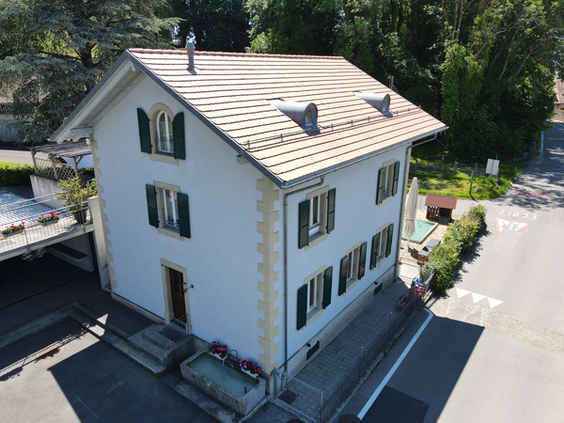} &
        \includegraphics[width=\linewidth]{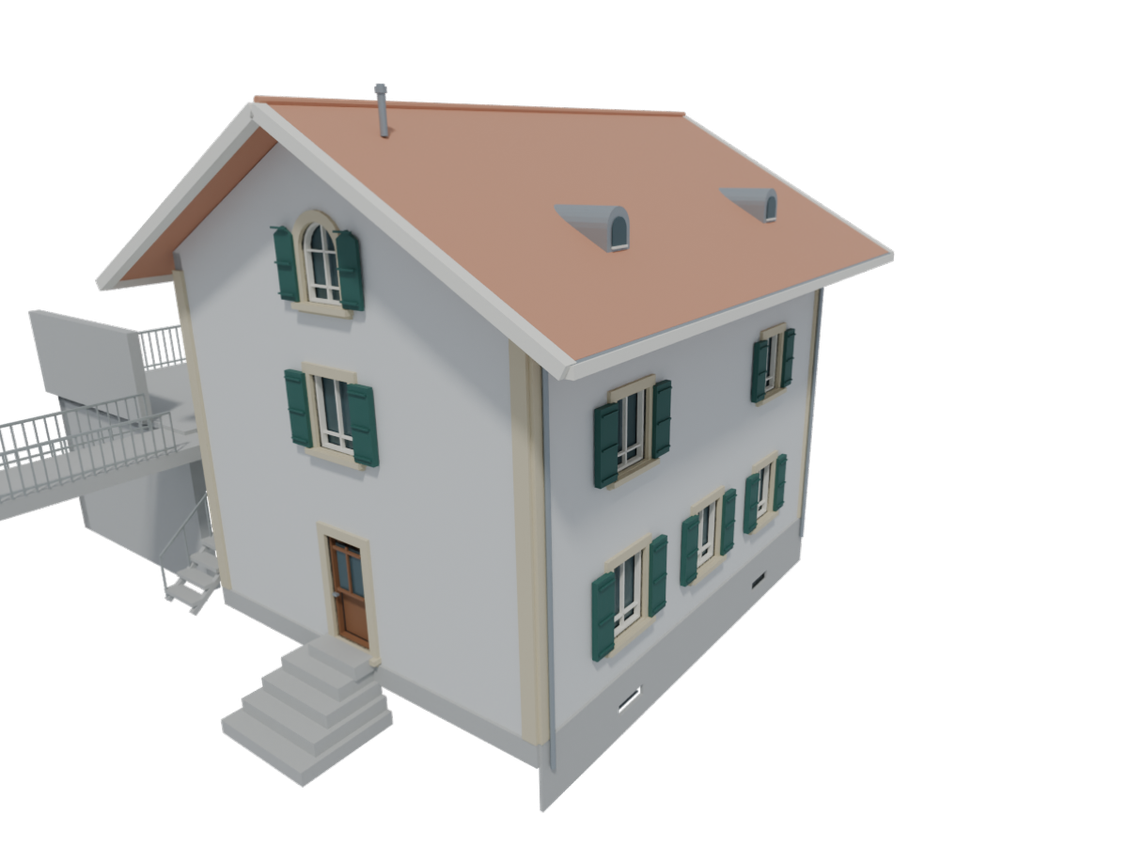} &
        \includegraphics[width=\linewidth]{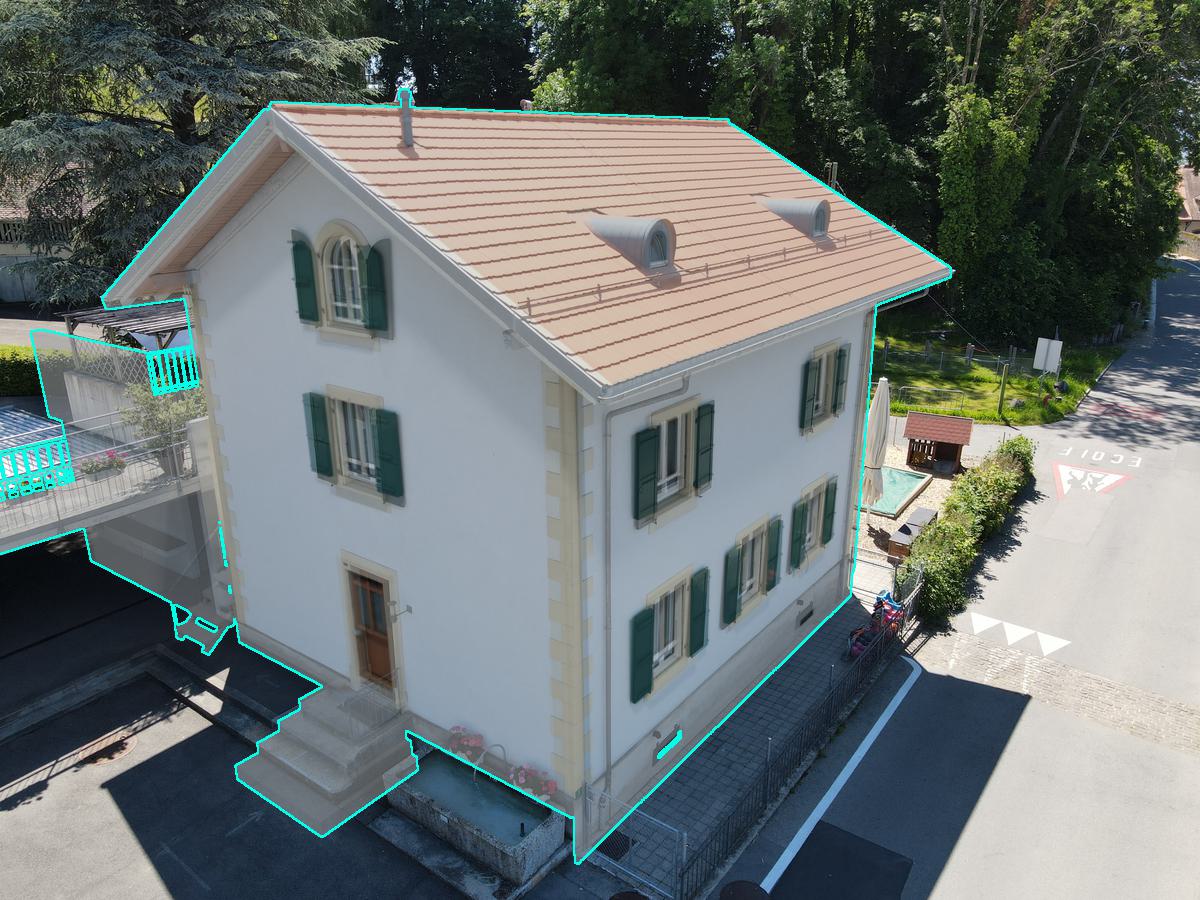} &
        \includegraphics[width=\linewidth]{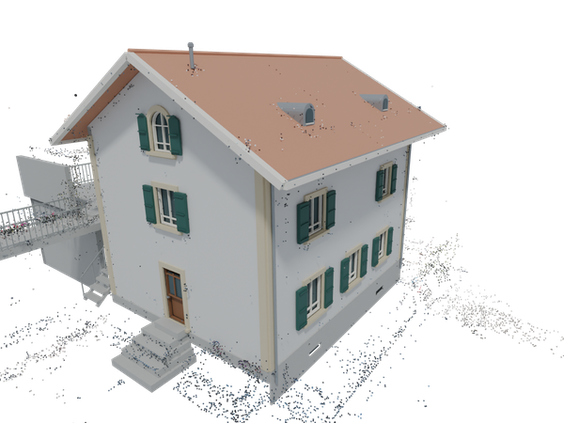}\\

        11 &
        \includegraphics[width=\linewidth]{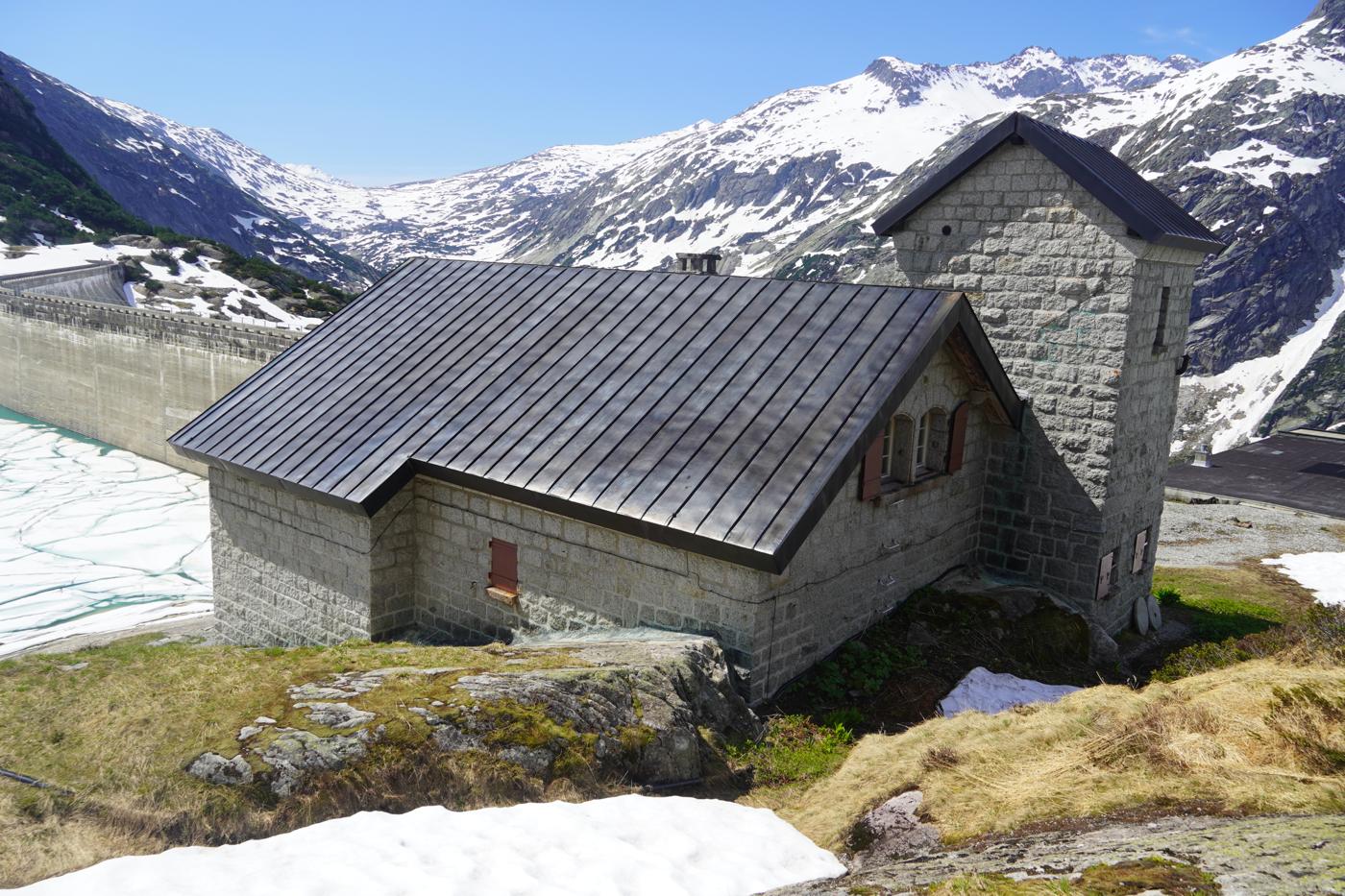} &
        \includegraphics[width=\linewidth]{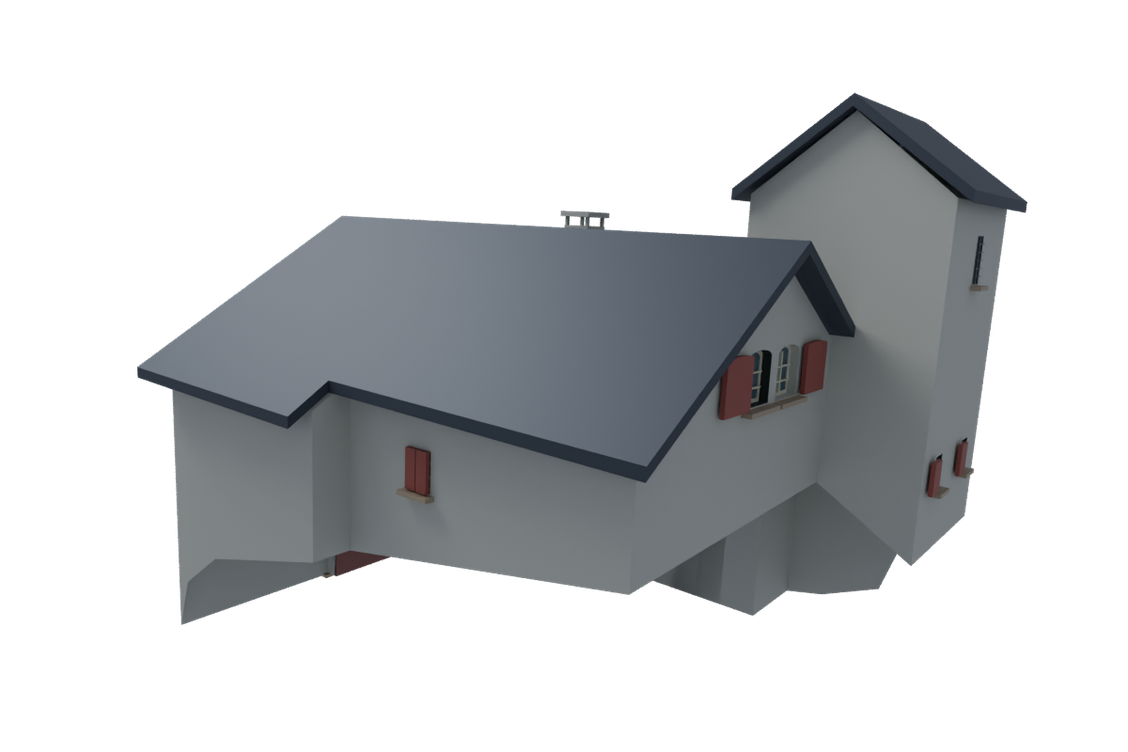} &
        \includegraphics[width=\linewidth]{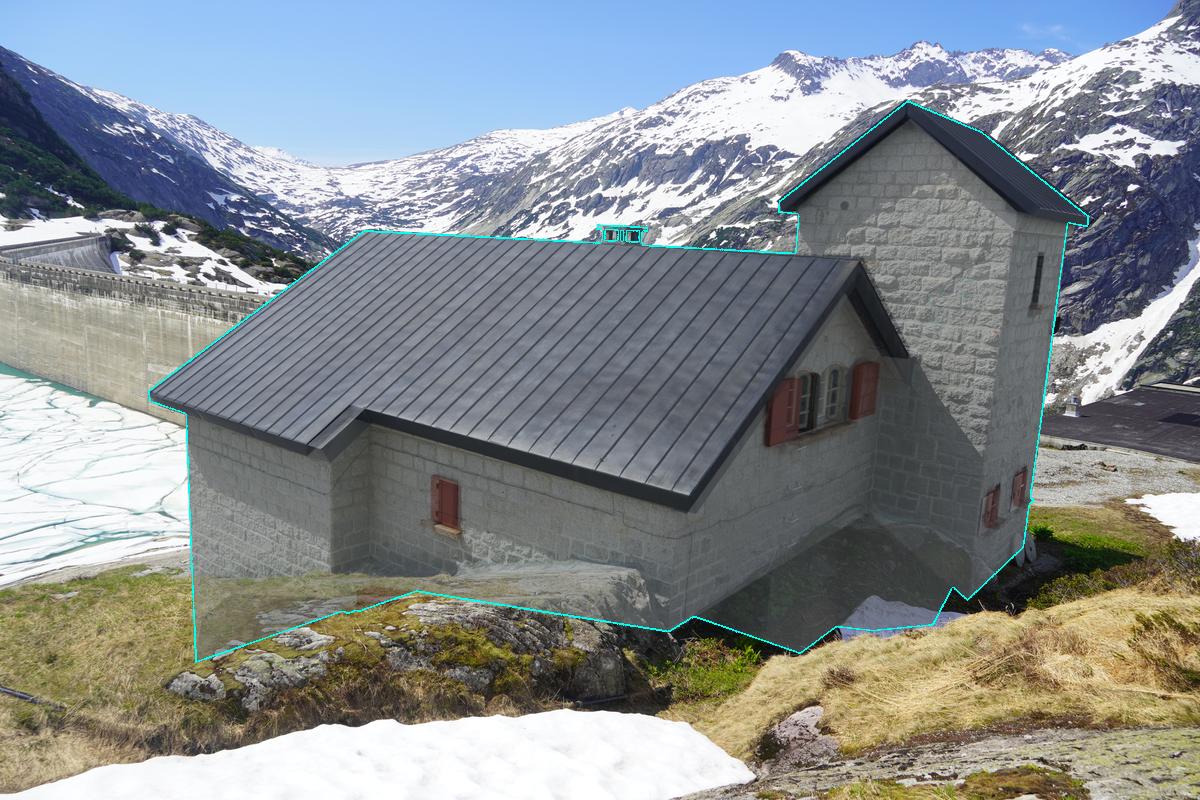} &
        \includegraphics[width=\linewidth]{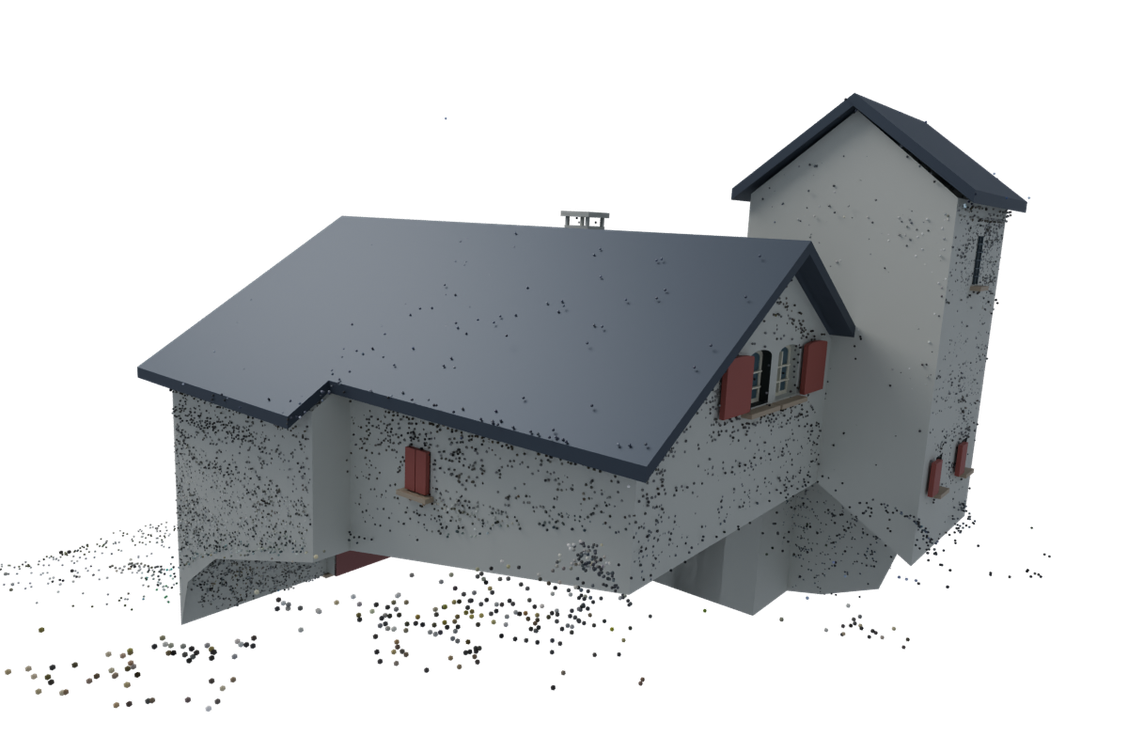}\\

        16 &
        \includegraphics[width=\linewidth]{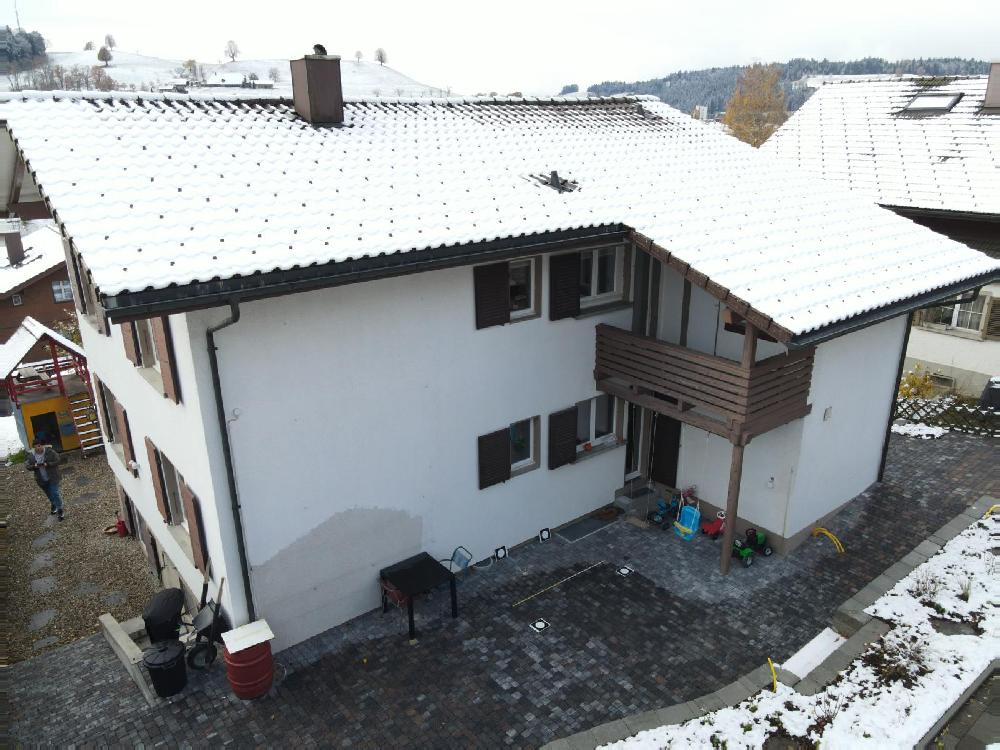} &
        \includegraphics[width=\linewidth]{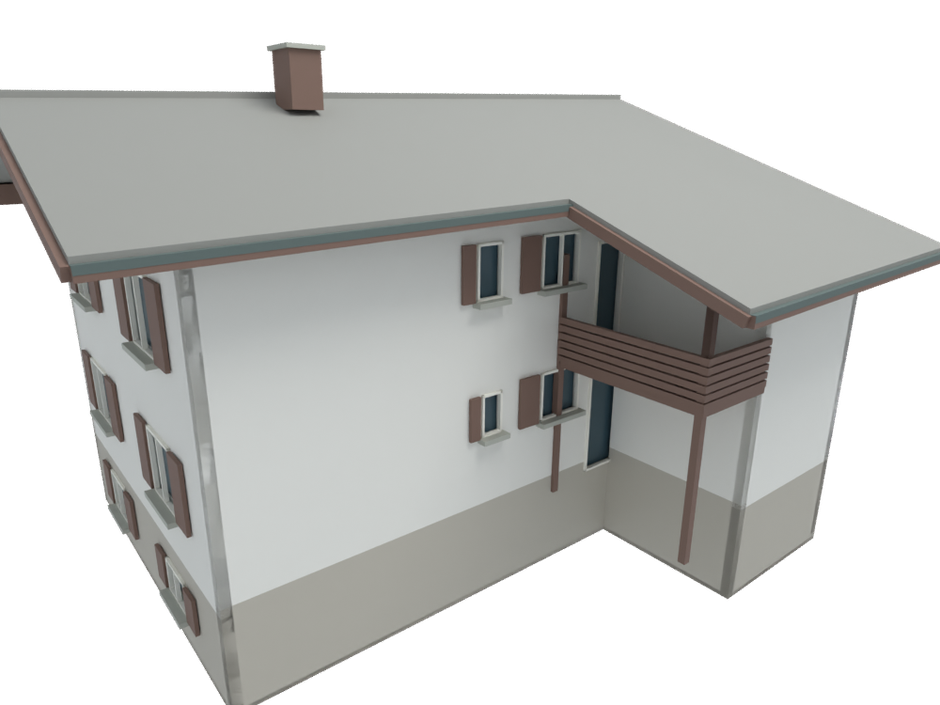} &
        \includegraphics[width=\linewidth]{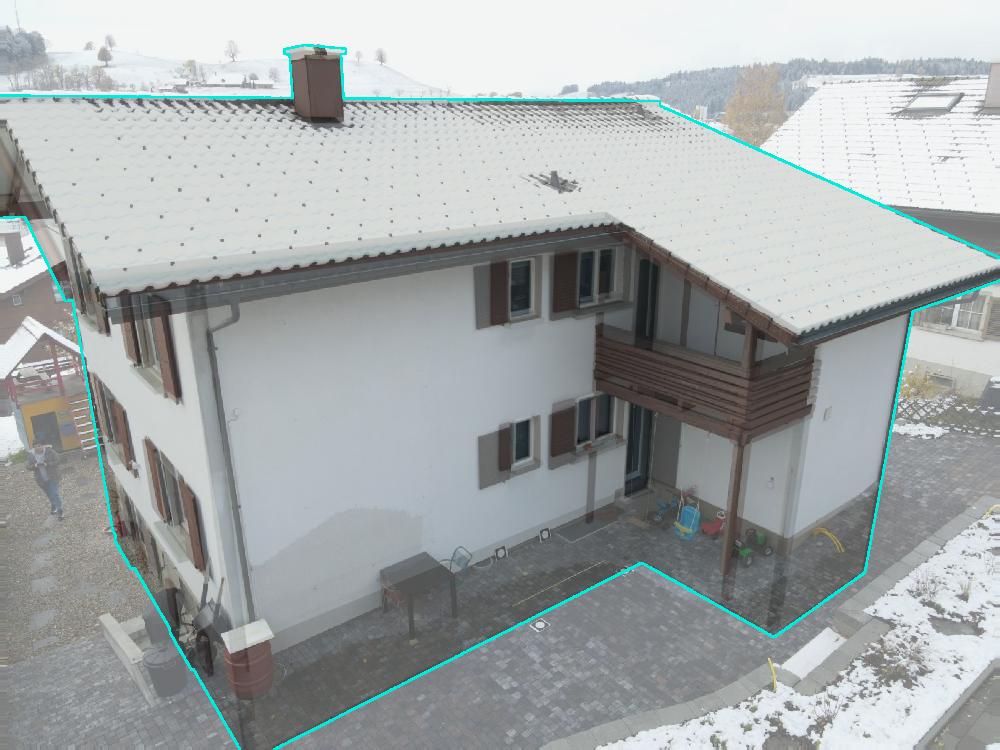} &
        \includegraphics[width=\linewidth]{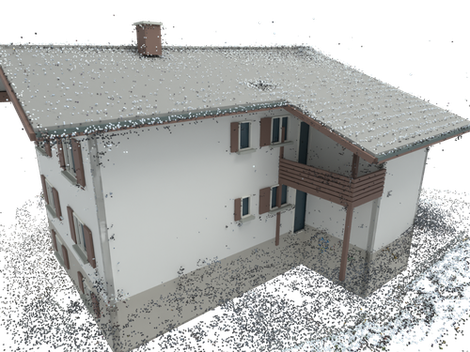}\\

        24 &
        \includegraphics[width=\linewidth]{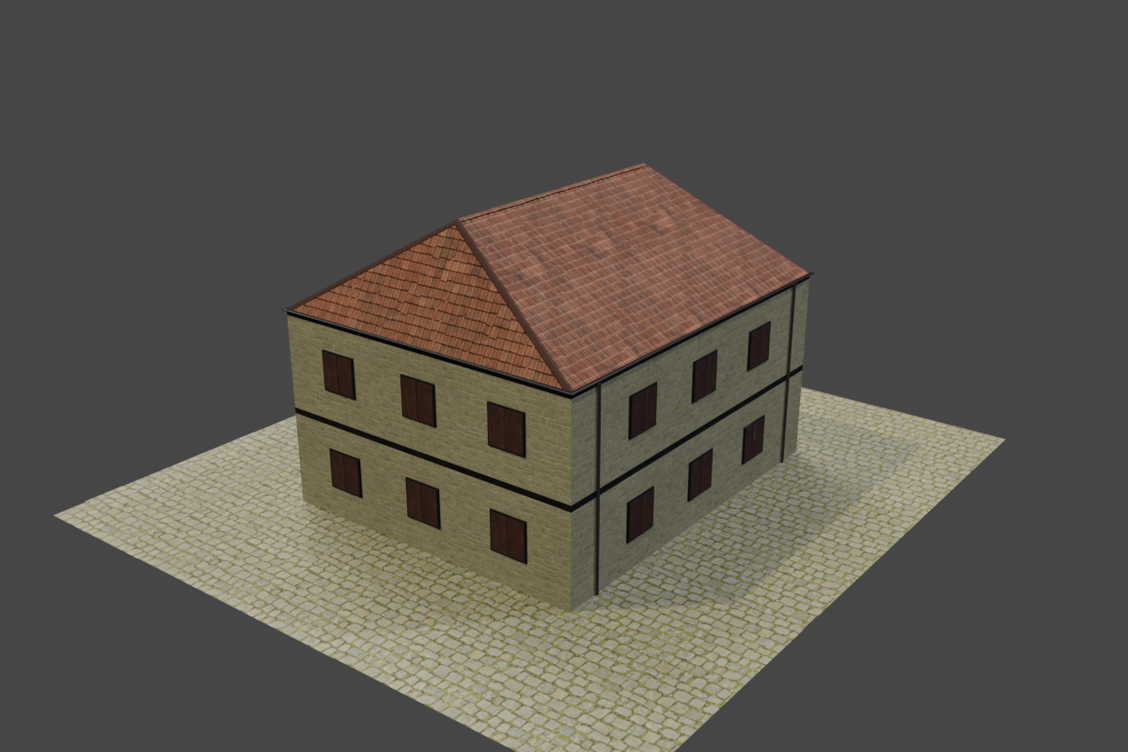} &
        \includegraphics[width=\linewidth]{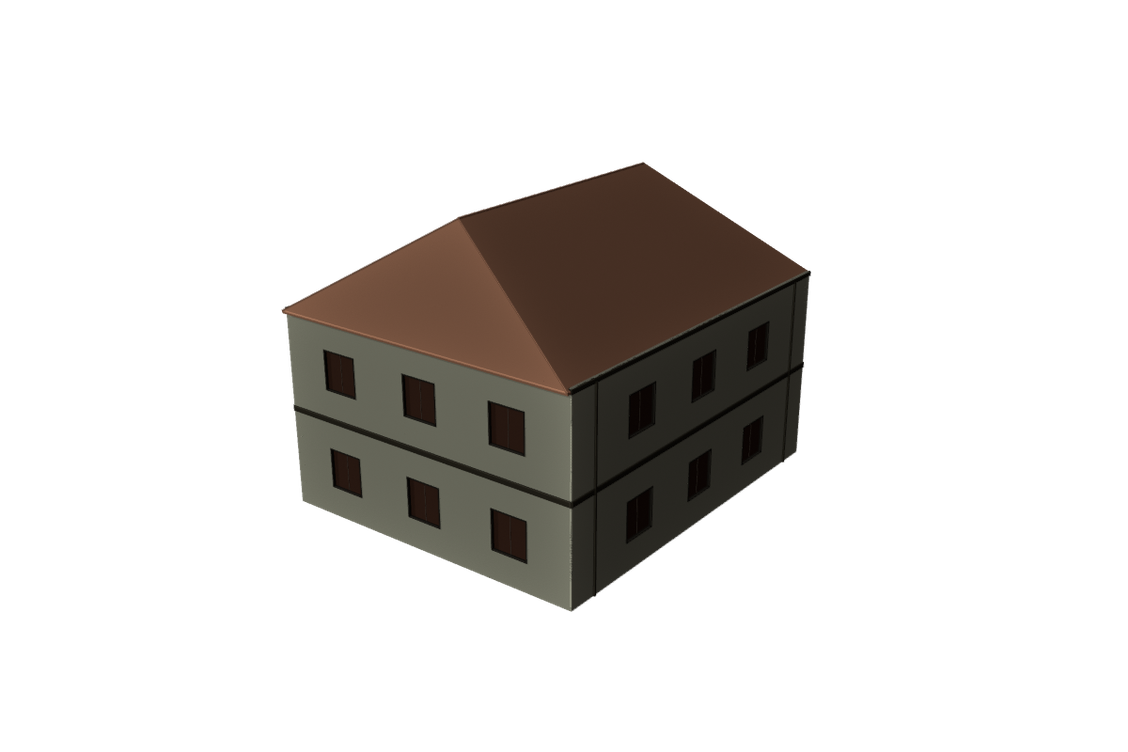} &
        \includegraphics[width=\linewidth]{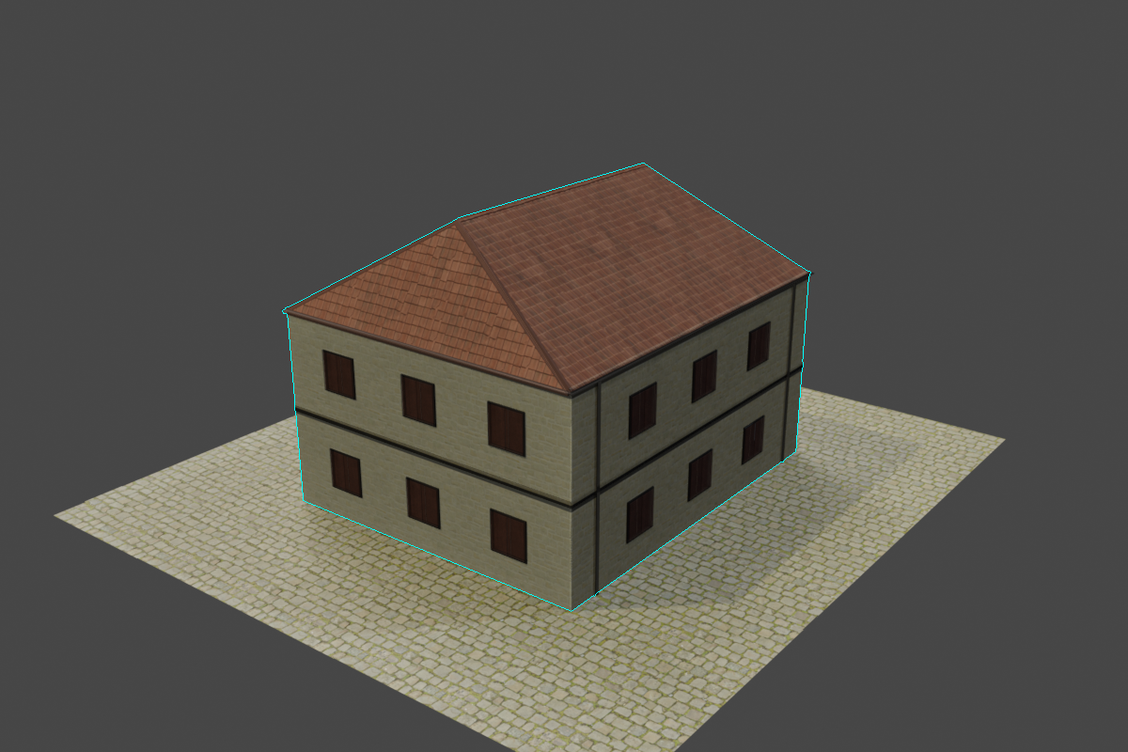} &
        \includegraphics[width=\linewidth]{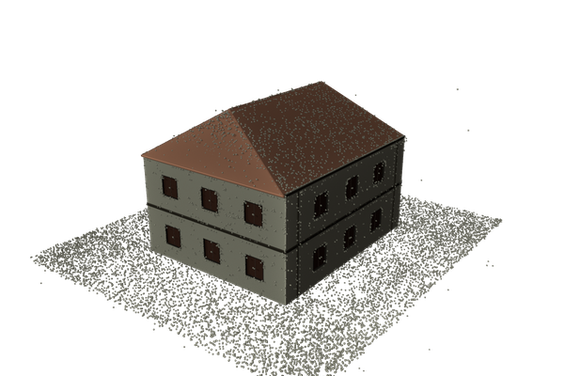}\\

    \end{tabular}

    \caption{Representative qualitative validation of AstraLOD3 reconstructions. Each row shows a source image, the model rendered from the corresponding calibrated camera, the model-mask contour overlaid on the source image, and the model overlaid with the filtered sparse SfM point cloud. The contour and point-cloud comparisons provide qualitative counterparts to FRDS and the geometric metrics, respectively.}
    \label{fig:qualitative_validation}
\end{figure}

Figure~\ref{fig:qualitative_validation} provides representative qualitative examples supporting the quantitative evaluation. For each selected building, the reconstructed model is rendered from a calibrated camera and its projected contour is overlaid on the corresponding source image, providing a visual interpretation of the agreement measured by FRDS. The model is additionally shown together with the filtered sparse SfM point cloud, allowing the three-dimensional relationship underlying $D_{\mathrm{CD}}$ and IMF to be inspected qualitatively. The examples show that the reconstructed envelopes generally follow both the image-supported building boundaries and the dominant point-cloud geometry, while local differences remain around simplified, weakly observed, or occluded regions.

\subsection{Comparison with previous reconstruction approaches}
\label{sec:comparison_results}

Table~\ref{tab:previous_comparison_results} compares AstraLOD3 with the corresponding models from the previous LOD3~\cite{PantojaRosero2022LOD3} and DADT~\cite{PantojaRosero2023DADT} studies. As described in Section~\ref{sec:previous_work_comparison}, the geometric metrics are evaluated at a common scale for each corresponding building.

\begin{table}[H]
\centering
\caption{Quantitative comparison between AstraLOD3 and corresponding models from the previous LOD3~\cite{PantojaRosero2022LOD3} and DADT~\cite{PantojaRosero2023DADT} studies. Higher FRDS and lower $D_{\mathrm{CD}}$ and IMF indicate greater agreement. Point cloud metrics are compared at a common scale for each building.}
\label{tab:previous_comparison_results}
\scriptsize
\resizebox{\textwidth}{!}{
\begin{tabular}{clccccccc}
\hline
& & \multicolumn{3}{c}{\textbf{AstraLOD3}} &
\multicolumn{4}{c}{\textbf{Previous reconstruction}} \\
\textbf{ID} & \textbf{Building} &
\textbf{FRDS} & \textbf{$D_{\mathrm{CD}}$} & \textbf{IMF} &
\textbf{Method} & \textbf{FRDS} & \textbf{$D_{\mathrm{CD}}$} & \textbf{IMF} \\
\hline
01 & Petrinja building & \textbf{0.9766} & 0.0424 & 0.0476 & DADT & 0.9600 & \textbf{0.0380} & \textbf{0.0284} \\
02 & Parish house      & \textbf{0.9841} & \textbf{0.0199} & 0.0228 & DADT & 0.9700 & 0.0312 & \textbf{0.0179} \\
03 & Petrinja school   & \textbf{0.9873} & \textbf{0.0188} & \textbf{0.0234} & DADT & 0.9400 & 0.2900 & 0.0744 \\
04 & Country house     & \textbf{0.9718} & \textbf{0.0123} & 0.0400 & DADT & 0.9100 & 0.0253 & \textbf{0.0330} \\
05 & School            & \textbf{0.9646} & 0.0426 & 0.0515 & LOD3 & 0.9533 & \textbf{0.0148} & \textbf{0.0416} \\
06 & UNIL              & \textbf{0.9814} & \textbf{0.0026} & \textbf{0.0222} & LOD3 & 0.9500 & 0.0242 & 0.0306 \\
07 & Ozcan             & 0.9126 & \textbf{0.0272} & \textbf{0.0515} & LOD3 & \textbf{0.9150} & 0.0273 & 0.0724 \\
08 & Bianco            & \textbf{0.9581} & 0.0723 & \textbf{0.0584} & LOD3 & 0.8750 & \textbf{0.0341} & 0.0766 \\
09 & Paiano            & \textbf{0.9644} & 0.0163 & \textbf{0.0426} & LOD3 & 0.9413 & \textbf{0.0162} & 0.0456 \\
10 & Wyss              & \textbf{0.9473} & 0.0380 & 0.0458 & LOD3 & 0.8200 & \textbf{0.0230} & \textbf{0.0299} \\
11 & Gerlmerbahn       & \textbf{0.9556} & \textbf{0.0161} & \textbf{0.0177} & LOD3 & 0.8375 & 0.0350 & 0.0219 \\
14 & Church Gora       & \textbf{0.9763} & \textbf{0.0274} & 0.0377 & DADT & 0.9500 & 0.0366 & \textbf{0.0351} \\
19 & Country house 2   & \textbf{0.9346} & \textbf{0.0372} & \textbf{0.0364} & DADT & 0.8600 & 0.0628 & 0.0474 \\
23 & Country house 3   & \textbf{0.9448} & 0.0071 & 0.0265 & DADT & 0.9200 & \textbf{0.0066} & \textbf{0.0209} \\
\hline
\end{tabular}
}
\end{table}

\begin{figure}[H]
    \centering
    \resizebox{\textwidth}{!}{
    \begin{tabular}{cccccc}
        \textbf{Source} &
        \textbf{AstraLOD3} &
        \textbf{DADT~\cite{PantojaRosero2023DADT}} &
        \textbf{Source} &
        \textbf{AstraLOD3} &
        \textbf{LOD3~\cite{PantojaRosero2022LOD3}} \\[2mm]

        \includegraphics[width=0.15\textwidth]{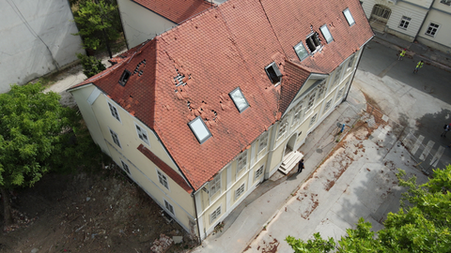} &
        \includegraphics[width=0.15\textwidth]{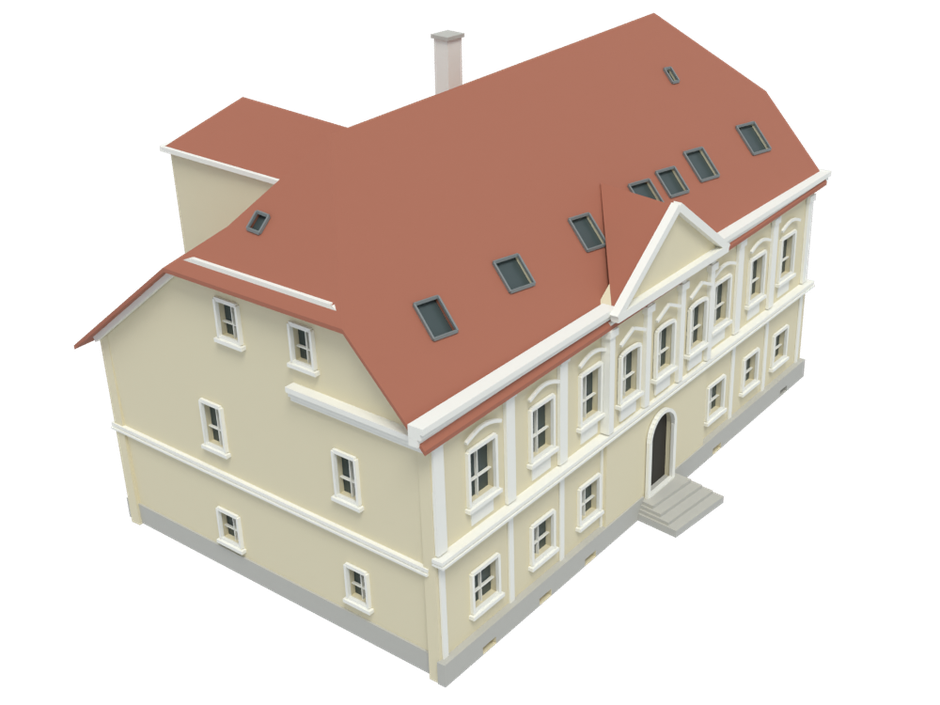} &
        \includegraphics[width=0.15\textwidth]{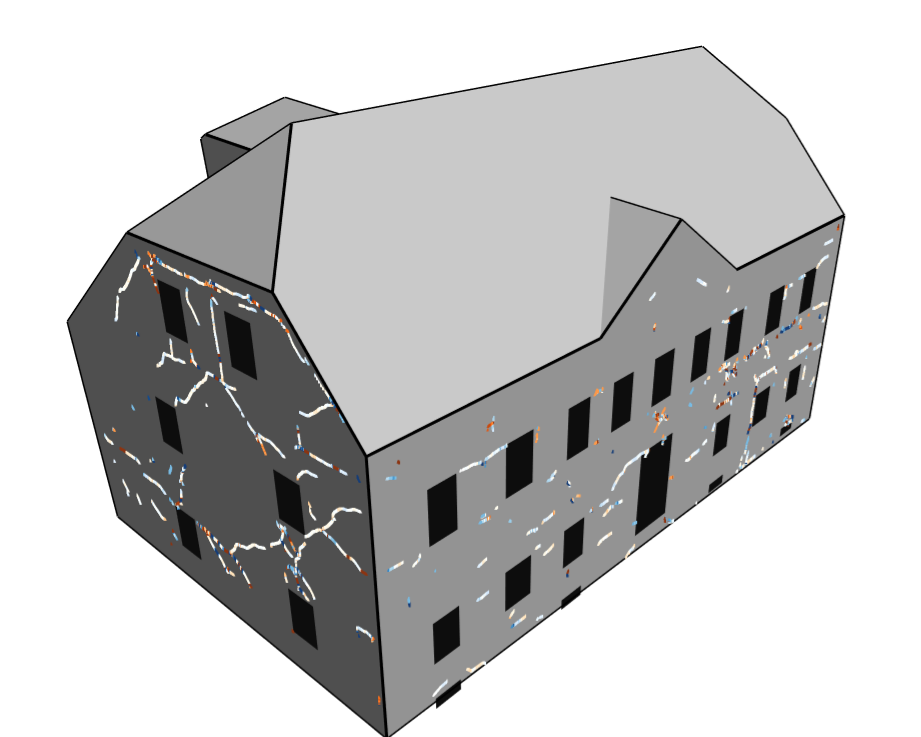} &
        \includegraphics[width=0.15\textwidth]{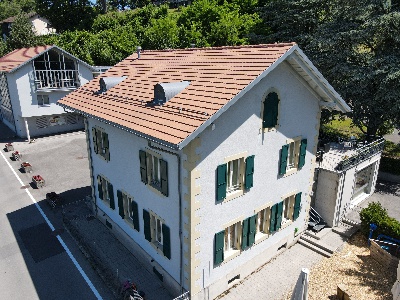} &
        \includegraphics[width=0.15\textwidth]{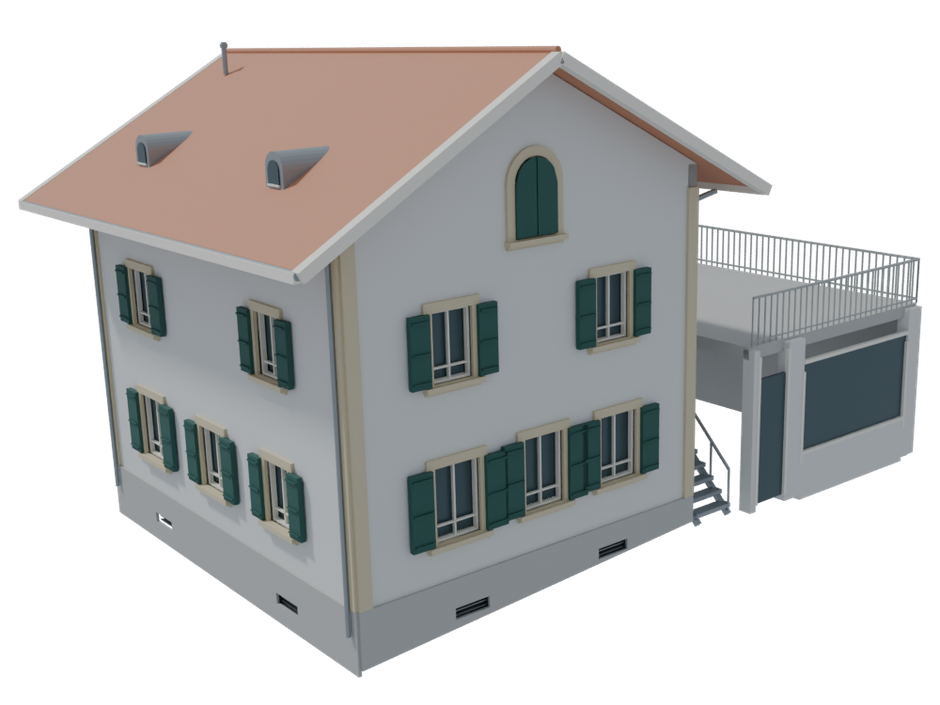} &
        \includegraphics[width=0.12\textwidth]{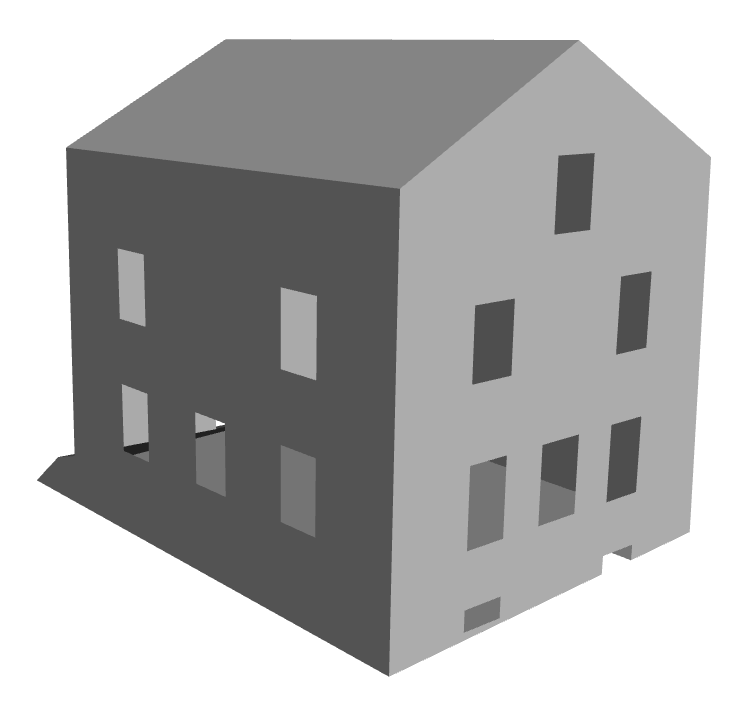} \\
        \multicolumn{3}{c}{\scriptsize\textbf{01}} &
        \multicolumn{3}{c}{\scriptsize\textbf{05}} \\[2mm]

        \includegraphics[width=0.15\textwidth]{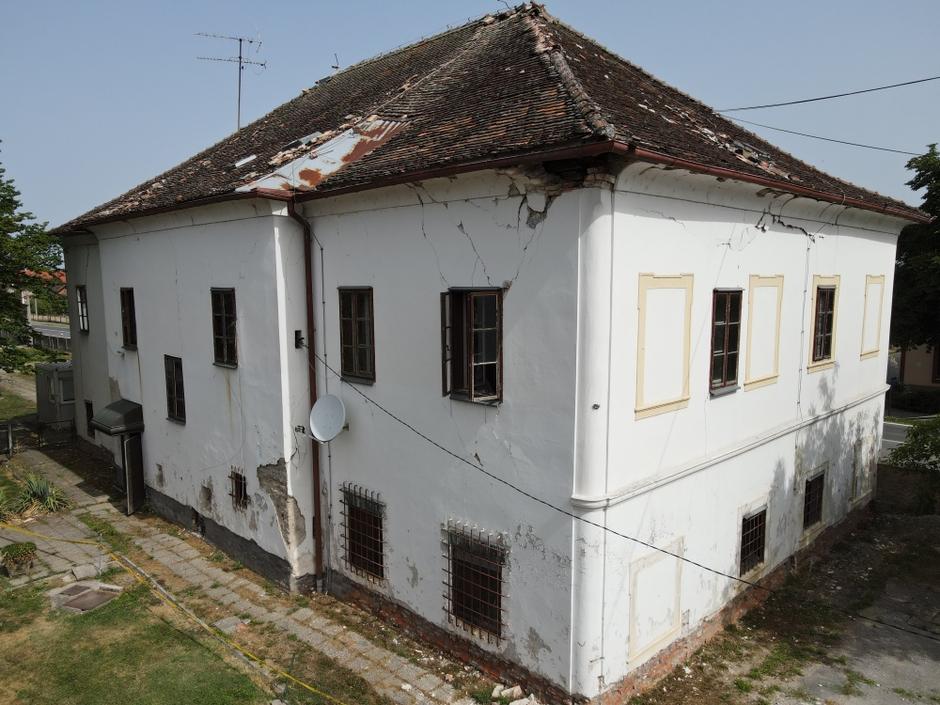} &
        \includegraphics[width=0.15\textwidth]{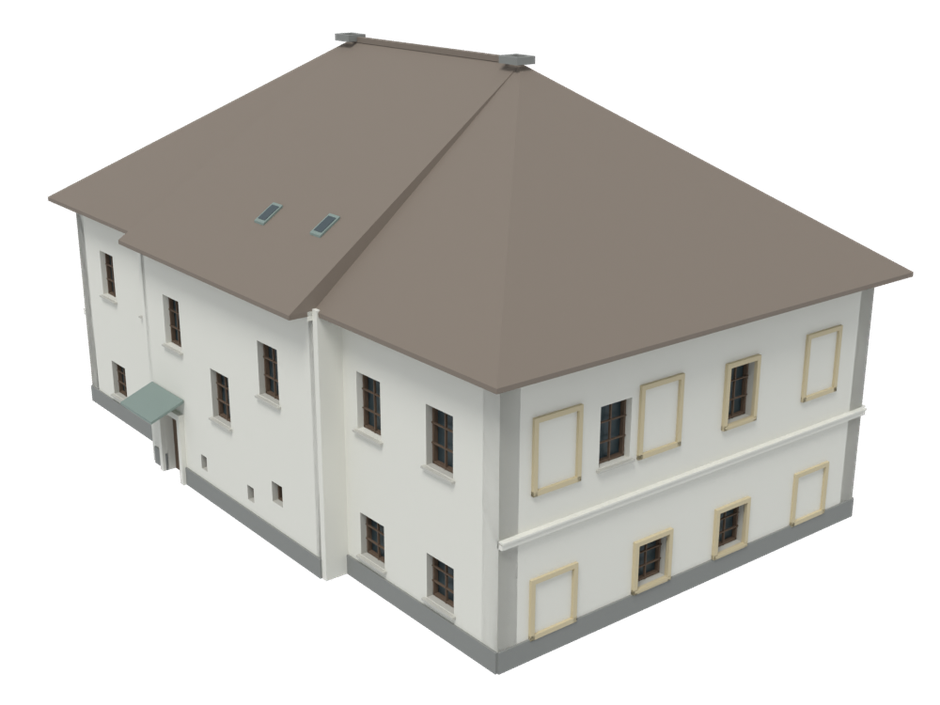} &
        \includegraphics[width=0.16\textwidth]{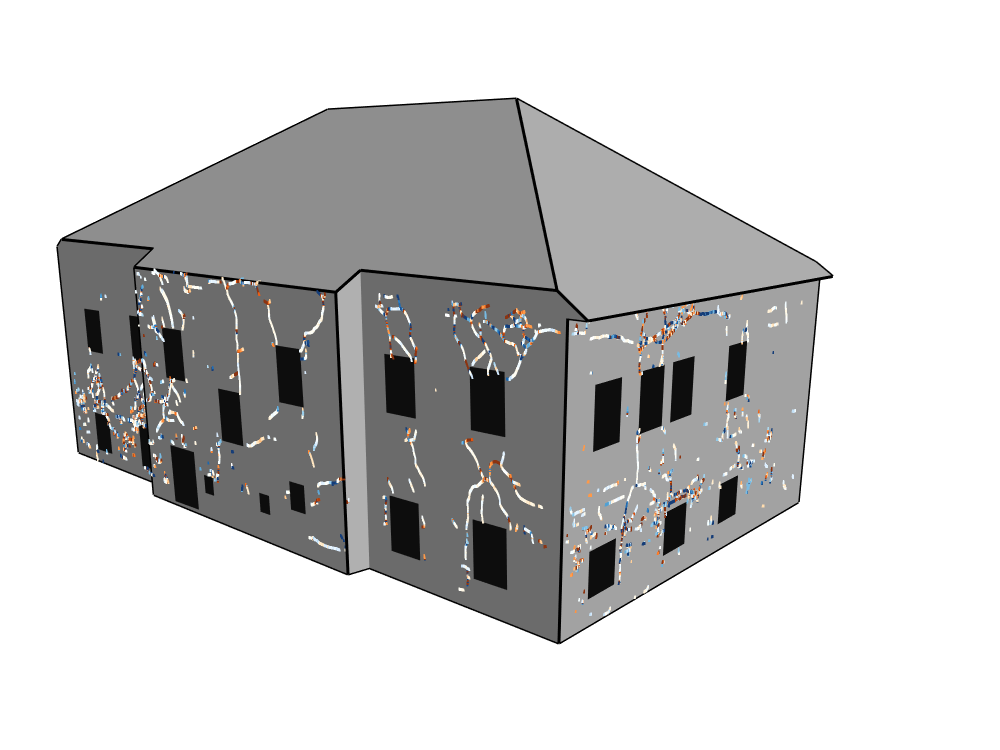} &
        \includegraphics[width=0.15\textwidth]{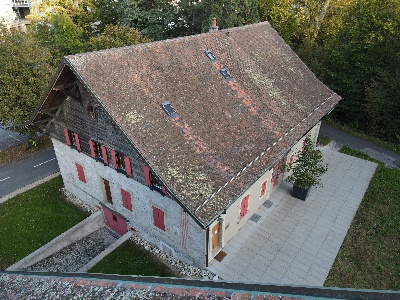} &
        \includegraphics[width=0.15\textwidth]{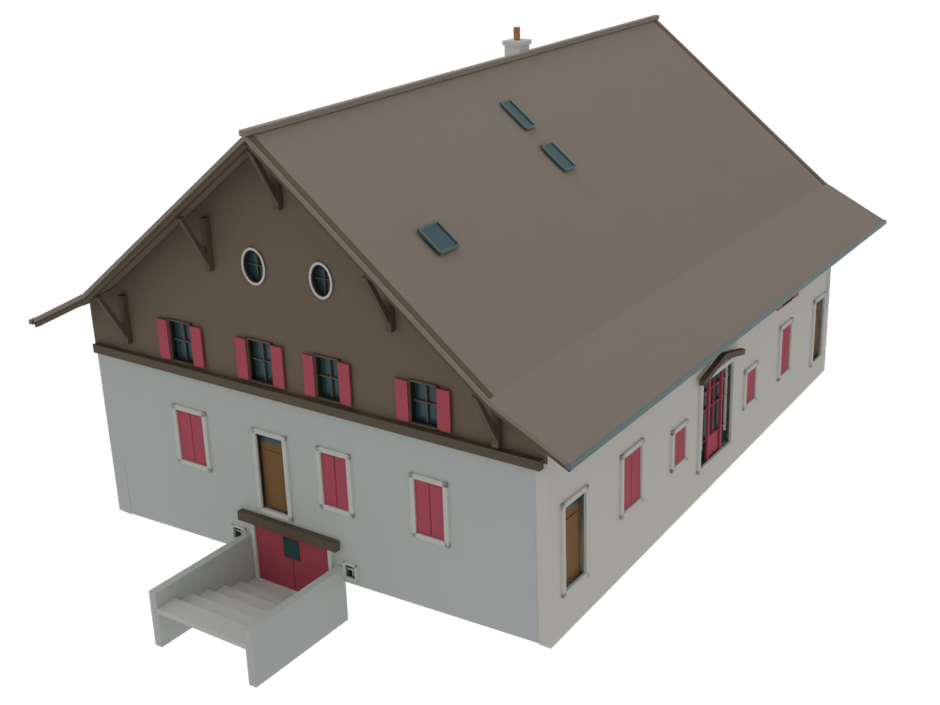} &
        \includegraphics[width=0.15\textwidth]{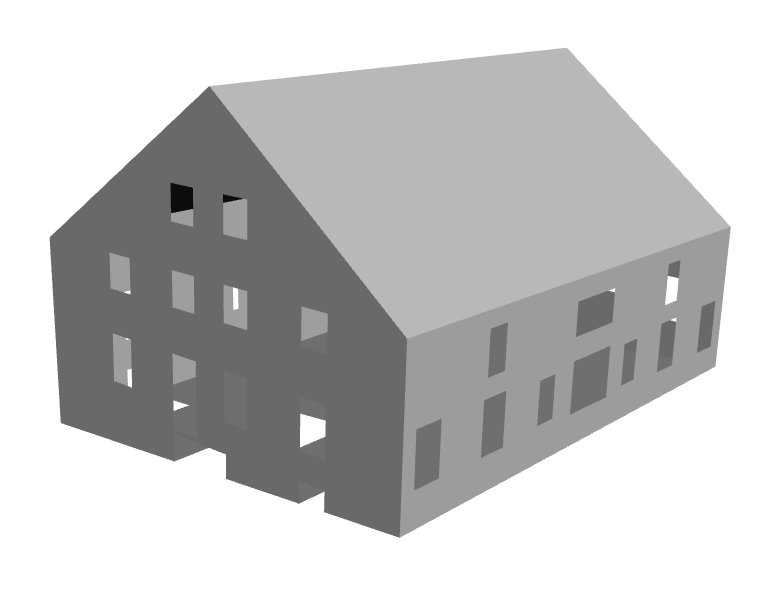} \\
        \multicolumn{3}{c}{\scriptsize\textbf{02}} &
        \multicolumn{3}{c}{\scriptsize\textbf{06}} \\[2mm]

        \includegraphics[width=0.15\textwidth]{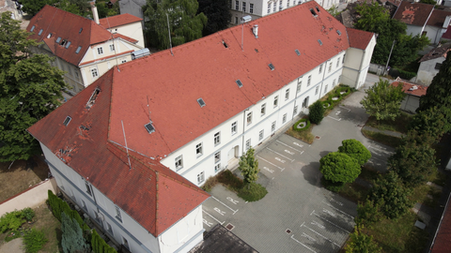} &
        \includegraphics[width=0.15\textwidth]{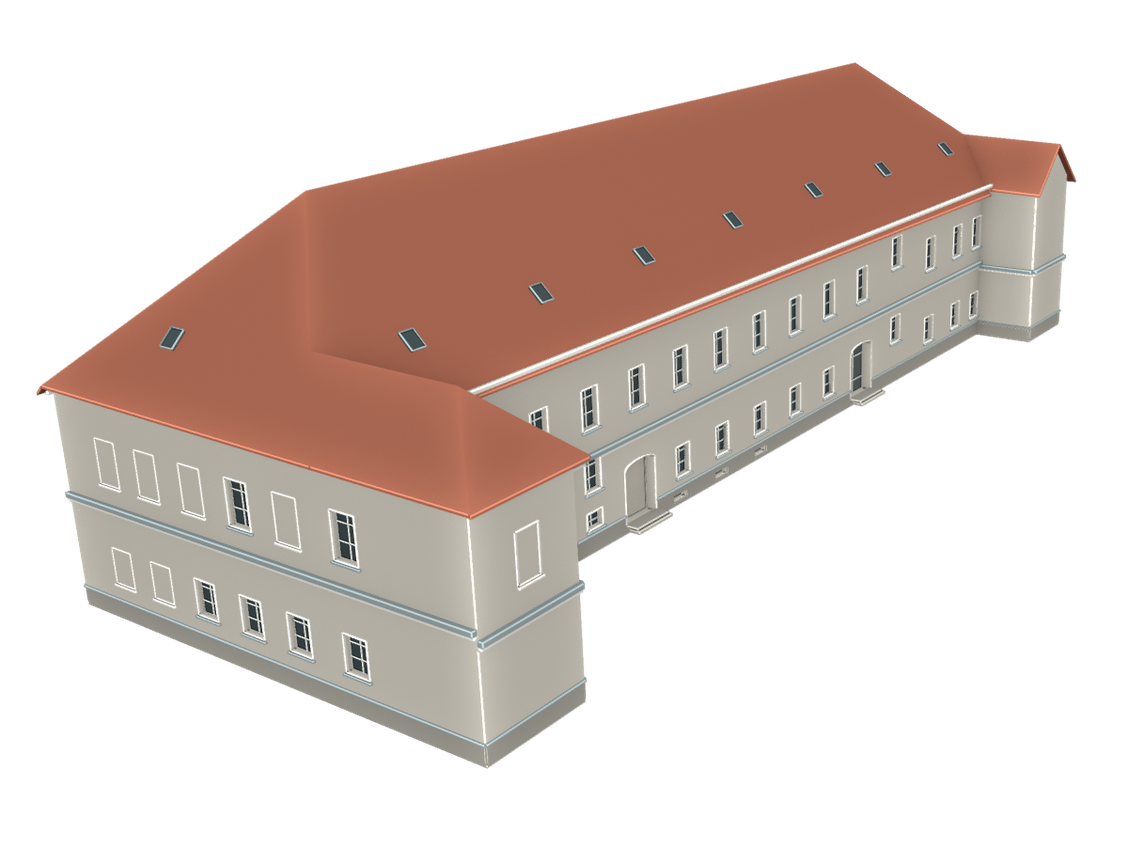} &
        \includegraphics[width=0.15\textwidth]{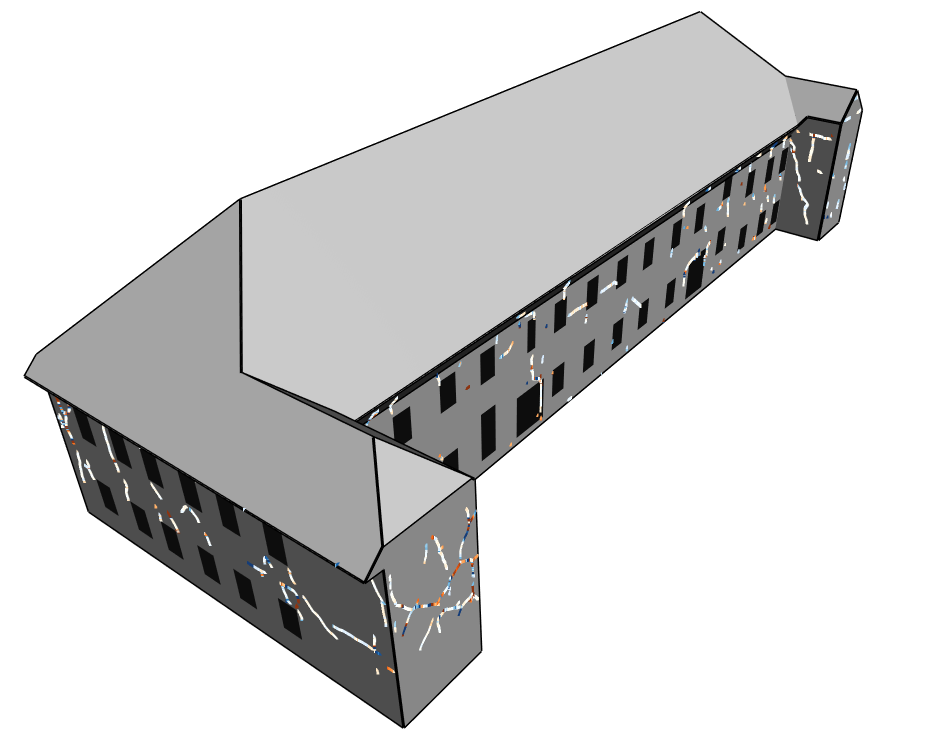} &
        \includegraphics[width=0.15\textwidth]{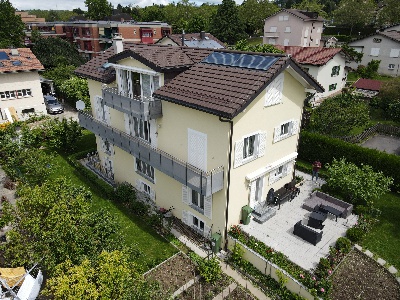} &
        \includegraphics[width=0.15\textwidth]{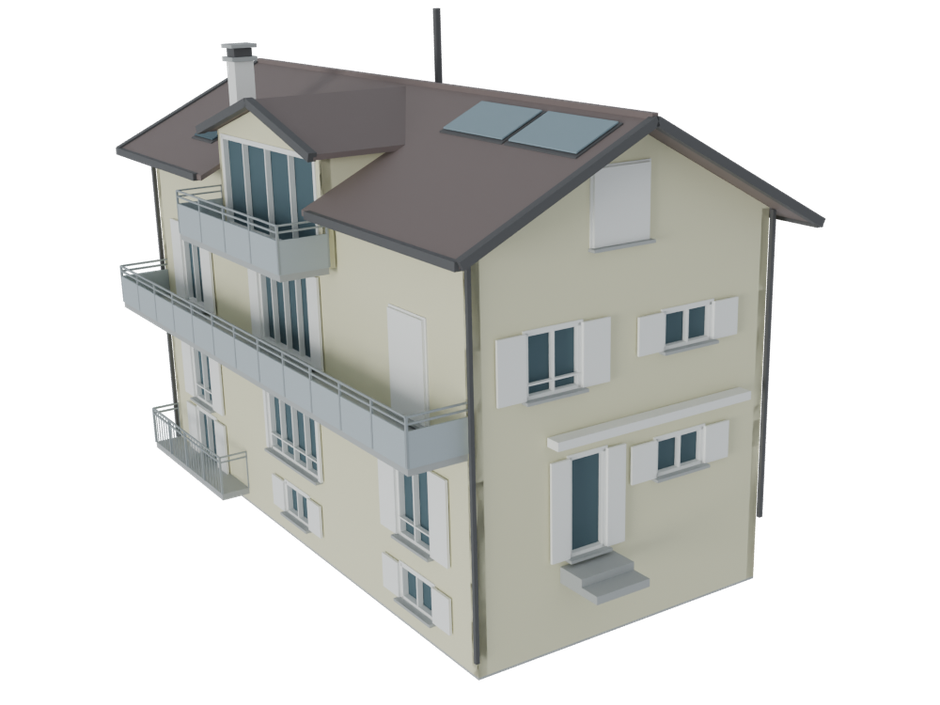} &
        \includegraphics[width=0.12\textwidth]{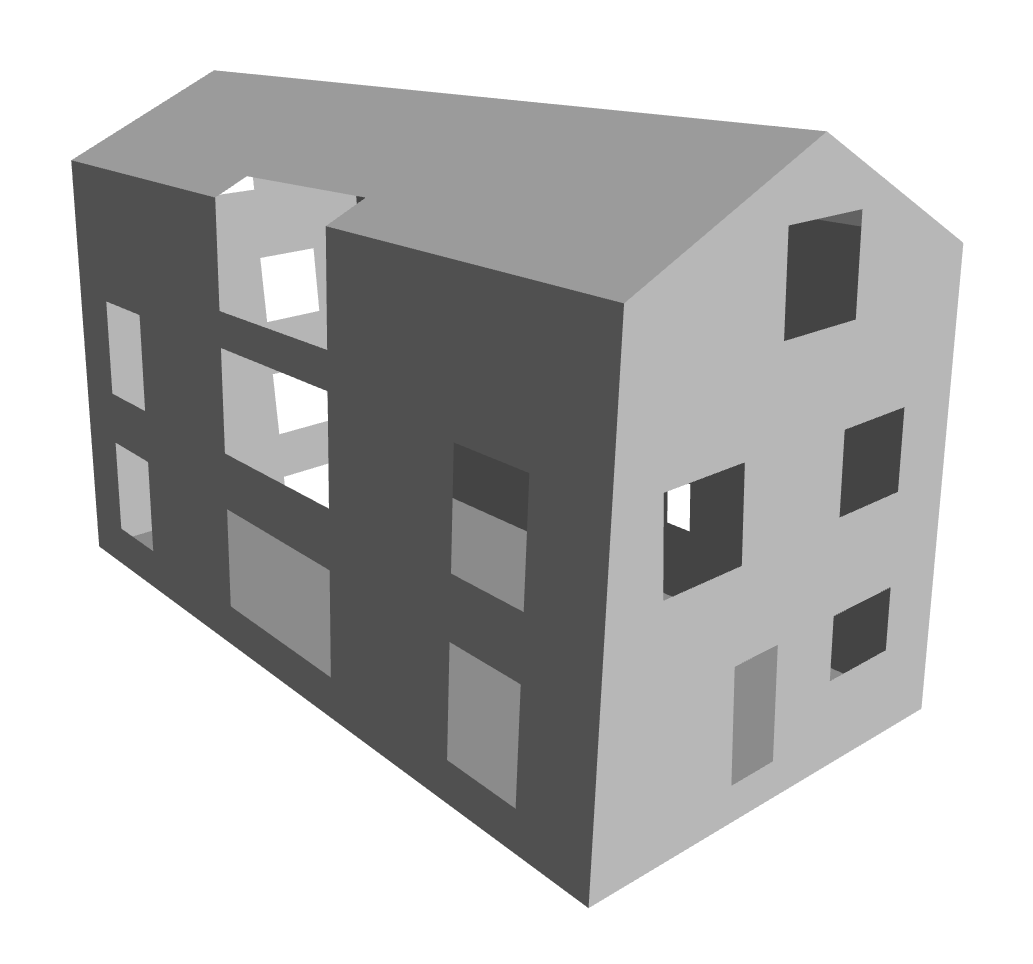} \\
        \multicolumn{3}{c}{\scriptsize\textbf{03}} &
        \multicolumn{3}{c}{\scriptsize\textbf{07}} \\[2mm]

        \includegraphics[width=0.15\textwidth]{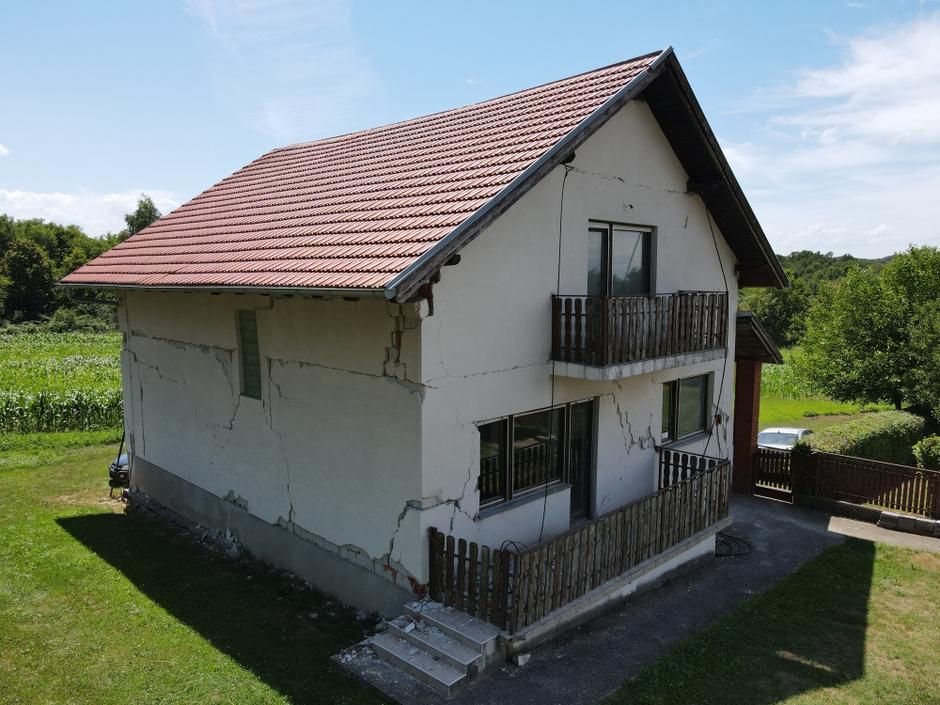} &
        \includegraphics[width=0.15\textwidth]{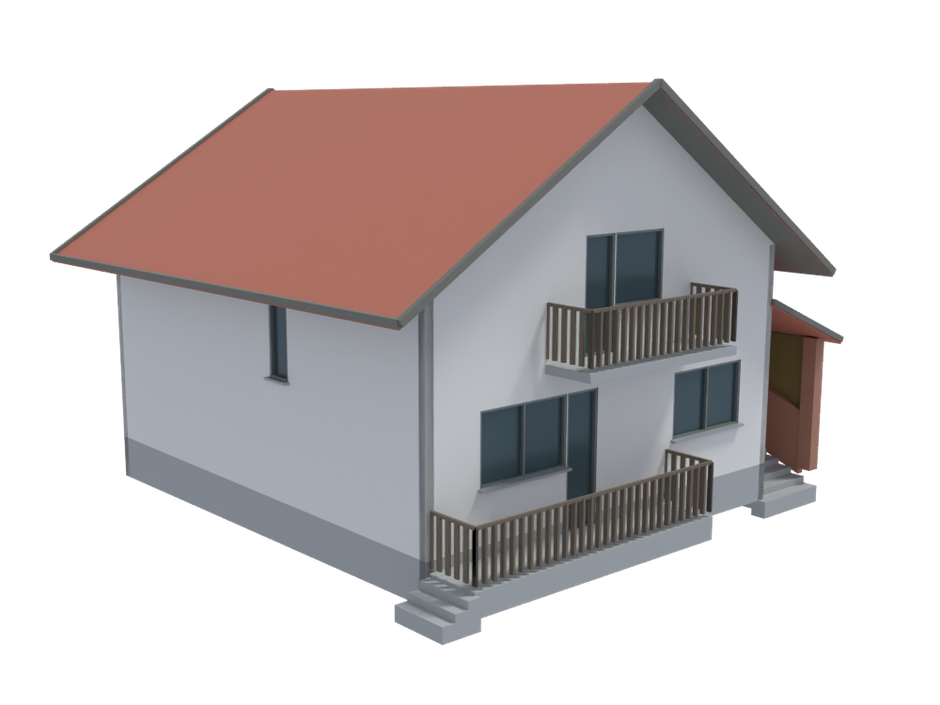} &
        \includegraphics[width=0.13\textwidth]{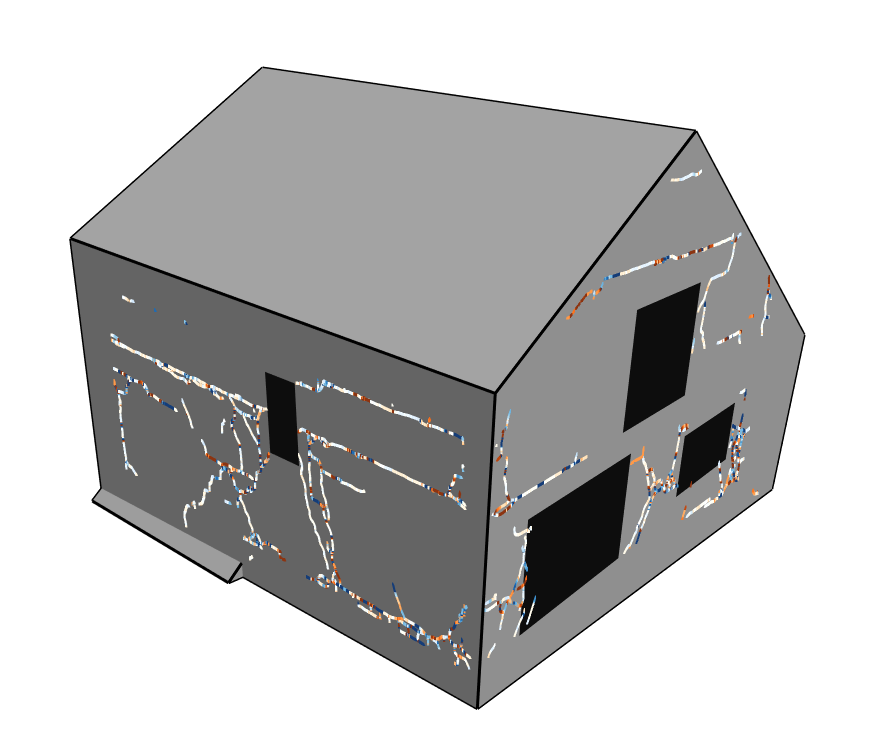} &
        \includegraphics[width=0.15\textwidth]{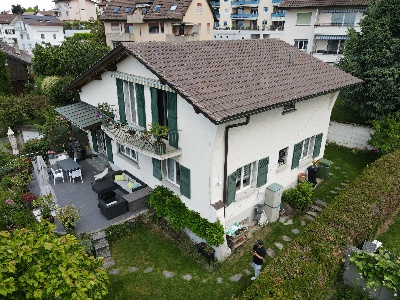} &
        \includegraphics[width=0.15\textwidth]{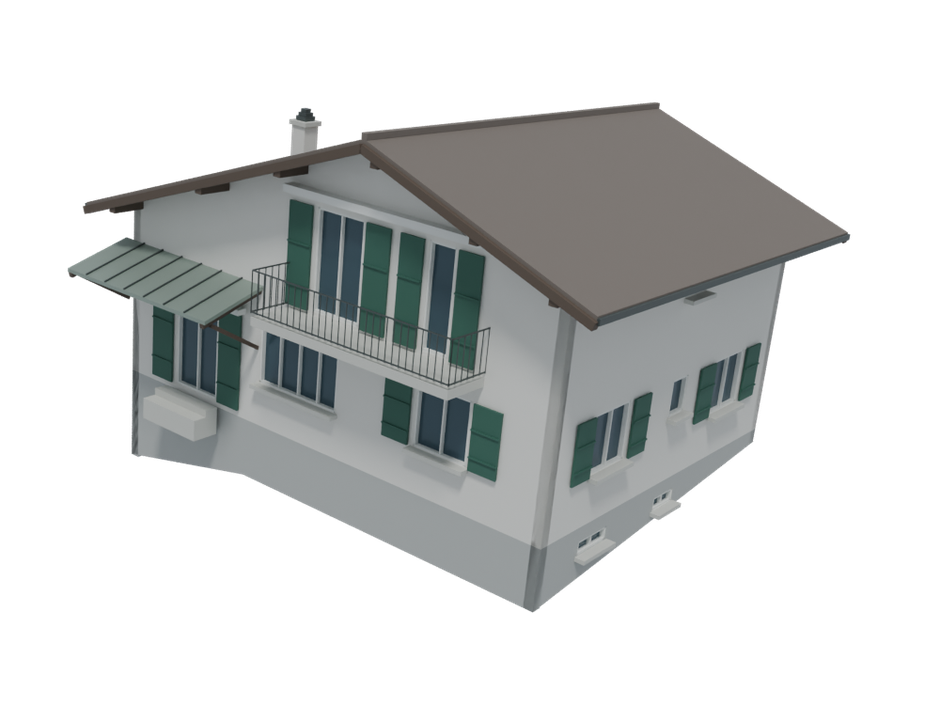} &
        \includegraphics[width=0.13\textwidth]{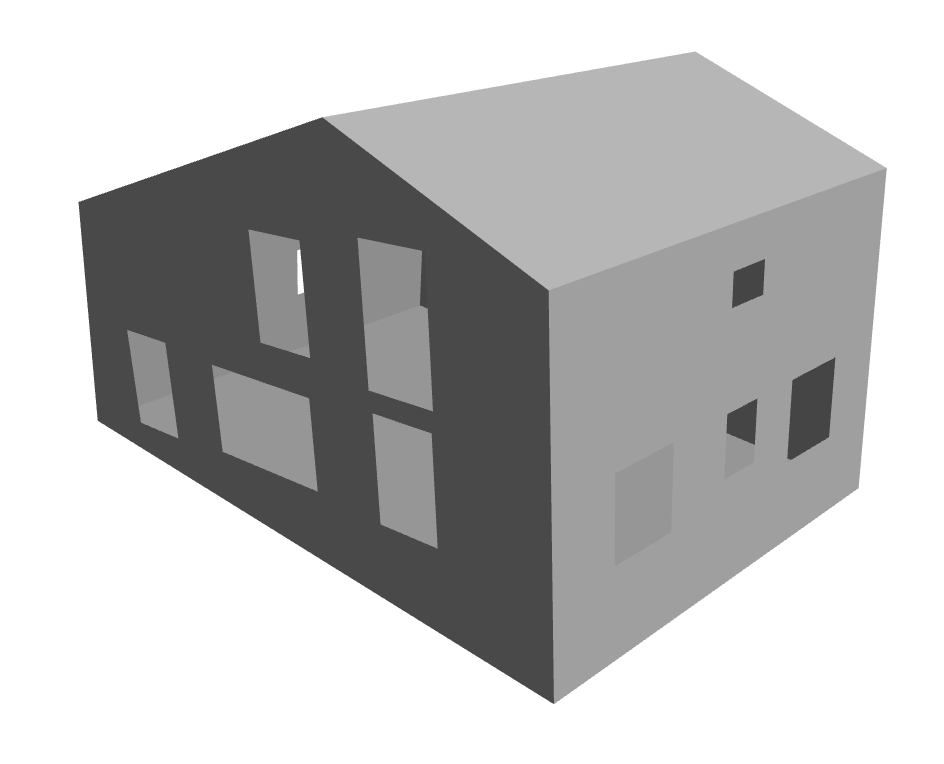} \\
        \multicolumn{3}{c}{\scriptsize\textbf{04}} &
        \multicolumn{3}{c}{\scriptsize\textbf{08}} \\[2mm]

        \includegraphics[width=0.15\textwidth]{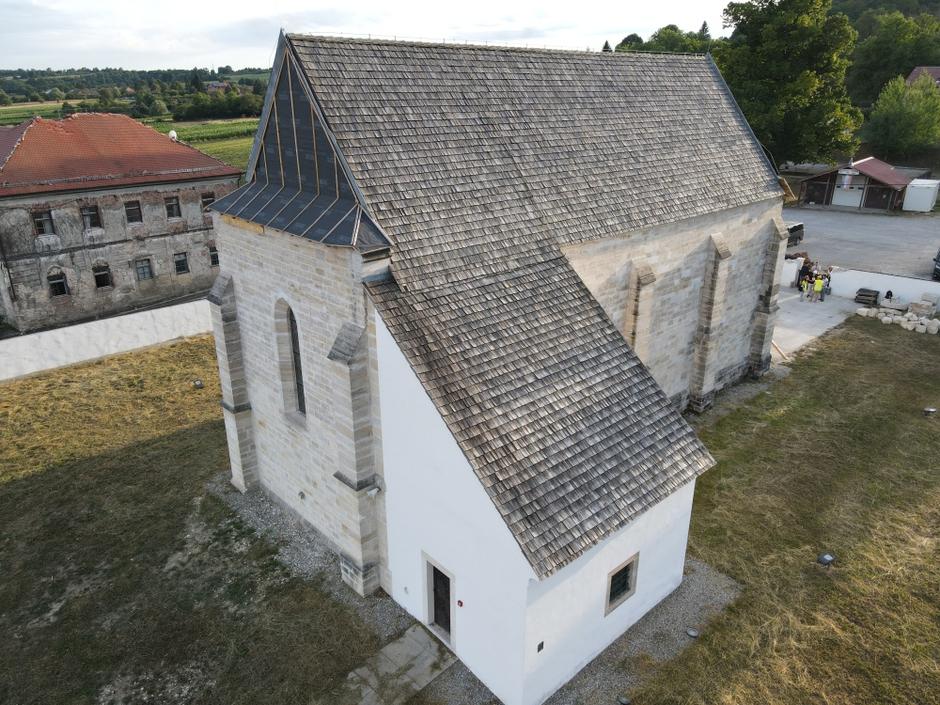} &
        \includegraphics[width=0.15\textwidth]{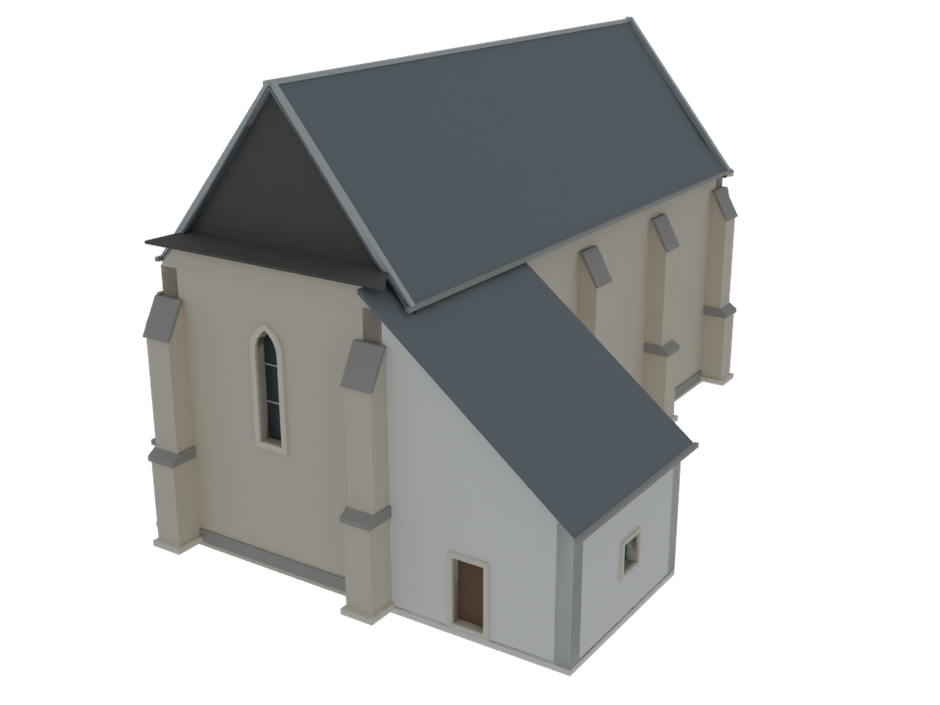} &
        \includegraphics[width=0.13\textwidth]{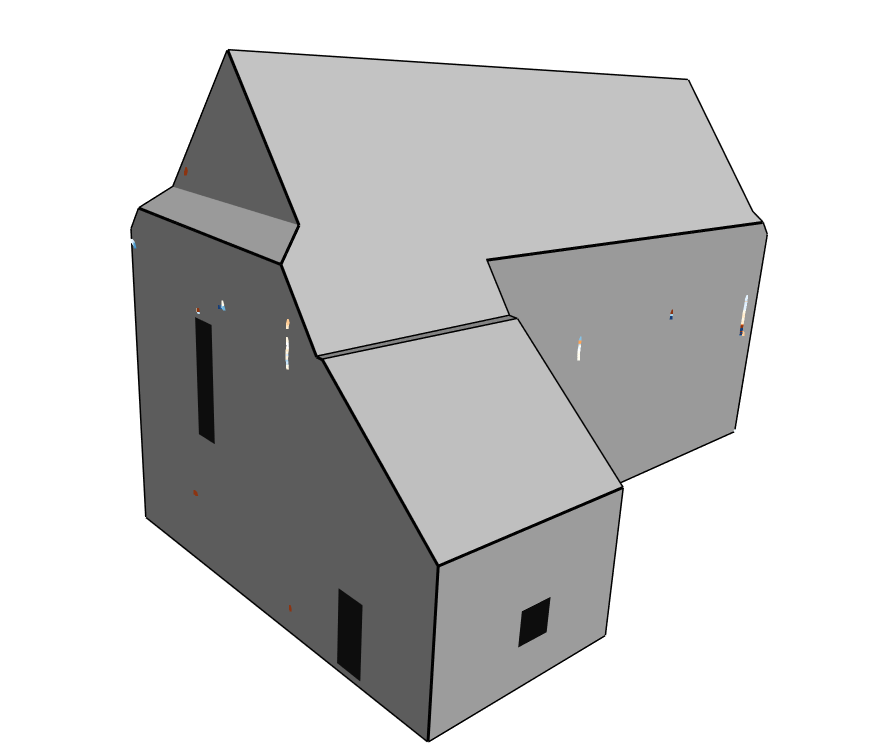} &
        \includegraphics[width=0.15\textwidth]{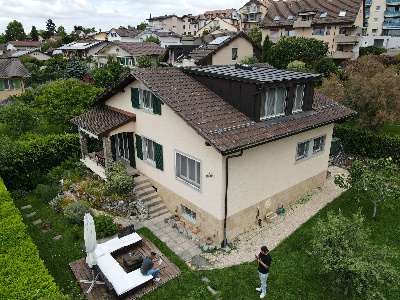} &
        \includegraphics[width=0.15\textwidth]{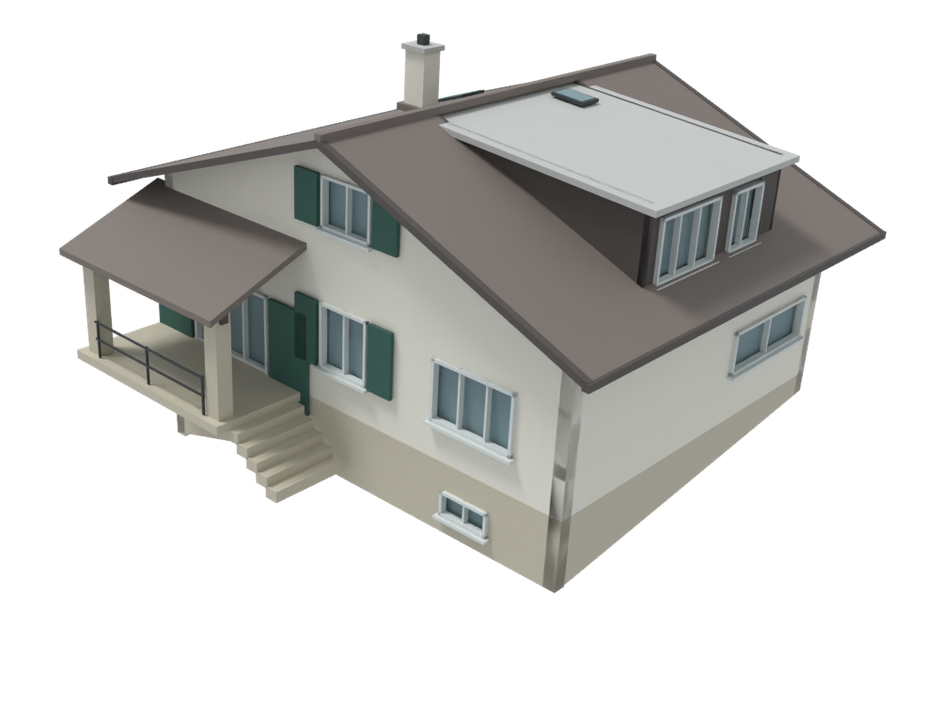} &
        \includegraphics[width=0.13\textwidth]{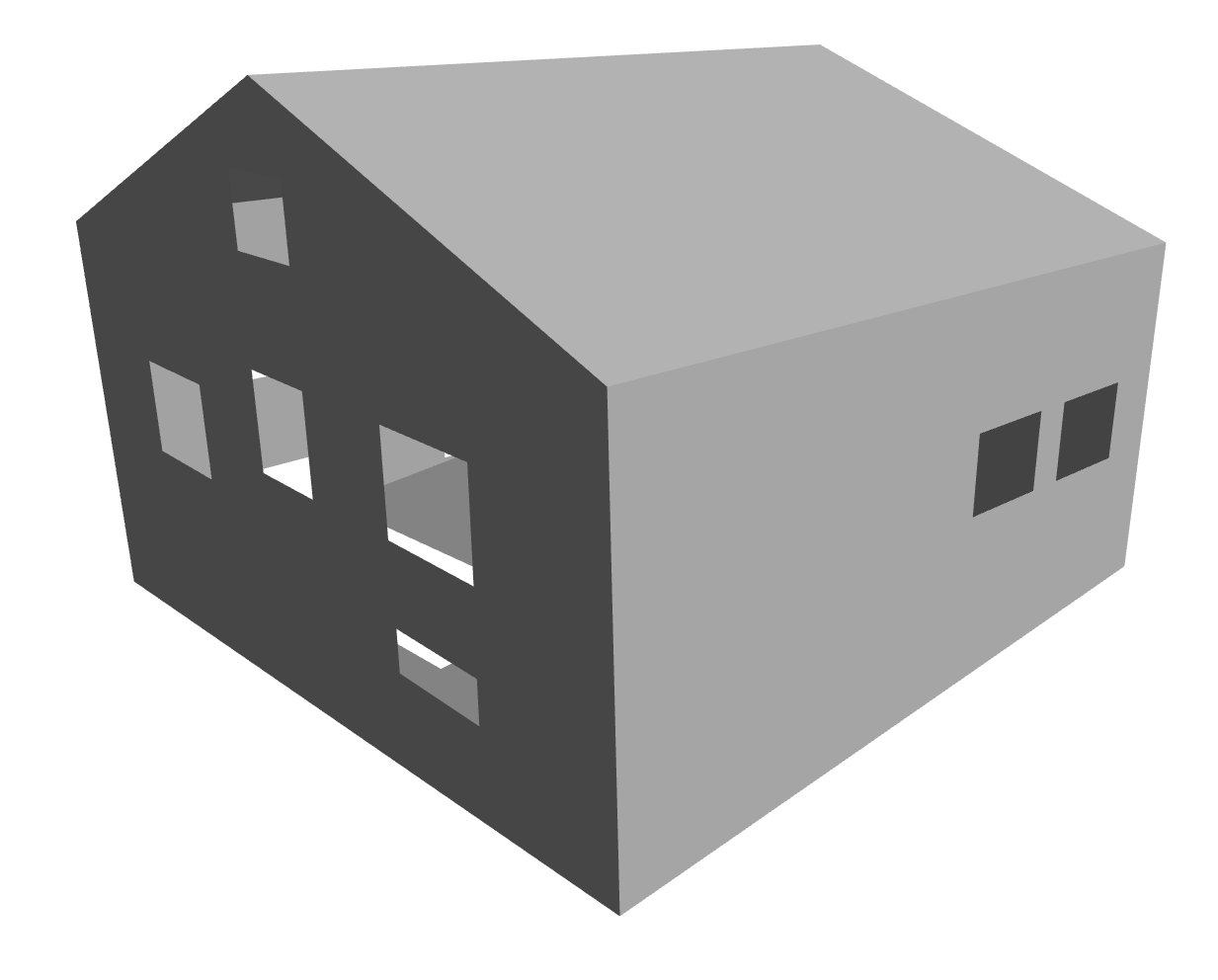} \\
        \multicolumn{3}{c}{\scriptsize\textbf{14}} &
        \multicolumn{3}{c}{\scriptsize\textbf{09}} \\[2mm]

        \includegraphics[width=0.15\textwidth]{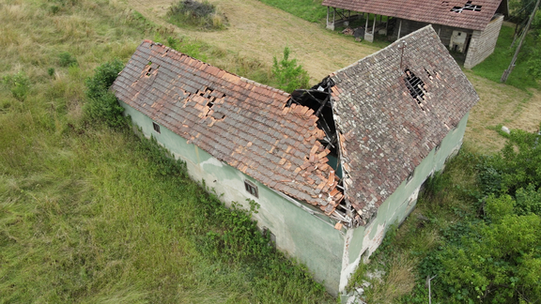} &
        \includegraphics[width=0.15\textwidth]{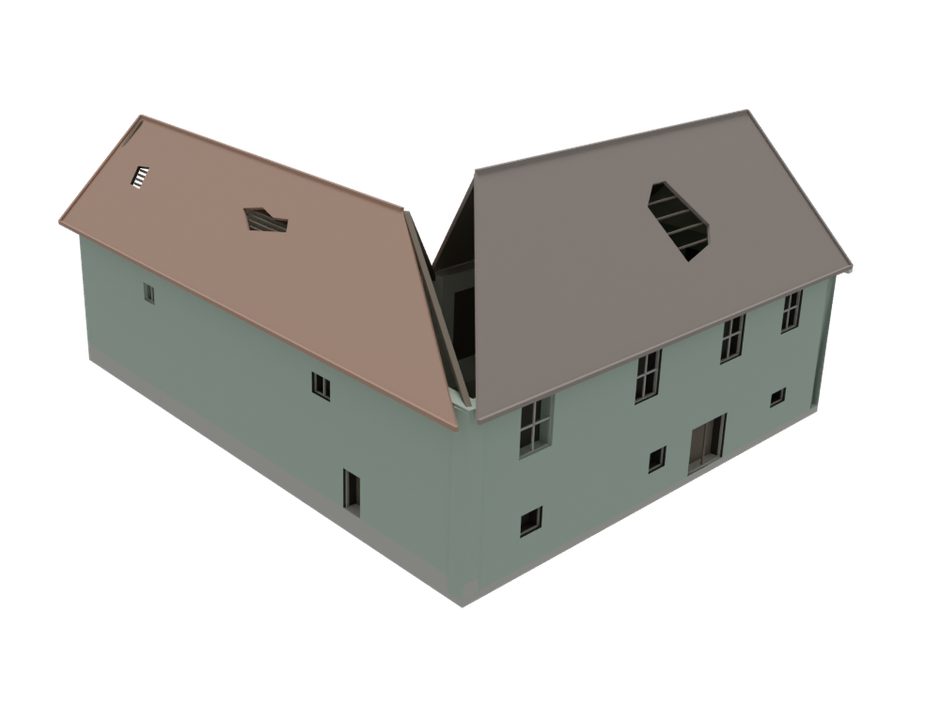} &
        \includegraphics[width=0.15\textwidth]{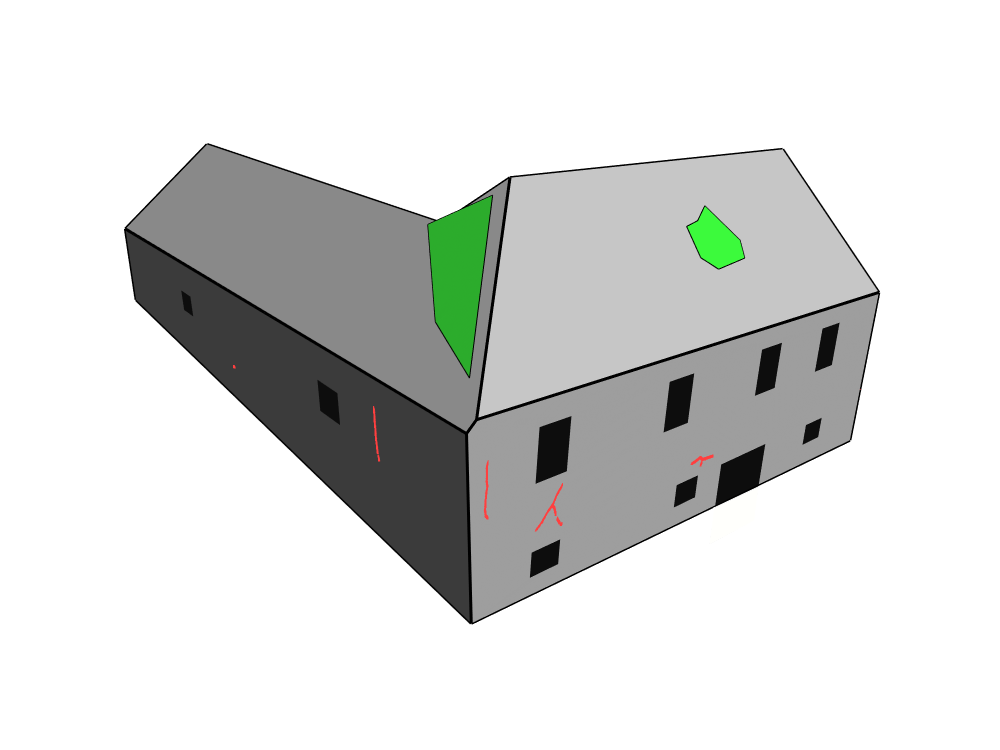} &
        \includegraphics[width=0.15\textwidth]{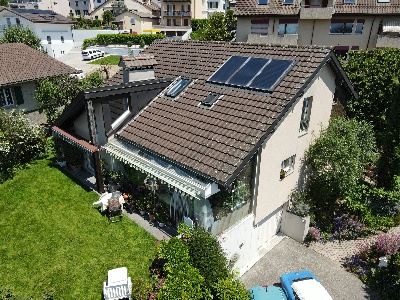} &
        \includegraphics[width=0.15\textwidth]{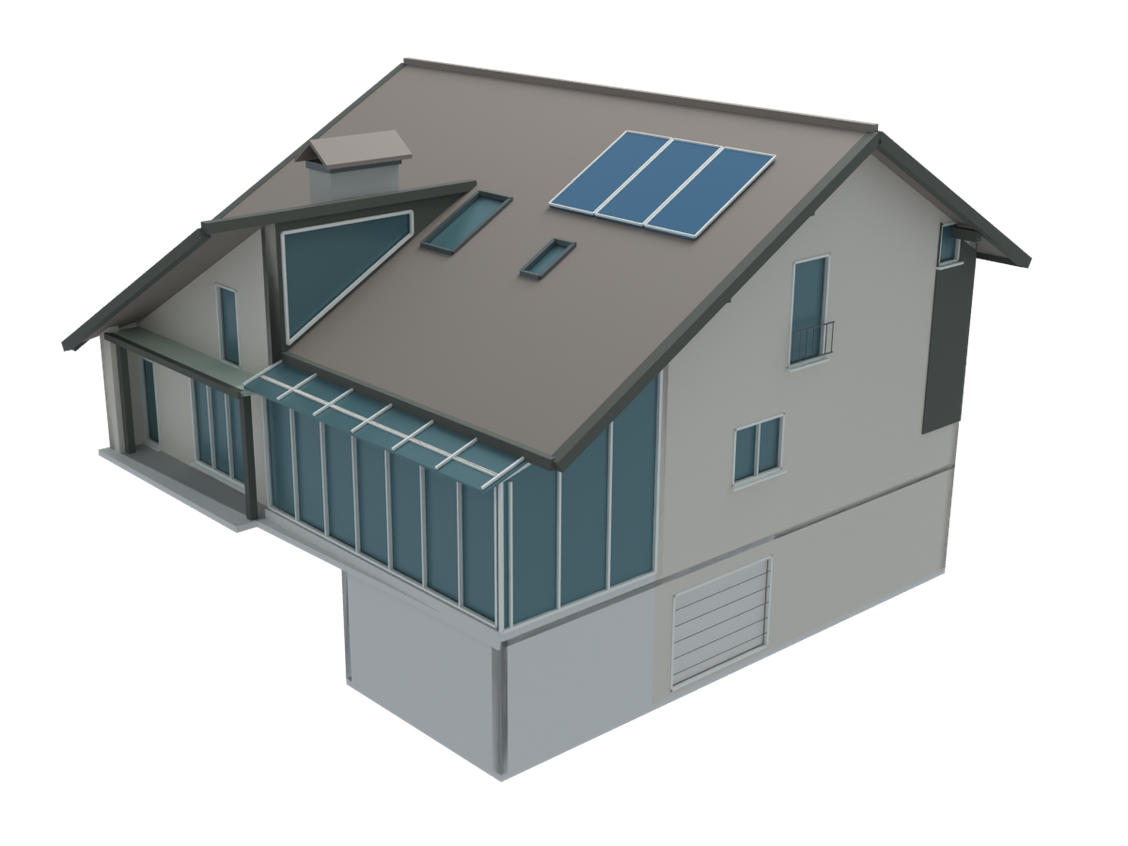} &
        \includegraphics[width=0.13\textwidth]{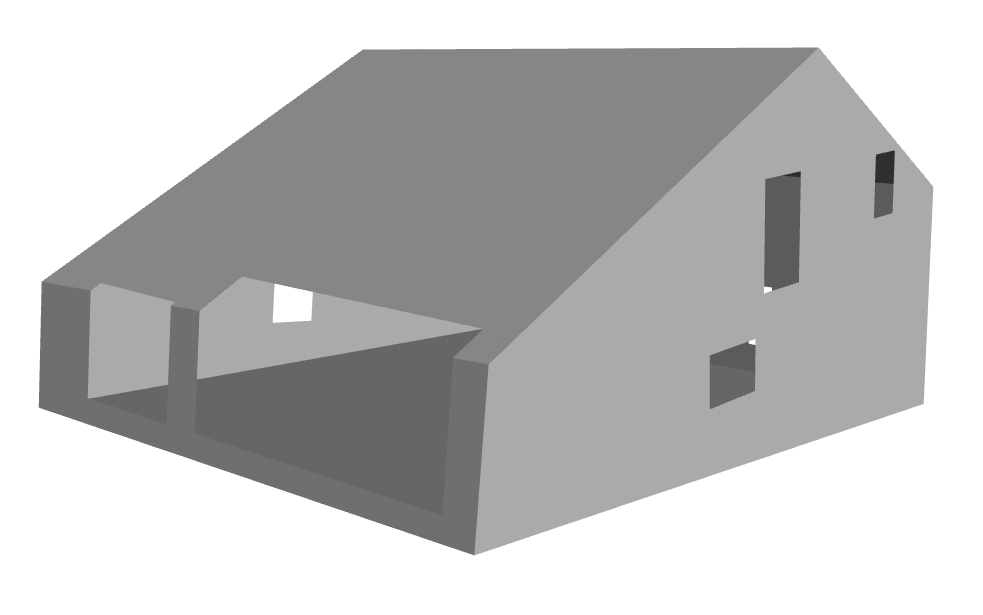} \\
        \multicolumn{3}{c}{\scriptsize\textbf{19}} &
        \multicolumn{3}{c}{\scriptsize\textbf{10}} \\[2mm]

        \includegraphics[width=0.15\textwidth]{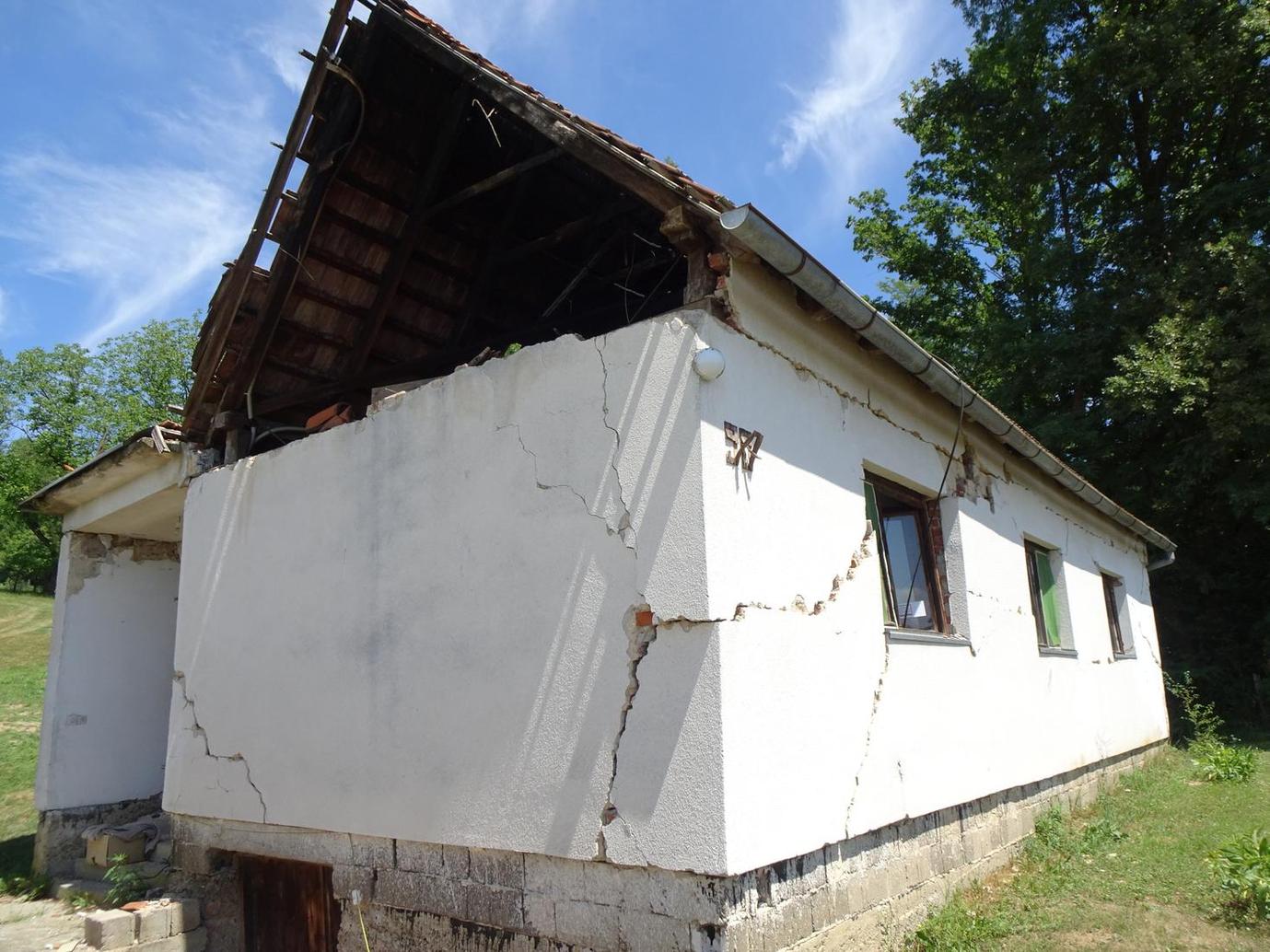} &
        \includegraphics[width=0.15\textwidth]{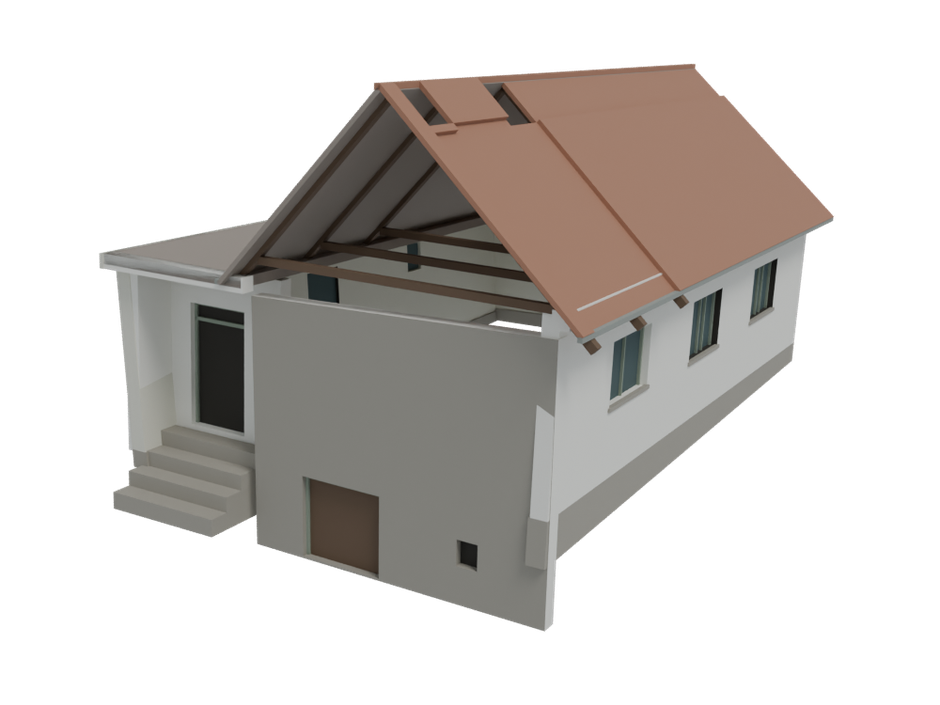} &
        \includegraphics[width=0.15\textwidth]{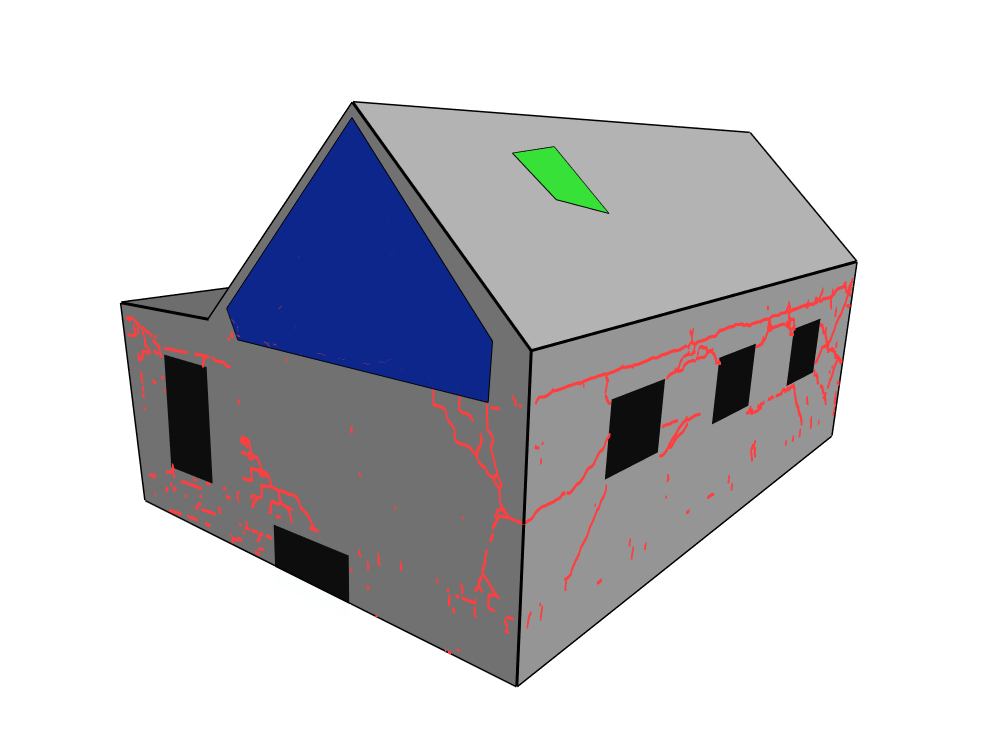} &
        \includegraphics[width=0.15\textwidth]{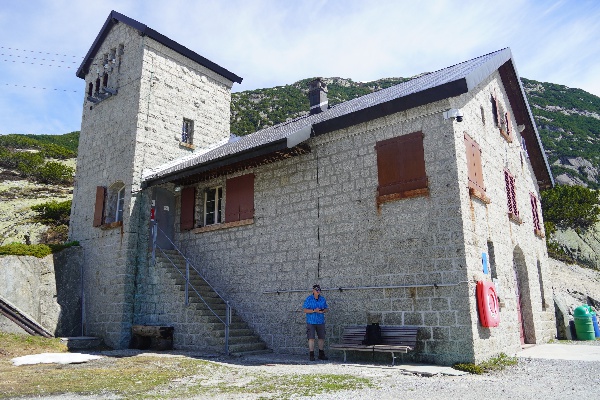} &
        \includegraphics[width=0.15\textwidth]{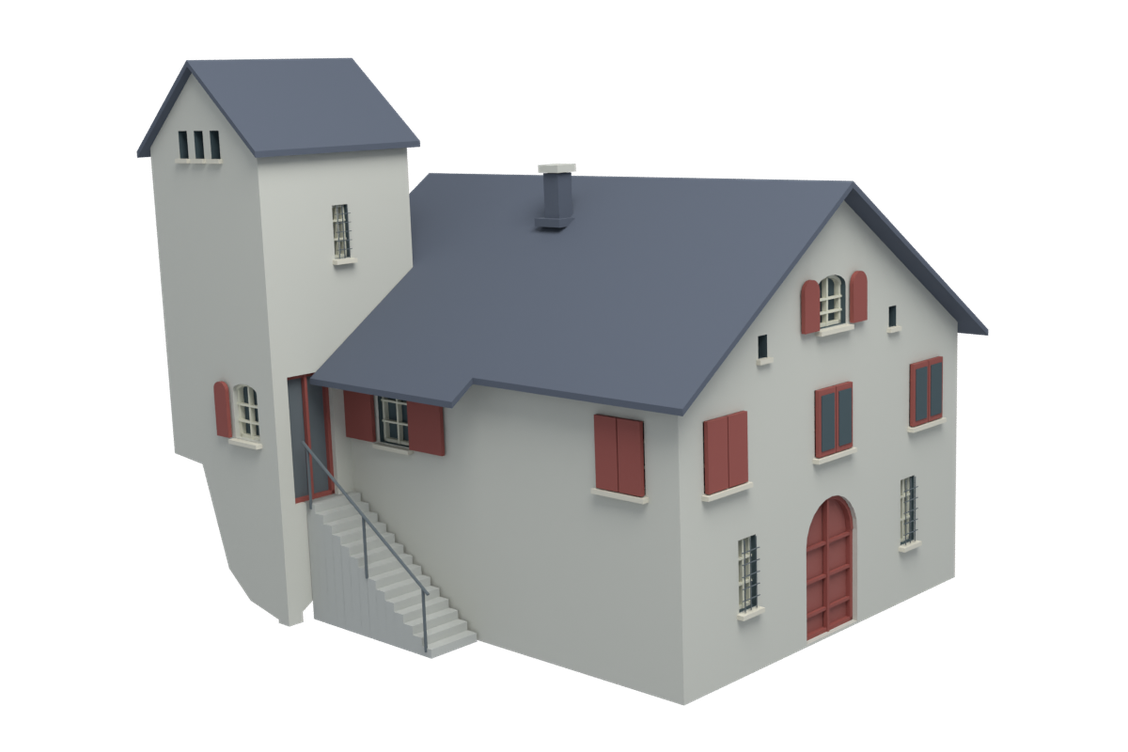} &
        \includegraphics[width=0.13\textwidth]{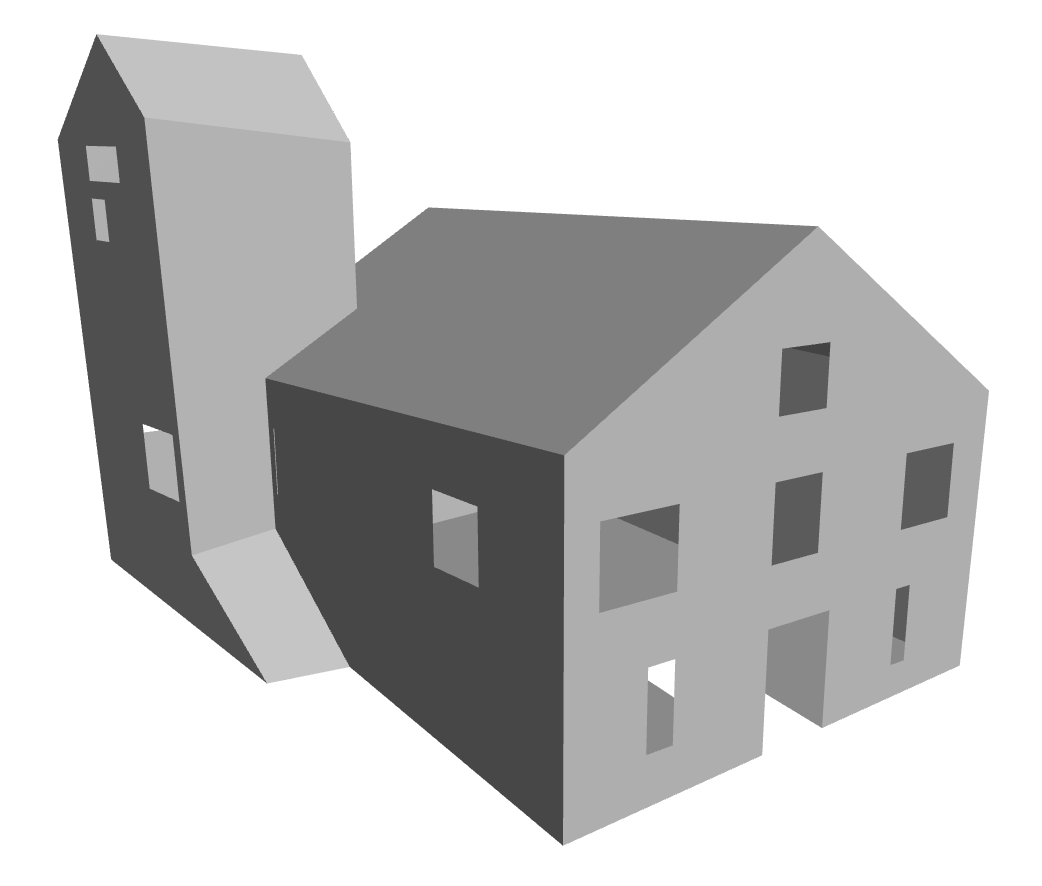} \\
        \multicolumn{3}{c}{\scriptsize\textbf{23}} &
        \multicolumn{3}{c}{\scriptsize\textbf{11}} \\
    \end{tabular}
    }
    \caption{Qualitative comparison of AstraLOD3 with the corresponding DADT~\cite{PantojaRosero2023DADT} reconstructions (left) and previous LOD3~\cite{PantojaRosero2022LOD3} reconstructions (right). For each building, a representative source image is shown together with the AstraLOD3 and previous reconstruction.}
    \label{fig:previous_work_comparison}
\end{figure}

The comparison shows stronger image-space agreement overall. AstraLOD3 obtains a higher FRDS for nearly all overlapping cases, while the geometric metrics remain more mixed. The only exception in FRDS is Ozcan, for which the two values are almost identical (0.9126 and 0.9150). As illustrated in Figure~\ref{fig:previous_work_comparison}, AstraLOD3 reconstructs several architectural elements at a finer level of geometric detail than the previous models, including balconies, railings, roof elements, openings, chimneys, and local façade components. In cases such as Ozcan, these thin or open components can introduce small differences in the projected binary mask compared with the more compact representation used in the previous reconstruction, even when the resulting geometry is visually consistent with the source images.

The point-cloud-based metrics require a more careful interpretation. In several cases, AstraLOD3 provides geometric agreement comparable to or better than the previous reconstructions, while in others the previous models obtain lower distances. Part of this variation arises from differences in both the reconstructed representation and the point clouds used for evaluation. In the previous studies, the SfM point clouds were manually cleaned to retain observations closely associated with the building, whereas the present workflow uses automatic statistical and spatial filtering. Although this removes most surrounding observations without manual intervention, residual ground, vegetation, clutter, damaged material, or other non-building points can remain and contribute to the measured distances. For IMF, the inlier criterion reduces the influence of the larger-distance tail but does not completely eliminate this effect.

The symmetric Chamfer distance is additionally sensitive to architectural surfaces reconstructed from the images but weakly represented in the sparse SfM cloud. Components such as chimneys, roof regions, balconies, railings, or other detailed elements can therefore increase the model-to-cloud term despite being supported by the visual evidence. The relatively small differences observed for many cases should consequently be interpreted in the context of both the richer geometric representation produced by AstraLOD3 and the less manually curated point-cloud preprocessing used in the present workflow.

The quantitative comparison should therefore be interpreted together with the qualitative results in Figure~\ref{fig:previous_work_comparison}. The previous LOD3 models generally provide more compact representations of the main building envelope, while AstraLOD3 recovers a larger number of image-supported architectural components. A similar trend is visible in the DADT cases, where AstraLOD3 reconstructs the principal geometry together with additional architectural detail while also preserving irregular or damaged building configurations when these are supported by the observations. Particularly favorable geometric agreement is obtained in cases such as Petrinja school, UNIL, and Gerlmerbahn, whereas other cases illustrate the expected trade-off between geometric detail and support from a sparse point cloud.

Overall, the comparison indicates that AstraLOD3 achieves comparable geometric agreement and generally stronger image-space agreement than the previous purpose-built approaches, while recovering a more detailed architectural representation in several of the evaluated cases. Importantly, these models are obtained without relying on the task-specific reconstruction procedures developed for the previous LOD3 and DADT pipelines.

\subsection{Effect of reconstruction-specification guidance}
\label{sec:guidance_results}

The effect of the guidance encoded in $\mathcal{R}$ is examined using the \emph{Normal}, \emph{Short}, and \emph{Nano} specifications while retaining the same Gerlmerbahn evidence $\mathcal{I}_{\mathrm{VCP}}$ and Astra \emph{High} model configuration. As described in Section~\ref{sec:controlled_experiments}, the \emph{Normal} specification provides the complete reconstruction requirements together with the prescribed six-stage workflow. The \emph{Short} specification retains the reconstruction objective, evidence and representation rules, camera constraints, and output requirements, but removes the explicit workflow. The \emph{Nano} specification further reduces $\mathcal{R}$ to the reconstruction task, available inputs, principal execution restrictions, and required outputs, leaving most geometric, semantic, and procedural decisions to the agent.

\begin{table}[H]
\centering
\caption{Effect of the guidance encoded in the reconstruction specification $\mathcal{R}$ on the Gerlmerbahn reconstruction.}
\label{tab:guidance_results}
\small
\begin{tabular}{lcccc}
\hline
\textbf{Specification} & \textbf{Time} &
\textbf{FRDS $\uparrow$} &
\textbf{$D_{\mathrm{CD}}\downarrow$} &
\textbf{IMF $\downarrow$} \\
\hline
\emph{Normal} & \textbf{20:00} & 0.9556 & 0.0161 & 0.0177 \\
\emph{Short}  & 28:11 & \textbf{0.9752} & \textbf{0.0070} & \textbf{0.0144} \\
\emph{Nano}   & 20:57 & 0.9108 & 0.0400 & 0.0191 \\
\hline
\end{tabular}
\end{table}

The results show that reconstruction performance does not vary monotonically with the amount of guidance provided through $\mathcal{R}$. The \emph{Short} specification achieves the strongest quantitative agreement, with the highest FRDS and the lowest $D_{\mathrm{CD}}$ and IMF, although it also requires the longest processing time. This result suggests that removing the prescribed workflow can provide useful flexibility when the reconstruction objective and geometric and evidence constraints remain clearly defined. In this run, the agent independently organized the reconstruction around plane fitting, calibrated image measurements, roof and opening refinement, and multi-view correction of the chimney and lower building boundaries. The final model retained open ground boundaries where the building was concealed by the slope rather than completing unsupported foundation geometry.

\begin{figure}[H]
    \centering
    \captionsetup[subfigure]{
        justification=centering,
        font=small,
        skip=2mm
    }

    \begin{subfigure}[t]{0.31\textwidth}
        \centering
        \includegraphics[width=\linewidth]
            {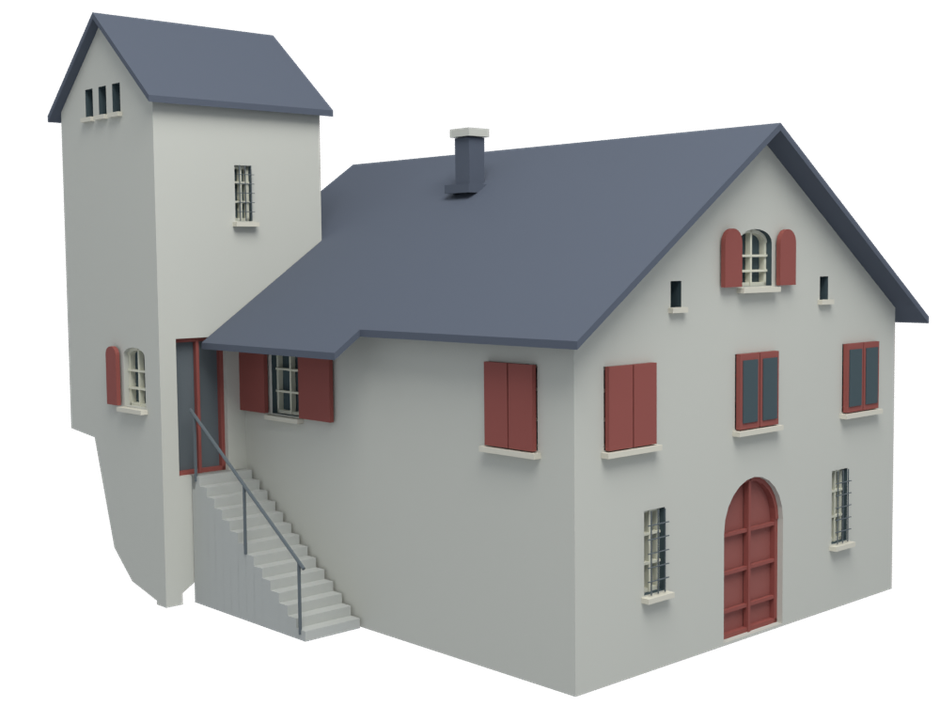}
        \par\vspace{-1mm}
        \includegraphics[width=\linewidth]
            {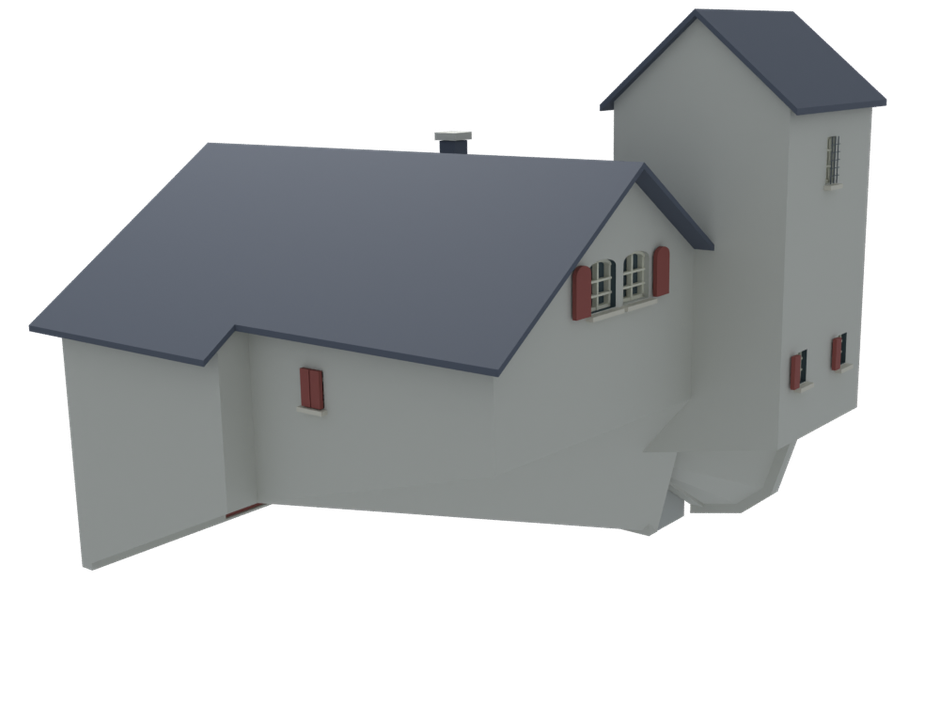}
        \caption*{\emph{Normal}}
        \label{fig:guidance_normal}
    \end{subfigure}
    \hfill
    \begin{subfigure}[t]{0.31\textwidth}
        \centering
        \includegraphics[width=\linewidth]
            {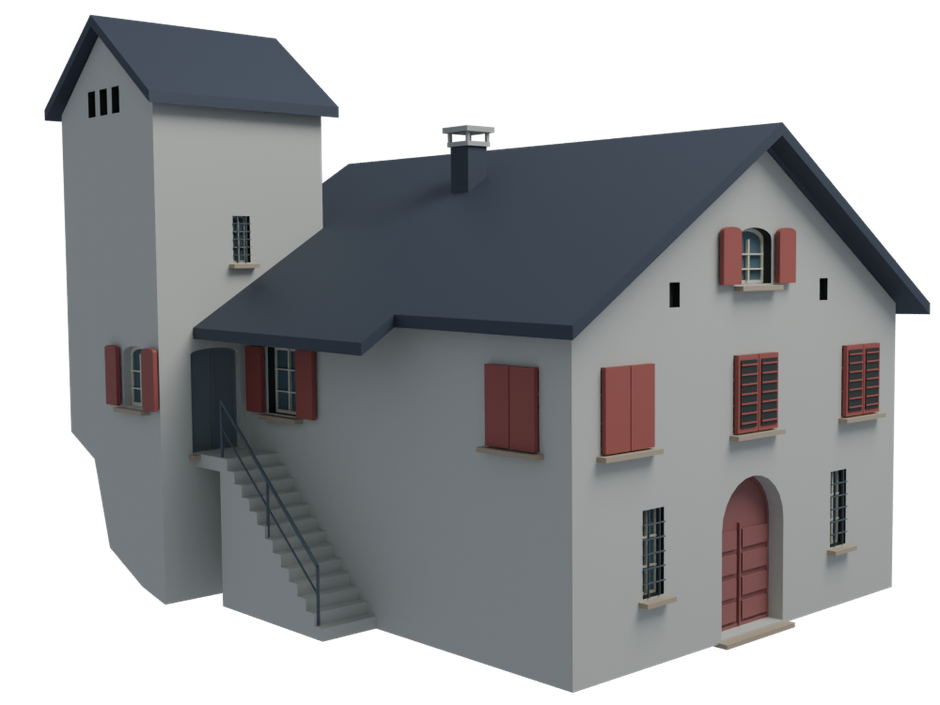}
        \par\vspace{-1mm}
        \includegraphics[width=\linewidth]
            {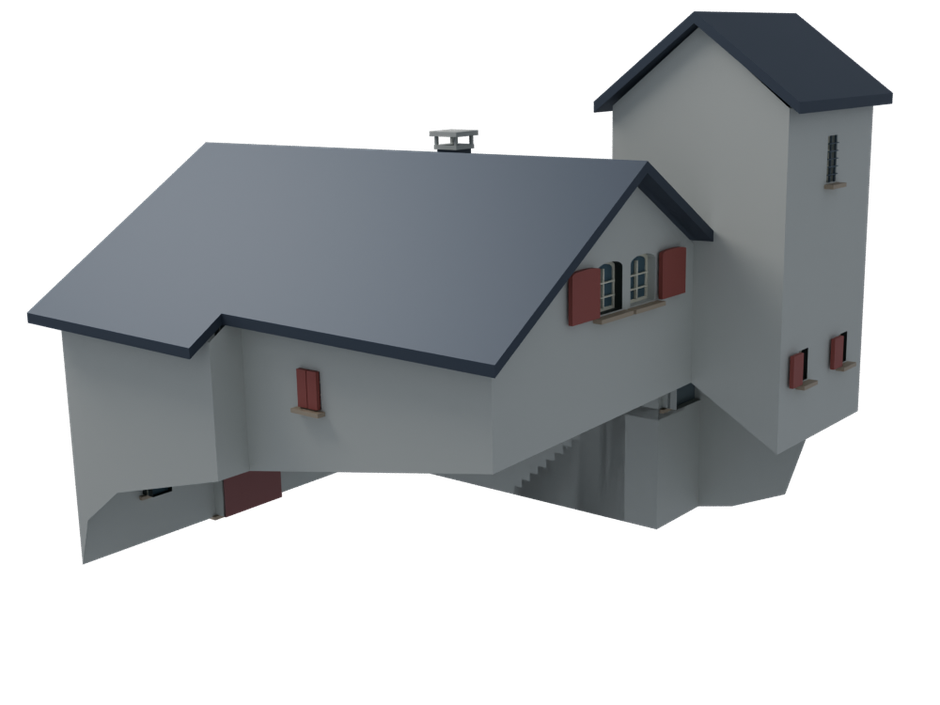}
        \caption*{\emph{Short}}
        \label{fig:guidance_short}
    \end{subfigure}
    \hfill
    \begin{subfigure}[t]{0.31\textwidth}
        \centering
        \includegraphics[width=\linewidth]
            {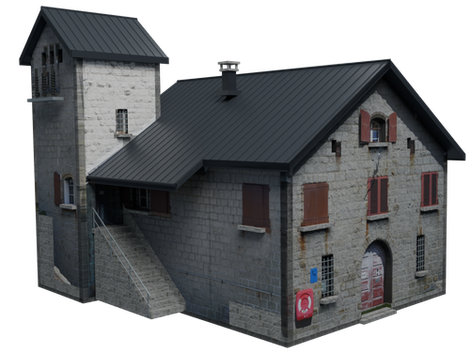}
        \par\vspace{-1mm}
        \includegraphics[width=\linewidth]
            {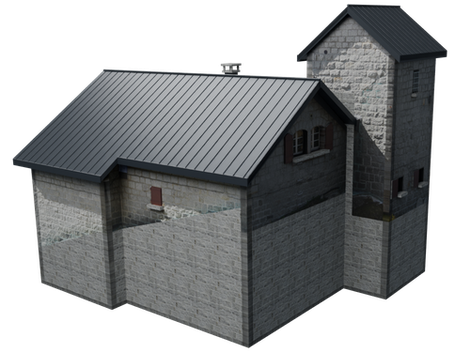}
        \caption*{\emph{Nano}}
        \label{fig:guidance_nano}
    \end{subfigure}

    \caption{Effect of the guidance encoded in $\mathcal{R}$ on the Gerlmerbahn reconstruction. \emph{Normal}, \emph{Short}, and \emph{Nano} are shown from the same two viewpoints using the same reconstruction evidence and Astra \emph{High} model configuration. Visual appearance is retained as generated by each run; the \emph{Nano} reconstruction independently introduced source-derived façade textures.}
    \label{fig:guidance_ablation}
\end{figure}

The \emph{Nano} experiment illustrates a different effect of increased autonomy. With most geometric and procedural guidance removed, the agent still reconstructed the main building and architectural components, but it also made additional representation choices not prescribed in the \emph{Normal} or \emph{Short} specifications. In particular, it generated source-derived façade textures and included additional geometric interpretation of concealed lower portions of the building. The photographic appearance visible in Figure~\ref{fig:guidance_ablation} therefore emerged from the agent's own interpretation of the reconstruction task rather than from an explicit texture-generation requirement.

The additional freedom does not necessarily improve the evaluation metrics. In the \emph{Nano} reconstruction, a shallow lower closure and rear wall surfaces were introduced in regions concealed by the rising terrain. Such completion can be architecturally plausible while reducing agreement with the evaluation evidence: surfaces visible in the rendered model but occluded in the source images can reduce FRDS, while reconstructed surfaces with little or no support in the sparse point cloud can increase the model-to-cloud contribution of the symmetric Chamfer distance. The comparatively small change in IMF is consistent with its one-sided formulation, which is less directly affected by additional model surfaces.

The three experiments therefore indicate that increasing agent autonomy changes not only the computational procedure but also how the reconstruction problem itself is interpreted. The \emph{Short} specification appears to provide a useful balance for this case: it preserves the geometric and evidence constraints of the full specification while allowing the agent to determine its own reconstruction procedure. The \emph{Nano} case demonstrates that substantially weaker guidance can still produce a plausible and richer representation, but also allows reasonable inferred content that is not necessarily favored by the selected evaluation metrics. 

Figure~\ref{fig:guidance_ablation} shows the corresponding reconstructions from two common viewpoints.

\subsection{Effect of reconstruction evidence}
\label{sec:evidence_results}

Table~\ref{tab:evidence_results} reports the Gerlmerbahn reconstructions obtained using the six evidence configurations defined in Eq.~\ref{eq:evidence_configurations}. The complete configuration $\mathcal{I}_{\mathrm{VCP}}$ provides the strongest overall quantitative agreement, with the highest FRDS and the lowest symmetric Chamfer distance. The lowest IMF is obtained with $\mathcal{I}_{\mathrm{CP}}$, although its value is close to that of the complete configuration.

\begin{table}[H]
\centering
\caption{Effect of the reconstruction evidence available to the Astra agent for Building~11. A dash indicates that the image-only reconstruction does not share the SfM coordinate system and scale required for direct quantitative comparison.}
\label{tab:evidence_results}
\small
\begin{tabular}{lcccc}
\hline
\textbf{Evidence} & \textbf{Time} &
\textbf{FRDS $\uparrow$} &
\textbf{$D_{\mathrm{CD}}\downarrow$} &
\textbf{IMF $\downarrow$} \\
\hline
$\mathcal{I}_{\mathrm{VCP}}$ & 20:00 & \textbf{0.9556} & \textbf{0.0161} & 0.0177 \\
$\mathcal{I}_{\mathrm{VC}}$  & 22:26 & 0.9123 & 0.0412 & 0.0184 \\
$\mathcal{I}_{\mathrm{VP}}$  & 17:28 & 0.8869 & 0.0447 & 0.0186 \\
$\mathcal{I}_{\mathrm{CP}}$  & 22:13 & 0.9354 & 0.0198 & \textbf{0.0164} \\
$\mathcal{I}_{\mathrm{P}}$   & 15:27 & 0.8759 & 0.0393 & 0.0167 \\
$\mathcal{I}_{\mathrm{V}}$   & \textbf{14:28} & --     & --     & --     \\
\hline
\end{tabular}
\end{table}

The quantitative and qualitative results show complementary roles for the three evidence modalities. Photographs provide important information for identifying architectural components and finer details, whereas the sparse point cloud directly constrains the three-dimensional building geometry. Accordingly, the reconstructions containing $\mathcal{V}$ generally preserve features such as the chimney, shutters, exterior stair, and more detailed window and door configurations. When photographs are unavailable, as in $\mathcal{I}_{\mathrm{CP}}$ and $\mathcal{I}_{\mathrm{P}}$, the principal envelope, tower, roof configuration, and several openings are still recovered, but the resulting models are more simplified. The point-cloud-only case is particularly notable: despite relying only on sparse geometric evidence, the agent recovers the main building configuration and several architectural openings, while omitting components such as the chimney for which sufficient geometric support is not found. 

Calibrated cameras provide the link between visual observations and the three-dimensional reference frame. This is particularly evident for $\mathcal{I}_{\mathrm{VC}}$: in the absence of a supplied point cloud, according the output report, the agent generated image correspondences and triangulated new three-dimensional observations from the calibrated photographs before fitting the building geometry. In contrast, $\mathcal{I}_{\mathrm{VP}}$ provides both photographs and sparse geometry but no calibrated relationship between them. The agent can therefore use the images to interpret architectural components and the point cloud to constrain the main geometry, but cannot directly transfer image observations into the SfM coordinate system. This helps explain why $\mathcal{I}_{\mathrm{VP}}$ remains visually detailed while producing lower FRDS and a larger $D_{\mathrm{CD}}$ than the complete configuration.

The behavior of the metrics is also consistent with the effects observed in Sections~\ref{sec:comparison_results} and~\ref{sec:guidance_results}. Architectural components recovered from the images can be weakly represented in the sparse point cloud and therefore increase the model-to-cloud contribution of $D_{\mathrm{CD}}$. In addition, some runs infer portions of the building hidden by the surrounding slope, while others terminate the geometry closer to the directly observed envelope. Such inferred surfaces can be architecturally plausible but are not represented in the visible image masks and may have little or no support in the sparse cloud, thereby reducing FRDS or increasing $D_{\mathrm{CD}}$. Conversely, the simpler $\mathcal{I}_{\mathrm{CP}}$ and $\mathcal{I}_{\mathrm{P}}$ reconstructions remain close to the dominant point-supported surfaces and consequently obtain low IMF values despite containing fewer architectural details. 

\begin{figure}[H]
    \centering
    \captionsetup[subfigure]{
        justification=centering,
        font=small,
        skip=2mm
    }

    \begin{subfigure}[t]{0.31\textwidth}
        \centering
        \includegraphics[width=\linewidth]
            {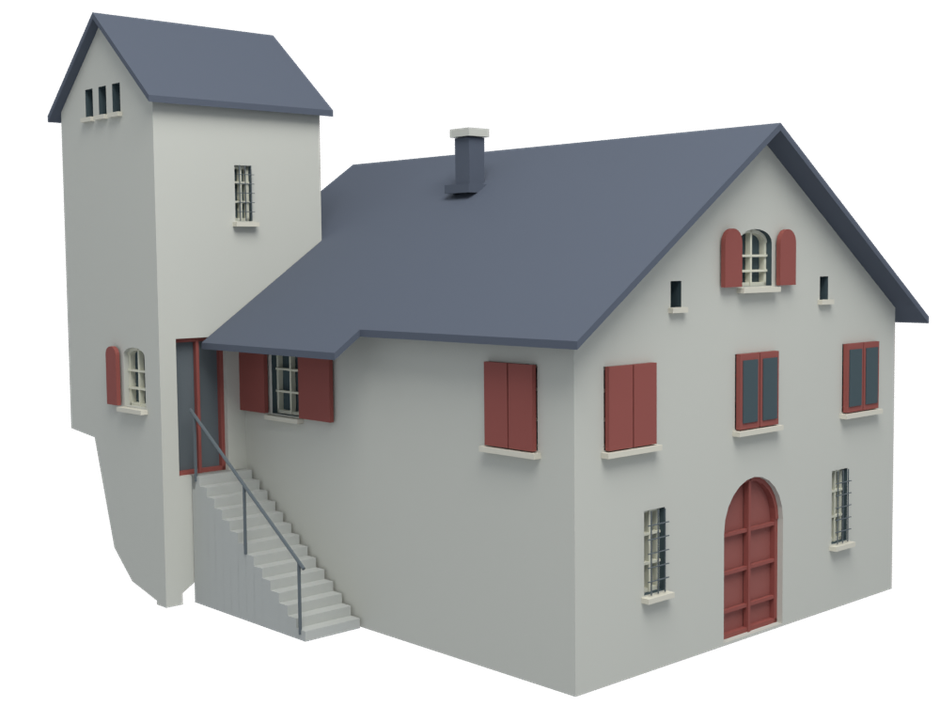}
        \par\vspace{-1mm}
        \includegraphics[width=\linewidth]
            {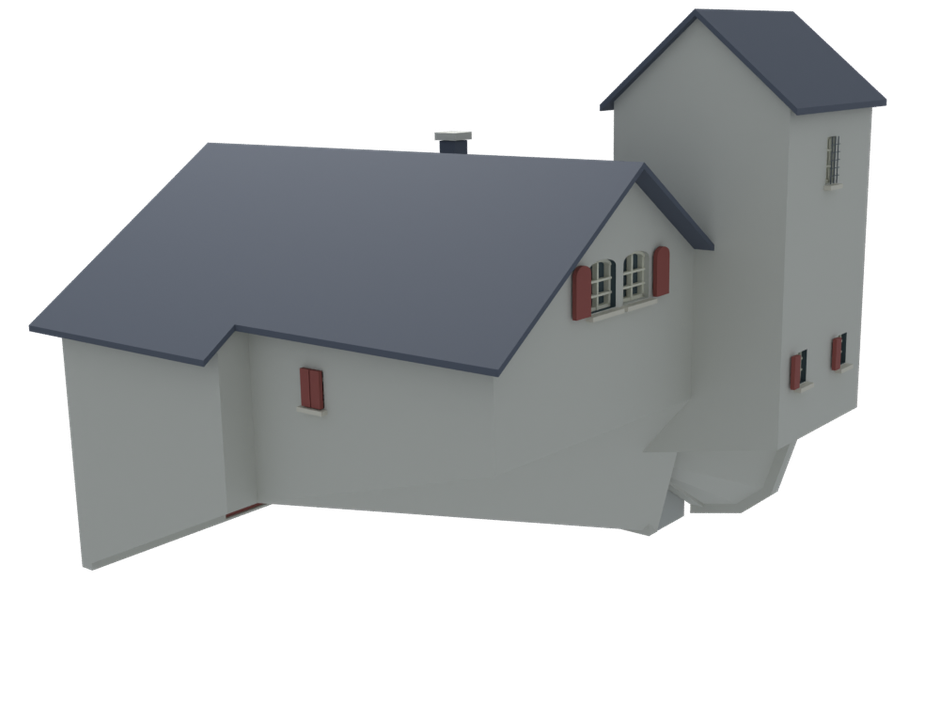}
        \caption*{$\mathcal{I}_{\mathrm{VCP}}$}
        \label{fig:evidence_vcp}
    \end{subfigure}
    \hfill
    \begin{subfigure}[t]{0.31\textwidth}
        \centering
        \includegraphics[width=\linewidth]
            {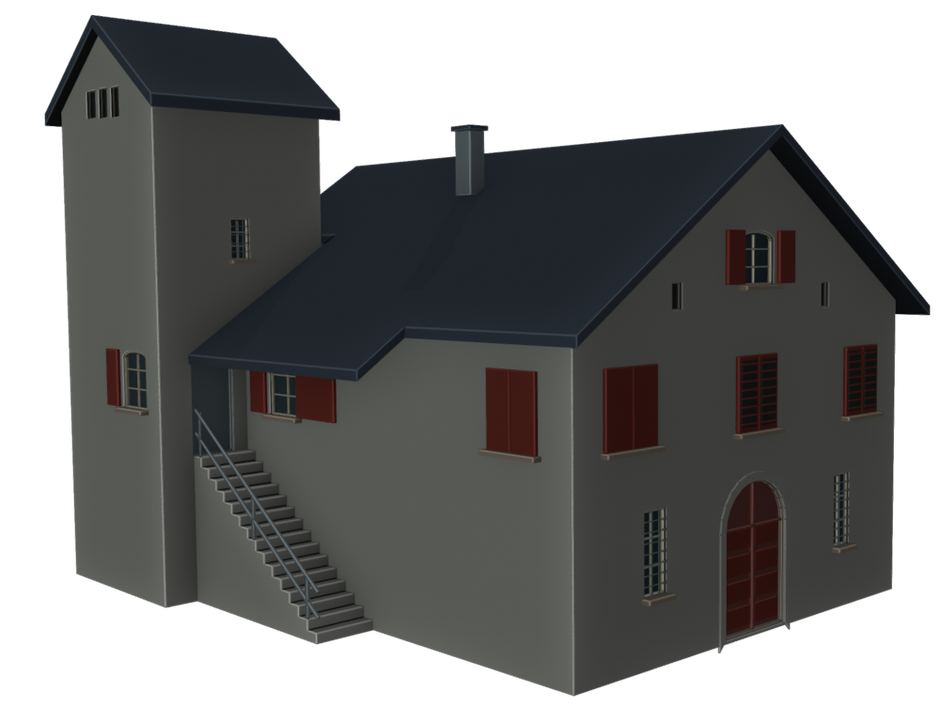}
        \par\vspace{-1mm}
        \includegraphics[width=\linewidth]
            {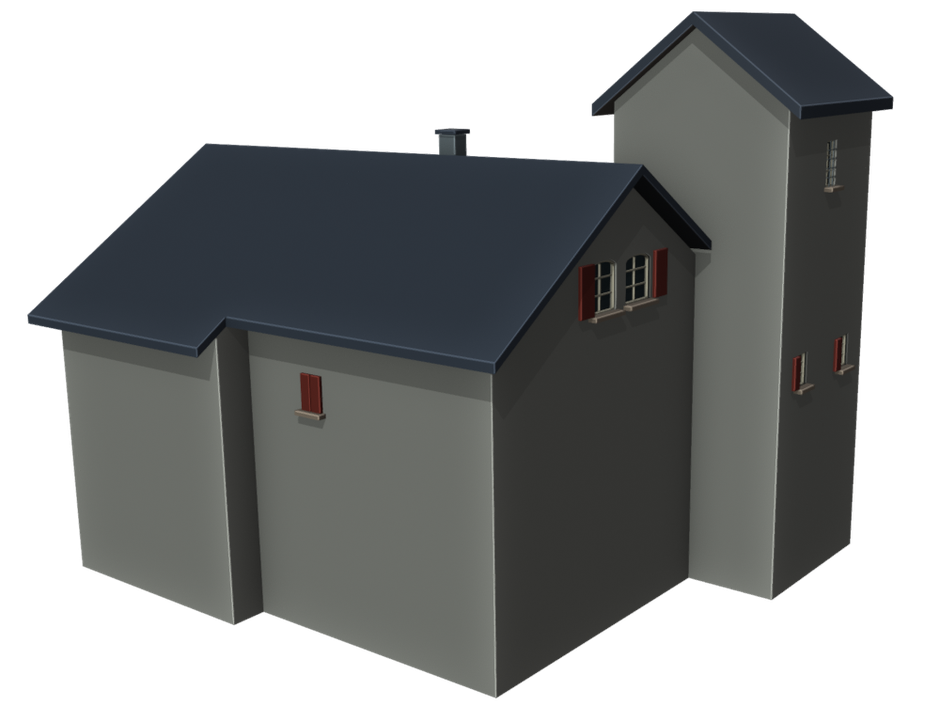}
        \caption*{$\mathcal{I}_{\mathrm{VC}}$}
        \label{fig:evidence_vc}
    \end{subfigure}
    \hfill
    \begin{subfigure}[t]{0.31\textwidth}
        \centering
        \includegraphics[width=\linewidth]
            {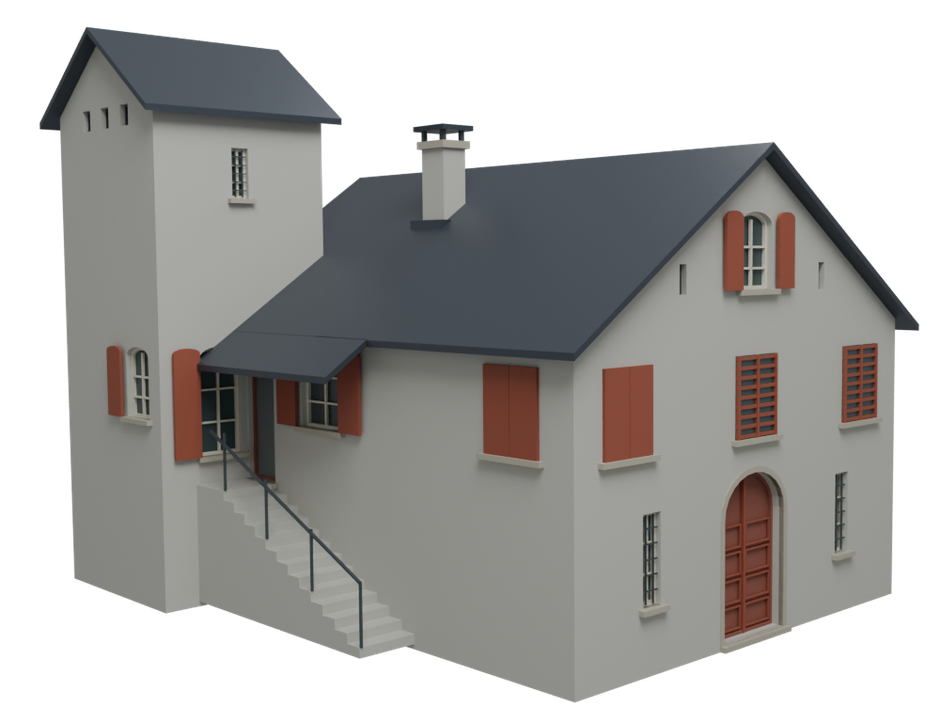}
        \par\vspace{-1mm}
        \includegraphics[width=\linewidth]
            {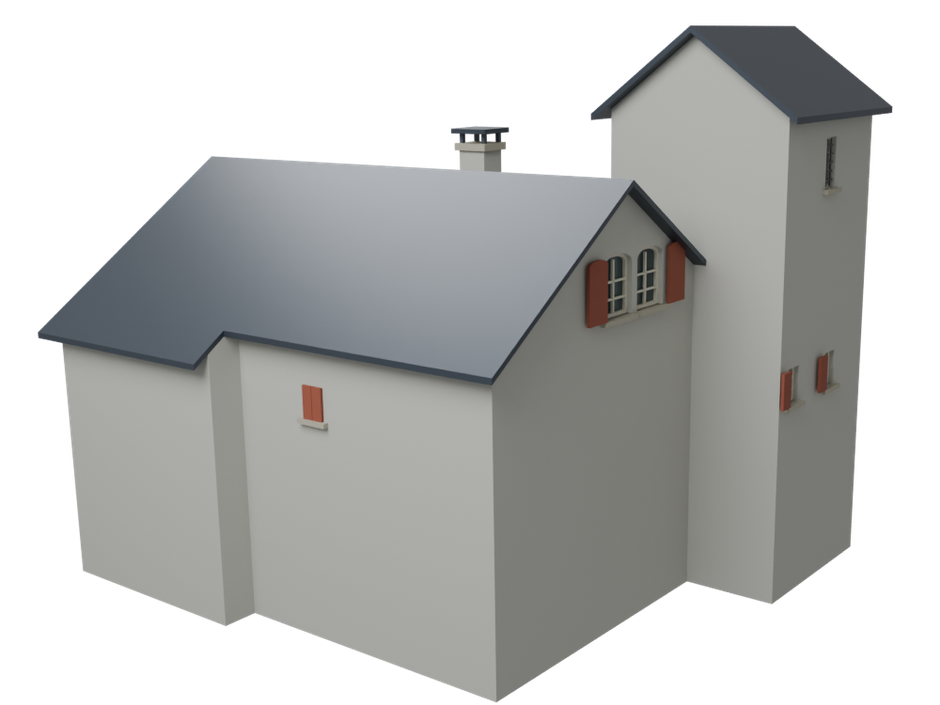}
        \caption*{$\mathcal{I}_{\mathrm{VP}}$}
        \label{fig:evidence_vp}
    \end{subfigure}

    \par\vspace{4mm}

    \begin{subfigure}[t]{0.31\textwidth}
        \centering
        \includegraphics[width=\linewidth]
            {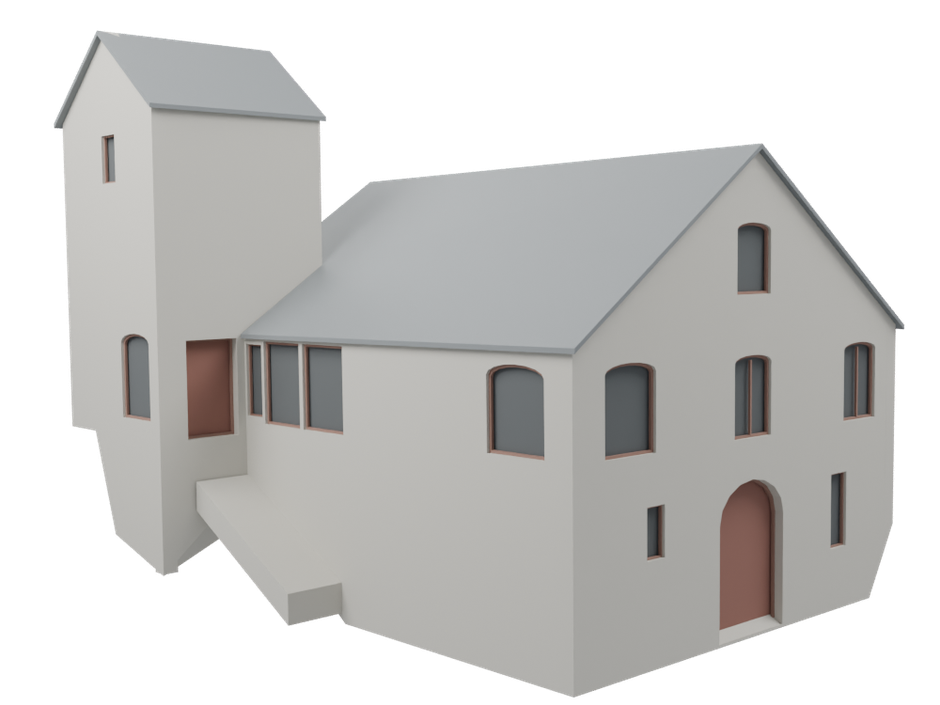}
        \par\vspace{-1mm}
        \includegraphics[width=\linewidth]
            {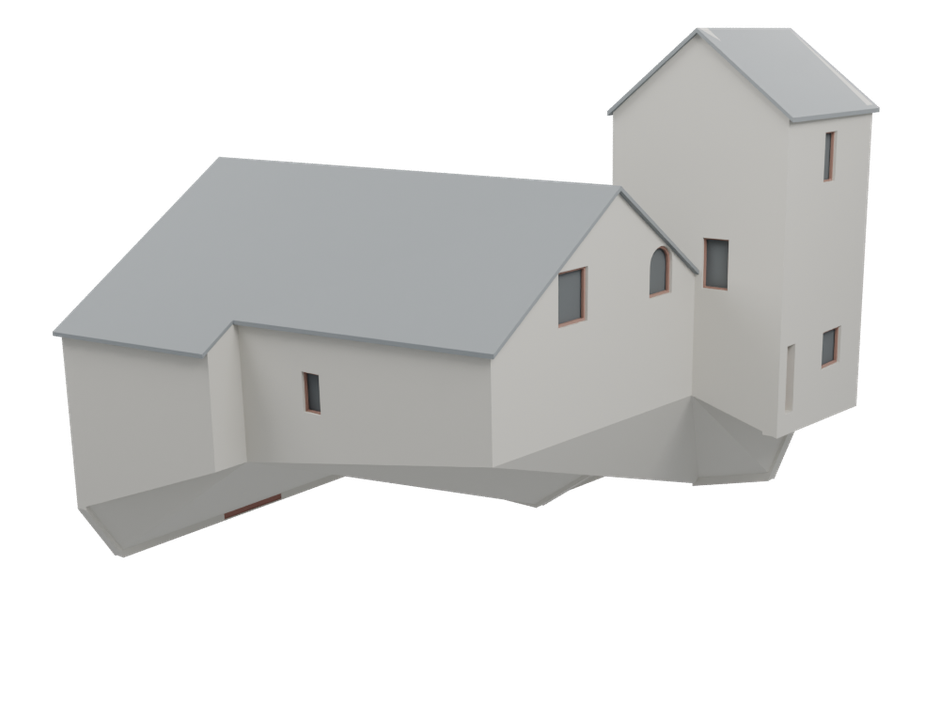}
        \caption*{$\mathcal{I}_{\mathrm{CP}}$}
        \label{fig:evidence_cp}
    \end{subfigure}
    \hfill
    \begin{subfigure}[t]{0.31\textwidth}
        \centering
        \includegraphics[width=\linewidth]
            {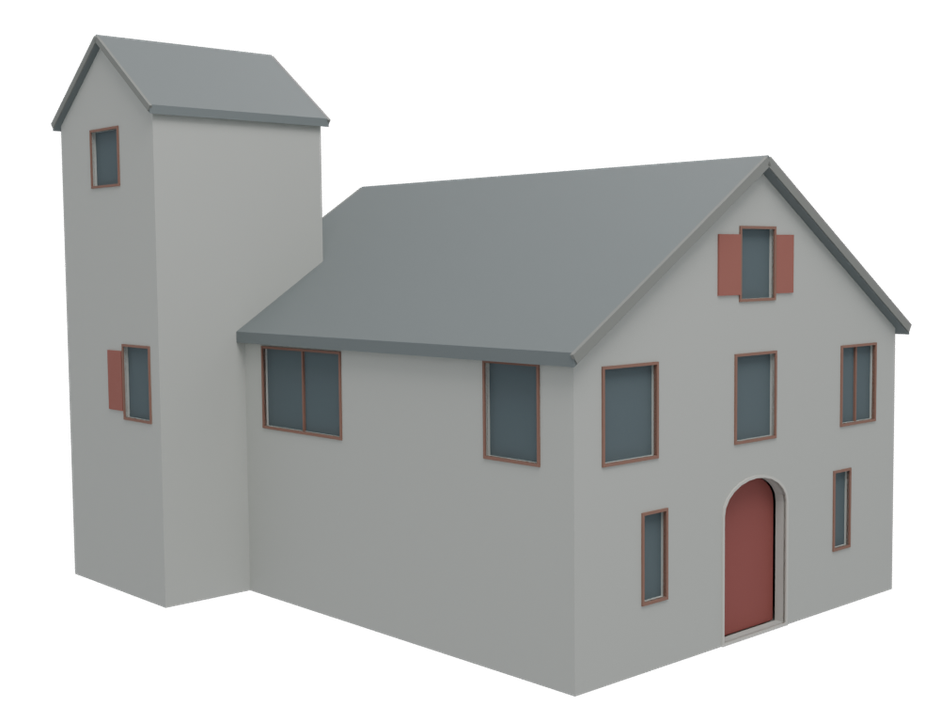}
        \par\vspace{-1mm}
        \includegraphics[width=\linewidth]
            {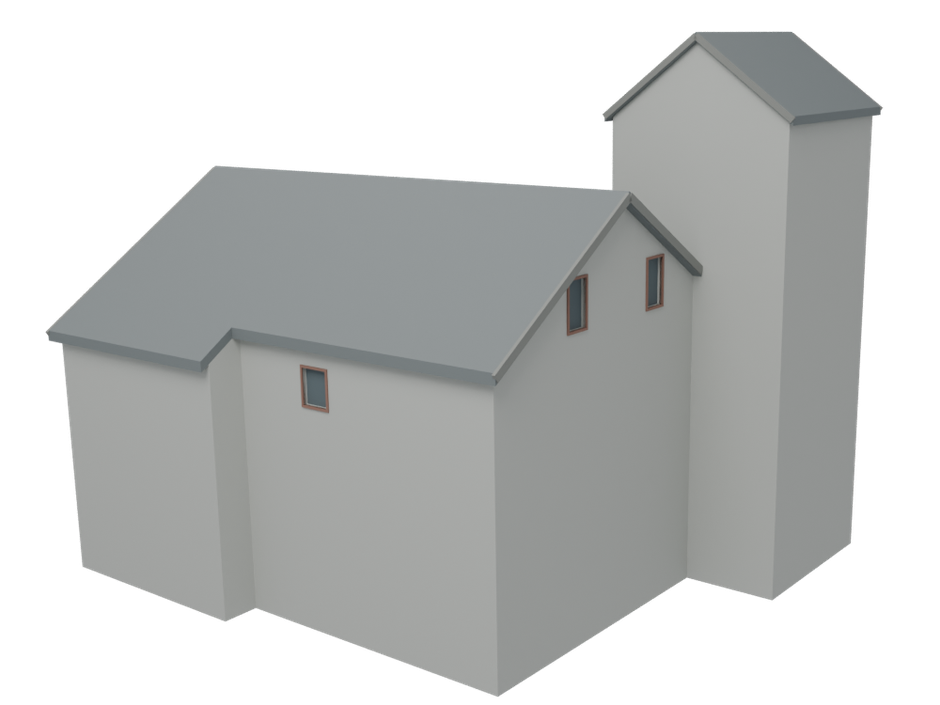}
        \caption*{$\mathcal{I}_{\mathrm{P}}$}
        \label{fig:evidence_p}
    \end{subfigure}
    \hfill
    \begin{subfigure}[t]{0.31\textwidth}
        \centering
        \includegraphics[width=\linewidth]
            {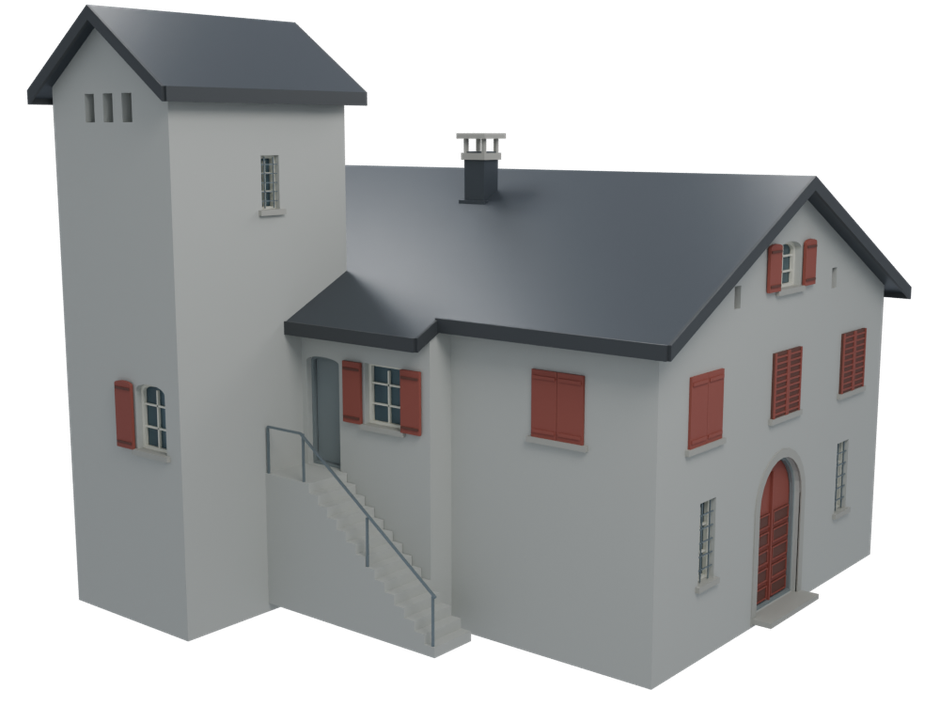}
        \par\vspace{-1mm}
        \includegraphics[width=\linewidth]
            {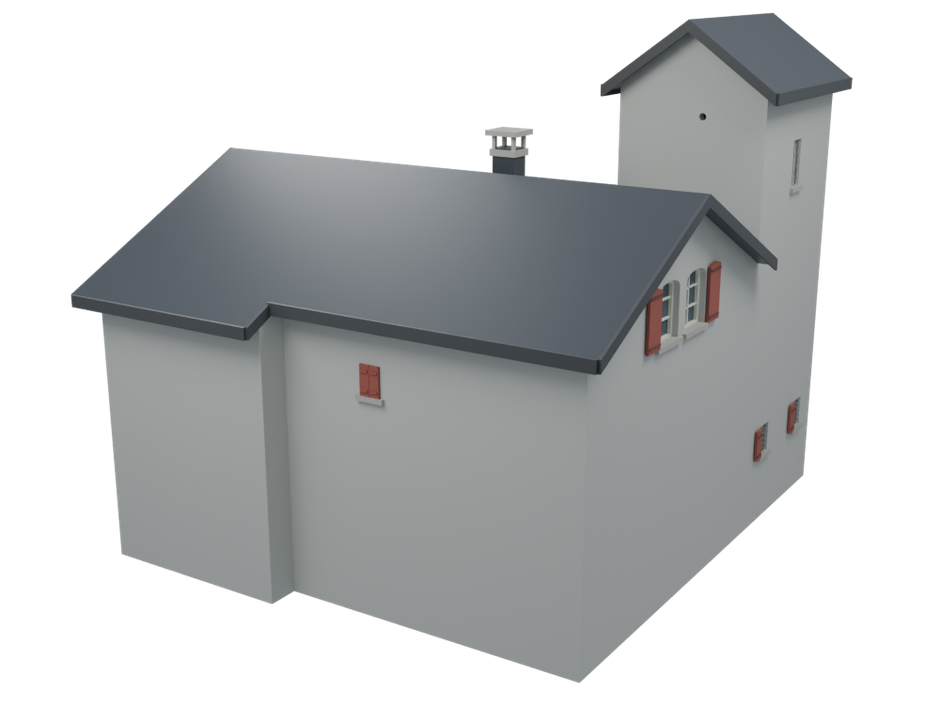}
        \caption*{$\mathcal{I}_{\mathrm{V}}$}
        \label{fig:evidence_v}
    \end{subfigure}

    \caption{Effect of the reconstruction evidence provided to the Astra agent. Each panel shows two representative views of the Gerlmerbahn reconstruction obtained using a different combination of images $\mathcal{V}$, calibrated camera information $\mathcal{C}$, and sparse geometric evidence $\mathcal{P}$. Rendering appearance is retained as generated by each run. The image-only case $\mathcal{I}_{\mathrm{V}}$ has an independently inferred scale and coordinate system and is shown for qualitative comparison only.}
    \label{fig:evidence_ablation}        
\end{figure}

The image-only configuration $\mathcal{I}_{\mathrm{V}}$ represents the limiting case. Without calibrated cameras or an SfM point cloud, the agent still produces a recognizable reconstruction, including the main building, tower, roof, chimney, openings, and exterior stair. However, its scale and coordinate system are independently inferred, and Figure~\ref{fig:evidence_ablation} shows larger differences in the relative geometry of the tower and main building. A direct FRDS or point-cloud comparison in the common SfM frame would therefore require an additional external alignment and is not reported here.

Overall, the ablation shows that removing an evidence modality changes not only the information available to the agent but also the computational procedure used to reconstruct the building. Depending on the available inputs, the agent shifts between calibrated multi-view triangulation, point-cloud fitting and geometric analysis, or image-based architectural interpretation. The complete $\mathcal{I}_{\mathrm{VCP}}$ configuration provides the strongest overall result because visual, geometric, and camera information can be jointly used during reconstruction, validation, and refinement. Figure~\ref{fig:evidence_ablation} illustrates the corresponding differences in the reconstructed models.

\subsection{Effect of Astra model configuration}
\label{sec:model_configuration_results}

The model-configuration sensitivity analysis compares the Astra \emph{Light}, \emph{Medium}, \emph{High}, and \emph{Extra-High} configurations while retaining the \emph{Normal} reconstruction specification and the complete evidence $\mathcal{I}_{\mathrm{VCP}}$. The quantitative results are reported in Table~\ref{tab:model_configuration_results}.

\begin{table}[H]
\centering
\caption{Sensitivity of the Gerlmerbahn reconstruction to the Astra model configuration.}
\label{tab:model_configuration_results}
\small
\begin{tabular}{lcccc}
\hline
\textbf{Astra configuration} & \textbf{Time} &
\textbf{FRDS $\uparrow$} &
\textbf{$D_{\mathrm{CD}}\downarrow$} &
\textbf{IMF $\downarrow$} \\
\hline
\emph{Light}      & 23:27 & 0.8915 & 0.0473 & 0.0200 \\
\emph{Medium}     & \textbf{11:33} & 0.8991 & 0.0420 & 0.0198 \\
\emph{High}       & 20:00 & 0.9556 & \textbf{0.0161} & 0.0177 \\
\emph{Extra-High} & 29:28 & \textbf{0.9646} & 0.0190 & \textbf{0.0175} \\
\hline
\end{tabular}
\end{table}

\begin{figure}[H]
    \centering

    \begin{subfigure}[t]{0.24\textwidth}
        \centering
        \includegraphics[width=\linewidth]{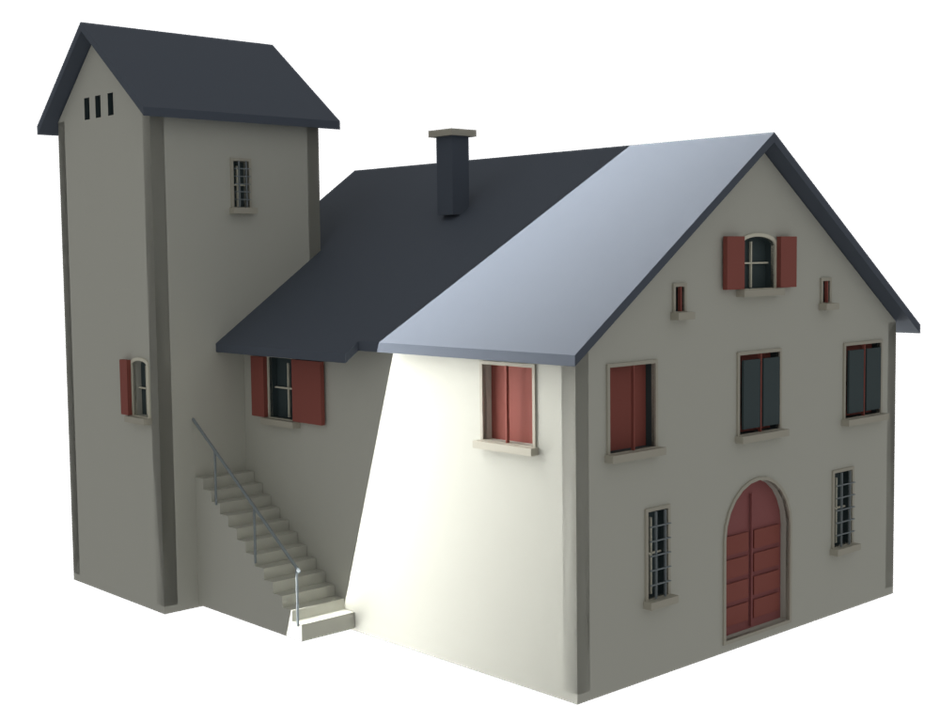}
        \includegraphics[width=\linewidth]{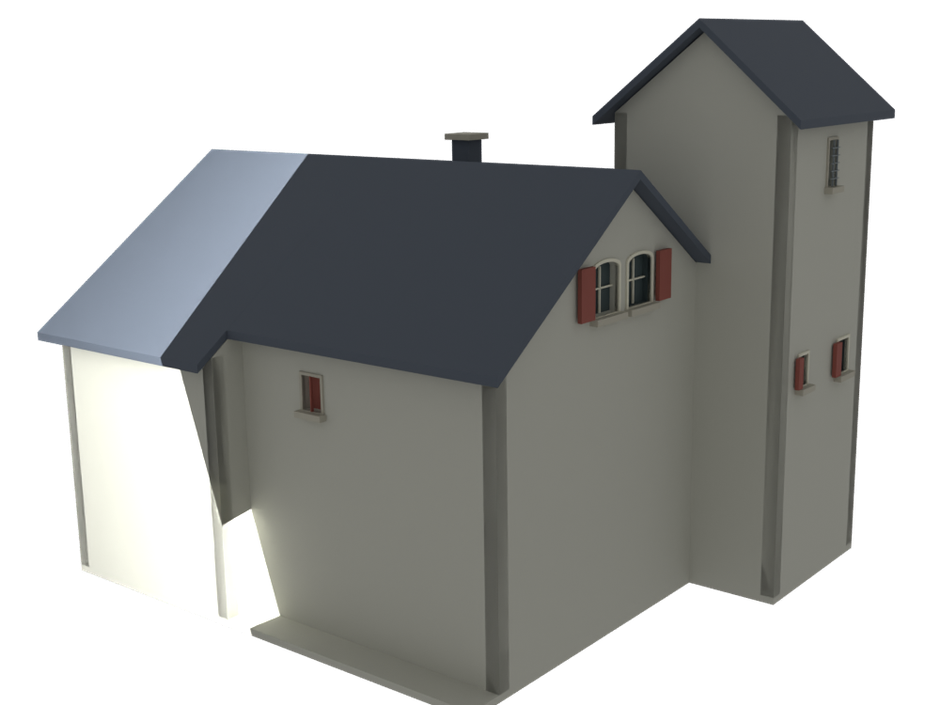}
        \caption*{Light}
    \end{subfigure}
    \begin{subfigure}[t]{0.24\textwidth}
        \centering
        \includegraphics[width=\linewidth]{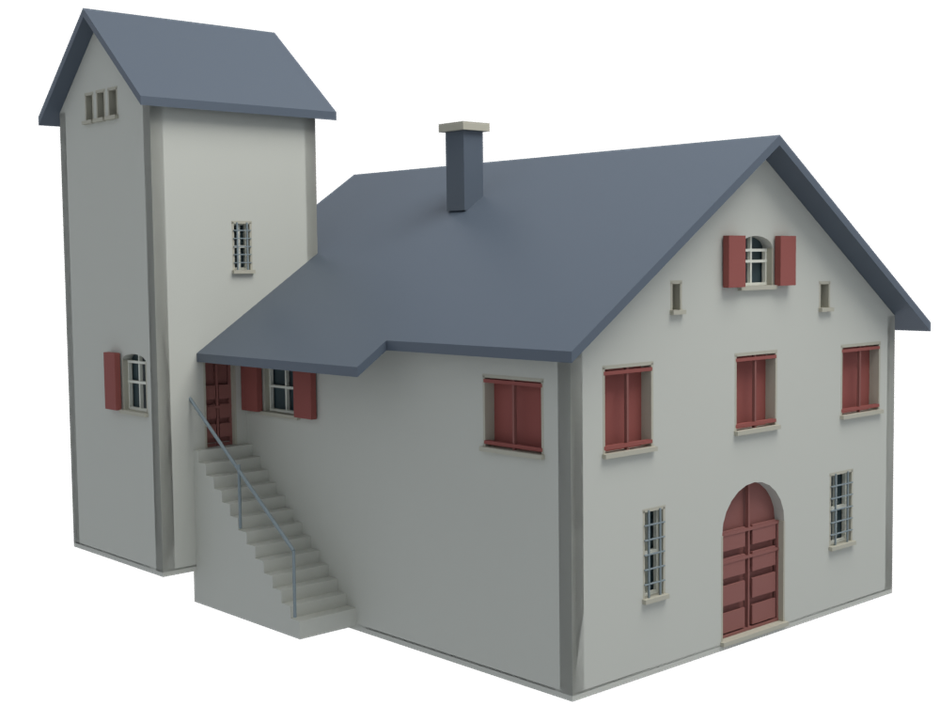}
        \includegraphics[width=\linewidth]{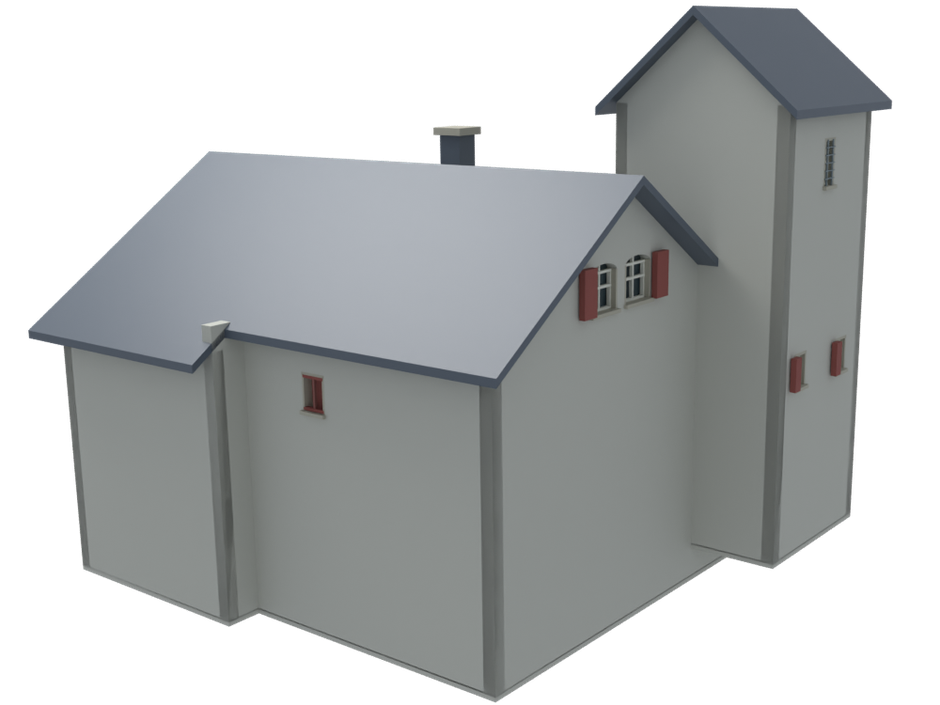}
        \caption*{Medium}
    \end{subfigure}
    \begin{subfigure}[t]{0.24\textwidth}
        \centering
        \includegraphics[width=\linewidth]{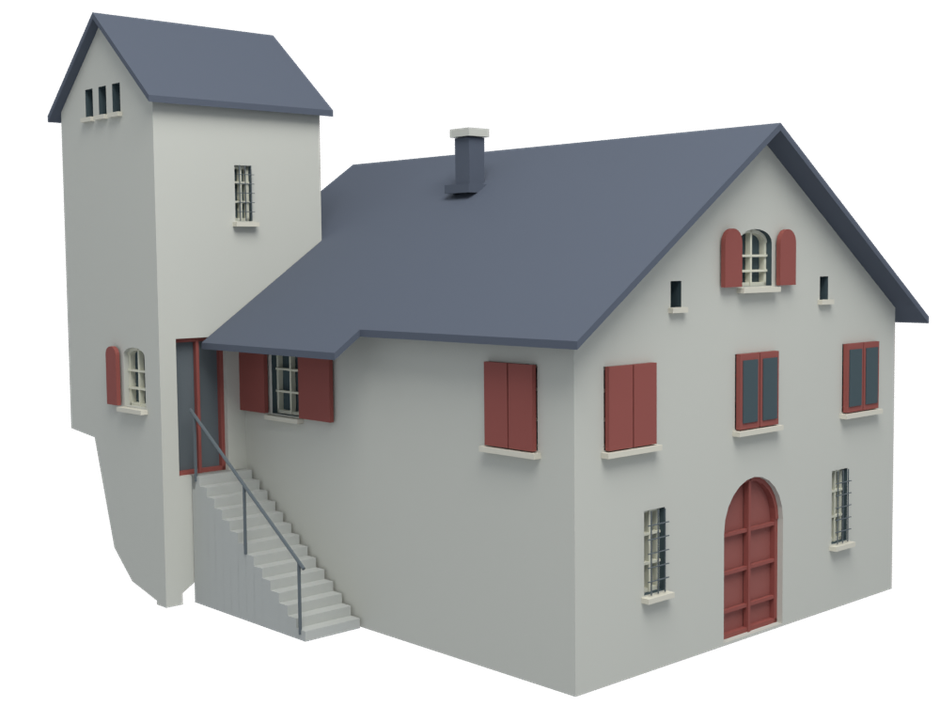}
        \includegraphics[width=\linewidth]{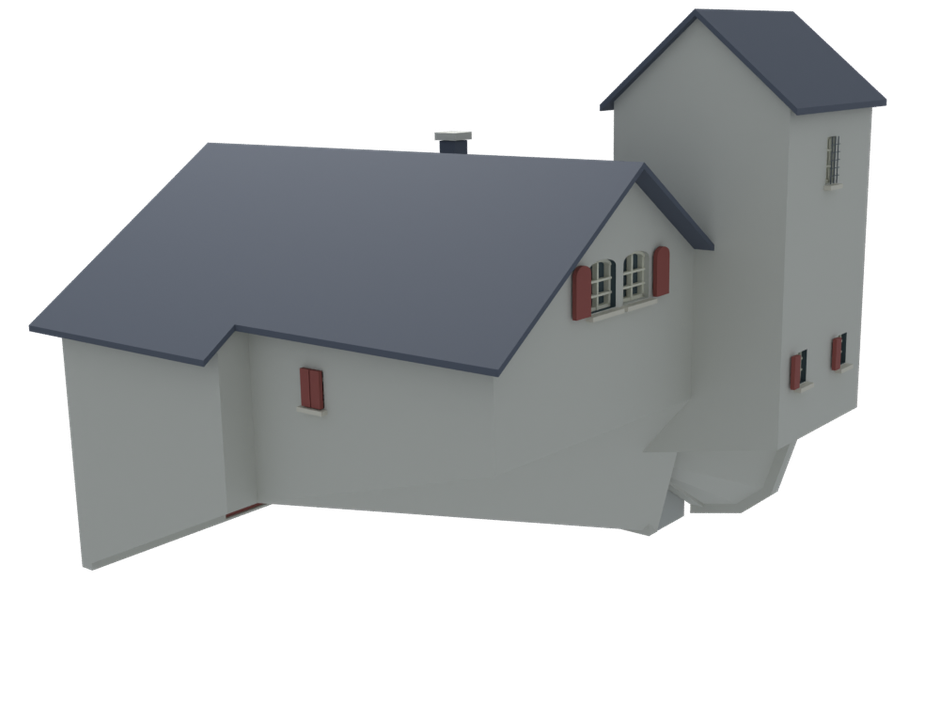}
        \caption*{High}
    \end{subfigure}
    \begin{subfigure}[t]{0.24\textwidth}
        \centering
        \includegraphics[width=\linewidth]{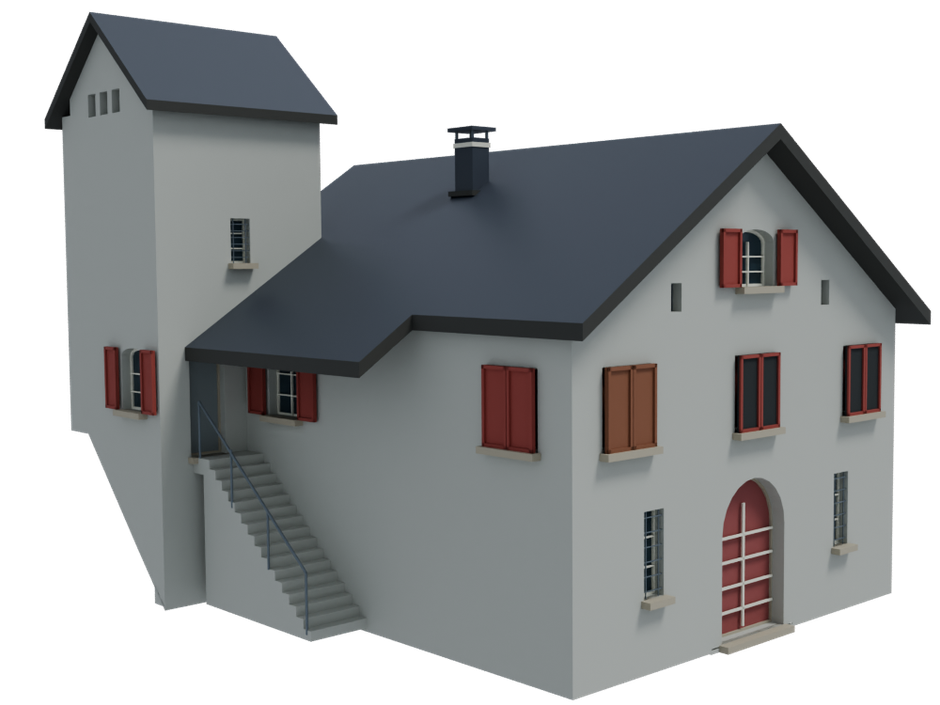}
        \includegraphics[width=\linewidth]{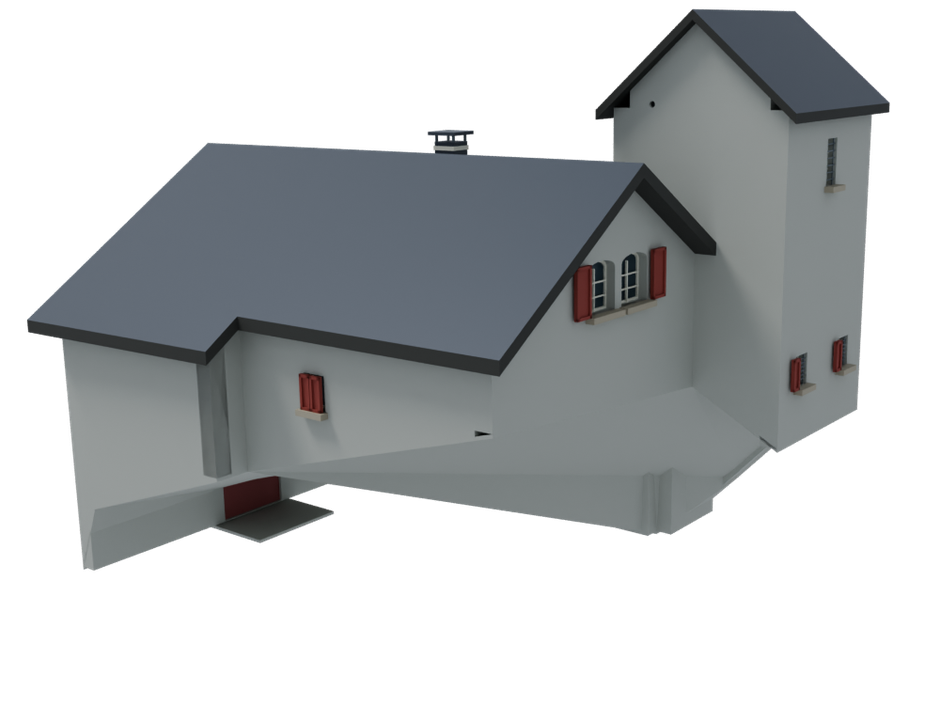}
        \caption*{Extra}
    \end{subfigure}

    \caption{Sensitivity of the Gerlmerbahn reconstruction to the Astra model configuration. \emph{Light}, \emph{Medium}, \emph{High}, and \emph{Extra-High} are shown from the same two viewpoints using the same reconstruction evidence and \emph{Normal} specification. Rendering appearance is retained as generated by each run.}
    \label{fig:model_configuration_sensitivity}

\end{figure}

The four configurations recover the same principal architectural organization, and the qualitative differences in Figure~\ref{fig:model_configuration_sensitivity} are relatively small compared with those observed in the evidence ablation. The main building, tower, roof configuration, chimney, openings, and exterior stair are reproduced across the four runs, although local differences remain in the placement and representation of individual components and in the treatment of the lower envelope.

The quantitative results show a clearer separation. \emph{High} and \emph{Extra-High} provide substantially higher FRDS and lower geometric discrepancies than \emph{Light} and \emph{Medium}. \emph{Extra-High} obtains the highest FRDS and lowest IMF, whereas \emph{High} produces the lowest symmetric Chamfer distance. As observed in the preceding experiments, part of this difference is influenced by how each run treats geometry hidden by the surrounding slope. More extensive inferred lower surfaces have limited support in the visible image masks and sparse point cloud and can consequently reduce FRDS or increase $D_{\mathrm{CD}}$ even when the completion is geometrically plausible.

Processing time does not vary monotonically with the Astra configuration. \emph{Medium} is the fastest run at 11:33, whereas \emph{Light} requires 23:27 despite using a lower configuration. \emph{High} completes in 20:00, while \emph{Extra-High} requires the longest time, 29:28. The run reports also show different use of the permitted refinement budget: \emph{Light} performs one geometry-refinement iteration, \emph{Medium} two, and \emph{High} and \emph{Extra-High} three. These observations suggest that the model configuration can influence not only the resulting geometry but also the computational trajectory and the extent of refinement performed during the run.

For this case, the qualitative similarity among the four reconstructions indicates that the lower configurations can already recover the principal LOD3 building structure, while \emph{High} and \emph{Extra-High} provide stronger quantitative agreement and somewhat more refined local geometry. \emph{Medium} additionally demonstrates that a comparatively short execution can still produce a plausible reconstruction. However, these results correspond to one independent execution per configuration, and processing time and reconstruction quality can also vary between repeated agentic runs. 

Figure~\ref{fig:model_configuration_sensitivity} compares the four reconstructions from the same two viewpoints.

\subsection{Run-to-run variability}
\label{sec:run_variability}

Building~03, Petrinja school, was selected for the repeated-run experiment because it required the longest processing time among the 24 reference reconstructions and represents one of the more architecturally extensive cases, with long façades, numerous repeated openings, projecting entrances, courtyard returns, and a complex roof configuration. It therefore provides a demanding case for examining whether independent executions under unchanged conditions follow similar reconstruction trajectories and produce comparable final models. Both runs use the same complete evidence $\mathcal{I}_{\mathrm{VCP}}$, \emph{Normal} reconstruction specification, and Astra \emph{High} model configuration. Their quantitative results are reported in Table~\ref{tab:run_variability_results}.

\begin{table}[H]
\centering
\caption{Two independent AstraLOD3 reconstructions of Building~03 under the same experimental conditions.}
\label{tab:run_variability_results}
\small
\begin{tabular}{lcccc}
\hline
\textbf{Run} & \textbf{Time} &
\textbf{FRDS $\uparrow$} &
\textbf{$D_{\mathrm{CD}}\downarrow$} &
\textbf{IMF $\downarrow$} \\
\hline
Run 1 & 38:47 & 0.9873 & 0.0188 & 0.0234 \\
Run 2 & 42:21 & 0.9609 & 0.0218 & 0.0294 \\
\hline
\end{tabular}
\end{table}

Both executions recover the same principal building organization and major architectural components, and both use the maximum of three permitted geometry-refinement iterations. Nevertheless, the computational trajectories differ. According to the run reports, Run~1 uses six representative calibrated views during validation, whereas Run~2 selects seven validation views. Run~2 also requires approximately 3.5 minutes more processing time. These differences show that identical evidence and reconstruction requirements do not lead the agent through an identical sequence of analysis, validation, and refinement operations.

\begin{figure}[H]
    \centering
    \captionsetup[subfigure]{
        justification=centering,
        font=small,
        skip=2mm
    }

    \begin{subfigure}[t]{0.46\textwidth}
        \centering
        \includegraphics[width=\linewidth]
            {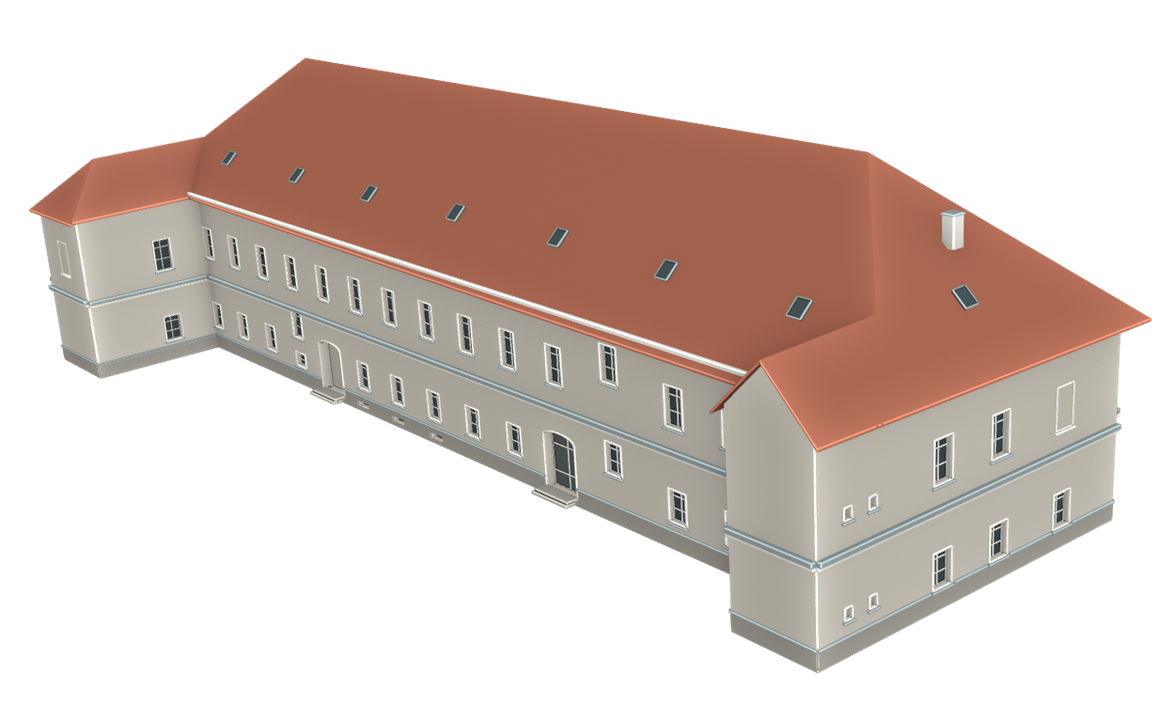}
        \par\vspace{-1mm}
        \includegraphics[width=\linewidth]
            {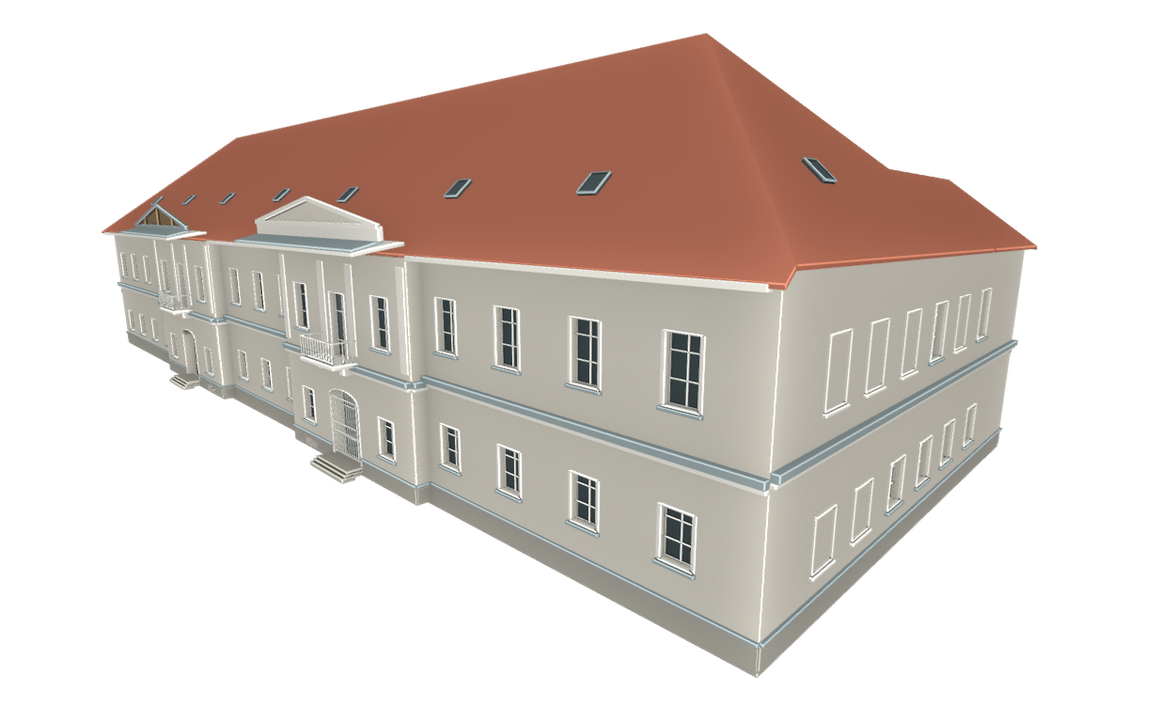}
        \caption*{\emph{Run 1}}
        \label{fig:run_variability_run1}
    \end{subfigure}
    \hfill
    \begin{subfigure}[t]{0.46\textwidth}
        \centering
        \includegraphics[width=\linewidth]
            {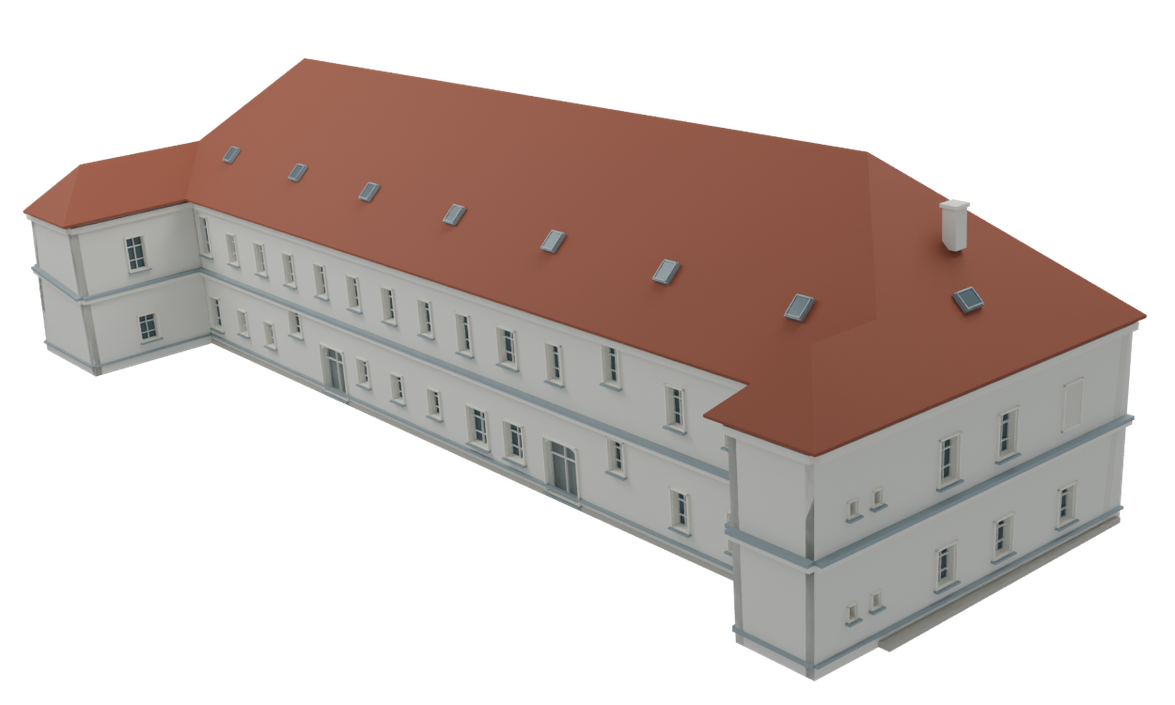}
        \par\vspace{-1mm}
        \includegraphics[width=\linewidth]
            {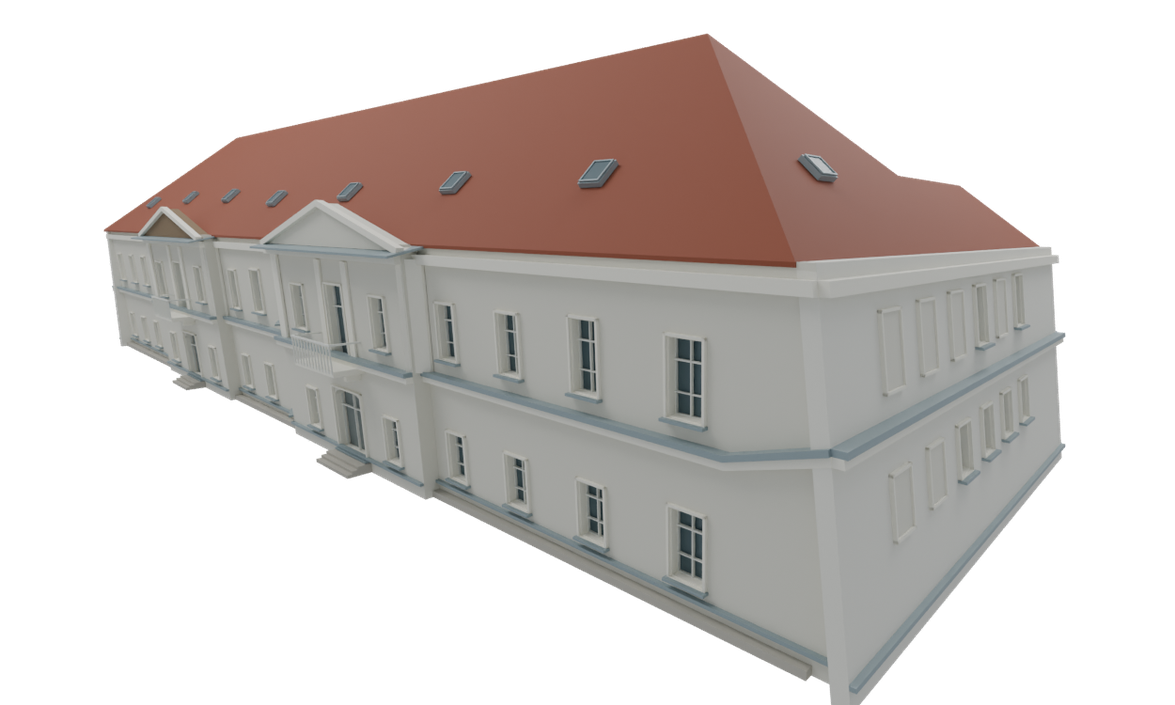}
        \caption*{\emph{Run 2}}
        \label{fig:run_variability_run2}
    \end{subfigure}

    \caption{Run-to-run variability for two independent Petrinja-school reconstructions obtained using the same evidence $\mathcal{I}_{\mathrm{VCP}}$, \emph{Normal} reconstruction specification, and Astra \emph{High} model configuration. Both results are shown from two common viewpoints.}
    \label{fig:run_variability}
\end{figure}

The final quantitative results also vary, with Run~1 obtaining higher FRDS and lower $D_{\mathrm{CD}}$ and IMF. With only two independent executions, these differences should not be interpreted as a systematic advantage of either trajectory. Qualitatively, both runs reproduce the overall massing, roof configuration, opening rhythm, entrances, and other principal components, as shown in Figure~\ref{fig:run_variability}. Local differences remain in the representation and placement of individual elements, and Run~2 contains a visible inclination in one façade region that is not present in Run~1. Such a local error could potentially be addressed through additional autonomous refinement or through a subsequent user-directed correction, but neither was permitted in this experiment in order to maintain the same bounded reconstruction protocol.

This variability is distinct from the re-executability of an individual generated solution. Once a run produces its final \texttt{build\_model.py}, the corresponding model can be regenerated from an empty Blender scene using the fixed script and parameters. A new Astra execution, however, can select different intermediate analyses, validation views, geometric parameterizations, and refinement decisions. The experiment therefore shows that the procedural output of an individual run is re-executable, while the complete agentic reconstruction process is not deterministic.

\subsection{Agentic behavior and limitations}
\label{sec:agentic_behavior_results}

Taken together, the experiments show a consistent distinction between the reconstruction requirements defined before execution and the computational trajectory followed by the agent. The required final outputs were produced across all runs, whereas the intermediate analyses, reconstruction plans, supporting files, validation choices, and refinement operations varied between buildings and experimental configurations. The computational procedures also adapted to the observed architecture: predominantly planar buildings were commonly treated using plane fitting, local architectural coordinates, rectified façade views, and image--geometry intersections, whereas more irregular, curved, or damaged buildings required different geometric interpretations and representations. This behavior is consistent with the bounded-autonomy formulation of AstraLOD3: the objective, evidence, constraints, and output contract are specified through $\mathcal{R}$, while the computational procedures used to satisfy them are determined during execution.

Iterative validation and refinement are an important part of this process. Across the experiments, subsequent iterations corrected issues involving roof intersections, façade extents, opening dimensions and placement, stairs or balconies, lower-envelope geometry, and implementation errors in the generated procedural models. The maximum of three geometry-refinement iterations was imposed as an experimental constraint to bound processing time and maintain comparable conditions across runs, rather than as an indication that the reconstruction had fully converged. Some remaining local discrepancies, including the inclined façade observed in the second Petrinja-school run, suggest that additional autonomous iterations could further improve individual models. A complementary extension would be user-directed refinement, in which the generated model and calibrated evidence are retained and the agent is asked to correct specific components after the autonomous reconstruction has finished.

The results are also limited by the information contained in the reconstruction evidence and by the evaluation itself. The SfM datasets do not generally provide an absolute physical scale, sparse point clouds can contain residual ground, vegetation, clutter, or damaged material, and roofs or other weakly textured surfaces may contain limited geometric support. Similarly, occlusion by terrain, vegetation, or other objects leaves portions of the building unobserved. In such cases, the agent may simplify the geometry or infer plausible hidden surfaces; these inferred elements are recorded through the observed/inferred provenance but should not be interpreted as survey-grade measurements of unobserved construction. The reported FRDS, $D_{\mathrm{CD}}$, and IMF values should therefore be interpreted as complementary measures of consistency with the available image and SfM evidence rather than as complete measures of three-dimensional architectural accuracy.

Overall, the contribution of AstraLOD3 is not a new geometric primitive or a fixed reconstruction algorithm, but a constrained agentic formulation in which a general-purpose multimodal model can select and combine computational procedures to produce a common structured LOD3 representation across heterogeneous building cases. The results show that this formulation can operate under different levels of guidance and reduced evidence, while the ablation and repeated-run experiments also reveal variability in execution time, intermediate decisions, and final geometric details. These characteristics define both the capabilities and the current limitations of the proposed approach and motivate further work on adaptive refinement, user-guided correction, and more explicit reconstruction of damage and other application-specific architectural information.

\section{Conclusions}
\label{sec:conclusions}

This study introduced AstraLOD3, a zero-shot multimodal agentic framework for automated reconstruction of lightweight and semantically structured LOD3 building models. The framework combines reconstruction evidence $\mathcal{I}=\{\mathcal{V},\mathcal{C},\mathcal{P}\}$, comprising multi-view images, calibrated camera information, and a filtered sparse SfM point cloud, with a human-defined reconstruction specification $\mathcal{R}$. Rather than prescribing a complete reconstruction algorithm, $\mathcal{R}$ defines the objective, evidence rules, representation requirements, execution constraints, procedural guidance, and required outputs. Operating under bounded autonomy, the Astra agent determines the computational procedures required for each case, including data inspection, geometric analysis, code generation, Blender-based model construction, validation, and iterative refinement. The reconstruction task is performed without task-specific training or fine-tuning, building-specific pretrained reconstruction pipelines, external 3D assets, ground-truth geometry, or manual correction during execution.

The framework was evaluated through 35 reconstruction runs, including a reference benchmark of 24 buildings and controlled experiments on reconstruction-specification guidance, input evidence, Astra model configuration, and run-to-run variability. Across the 24 reference cases, FRDS ranged from 0.9126 to 0.9937, with a mean of 0.9647, showing consistently strong agreement between the projected reconstructions and the observed building regions. Comparisons with previous purpose-built LOD3 and DADT approaches showed generally stronger image-space agreement and geometric consistency of a similar order, while AstraLOD3 recovered a richer set of architectural components in several cases. The results are particularly relevant because these models were obtained through a general-purpose multimodal agent rather than a reconstruction pipeline explicitly designed for the considered building datasets.

The controlled experiments further clarify the role of the different elements of the framework. Reducing the amount of guidance in $\mathcal{R}$ changed both the computational trajectory and the interpretation of the reconstruction task, with the \emph{Short} specification providing the strongest quantitative result for the investigated case and the less constrained \emph{Nano} specification introducing additional inferred geometry and image-derived textures. The evidence ablation showed that images, calibrated cameras, and sparse geometry provide complementary information: photographs support architectural interpretation and fine detail, point clouds constrain the three-dimensional building geometry, and calibrated cameras connect visual observations to the SfM reference frame. Meaningful reconstructions were nevertheless obtained under substantially reduced evidence, including from the sparse point cloud alone. The Astra model-configuration experiment showed comparatively small qualitative differences between \emph{Light}, \emph{Medium}, \emph{High}, and \emph{Extra-High} for the principal building geometry, although processing time, refinement behavior, and quantitative agreement varied between executions. Finally, the repeated reconstruction demonstrated that the generated procedural solution is re-executable, but the complete agentic reconstruction process is not deterministic and can follow different computational paths under identical input conditions.

These findings also define the current limitations of AstraLOD3. The quality of the reconstruction remains dependent on the available evidence, particularly for weakly textured roofs, occluded façades, concealed foundations, and other regions that are poorly observed in images or sparse geometry. Individual SfM datasets have arbitrary scale, and residual ground, vegetation, clutter, or damaged material can remain in automatically filtered point clouds and influence geometric evaluation. Similarly, FRDS, $D_{\mathrm{CD}}$, and IMF measure consistency with the available image and SfM evidence rather than complete three-dimensional accuracy against independent geometric ground truth. The maximum of three geometry-refinement iterations used in this study was introduced to bound computational effort and provide comparable experimental conditions; it should therefore not be interpreted as an intrinsic convergence criterion. Some remaining local errors indicate that additional autonomous refinement could improve individual reconstructions. The controlled ablations are also based primarily on a single building and one independent execution per configuration, and broader repeated experiments are required to characterize statistically the variability associated with reconstruction specifications and model configurations.

Several extensions follow directly from these limitations. Adaptive refinement and stopping criteria could allow the agent to continue reconstruction only when measurable improvements are obtained, reducing unnecessary computation while permitting difficult cases to receive additional attention. A user-in-the-loop mode could retain the generated model, evidence, and reconstruction history and allow targeted instructions for correcting individual openings, roof elements, façades, or other local discrepancies after the autonomous stage. Explicit uncertainty estimates could also be associated with reconstructed components, complementing the current observed/inferred provenance and providing a clearer distinction between directly supported geometry and geometric completion. Additional sensing modalities, including denser geometric observations, could further constrain weakly observed surfaces and reduce ambiguities associated with sparse SfM evidence.

A further research direction is the specialization of the underlying agent for building reconstruction. The present work deliberately establishes a zero-shot baseline using a general-purpose foundation model. Task-specific post-training, fine-tuning, or specialized learned components could subsequently be investigated to improve geometric interpretation, architectural semantics, refinement efficiency, and consistency across repeated runs. Such developments would no longer constitute the same zero-shot setting studied here, but comparison against AstraLOD3 would provide a useful basis for determining how much performance is gained through specialization relative to the general-purpose agentic approach. The framework could also be extended toward damage-aware reconstruction, including explicit representation of damaged roofs, cracks, displaced components, and other post-disaster conditions, as well as integration with structural assessment and digital-twin workflows.

To support further research and reproducibility, the experimental configurations, reconstruction specifications, generated models, procedural scripts, run reports, intermediate results, and evaluation outputs produced in this study will be released publicly, together with the associated benchmark data where redistribution is permitted. Beyond providing the individual reconstruction results, this resource documents how a general-purpose multimodal agent behaves across different buildings, evidence configurations, levels of guidance, and model configurations, and can therefore support future comparisons of agentic reconstruction strategies.

Overall, AstraLOD3 demonstrates that building-scale LOD3 reconstruction can be formulated as a constrained agentic process rather than only as a fixed geometric pipeline or task-specific learning problem. Within this formulation, a general-purpose multimodal model dynamically selects and combines computational procedures according to the available evidence and reconstruction requirements. The study therefore contributes both a reconstruction framework and an experimental basis for investigating agentic methods in architectural and infrastructure modeling. This direction opens opportunities for more adaptive, interactive, and application-specific generation of structured digital representations for building documentation, inspection, damage assessment, and digital-twin applications.


\section*{Acknowledgments}

This research is supported by the Ministry of Education, Singapore, under its Academic Research Fund Tier 1 (Grant No. RS52/25) for the project ``Spatio-temporal visual perception and AI-agent-based scene representation for robotic support in construction''.\\

\noindent\textbf{Declaration of generative AI and AI-assisted technologies in the writing process}

During the preparation of this work the author(s) used ChatGPT in order to check grammar and improve readability. After using this tool/service, the author(s) reviewed and edited the content as needed and take(s) full responsibility for the content of the publication.\\

\noindent\textbf{\hl{Data availability statement}}

The source code, reconstruction specifications, generated models, evaluation resources, and associated data will be made publicly available at
\url{https://github.com/disc-laboratory/AstraLOD3}.
A project webpage containing an overview of AstraLOD3, visual results, videos, and links to the associated resources will be available at
\url{https://disc-laboratory.github.io/AstraLOD3/}.

\clearpage

\appendix
\section{Reference reconstruction specification}
\label{app:reconstruction_specification}

The complete reconstruction specification $\mathcal{R}$ used for the
reference AstraLOD3 experiments is reproduced below. This specification
corresponds to the \emph{Normal} configuration introduced in
Section~\ref{sec:controlled_experiments}. For presentation, the
building-specific directory name used in each experiment is replaced by
the generic placeholder \texttt{<BUILDING>}; during execution, this
placeholder corresponds to the directory of the building being
reconstructed.

\begin{PromptBlock}
You are acting as an autonomous 3D building-reconstruction agent.

Use the conda environment 'astralod3'

Your task is to reconstruct the building contained in this dataset as a geometrically consistent, lightweight architectural 3D model using Blender.

This is a ZERO-SHOT reconstruction experiment. Do not use external web resources, pretrained building-specific reconstruction pipelines, online assets, or manually downloaded 3D models.

You may inspect all files contained in the 'input/<BUILDING>' directory.

INPUTS
------
1. Multiple calibrated photographs of the same building
   (input/<BUILDING>/images_sfm/).
2. COLMAP camera intrinsics and extrinsics for these photographs
   (input/<BUILDING>/sparse).
3. A filtered sparse Structure-from-Motion point cloud representing
   only the building (input/<BUILDING>/sparse_filtered.ply).

OBJECTIVE
---------
Generate a detail-enriched but lightweight LOD3 geometric representation of the observed building.

The reconstructed model must prioritize:

1. consistency with the supplied SfM geometry;
2. consistency with the supplied images when projected from their
   corresponding camera poses;
3. preservation of architecturally relevant components;
4. compact geometric representation.

ARCHITECTURAL COMPONENTS
------------------------
Model when supported by visual or geometric evidence:

- main building envelope;
- individual facade planes;
- roof geometry;
- window openings;
- door openings;
- simplified window and door frames;
- recess depth when visually evident;
- balconies;
- parapets;
- major roof overhangs;
- chimneys;
- major cornices and facade projections.

DO NOT model:

- individual bricks;
- image textures as geometry;
- vegetation;
- cables;
- furniture;
- vehicles;
- people;
- tiny decorative objects;
- geometry that is not relevant to the architectural shape.

EVIDENCE RULE
-------------
Do not invent architectural components merely because they are common for this architectural style.

Every generated architectural component should be supported by at least one input image or by the supplied 3D geometry.

Where geometry is genuinely unobservable, use the simplest geometrically reasonable representation.

GEOMETRIC REPRESENTATION
------------------------
Use low-poly procedural geometry whenever possible.

Prefer:

- planar surfaces;
- extrusions;
- boolean openings;
- repeated/instanced components;
- simple profile geometry.

Avoid unnecessarily dense meshes.

CAMERA GEOMETRY
---------------
Carefully interpret the COLMAP camera convention before creating Blender cameras. Confirm the world-to-camera / camera-to-world transformation mathematically before using it.

Keep all reconstructed geometry in the original SfM coordinate system. Do not independently rescale or arbitrarily rotate the building unless the transformation is explicitly recorded.

WORKFLOW
--------
Perform the reconstruction using the following stages.

STAGE 1 - DATA INSPECTION

Inspect:
- images;
- camera parameters;
- point-cloud bounds and distribution.

Determine:
- principal facade planes;
- approximate building envelope;
- roof configuration;
- visible architectural components.

STAGE 2 - RECONSTRUCTION PLAN

Before generating final geometry, create:

outputs/<BUILDING>/reconstruction_plan.json

containing:
- inferred building planes;
- building dimensions;
- roof interpretation;
- architectural component inventory;
- which input images support each major component.

STAGE 3 - BLENDER GENERATION

Write a reproducible Blender Python program:

outputs/<BUILDING>/build_model.py

The entire geometry must be regenerable by running this script from an empty Blender scene.

Do not manually edit the resulting Blender file.

STAGE 4 - CAMERA VALIDATION

Import the supplied cameras into Blender.

Render the reconstructed building from representative supplied camera poses.

Store validation renders under:

outputs/<BUILDING>/validation/iteration_01/

Inspect these renders against the corresponding source photographs.

STAGE 5 - SELF-REFINEMENT

You may make at most THREE geometry-refinement iterations.

For every iteration:
1. modify build_model.py;
2. regenerate the model from an empty scene;
3. render the same validation cameras;
4. inspect geometric inconsistencies.

Do not exceed three refinement iterations.

STAGE 6 - FINAL OUTPUT

Generate:

outputs/<BUILDING>/final/model.blend
outputs/<BUILDING>/final/model.glb
outputs/<BUILDING>/final/model.obj
outputs/<BUILDING>/final/build_model.py
outputs/<BUILDING>/final/reconstruction.json
outputs/<BUILDING>/final/run_report.md

reconstruction.json must list every major architectural component with:

- component ID;
- semantic class;
- dimensions;
- position;
- source image(s) supporting the component;
- whether its geometry is directly observed or inferred.

run_report.md must document:

- files used;
- decisions taken;
- number of refinement iterations;
- geometry statistics;
- any unresolved ambiguities.

IMPORTANT
---------
Do not access files outside the experiment input directory except for software/library dependencies.

Do not inspect any ground-truth or evaluation geometry.

Do not ask the user to manually correct the model.

Complete the reconstruction autonomously using the supplied evidence.
\end{PromptBlock}

\clearpage

\bibliography{literature}

\end{document}